\documentclass[11pt]{article}
\usepackage{arxiv}

\usepackage[utf8]{inputenc}
\usepackage[T1]{fontenc}
\usepackage{amsmath,amssymb,amsfonts}
\usepackage{dsfont}
\usepackage{graphicx}
\usepackage{hyperref}
\usepackage{url}
\usepackage{booktabs}
\usepackage{multirow}
\usepackage{xcolor}
\usepackage{tikz}
\usepackage{pgfplots}
\pgfplotsset{compat=1.18}
\usetikzlibrary{trees,positioning,arrows.meta,shapes.geometric,fit,calc,decorations.pathreplacing}
\usepackage{caption}
\usepackage{subcaption}
\usepackage{algorithm}
\usepackage{algorithmic}
\usepackage{tcolorbox}
\tcbuselibrary{breakable}
\usepackage{listings}
\usepackage[numbers]{natbib}
\usepackage{multicol}
\usepackage{enumitem}

\usepackage{afterpage}
\usepackage{float}

\AtBeginDocument{%
  \setlength{\abovedisplayskip}{4pt plus 2pt minus 2pt}%
  \setlength{\belowdisplayskip}{4pt plus 2pt minus 2pt}%
  \setlength{\abovedisplayshortskip}{0pt plus 2pt}%
  \setlength{\belowdisplayshortskip}{2pt plus 2pt minus 1pt}%
  \setlength{\parskip}{3pt plus 1pt minus 0.5pt}%
}

\definecolor{codebg}{HTML}{F7F7F7}
\definecolor{codeframe}{HTML}{DDDDDD}
\definecolor{codegreen}{HTML}{2E7D32}
\definecolor{codeblue}{HTML}{1565C0}
\definecolor{codegray}{HTML}{757575}
\definecolor{codepurple}{HTML}{7B1FA2}

\lstdefinestyle{pythonstyle}{
    language=Python,
    backgroundcolor=\color{codebg},
    frame=single,
    rulecolor=\color{codeframe},
    basicstyle=\ttfamily\small,
    keywordstyle=\color{codeblue}\bfseries,
    stringstyle=\color{codegreen},
    commentstyle=\color{codegray}\itshape,
    numberstyle=\tiny\color{codegray},
    numbers=left,
    numbersep=6pt,
    breaklines=true,
    showstringspaces=false,
    tabsize=4,
    xleftmargin=12pt,
    framexleftmargin=10pt,
    aboveskip=10pt,
    belowskip=8pt,
    morekeywords={async,await,match,case},
}
\usepackage{fontawesome5}
\newtcolorbox{tipbox}[1][]{colback=green!5!white, colframe=green!60!black, fonttitle=\bfseries, title={\small\textsf{\faLightbulb\; Practitioner Tip}}, #1}
\newtcolorbox{warningbox}[1][]{colback=red!5!white, colframe=red!60!black, fonttitle=\bfseries, title={\small\textsf{\faExclamationTriangle\; Warning}}, #1}
\newtcolorbox{questionbox}[1][]{colback=blue!4!white, colframe=blue!50!black, fonttitle=\bfseries, title={\small\textsf{\faQuestionCircle\; #1}}, breakable}
\newtcolorbox{examplebox}[1][]{colback=yellow!5!white, colframe=yellow!60!black, fonttitle=\bfseries, title={\small\textsf{\faBookOpen\; #1}}}
\usepackage{float}
\usepackage[section]{placeins}

\hypersetup{
    colorlinks=true,
    linkcolor=blue!70!black,
    citecolor=blue!70!black,
    urlcolor=blue!70!black
}

\title{The Hitchhiker's Guide to System~1 Models:\\[4pt] \large Fast Decision-Making Beyond LLMs}

\author{}

\date{}

\renewcommand{\undertitle}{}
\renewcommand{\headeright}{}
\renewcommand{\shorttitle}{The Hitchhiker's Guide to System~1 Models}

\begin{document}
\maketitle

\vspace{0.8in}
\begin{center}
\includegraphics[width=0.22\textwidth]{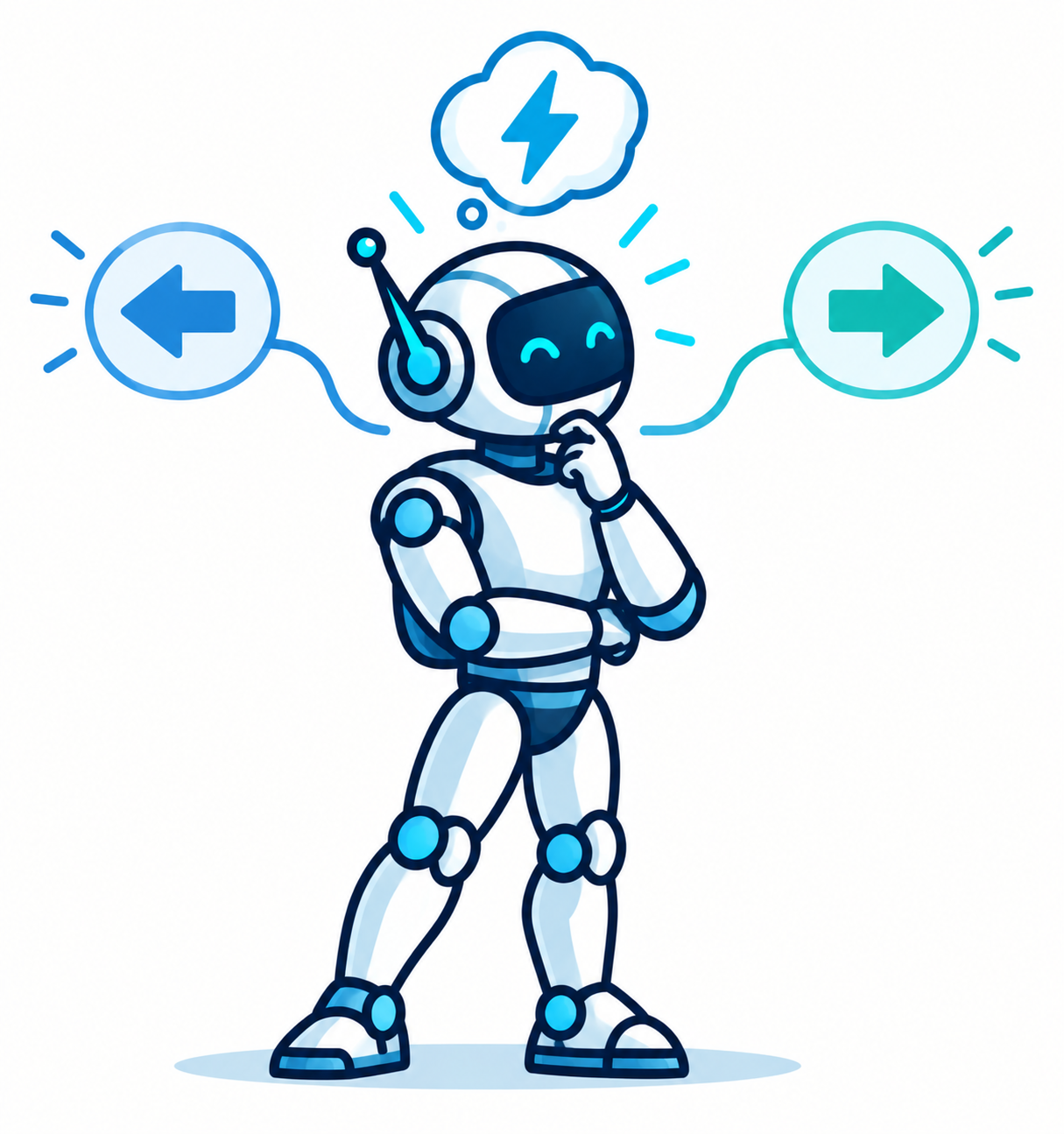}
\end{center}
\vspace{0.5in}
\begin{center}
{\large Haggai Roitman}\\[0.3em]
{2026}
\end{center}
\vfill
\begin{center}
{\normalsize Version 1.1}
\end{center}
\thispagestyle{empty}
\newpage

\thispagestyle{empty}
\vspace*{\fill}
\begin{center}
\Large\itshape
To my beloved wife Janna and daughters Inbar and Einav who make my daily decisions easier
\end{center}
\vspace*{\fill}
\newpage

\setcounter{tocdepth}{2}
\tableofcontents
\newpage

\thispagestyle{empty}
\section*{Disclaimer}
\addcontentsline{toc}{section}{Disclaimer}

This document is an independent practical guide and educational resource prepared solely by the author. The views, opinions, and conclusions expressed herein are those of the author alone and do not necessarily reflect the views of any employer, organization, or institution with which the author is or has been affiliated.

This work does not contain any proprietary, confidential, or trade-secret information. All directly referenced material is drawn from publicly available sources, including peer-reviewed publications, open-access preprints, official documentation, and open-source repositories. Some technical details discussed in this guide are noted by their vendors as proprietary or unpublished; in such cases, only publicly disclosed information is presented.

The content is provided ``as is'' without warranty of any kind, express or implied. The author makes no representations regarding the accuracy, completeness, or suitability of the information for any particular purpose. Readers should independently verify any claims, formulas, or implementation details before applying them in production systems.

\textbf{Feedback welcome.} If you find any mistakes, inaccuracies, or have suggestions for improvement, you are encouraged to send feedback to \texttt{[first\_name]r@gmail.com}.

\textbf{AI Disclosure.} Large language models were used as a research and drafting aid. All content was edited and verified by the author on a best-effort basis.

\textbf{License.} This work is licensed under the Creative Commons Attribution-ShareAlike 4.0 International License (CC BY-SA 4.0). You are free to share (copy and redistribute) and adapt (remix, transform, and build upon) this material for any purpose, including commercially, provided you give appropriate credit, provide a link to the license, indicate if changes were made, and distribute any derivative works under the same license. Full license text: \url{https://creativecommons.org/licenses/by-sa/4.0/}.
\newpage

\thispagestyle{empty}
\section*{About the Author}
\addcontentsline{toc}{section}{About the Author}

Haggai Roitman has spent over two decades at the intersection of AI research and large-scale production systems. His work bridges theory and practice -- from publishing foundational research to shipping systems that serve millions of users.

His research interests span information retrieval, recommender systems, natural language processing, large language models, reinforcement learning for LLMs, and agentic AI. He has authored more than 100 peer-reviewed publications and holds approximately 100 patents. He earned his BSc (\emph{Cum Laude}) and PhD from the Technion -- Israel Institute of Technology.

When not needing to make hard AI decisions, he can be found behind a set of turntables, mixing progressive trance and deep house.
\newpage

\section*{Preface}
\addcontentsline{toc}{section}{Preface}

\subsection*{Why This Guide Exists}

Large Language Models have achieved remarkable capabilities in text generation, reasoning, and decision-making. Yet their autoregressive, token-by-token generation paradigm -- analogous to \emph{System~2} (slow, deliberate) thinking in cognitive science -- incurs substantial latency, cost, and calibration challenges that render them ill-suited for the many software control-flow decisions that demand sub-second, well-calibrated responses.

A rapidly growing body of work addresses this gap through \emph{System~1 decision models}: architectures that map an input state directly to a typed, structured decision in a single (or few) forward pass(es), trading generative flexibility for speed, cost efficiency, and calibration. This knowledge is scattered across hundreds of papers, preprints, blog posts, and tribal knowledge within a handful of labs. This guide exists because practitioners need a single, unified reference that covers the entire landscape -- not just the theory, but the implementation details that make things actually work.

The field has seen a recent explosion of interest, with a rapidly growing number of papers, open-source models, and commercial offerings appearing in a short span of time.

\subsection*{Companion Guide}

This guide focuses specifically on System~1 decision models. Readers seeking broader coverage of LLM internals, reinforcement learning for language models, reasoning architectures, and agentic AI systems are referred to the companion volume, \emph{The Hitchhiker's Guide to Agentic AI: From Foundations to Systems}~\citep{roitman2026agentic}, which provides a comprehensive treatment of the full stack -- from transformer architectures and training infrastructure through alignment algorithms to multi-agent orchestration.

\subsection*{Who This Guide Is For}

This document is written for practitioners who build things:

\begin{itemize}
\item \textbf{ML engineers} who need to deploy fast, calibrated decision models in production systems.
\item \textbf{Applied researchers} evaluating System~1 architectures, training methodologies, and calibration strategies for their specific domains.
\item \textbf{Agent developers} building agentic systems that require sub-second routing, gating, and scoring decisions.
\item \textbf{Systems engineers} responsible for inference infrastructure who need to understand latency, throughput, and cost trade-offs.
\item \textbf{Technical leaders} making architectural decisions about when to use System~1 vs.\ System~2 models.
\end{itemize}

We assume familiarity with neural networks and basic probability. No prior knowledge of decision models, RLCD, or non-autoregressive architectures is required -- the guide builds from first principles.

\subsection*{What You Will Gain}

After reading this guide, you will be able to:

\begin{itemize}
\item \textbf{Understand the landscape} -- the five families of System~1 decision models, from encoder-based classifiers to extraction models, and when each is appropriate.
\item \textbf{Reason about calibration} -- why it matters, how to measure it, and the training techniques (RLCD, proper scoring rules) that achieve it.
\item \textbf{Train your own models} -- using RLHF, RLCD, contrastive learning, and knowledge distillation, with practical recipes and code.
\item \textbf{Deploy in production} -- with confidence estimation, System~1\,+\,System~2 cascades, and agentic workflow integration.
\item \textbf{Evaluate rigorously} -- using appropriate benchmarks, robustness tests, and calibration metrics.
\end{itemize}

\subsection*{How This Guide Is Organized}

The guide spans 18 sections and two appendices, organized in seven chapters:

\begin{enumerate}
\item \textbf{Chapter~I -- Foundations} (Sections~1--3): Introduction to dual-process theory, the calibration problem -- including the theory of proper scoring rules, calibration--sharpness decomposition, and decision-theoretic motivation -- and a unified taxonomy of System~1 decision model families.
\item \textbf{Chapter~II -- Model Families} (Sections~4--8): In-depth coverage of each family -- encoder-based classifiers, reward models, non-autoregressive decision engines, contrastive decision models, and extraction models -- with code examples and comparative analysis.
\item \textbf{Chapter~III -- Methods} (Sections~9--10): Calibration and confidence estimation techniques, and training methodologies including RLHF, RLCD, contrastive learning, and knowledge distillation.
\item \textbf{Chapter~IV -- Applications and Integration} (Sections~11--13): Application domains -- from agent routing and search reranking to text quantification and cybersecurity -- System~1 models in agentic workflows, and System~1\,+\,System~2 cascades.
\item \textbf{Chapter~V -- Evaluation and Robustness} (Sections~14--15): Benchmarks, evaluation methodology, and robustness analysis.
\item \textbf{Chapter~VI -- Open Challenges} (Section~16): Research frontiers including the expressiveness--speed trade-off, calibration under distribution shift, scaling laws, continual learning, and privacy.
\item \textbf{Chapter~VII -- Deployment and Reference} (Sections~17--18): Practitioner's checklist of common pitfalls and production tips, and conclusion. Two appendices provide an open-source model zoo and quiz questions with detailed answers.
\end{enumerate}

\subsection*{Design Philosophy}

Three principles guide this document:

\begin{enumerate}
\item \textbf{Intuition first, formalism second.} Nearly every equation is preceded by a plain-English explanation of what it means and why it matters.
\item \textbf{Implementation-aware.} Theory is useless without knowing how to make it work. We include code, training recipes, and debugging strategies throughout.
\item \textbf{Honest about what works.} We aim to distinguish approaches deployed at scale from those where evidence remains preliminary or unverified.
\end{enumerate}

\subsection*{Positioning Relative to Existing Work}

This guide is not a survey of any single topic -- each is well served by existing work -- but rather a \emph{practical reference} that integrates three dimensions that prior work treats separately: (1)~a common decision interface connecting encoder classifiers, reward models, non-autoregressive engines, contrastive models, and the emerging extraction models; (2)~a joint treatment of decision quality, calibration, latency, cost, and robustness as evaluated properties of these models; and (3)~practical guidance on how fast decision models compose into agent workflows and cascade with slower reasoning models. This extended version additionally provides the first comprehensive coverage of the open-source RLCD ecosystem (OpenJev-RLCD, Tachyone, DeepOpen), new calibration phenomena (IIA violations, probability contracts), and expanded application domains (cybersecurity, theory of mind, political science, low-resource languages). We acknowledge that our five-family taxonomy is one possible organization of a field that does not yet have a consensus structure; the proposed framing should be judged by whether it usefully connects disparate research threads, not as a claim that these communities view themselves as part of a single field.

\vspace{1em}
\hfill -- Haggai Roitman, 2026

\clearpage

\section*{Glossary of Acronyms}
\addcontentsline{toc}{section}{Glossary of Acronyms}
\label{sec:glossary}

This guide uses many acronyms drawn from machine learning, calibration theory, information retrieval, and agent research. The following reference covers the most common ones; each entry appears in the chapter where the concept is first introduced in full.

\begin{multicols}{2}
\small
\begin{description}[style=nextline, leftmargin=2.2cm, labelwidth=2cm]

\item[AE] Absolute Error (quantification metric)
\item[AI] Artificial Intelligence
\item[API] Application Programming Interface
\item[AUC] Area Under the Curve
\item[AUROC] Area Under the Receiver Operating Characteristic curve
\item[BERT] Bidirectional Encoder Representations from Transformers
\item[BM25] Best Matching 25 (term-weighting retrieval function)
\item[BT] Bradley--Terry (preference model)
\item[CDF] Cumulative Distribution Function
\item[CE] Cross-Entropy (loss function)
\item[CLIP] Contrastive Language--Image Pre-training
\item[CLM] Contrastive Language Model (System~1 model; also Causal Language Modeling)
\item[CLS] Classification token (\texttt{[CLS]})
\item[CPU] Central Processing Unit
\item[CRPS] Continuous Ranked Probability Score
\item[CROW] Classifier-Routed Organization of LLM Wikis
\item[CUDA] Compute Unified Device Architecture
\item[DPO] Direct Preference Optimization
\item[DPR] Dense Passage Retrieval
\item[ECE] Expected Calibration Error
\item[ELECTRA] Efficiently Learning an Encoder that Classifies Token Replacements Accurately
\item[EMQ] Expectation Maximization for Quantification (Saerens--Latinne--Decaestecker method)
\item[EWC] Elastic Weight Consolidation
\item[F1] F1 Score (harmonic mean of precision and recall)
\item[FFN] Feed-Forward Network
\item[FLOPs] Floating-Point Operations
\item[GLUE] General Language Understanding Evaluation (benchmark)
\item[GPU] Graphics Processing Unit
\item[GRPO] Group Relative Policy Optimization
\item[GTE] General Text Embeddings
\item[GUI] Graphical User Interface
\item[HNSW] Hierarchical Navigable Small World (graph index)
\item[IIA] Independence of Irrelevant Alternatives
\item[INT8] 8-bit Integer (quantization format)
\item[IPO] Identity Preference Optimization
\item[IR] Information Retrieval
\item[IVF] Inverted File Index (vector search)
\item[JSON] JavaScript Object Notation
\item[KD] Knowledge Distillation
\item[KL] Kullback--Leibler (divergence)
\item[KLD] KL Divergence (quantification metric)
\item[KTO] Kahneman--Tversky Optimization
\item[KV] Key--Value (cache)
\item[LLM] Large Language Model
\item[LPC] Label Prior Correction
\item[LSTM] Long Short-Term Memory (recurrent network)
\item[MAE] Mean Absolute Error
\item[MHSA] Multi-Head Self-Attention
\item[MLM] Masked Language Modeling
\item[MLP] Multi-Layer Perceptron
\item[MRR] Mean Reciprocal Rank
\item[NAR] Non-Autoregressive
\item[NDCG] Normalized Discounted Cumulative Gain
\item[NER] Named Entity Recognition
\item[NLI] Natural Language Inference
\item[NLL] Negative Log-Likelihood
\item[NLU] Natural Language Understanding
\item[NSP] Next Sentence Prediction
\item[ONNX] Open Neural Network Exchange
\item[ORM] Outcome Reward Model
\item[ORPO] Odds Ratio Preference Optimization
\item[PACC] Probabilistic Adjusted Classify and Count
\item[PCC] Probabilistic Classify and Count
\item[PII] Personally Identifiable Information
\item[PPO] Proximal Policy Optimization
\item[PRM] Process Reward Model
\item[PSR] Proper Scoring Rule
\item[QA] Question Answering
\item[RAE] Relative Absolute Error (quantification metric)
\item[RAG] Retrieval-Augmented Generation
\item[REST] Representational State Transfer
\item[RL] Reinforcement Learning
\item[RLCD] Reinforcement Learning for Calibrated Decisions
\item[RLCR] Reinforcement Learning with Calibration Rewards
\item[RLHF] Reinforcement Learning from Human Feedback
\item[RLVR] Reinforcement Learning with Verifiable Rewards
\item[RM] Reward Model
\item[ROC] Receiver Operating Characteristic
\item[RTD] Replaced Token Detection
\item[SFT] Supervised Fine-Tuning
\item[SGD] Stochastic Gradient Descent
\item[SLD] Saerens--Latinne--Decaestecker (quantification adjustment)
\item[SPLADE] SParse Lexical AnD Expansion (retrieval model)
\item[SVM] Support Vector Machine
\item[SWE-bench] Software Engineering Benchmark
\item[VLM] Vision-Language Model
\item[WER] Word Error Rate

\end{description}
\end{multicols}

\clearpage

\part{Foundations}

Before examining specific model architectures, we establish the conceptual and theoretical groundwork that the rest of the guide builds upon. This chapter contains three sections. Section~\ref{sec:intro} motivates the need for System~1 decision models by contrasting them with autoregressive LLMs on the dimensions that matter in production: latency, cost, calibration, and output structure. Section~\ref{sec:background} introduces the dual-process theory from cognitive science that inspired the System~1/System~2 distinction, develops the theory of calibration and proper scoring rules that underpins every model family discussed later, and provides a formal definition of what qualifies as a System~1 decision model. Section~\ref{sec:taxonomy} presents a unified taxonomy that organizes the field into five families -- encoder-based classifiers, reward models, non-autoregressive decision engines, contrastive decision models, and extraction models -- and traces the architectural lineage connecting them. Together, these three sections equip the reader with the vocabulary, theoretical tools, and organizing framework needed to navigate the model families, methods, and applications that follow.

\section{Introduction}
\label{sec:intro}

Daniel Kahneman's seminal work \emph{Thinking, Fast and Slow}~\citep{kahneman2011thinking} introduced the dual-process framework that distinguishes two modes of human cognition.
\textbf{System~1} is fast, automatic, intuitive, and parallel, it operates with minimal effort, recognizing faces, reading emotions, and making snap judgments. \textbf{System~2} is slow, deliberate, effortful, and sequential, it handles complex reasoning, planning, and novel problem-solving.

This framework, rooted in decades of cognitive science research~\citep{stanovich2000individual,evans2008dual,wason1974dual}, has become an increasingly influential lens for understanding artificial intelligence systems~\citep{booch2021thinking,anthony2017thinking,bengio2019system2}. Modern Large Language Models (LLMs) such as GPT-4~\citep{openai2023gpt4}, Claude~\citep{anthropic2024claude}, and Llama~\citep{touvron2023llama} are quintessential System~2 machines: they generate text token by token through autoregressive decoding, can engage in chain-of-thought reasoning~\citep{wei2022chain}, and excel at open-ended, complex tasks requiring deliberation.

However, the vast majority of decisions in real-world software systems are \emph{not} open-ended reasoning problems. They are structured, bounded, and require fast, well-calibrated answers: \emph{Should this email be flagged as spam?} (binary classification), \emph{Which of these five tools should the agent invoke?} (multi-class routing), \emph{How urgent is this support ticket, on a 1--5 scale?} (ordinal scoring), \emph{Is this generated response safe to display?} (safety gating), or \emph{Which of these ten retrieved documents is most relevant?} (reranking).

For such tasks, autoregressive LLMs are \emph{dramatically over-provisioned}: they generate verbose natural-language rationales when a single probability or class label would suffice; they typically incur 500ms--5s of latency when specialized models can achieve 10--70ms on comparable hardware; they can cost 40--400$\times$ more per decision depending on model size and serving configuration; and their verbalized confidence scores are notoriously poorly calibrated~\citep{geng2024survey,kadavath2022language}. These order-of-magnitude gaps (whose exact magnitude depends on model size, input length, hardware, batch size, and serving infrastructure) have motivated a new class of models that we term \textbf{System~1 decision models}, architectures explicitly designed for fast, structured, calibrated decision-making.

\section{Background}
\label{sec:background}

This section establishes the conceptual and technical foundations for the rest of the guide. We begin with the dual-process theory from cognitive science that motivates the System~1 vs.\ System~2 distinction (Section~\ref{sec:dual_process}), then formalize the calibration problem that is central to reliable decision-making (Section~\ref{sec:calibration_problem}), and finally present our unified taxonomy of System~1 decision model families (Section~\ref{sec:taxonomy}).

\subsection{Dual-Process Theory in Cognitive Science}
\label{sec:dual_process}

The distinction between fast and slow cognition has deep roots in psychology. The idea that human thought involves qualitatively different modes of processing has been proposed in various forms for over a century, but modern dual-process theory crystallized through a series of converging contributions from the 1970s onward.

\paragraph{Origins.} \citet{wason1974dual} proposed one of the first explicit dual-process accounts of reasoning. Working with Wason's selection task -- in which subjects consistently fail to apply logical rules they can articulate when asked -- they hypothesized that two distinct processes govern human inference: a fast, automatic process driven by superficial features of the stimulus (e.g., matching bias), and a slower, deliberate process that attempts logical analysis but is often overridden. Their key observation was that subjects' verbal justifications (the output of the deliberate process) frequently confabulated reasons for choices that had actually been driven by the automatic process -- an early demonstration that introspection does not reliably track the true causes of judgment.

\paragraph{Theoretical development.} Over the next two decades, several researchers independently converged on similar two-system frameworks:

\begin{itemize}
\item \textbf{Evans (1984, 1989)} developed the \emph{heuristic-analytic theory}, arguing that a heuristic process first selects what information to attend to, and an analytic process then reasons over that pre-filtered representation~\citep{evans2008dual}. Crucially, because the heuristic stage is opaque to introspection, people are often unaware that their ``deliberate'' reasoning operates on a biased subset of the available evidence.
\item \textbf{Epstein (1994)} reformulated his \emph{Cognitive-Experiential Self-Theory} (CEST) explicitly as a dual-process theory, distinguishing a rational system (deliberate, analytical, affect-free) from an experiential system (automatic, holistic, emotionally driven)~\citep{epstein1994cest}. CEST emphasized that the experiential system is not merely a source of errors -- it encodes adaptive knowledge learned through emotional experience and operates effectively in domains where analytical reasoning is too slow or data-poor.
\item \textbf{Sloman (1996)} presented ``The Empirical Case for Two Systems of Reasoning,'' distinguishing an \emph{associative system} (operating on similarity and temporal contiguity) from a \emph{rule-based system} (operating on symbolic, language-like representations)~\citep{sloman1996empirical}. His key evidence came from cases where people simultaneously hold two contradictory beliefs -- recognizing that a logical rule gives one answer while their intuition insists on another -- arguing that this conflict is only explicable if the beliefs arise from separate systems.
\item \textbf{Evans and Over (1996)} distinguished \emph{tacit} thought processes (implicit, pragmatic, achieving personal goals effectively) from \emph{explicit} thought processes (deliberate, rule-following, capable of hypothetical thinking)~\citep{evansover1996rationality}. They argued that humans are highly rational in the tacit sense -- we navigate the world effectively -- but only limitedly rational in the explicit, normative sense.
\end{itemize}

\paragraph{The System~1/System~2 synthesis.} \citet{stanovich2000individual} synthesized these converging lines of evidence into a unified framework, introducing the now-standard \emph{System~1} and \emph{System~2} terminology. They proposed that individual differences in reasoning performance arise primarily from variation in System~2 capacity (working memory, cognitive ability) and in the \emph{disposition} to engage System~2 when System~1 produces a default response. This framework was later popularized for a general audience by \citet{kahneman2011thinking} in \emph{Thinking, Fast and Slow}. Key properties of the two systems are summarized in Table~\ref{tab:dual_process}.

\begin{table}[H]
\centering
\caption{Properties of System~1 and System~2 cognition, adapted from \citet{kahneman2011thinking} and \citet{stanovich2000individual}.}
\label{tab:dual_process}
\begin{tabular}{lll}
\toprule
\textbf{Property} & \textbf{System~1} & \textbf{System~2} \\
\midrule
Speed & Fast (milliseconds) & Slow (seconds--minutes) \\
Effort & Effortless, automatic & Effortful, deliberate \\
Awareness & Unconscious, opaque & Conscious, introspectable \\
Parallelism & Highly parallel & Sequential \\
Evolutionary origin & Old, shared with animals & Recent, uniquely human \\
Knowledge & Associative, experiential & Rule-based, symbolic \\
Error profile & Biases, heuristics & Logical but resource-limited \\
Output & Impressions, intuitions & Judgments, plans \\
Examples & Face recognition, reading & Long division, tax planning \\
\bottomrule
\end{tabular}
\end{table}

\paragraph{Critiques and limitations.} Dual-process theory is not without controversy. \citet{osman2004evaluation} argued that the evidence is equally consistent with a single system whose representational properties vary continuously, rather than two discrete systems. \citet{kerenschul2009} challenged the notion that System~1 and System~2 have distinct \emph{clusters} of properties, noting that individual features (speed, consciousness, effort) dissociate experimentally -- a process can be fast yet deliberate, or slow yet automatic. \citet{kruglanski2011dual} proposed a unified model in which both ``intuitive'' and ``deliberative'' judgments are rule-based, differing only in the accessibility and complexity of the rules applied. These critiques do not invalidate the framework's utility as an organizing metaphor, but they caution against reifying System~1 and System~2 as sharply bounded cognitive modules.

\paragraph{Application to AI.} Recent work has increasingly applied this framework to AI systems. \citet{booch2021thinking} argued that achieving human-level AI requires both fast pattern matching (System~1) and deliberate reasoning (System~2). \citet{bengio2019system2} proposed that deep learning primarily implements System~1 processing, and that System~2 capabilities require explicit architectural support for sequential, compositional reasoning. We adopt the System~1/System~2 terminology as a \emph{conceptual analogy} for organizing AI architectures by their computational profile (single-pass vs.\ multi-step, structured output vs.\ free-form generation), not as a claim that neural network architectures implement human dual-process cognition in any mechanistic sense. Indeed, the critiques above reinforce this position: the value of the metaphor lies in its ability to organize a diverse landscape of model families by their computational properties, not in any implied isomorphism with human neurocognition.

\subsection{LLMs as System~2 Machines}

Modern LLMs operate through autoregressive decoding: given an input sequence $x = (x_1, \ldots, x_n)$, the model generates output tokens $y_t$ one at a time, each conditioned on all preceding tokens:
\begin{equation}
    p(y \mid x) = \prod_{t=1}^{T} p(y_t \mid x, y_{<t}).
    \label{eq:autoregressive}
\end{equation}

This sequential process is inherently slow: generating $T$ tokens requires $T$ forward passes through the model. The emergence of chain-of-thought (CoT) reasoning~\citep{wei2022chain}, tree-of-thoughts~\citep{yao2023tree}, and reasoning models like DeepSeek-R1~\citep{deepseek2025r1} and OpenAI o1~\citep{openai2024o1} has pushed LLMs further toward System~2 processing, deliberately trading speed for reasoning quality through extended ``thinking'' token sequences.

\paragraph{Why System~2 is the wrong tool for many decisions.} The autoregressive architecture creates a fundamental mismatch when applied to structured decision tasks. Consider a binary safety gate (``is this response safe?''). The answer is a single bit of information -- yes or no -- plus a calibrated probability. Yet an LLM must generate a \emph{sequence of tokens} to express this answer, navigating a combinatorially vast output space (every possible string) when the decision space contains exactly two elements. This wastes compute on tokens that carry no decision-relevant information and introduces failure modes that single-pass models avoid entirely:

\begin{itemize}
    \item \textbf{Calibration is an afterthought.} LLM next-token probabilities are optimized for language modeling, not for decision calibration. Extracting a calibrated decision confidence requires either parsing verbalized numbers (``I am 85\% confident...''), which are notoriously unreliable~\citep{kadavath2022language}, or marginalizing token-level probabilities over all sequences that express the same decision -- a computation that is expensive and often intractable.
    \item \textbf{Irreducible latency floor.} Even with speculative decoding~\citep{leviathan2023fast} and KV-cache optimizations, autoregressive generation imposes a sequential latency floor proportional to the output length. A System~1 model producing a scalar output in a single forward pass faces no such floor.
    \item \textbf{Prompt fragility.} LLM decisions are sensitive to prompt phrasing, option ordering, and few-shot example selection~\citep{lu2022fantastically}. Minor reformulations of the same question can flip the output, a brittleness that encoder-based models trained on structured decision objectives largely avoid.
\end{itemize}

\noindent These are not engineering inconveniences that better prompting or cheaper inference will resolve -- they are \emph{architectural} consequences of using a sequential text generator for tasks that require a single, calibrated judgment. This structural mismatch motivates the models covered in the rest of this guide.

The recent survey by \citet{zhang2025system1to2} provides a comprehensive treatment of the System~1$\to$System~2 progression in reasoning LLMs. Complementarily, our guide focuses on the \emph{other direction}: models optimized for System~1 decision-making, where speed, cost, and calibration are paramount.

\subsection{The Calibration Problem}
\label{sec:calibration_problem}

A model is \emph{well-calibrated} if its predicted probabilities match empirical frequencies: when it assigns probability $p$ to an event, that event should occur with frequency $p$ in practice~\citep{guo2017calibration,niculescu2005predicting}. Formally, for a model $f$ and true labels $y$:
\begin{equation}
    \mathbb{P}(y = 1 \mid f(x) = p) = p, \quad \forall p \in [0, 1].
    \label{eq:calibration_def}
\end{equation}

In plain terms: gather every instance where the model said ``70\% chance this is positive.'' If the model is calibrated, roughly 70\% of those instances should actually be positive -- not 50\%, not 90\%, but 70\%. The left-hand side of Equation~\ref{eq:calibration_def} measures this empirical frequency (``among all inputs where the model output $p$, how often was $y$ actually 1?''), and calibration demands that this frequency equal the model's stated confidence $p$, for \emph{every} confidence level simultaneously. A model that says ``90\% confident'' and is right only 60\% of the time is \emph{overconfident}; one that says ``60\% confident'' but is right 90\% of the time is \emph{underconfident}. Both are miscalibrated, and both lead to poor downstream decisions -- the first by acting when it should hesitate, the second by hesitating when it should act.

To illustrate how this relates to System~2 models: suppose you ask an LLM ``Is this email spam?'' 100 times (with nonzero temperature), and it answers ``yes'' on 70 of those runs. In a frequentist sense, you could say the model's implicit probability is 0.7 -- and if the email is indeed spam, that estimate is reasonably calibrated. But extracting this probability required \emph{100 forward passes} through a billion-parameter model. A System~1 decision model returns $p = 0.7$ in a \emph{single} forward pass, because it is architecturally designed to output a calibrated probability rather than a yes/no token that must be aggregated over many stochastic samples. This is the core efficiency argument for System~1 models: they make the implicit explicit, collapsing what would be a Monte Carlo estimation problem into a single deterministic computation.

\paragraph{An efficiency bound.} We can make this argument precise. Suppose a System~2 model has an implicit (unknown) probability $p^*$ of answering ``yes'' to a given input, and we wish to estimate $p^*$ by repeated sampling. Each retry is an independent Bernoulli trial with success probability $p^*$, so the sample proportion $\hat{p} = k/N$ after $N$ retries has standard error $\text{SE} = \sqrt{p^*(1-p^*)/N}$. To ensure the estimate is within $\pm\varepsilon$ of the true probability with $\geq 95\%$ confidence, the normal approximation requires:
\begin{equation}
    N \;\geq\; \frac{z_{0.975}^2 \;\cdot\; p^*(1-p^*)}{{\varepsilon}^2}.
    \label{eq:retry_bound}
\end{equation}
Since $p^*$ is unknown and $p^*(1-p^*) \leq 1/4$ (maximized at $p^* = 0.5$), the worst-case (most conservative) requirement is:
\begin{equation}
    N \;\geq\; \frac{1.96^2}{4\varepsilon^2} \;=\; \frac{0.9604}{\varepsilon^2}.
    \label{eq:retry_bound_worstcase}
\end{equation} A System~1 model produces its estimate in a single forward pass ($N = 1$). Denoting the per-pass cost (in FLOPs, latency, or dollars) of the System~2 model as $C_2$ and that of the System~1 model as $C_1$, the cost ratio for achieving precision $\varepsilon$ is:
\begin{equation}
    \text{Cost ratio} \;=\; \frac{N \cdot C_2}{C_1} \;\geq\; \frac{0.96}{\varepsilon^2} \cdot \frac{C_2}{C_1}.
    \label{eq:cost_ratio}
\end{equation}
Table~\ref{tab:retry_cost} illustrates concrete numbers. Even at coarse precision ($\varepsilon = 0.10$), the System~2 model needs ${\sim}100$ retries; at fine precision ($\varepsilon = 0.01$), it needs ${\sim}10{,}000$. Multiplied by the per-call cost ratio $C_2/C_1$, which is typically $10$--$100\times$ for System~2 vs.\ System~1 models on comparable hardware, the total efficiency gap ranges from \textbf{three to six orders of magnitude}. This is not an argument that System~2 models are \emph{never} the right choice -- for tasks requiring genuine reasoning, they are indispensable. It is an argument that using them to \emph{estimate a probability} is architecturally wasteful when models exist that output calibrated probabilities natively.

\begin{table}[H]
\centering
\caption{Minimum System~2 retries $N$ for probability estimation at precision $\pm\varepsilon$ (95\% confidence, worst case $p^* = 0.5$), and total cost ratio assuming $C_2/C_1 = 50$ (a conservative mid-range for LLM vs.\ encoder-based System~1 models).}
\label{tab:retry_cost}
\begin{tabular}{cccc}
\toprule
\textbf{Precision $\varepsilon$} & \textbf{Min retries $N$} & \textbf{Cost ratio ($C_2/C_1 = 50$)} & \textbf{Practical meaning} \\
\midrule
$\pm 0.10$ & 97 & $4{,}802\times$ & Coarse: ``roughly 70\%'' \\
$\pm 0.05$ & 385 & $19{,}208\times$ & Moderate: ``between 65--75\%'' \\
$\pm 0.01$ & 9{,}604 & $480{,}200\times$ & Fine: ``70.0\% $\pm$ 1\%'' \\
\bottomrule
\end{tabular}
\end{table}

Calibration is critical for decision-making: a routing agent that believes ``Tool~A is appropriate with 85\% confidence'' must be correct 85\% of the time for downstream systems to make sound expected-value calculations. This subsection introduces the key theoretical concepts needed to reason about calibration throughout the guide; Section~\ref{sec:calibration} provides the full technical treatment, including detailed metrics, training methods, and empirical results.

\paragraph{Proper scoring rules.} A \emph{scoring rule} $S(p, y)$ assigns a numerical score to a predicted probability $p$ given the observed outcome $y \in \{0,1\}$. A scoring rule is \emph{proper} if the expected score is minimized (or maximized, depending on orientation) when the forecaster reports the true probability; it is \emph{strictly proper} if that minimum is unique~\citep{gneiting2007strictly}. Two strictly proper scoring rules appear throughout this guide:
\begin{itemize}
    \item \textbf{Brier score:} $\text{BS}(p, y) = (p - y)^2$, the squared distance between the prediction and the outcome.
    \item \textbf{Log loss:} $\text{LL}(p, y) = -[y \log p + (1 - y) \log(1 - p)]$, the negative log-likelihood of the observed outcome under the predicted distribution.
\end{itemize}

\noindent \textbf{A worked example.} Suppose the true probability that an email is spam is $p^* = 0.7$. A model can report any value $\hat{p} \in [0,1]$. Since the outcome $y$ is random (spam with probability $p^*$, not spam with probability $1 - p^*$), we evaluate each scoring rule \emph{in expectation} over $y$:
\begin{align}
    \mathbb{E}[\text{BS}(\hat{p}, y)] &= p^* \cdot (\hat{p} - 1)^2 + (1 - p^*) \cdot (\hat{p} - 0)^2, \label{eq:expected_brier} \\
    \mathbb{E}[\text{LL}(\hat{p}, y)] &= p^* \cdot [-\ln \hat{p}] \;+\; (1 - p^*) \cdot [-\ln(1 - \hat{p})]. \label{eq:expected_logloss}
\end{align}
In words: the expected loss is a weighted average of two scenarios -- the outcome is positive (weight $p^*$) and the outcome is negative (weight $1 - p^*$) -- with the per-scenario loss determined by how far $\hat{p}$ is from the realized outcome.

\noindent \textbf{Example: $\hat{p} = 0.9$ (overconfident).} With $p^* = 0.7$:
\begin{itemize}[nosep]
    \item Brier: $0.7 \cdot \underbrace{(0.9 - 1)^2}_{0.01} + 0.3 \cdot \underbrace{(0.9 - 0)^2}_{0.81} = 0.007 + 0.243 = 0.250$.
    \item Log loss: $0.7 \cdot \underbrace{[-\ln 0.9]}_{0.105} + 0.3 \cdot \underbrace{[-\ln 0.1]}_{2.303} = 0.074 + 0.691 = 0.764$.
\end{itemize}
The overconfident prediction pays a small penalty when correct ($0.9$ is close to $1$), but a large penalty when wrong ($0.9$ is far from $0$) -- and the ``when wrong'' case happens 30\% of the time. Applying the same formulas to three predictions:

\begin{center}
\small
\begin{tabular}{lccc}
\toprule
\textbf{Prediction $\hat{p}$} & \textbf{Expected Brier score} & \textbf{Expected log loss} & \textbf{Honest?} \\
\midrule
$0.7$ (truthful) & $0.7 \cdot 0.09 + 0.3 \cdot 0.49 = \mathbf{0.210}$ & $0.7 \cdot 0.357 + 0.3 \cdot 1.204 = \mathbf{0.611}$ & Yes \\
$0.9$ (overconfident) & $0.7 \cdot 0.01 + 0.3 \cdot 0.81 = 0.250$ & $0.7 \cdot 0.105 + 0.3 \cdot 2.303 = 0.764$ & No \\
$0.5$ (hedging) & $0.7 \cdot 0.25 + 0.3 \cdot 0.25 = 0.250$ & $0.7 \cdot 0.693 + 0.3 \cdot 0.693 = 0.693$ & No \\
\bottomrule
\end{tabular}
\end{center}

Under both scoring rules, the truthful report $\hat{p} = 0.7$ achieves the \emph{lowest} expected loss. Overstating confidence ($\hat{p} = 0.9$) incurs a large penalty when the event does not occur, and hedging ($\hat{p} = 0.5$) wastes the model's knowledge. This is what ``proper'' means in practice: the only way to minimize expected loss is to report what you actually believe. Note that log loss penalizes confident errors more harshly than the Brier score -- predicting $\hat{p} = 0.99$ when $y = 0$ incurs a Brier penalty of 0.98 but a log loss of 4.6 -- which makes it the preferred objective when avoiding catastrophic false confidence matters most.

\begin{figure}[H]
\centering
\begin{tikzpicture}
\begin{axis}[
    width=0.82\textwidth,
    height=6.5cm,
    xlabel={Predicted probability $\hat{p}$},
    ylabel={Loss},
    xmin=0, xmax=1,
    ymin=0, ymax=2.5,
    legend style={font=\small, at={(0.5,0.97)}, anchor=north, fill=white, fill opacity=0.95, draw=gray!40, legend columns=2},
    grid=major,
    grid style={gray!15},
    tick label style={font=\small},
    label style={font=\small},
    clip=true,
]
\addplot[blue!70!black, very thick, domain=0.01:0.99, samples=80] {(x-1)^2};
\addlegendentry{Brier ($y{=}1$)}

\addplot[blue!70!black, very thick, dashed, domain=0.01:0.99, samples=80] {x^2};
\addlegendentry{Brier ($y{=}0$)}

\addplot[red!75!black, very thick, domain=0.05:0.99, samples=80] {-ln(x)};
\addlegendentry{Log loss ($y{=}1$)}

\addplot[red!75!black, very thick, dashed, domain=0.01:0.95, samples=80] {-ln(1-x)};
\addlegendentry{Log loss ($y{=}0$)}

\end{axis}
\end{tikzpicture}
\caption{Penalty profiles of the Brier score and log loss. Solid lines: loss when $y=1$; dashed lines: loss when $y=0$. Both are minimized at the correct prediction ($\hat{p} = y$), but they differ sharply at the extremes: the Brier score is bounded by~1, while the log loss diverges to $+\infty$ as the model becomes confidently wrong (visible as the steep rise at the edges). This makes log loss the more punitive objective for avoiding catastrophic false confidence.}
\label{fig:scoring_rules}
\end{figure}
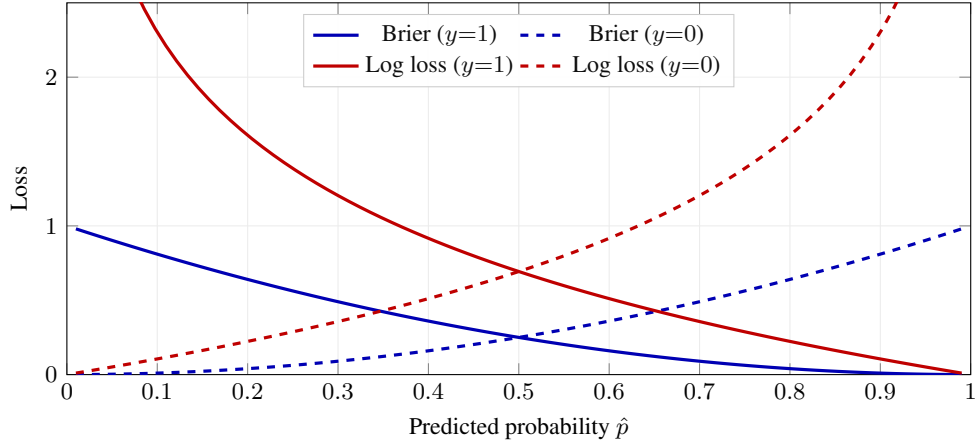

\paragraph{Why accuracy is not a proper scoring rule.} Classification accuracy -- the fraction of correct hard predictions -- is the most widely used metric but is \emph{not} a proper scoring rule. To see why, consider a dataset where 80\% of emails are spam. A model that predicts $\hat{p} = 0.8$ for every input (the base rate) and thresholds at 0.5 achieves 80\% accuracy by always predicting ``spam.'' A more sophisticated model that produces varying probabilities ($\hat{p} = 0.95$ for obvious spam, $\hat{p} = 0.3$ for likely ham) might also achieve 80\% accuracy but carries far more information. Accuracy cannot distinguish between them. Worse, accuracy depends on the \emph{threshold} used to convert probabilities to hard predictions, a free parameter that obscures the underlying calibration quality. Proper scoring rules evaluate the \emph{probability itself}, not the hard decision derived from it, which is why they are the right objective for training System~1 models.

Strict propriety is the theoretical guarantee that \emph{the only way to minimize the loss is to report the truth}. This property makes proper scoring rules the natural training objective for System~1 decision models: a model trained to minimize Brier score or log loss is directly incentivized to produce calibrated probabilities, with no auxiliary calibration step required.

\paragraph{Calibration vs.\ sharpness.} Calibration alone is insufficient: a model that always predicts the base rate $\bar{y} = \mathbb{E}[Y]$ is perfectly calibrated but completely uninformative -- it assigns the same probability to every input. The missing ingredient is \emph{sharpness} (also called \emph{refinement} or \emph{discrimination}): the model's ability to separate high-probability instances from low-probability ones by producing predictions that deviate meaningfully from the base rate.

\noindent \textbf{Example: three spam classifiers.} Suppose 60\% of emails are spam ($\bar{y} = 0.6$). Consider three models evaluated on 1{,}000 emails:

\begin{itemize}
    \item \textbf{Model A (calibrated, not sharp):} predicts $\hat{p} = 0.6$ for every email. Among any group of its predictions, 60\% are indeed spam. Perfectly calibrated, but useless for ranking -- it cannot distinguish obvious spam from legitimate emails.
    \item \textbf{Model B (sharp, not calibrated):} predicts $\hat{p} = 0.95$ for 500 emails and $\hat{p} = 0.05$ for 500. Among the 500 flagged at 0.95, only 400 are actually spam (80\%, not 95\% -- overconfident). Among the 500 flagged at 0.05, only 200 are spam (40\%, not 5\% -- underconfident). Model~B is highly informative -- its high-confidence group is much more likely to be spam than its low-confidence group -- but its probabilities cannot be trusted at face value.
    \item \textbf{Model C (calibrated and sharp):} predicts $\hat{p} = 0.95$ for 400 emails (of which ${\sim}380$ are spam), $\hat{p} = 0.20$ for 400 (of which ${\sim}80$ are spam), and intermediate values for the rest, with empirical frequencies matching throughout. Model~C is both honest \emph{and} informative -- the gold standard.
\end{itemize}

Every strictly proper scoring rule admits a decomposition into these two orthogonal components~\citep{murphy1973new,degroot1983comparison}:
\begin{equation}
    \text{Expected Loss} = \underbrace{\text{Calibration loss}}_{\text{do probabilities match reality?}} + \text{Uncertainty} - \underbrace{\text{Refinement}}_{\text{are predictions informative?}},
    \label{eq:cal_sharp_intro}
\end{equation}
where the uncertainty term depends only on the data and is irreducible. In the example above, Model~A has zero calibration loss but zero refinement (all predictions equal the base rate). Model~B has high refinement but nonzero calibration loss (predictions deviate from observed frequencies). Model~C minimizes both calibration loss \emph{and} maximizes refinement. This decomposition explains why raw accuracy is a poor proxy for decision quality: a model can be highly accurate yet poorly calibrated (Model~B after thresholding), or well-calibrated yet uninformative (Model~A).

\paragraph{Reliability diagrams.} The primary visual tool for assessing calibration is the \emph{reliability diagram} (also called a \emph{calibration plot}): a plot of observed accuracy against predicted confidence across bins~\citep{degroot1983comparison}.

\noindent \textbf{How to read one.} The construction is straightforward: (1)~sort all $N$ predictions by predicted confidence $\hat{p}$; (2)~divide them into $B$ bins (e.g., $[0, 0.1), [0.1, 0.2), \ldots, [0.9, 1.0]$); (3)~for each bin, compute the average predicted confidence (x-axis) and the fraction of positive outcomes (y-axis). A perfectly calibrated model traces the diagonal: every bin's observed frequency matches its average prediction. Deviations reveal systematic errors:

\begin{itemize}
    \item \textbf{Curve below the diagonal} $\to$ \emph{overconfidence}: the model claims higher confidence than warranted. For instance, among predictions in the 0.8--0.9 bin, only 65\% are actually positive. This is the most common failure mode in deep neural networks~\citep{guo2017calibration} and is particularly prevalent in LLMs after RLHF training.
    \item \textbf{Curve above the diagonal} $\to$ \emph{underconfidence}: the model is more accurate than it claims. Among predictions in the 0.5--0.6 bin, 75\% are positive. Underconfidence is less common but can occur after aggressive temperature scaling or in distilled models.
    \item \textbf{S-shaped curve}: overconfident at high $\hat{p}$ and underconfident at low $\hat{p}$ (or vice versa), indicating that the model's probability scale is compressed or stretched relative to reality.
\end{itemize}

The \emph{Expected Calibration Error} (ECE) summarizes the diagram as a single number: the weighted average of per-bin absolute deviations from the diagonal, $\text{ECE} = \sum_{b=1}^{B} \frac{n_b}{N} \left| \text{acc}(b) - \text{conf}(b) \right|$, where $n_b$ is the number of samples in bin $b$, $\text{acc}(b)$ is the observed accuracy, and $\text{conf}(b)$ is the mean predicted confidence. Section~\ref{sec:calibration} provides a detailed reliability diagram (Figure~\ref{fig:reliability}) comparing overconfident LLMs with well-calibrated System~1 models, along with a full code listing for computing ECE (Listing~\ref{lst:ece}).

\begin{figure}[H]
\centering
\begin{tikzpicture}
\begin{axis}[
    width=0.78\textwidth,
    height=7cm,
    xlabel={Mean predicted confidence},
    ylabel={Observed accuracy (fraction of positives)},
    xmin=0, xmax=1,
    ymin=0, ymax=1,
    xtick={0,0.2,0.4,0.6,0.8,1.0},
    ytick={0,0.2,0.4,0.6,0.8,1.0},
    grid=major,
    grid style={gray!15},
    tick label style={font=\small},
    label style={font=\small},
    legend style={font=\small, at={(0.5,-0.18)}, anchor=north, draw=gray!40, legend columns=2},
]

\addplot[black, thin, dashed, domain=0:1] {x};
\addlegendentry{Perfect calibration}

\addplot[green!60!black, very thick, mark=square*, mark size=2.5pt]
    coordinates {(0.05,0.04) (0.15,0.16) (0.25,0.23) (0.35,0.36) (0.45,0.44)
                 (0.55,0.57) (0.65,0.63) (0.75,0.76) (0.85,0.84) (0.95,0.96)};
\addlegendentry{System~1 (ECE\,$=$\,0.015)}

\addplot[red!80!black, very thick, mark=triangle*, mark size=2.5pt]
    coordinates {(0.05,0.08) (0.15,0.18) (0.25,0.25) (0.35,0.32) (0.45,0.38)
                 (0.55,0.44) (0.65,0.52) (0.75,0.58) (0.85,0.66) (0.95,0.74)};
\addlegendentry{Overconfident LLM (ECE\,$=$\,0.12)}

\addplot[blue!70!black, very thick, mark=o, mark size=2.5pt]
    coordinates {(0.05,0.12) (0.15,0.28) (0.25,0.38) (0.35,0.48) (0.45,0.56)
                 (0.55,0.64) (0.65,0.73) (0.75,0.82) (0.85,0.90) (0.95,0.97)};
\addlegendentry{Underconfident (ECE\,$=$\,0.09)}

\draw[thin, red!50!black] (axis cs:0.85,0.85) -- (axis cs:0.85,0.66);
\draw[thin, red!50!black] (axis cs:0.83,0.85) -- (axis cs:0.87,0.85);
\draw[thin, red!50!black] (axis cs:0.83,0.66) -- (axis cs:0.87,0.66);
\node[font=\scriptsize, text=red!60!black, anchor=west] at (axis cs:0.88,0.755) {gap};

\end{axis}
\end{tikzpicture}
\caption{Reliability diagram (illustrative). A perfectly calibrated model follows the dashed diagonal. The System~1 model (green) hugs the diagonal (ECE\,$=$\,0.015). The overconfident LLM (red) curves below -- at the 0.85 bin it claims 85\% confidence but is correct only 66\% of the time (bracket). The underconfident model (blue) curves above -- it is more accurate than it claims.}
\label{fig:reliability_intro}
\end{figure}
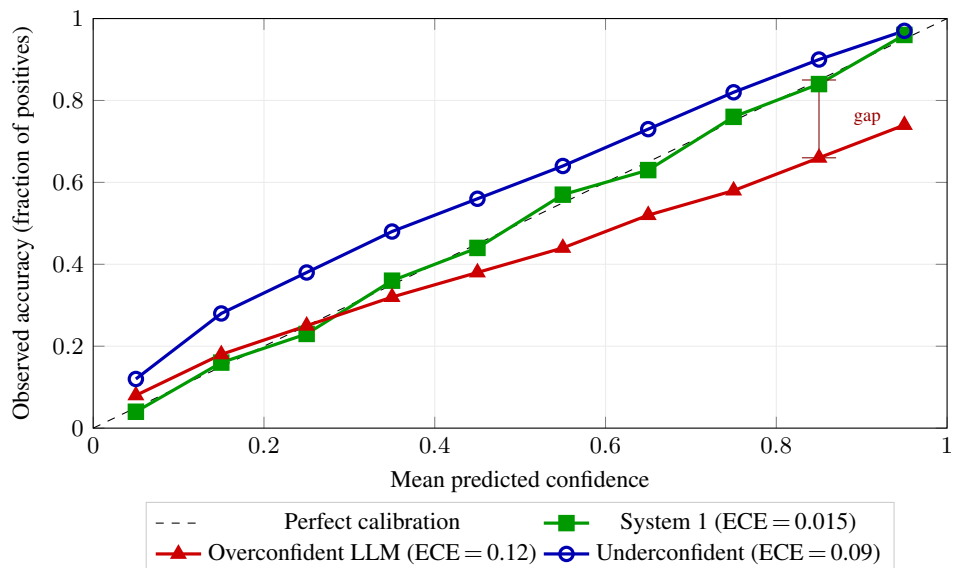

\paragraph{Post-hoc calibration.} When a model's raw probabilities are miscalibrated, post-hoc methods can adjust them using a held-out calibration set. The core idea is simple: learn a mapping from the model's raw scores to calibrated probabilities, using a separate validation set where you know the ground truth. Three techniques recur throughout this guide, ordered from simplest to most flexible:

\begin{itemize}
    \item \textbf{Temperature scaling}~\citep{guo2017calibration}: divides logits by a single learned temperature $T$ before softmax: $\hat{p}_{\text{cal}} = \operatorname{softmax}(z / T)$. When $T > 1$, the distribution is softened (reducing overconfidence); when $T < 1$, it is sharpened. Temperature scaling has one free parameter, is fast to fit (a few seconds on any dataset), and preserves the model's ranking -- if the model ranked input $A$ above input $B$ before calibration, it still does after. This makes it the default first choice for neural classifiers that are well-ranked but poorly scaled. Its limitation: it applies the same correction globally, so it cannot fix calibration that varies across confidence levels (e.g., overconfident at high $\hat{p}$ but underconfident at low $\hat{p}$).

    \item \textbf{Platt scaling}~\citep{platt1999probabilistic}: fits a logistic regression $\hat{p}_{\text{cal}} = \sigma(a \cdot z + b)$ on the model's logit $z$, learning both a scale $a$ and a bias $b$. With two parameters, Platt scaling can correct both the slope and the intercept of the calibration curve, handling cases where the model is systematically biased (e.g., predictions are centered around 0.6 instead of 0.5 on a balanced dataset). Originally developed for SVMs, it remains widely used for any model that produces a single scalar score.

    \item \textbf{Isotonic regression}~\citep{zadrozny2002transforming}: fits a non-parametric, monotone step function from predicted scores to calibrated probabilities. It makes no distributional assumptions and can correct arbitrarily complex calibration curves -- including the S-shaped distortions that temperature and Platt scaling cannot fix. The trade-off is data efficiency: isotonic regression requires more calibration samples (typically $>$1{,}000) to avoid overfitting, and its step-function output can be non-smooth.
\end{itemize}

\noindent \textbf{When to use which.} As a rule of thumb: start with temperature scaling (one parameter, hard to overfit). If the reliability diagram shows a shifted or tilted curve, try Platt scaling. If it shows a complex, non-monotone pattern, use isotonic regression. In all cases, evaluate on a held-out test set that was not used for fitting the calibration map -- calibrating and evaluating on the same data leads to optimistic ECE estimates.

These methods are powerful but have a fundamental limitation: they correct calibration \emph{after} training, which means the model's internal representations are not optimized for calibration. System~1 decision models take a different approach -- they incorporate proper scoring rules into the training objective itself (Section~\ref{sec:calibration}), producing probabilities that are calibrated \emph{by construction} rather than by post-hoc correction. Post-hoc methods remain valuable as a safety net (even a well-trained model can drift out of calibration under distribution shift), but the goal is to minimize the correction needed.

\paragraph{Decision-theoretic motivation.} The practical importance of calibration becomes clear through the lens of decision theory. Consider a binary decision with threshold $\theta$: take action $A$ if $p \geq \theta$, otherwise take action $B$. If action $A$ incurs cost $c_A$ when wrong and benefit $b_A$ when right, the optimal threshold is $\theta^* = c_A / (c_A + b_A)$. This calculation is valid \emph{only if $p$ is calibrated} -- if the model is overconfident, the effective threshold shifts and the decision-maker takes action $A$ too often, incurring unnecessary costs. In multi-action settings (routing among $K$ tools, triage across $N$ priority levels), miscalibration compounds: each decision node introduces error, and in an agent trajectory of $L$ steps the cumulative effect can make the overall policy substantially suboptimal. Calibrated probabilities are therefore not a theoretical luxury but a prerequisite for sound automated decision-making at scale.

\paragraph{LLMs are poorly calibrated.} Multiple studies have demonstrated that LLMs exhibit systematic calibration failures~\citep{geng2024survey,kadavath2022language,xiong2024llms}. Verbalized confidence -- asking the LLM to state its confidence -- shows near-chance correlation with actual accuracy~\citep{xiong2024llms}. Token-level probabilities from the softmax layer are better but still miscalibrated, especially after RLHF training, which distorts the probability landscape~\citep{openai2023gpt4}. Overconfidence is pervasive: LLMs assign high probability to incorrect answers, particularly on out-of-distribution inputs~\citep{kadavath2022language}.

This calibration gap is a primary motivation for System~1 decision models, which use training objectives designed to incentivize calibrated probability estimation.

\paragraph{Further reading.} The theoretical foundations of calibration and scoring rules are deep and well-studied. \citet{gneiting2007strictly} provide the definitive treatment of strictly proper scoring rules, their characterizations, and their role in probabilistic forecasting. \citet{degroot1983comparison} introduced reliability diagrams and the formal framework for comparing calibration across forecasters. \citet{murphy1973new} established the calibration--refinement decomposition that underpins modern calibration analysis. For post-hoc calibration methods, \citet{platt1999probabilistic} introduced Platt scaling, \citet{zadrozny2002transforming} developed isotonic regression for calibration, and \citet{guo2017calibration} demonstrated that modern deep networks are systematically miscalibrated and that temperature scaling is a surprisingly effective remedy. \citet{niculescu2005predicting} provide a comprehensive empirical comparison of calibration methods across classifiers. For the decision-theoretic perspective, \citet{berger1985statistical} remains the standard reference on statistical decision theory, including the role of calibrated probabilities in Bayes-optimal decision-making. Within this guide, Section~\ref{sec:calibration} provides the full technical treatment of calibration metrics, training methodologies, and empirical results specific to System~1 decision models.

\subsection{Defining System~1 Decision Models}

We adopt the following operational definition. A \textbf{System~1 decision model} is a model whose primary inference interface maps an input state to a bounded, structured decision (a class label, a bounded score, or a calibrated probability) without requiring free-form autoregressive generation of that decision. This \emph{defining characteristic} concerns the output interface and computation pattern, not the model's performance on any particular metric.

We distinguish the defining characteristic from \emph{evaluated properties} that vary across models and deployments:

\begin{table}[H]
\centering
\small
\begin{tabular}{lp{60mm}p{40mm}}
\toprule
\textbf{Dimension} & \textbf{Description} & \textbf{Status} \\
\midrule
Decision interface & Maps input to a bounded, structured output & Defining characteristic \\
Inference architecture & Non-autoregressive or few-pass computation & Defining characteristic \\
\midrule
Latency & Sub-100ms inference & Evaluated property \\
Calibration & Predicted probabilities match empirical frequencies & Evaluated property \\
Cost efficiency & Lower per-decision cost than LLM alternatives & Evaluated property \\
Determinism & Same input yields same output & Evaluated property \\
\bottomrule
\end{tabular}
\end{table}

\noindent A model that produces a structured decision via a single forward pass but is poorly calibrated still qualifies as a System~1 decision model under this definition; calibration is a quality dimension, not a membership criterion. Conversely, a fast autoregressive model that happens to generate short JSON responses does not qualify, because its output is generated token-by-token rather than produced directly by a decision head. This separation prevents ``System~1'' from becoming an umbrella label for any model used for fast inference.

\newpage
\section{Taxonomy of System~1 Decision Models}
\label{sec:taxonomy}

\begin{figure}[H]
\centering
\resizebox{\textwidth}{!}{%
\begin{tikzpicture}[
  root/.style={rectangle, rounded corners=4pt, draw=blue!60, fill=blue!12,
    font=\normalsize\bfseries, minimum height=8mm, inner sep=5pt},
  fam/.style 2 args={rectangle, rounded corners=3pt, draw=#1, fill=#2,
    font=\footnotesize\bfseries, minimum height=7mm, inner sep=3pt, text width=28mm, align=center},
  mem/.style={font=\tiny, text width=30mm, align=left, inner sep=1pt},
]
\node[root] (R) {System~1 Decision Models};

\node[fam={green!60!black}{green!8}, below=16mm of R, xshift=-88mm] (F1) {I.\ Encoder-Based\\[-1pt]Classifiers};
\node[fam={orange!80!black}{orange!8}, below=16mm of R, xshift=-44mm] (F2) {II.\ Reward\\[-1pt]Models};
\node[fam={red!70!black}{red!8}, below=16mm of R, xshift=0mm]  (F3) {III.\ Non-Autoregressive\\[-1pt]Decision Engines};
\node[fam={violet!90}{violet!8}, below=16mm of R, xshift=44mm]     (F4) {IV.\ Contrastive\\[-1pt]Decision Models};
\node[fam={cyan!70!black}{cyan!8}, below=16mm of R, xshift=88mm]     (F5) {V.\ Extraction\\[-1pt]Models$^*$};

\foreach \f in {F1,F2,F3,F4,F5}
  \draw[thick, -Stealth, gray!60] (R.south) -- (\f.north);

\node[mem, below=4mm of F1] (L1) {%
  BERT\\
  RoBERTa\\
  DeBERTa\\
  Cross-Encoders\\
  SPLADE\\
  ColBERT};

\node[mem, below=4mm of F2] (L2) {%
  Bradley-Terry\\
  Plackett-Luce\\
  Process RM\\
  Outcome RM\\
  Generative RM\\
  Rule-Based};

\node[mem, below=4mm of F3] (L3) {%
  Jev 1.14\\
  Laya / Lia\\
  OpenJev-RLCD\\
  Tachyone\\
  THX-01\\
  DeepOpen};

\node[mem, below=4mm of F4] (L4) {%
  CLM-8B\\
  Sentence-BERT\\
  DPR\\
  CLIP\\
  E5 / GTE\\
  GLiNER2.5};

\node[mem, below=4mm of F5] (L5) {%
  Clairic\\
  Jev Box primitive\\
  VX-One\\[1.5mm]
  {\tiny\itshape $^*$Emerging; shares}\\
  {\tiny\itshape arch.\ DNA w/ Fam.~III}};

\foreach \f/\l in {F1/L1, F2/L2, F3/L3, F4/L4, F5/L5}
  \draw[gray!40] (\f.south) -- (\l.north);

\end{tikzpicture}%
}
\caption{Taxonomy of System~1 Decision Models. The field is organized into five families based on historical lineage and typical deployment role. Family~V (Extraction Models) is emerging and shares architectural DNA with Family~III. Boundaries between families are porous (see Section~\ref{sec:taxonomy} for discussion).}
\label{fig:taxonomy_tree}
\end{figure}
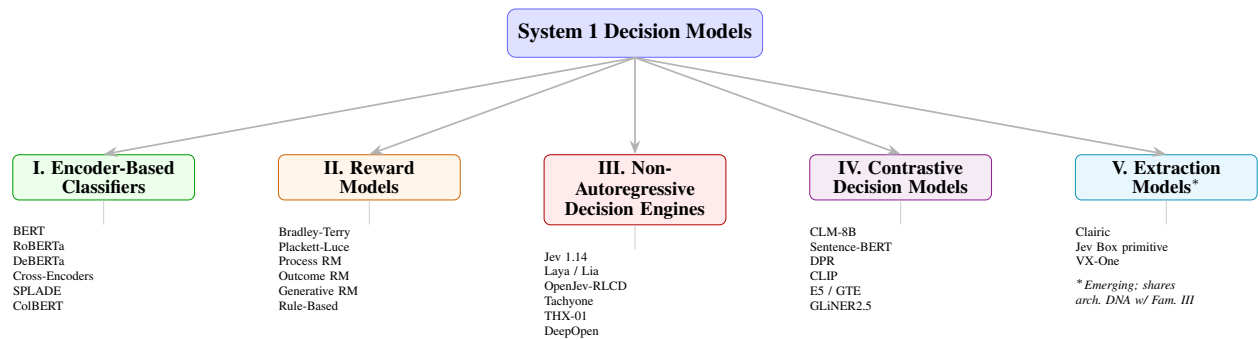

We organize the field into five major families, visualized in Figure~\ref{fig:taxonomy_tree}. The first four families represent distinct architectural approaches to fast, structured decision-making; the fifth (extraction models) is an emerging category that shares architectural DNA with decision models but targets typed value extraction rather than decisions. Below we briefly characterize each family; Chapter~II provides the full technical treatment.

\begin{itemize}
    \item \textbf{Family~I: Encoder-Based Classifiers} -- bidirectional transformer encoders (BERT, RoBERTa, DeBERTa) with task-specific heads. The oldest and most mature family; fast, well-tooled, post-hoc calibrated.
    \item \textbf{Family~II: Reward Models} -- scalar scoring heads trained on preference data to guide RL. Span the widest parameter range; produce ordinal scores rather than calibrated probabilities.
    \item \textbf{Family~III: Non-Autoregressive Decision Engines} -- typed decision primitives (Choice, Score, Noul) trained with RLCD. The family with the strongest emphasis on end-to-end calibration through proper scoring rule training on structured decision tasks.
    \item \textbf{Family~IV: Contrastive Decision Models} -- dual encoders producing aligned embeddings; decisions reduce to similarity computations. Fast via caching; require a calibration layer for probabilities.
    \item \textbf{Family~V: Extraction Models} (emerging) -- typed value extraction from text. Smallest and fastest; boundaries with Family~III remain fluid.
\end{itemize}

\paragraph{Taxonomy scope and overlaps.} We emphasize that these families are organized primarily by \emph{historical lineage and typical deployment role}, not by mutually exclusive architectural properties. In practice, the boundaries are porous: a reward model may use an encoder backbone (overlapping Family~I), a contrastive model may serve as a classifier (overlapping Family~I), a non-autoregressive decision engine may incorporate contrastive pretraining (overlapping Family~IV), and Jev~1.14's Box primitive blurs the line between Family~III and Family~V. We retain this organization because it reflects how the community has developed and discussed these models, but readers should understand the categories as partially overlapping clusters rather than strict partitions. An alternative organization along orthogonal axes (model architecture, output type, training objective, deployment role) would be equally valid; we discuss the relationship between these axes throughout the guide.

Table~\ref{tab:family_comparison} provides a high-level comparison of the five families across key dimensions.

\begin{table}[H]
\centering
\caption{Comparison of the five System~1 decision model families. Latency and calibration values are indicative ranges from representative systems; actual performance depends on model size, hardware, input length, and evaluation protocol.}
\label{tab:family_comparison}
\scriptsize
\begin{tabular}{@{}lcccccc@{}}
\toprule
\textbf{Family} & \textbf{Architecture} & \textbf{Latency} & \textbf{Output} & \textbf{Calibration} & \textbf{Training} & \textbf{Params} \\
\midrule
I. Encoder-Based & Bidir.\ encoder + head & 5--50ms & Logits/probs & Post-hoc & CE / NLI & 100M--600M \\
II. Reward Models & Enc./dec.\ + scalar head & 10--100ms & Scalar score & Task-dependent & BT / PL loss & 300M--70B \\
III. Non-Autoregressive & Encoder + typed heads & 10--70ms & Choice/Score/Noul & Incentivized (RLCD) & RLCD / RLCR & 150M--27B \\
IV. Contrastive & Dual encoder + projection & 5--30ms & Similarity score & Requires calib.\ layer & InfoNCE & 300M--8B \\
V. Extraction$^*$ & Encoder + typed extraction & 1--10ms & Typed values & Edit-distance based & Supervised / RLCD & 150M--1B \\
\bottomrule
\multicolumn{7}{l}{\scriptsize $^*$Emerging family; boundaries with Family~III are fluid.}
\end{tabular}
\end{table}

\paragraph{Key patterns.} Several patterns emerge from the comparison in Table~\ref{tab:family_comparison}. First, \emph{latency and parameter count are not strongly correlated}: Family~IV (contrastive) models achieve the lowest latency (5--30ms) despite having up to 8B parameters, because the dual-encoder architecture enables precomputation and caching of candidate embeddings. Family~V (extraction) models are the fastest (1--10ms) and smallest (150M--1B parameters), reflecting their narrow task scope. Family~II (reward models) spans the widest parameter range (300M--70B) because they inherit the backbone of whatever LLM they were designed to score.

Second, \emph{calibration strategy is the sharpest differentiator} between families. Families~I, II, and IV all require some form of post-hoc calibration -- their training objectives (cross-entropy, Bradley-Terry loss, InfoNCE) do not directly optimize for calibrated probabilities. Family~III is unique in achieving calibration \emph{by construction} through RLCD training with proper scoring rule rewards, eliminating the need for a separate calibration step. This architectural difference has profound practical implications: post-hoc calibration requires a held-out calibration set that matches the deployment distribution, and recalibration is needed whenever the distribution shifts. RLCD-trained models, by contrast, maintain calibration across distribution shifts (within the scope of their training) because the calibration property is embedded in the model's weights rather than applied as a post-processing layer.

Third, \emph{output type determines integration complexity}. Families~I and III produce the richest structured outputs (class probabilities, typed decisions with confidence), making them the most straightforward to integrate into threshold-based decision logic. Family~II produces ordinal scores that require careful interpretation. Family~IV produces similarity scores that are meaningful only relative to other candidates, not in absolute terms. Practitioners selecting a model family should consider not just accuracy and latency, but how naturally the model's output type maps onto their downstream decision logic.

Fourth, \emph{Family~V (extraction) stands apart} from the other four families in a fundamental way: its output is a typed \emph{value} (a date, a price, an entity name), not a calibrated probability or a decision. This means extraction models do not fit neatly into the calibration framework that unifies Families~I--IV. They are included in this taxonomy because their inference architecture satisfies our operational definition (single-pass, structured output, no autoregressive generation), and because they increasingly appear alongside decision models in production pipelines -- e.g., an extraction model identifies the dollar amount in a contract clause, and a Family~III decision engine determines whether that amount exceeds a risk threshold. However, their calibration properties remain largely unexplored: it is an open question how to define and measure ``calibration'' for a model whose output is a structured value rather than a probability. We mark Family~V as \emph{emerging} throughout the guide to reflect this less mature status.

\clearpage
\ifdim\pagetotal>0pt\thispagestyle{empty}\newpage\fi
\part{Model Families}

The taxonomy presented in Section~\ref{sec:taxonomy} organizes System~1 decision models into five families based on architecture, training methodology, and output semantics. This chapter examines each family in depth. We begin with the classical encoder-based classifiers (Family~I) that established the paradigm, move through reward models (Family~II) that serve as scoring oracles within RL pipelines, then cover the recent non-autoregressive decision engines (Family~III) that introduced typed decision primitives and RLCD training, the contrastive models (Family~IV) that trade interaction depth for cacheability, and finally the emerging extraction models (Family~V) that extend the typed-output paradigm from decisions to structured value extraction. For each family we cover architecture, training, inference, and practical trade-offs, with code examples where applicable.

\section{Family I: Encoder-Based Classifiers}
\label{sec:encoder}

The oldest and most established family of System~1 decision models builds on bidirectional transformer encoders. These models process the entire input in a single forward pass and produce a fixed-dimensional representation suitable for classification. Starting with BERT's \texttt{[CLS]} paradigm in 2018, this family has evolved through RoBERTa and DeBERTa (which improved training and attention mechanisms), cross-encoder rerankers (which became a workhorse of production search systems), and specialized retrieval architectures such as SPLADE and ColBERT. We cover each in turn, tracing how the single-forward-pass classification pattern was refined from a research curiosity into a production-grade decision architecture.

\subsection{BERT and the \texttt{[CLS]} Paradigm}
\label{sec:bert_cls}

BERT (Bidirectional Encoder Representations from Transformers)~\citep{devlin2019bert} introduced the paradigm of pretraining a bidirectional transformer encoder on masked language modeling (MLM) and next sentence prediction (NSP), then fine-tuning on downstream tasks.

\begin{figure}[H]
\centering
\begin{tikzpicture}[
    box/.style={draw, rounded corners=2pt, minimum height=0.7cm, minimum width=1.4cm, font=\small\sffamily, fill=#1},
    box/.default=blue!8,
    arr/.style={-{Stealth[length=4pt]}, thick},
    lbl/.style={font=\scriptsize\sffamily},
]
\foreach \i/\tok in {0/\texttt{[CLS]}, 1/$w_1$, 2/$w_2$, 3/\dots, 4/$w_n$} {
    \node[box=gray!12] (in\i) at (\i*1.7, 0) {\tok};
}

\foreach \i in {0,...,4} {
    \node[box=orange!15, minimum width=1.4cm] (emb\i) at (\i*1.7, 1.2) {};
    \draw[arr] (in\i) -- (emb\i);
}
\node[lbl, left=0.1cm of emb0] {Embeddings};

\foreach \i in {0,...,4} {
    \node[box=blue!12, minimum width=1.4cm] (t1\i) at (\i*1.7, 2.4) {};
    \draw[arr] (emb\i) -- (t1\i);
}
\node[lbl, left=0.1cm of t10] {Trm Layer 1};

\foreach \i in {0,...,3} {
    \pgfmathtruncatemacro{\j}{\i+1}
    \draw[{Stealth[length=3pt]}-{Stealth[length=3pt]}, gray!60, thin] (t1\i.east) -- (t1\j.west);
}

\foreach \i in {0,...,4} {
    \node[box=blue!20, minimum width=1.4cm] (t2\i) at (\i*1.7, 3.6) {};
    \draw[arr] (t1\i) -- (t2\i);
}
\node[lbl, left=0.1cm of t20] {Trm Layer $L$};

\foreach \i in {0,...,3} {
    \pgfmathtruncatemacro{\j}{\i+1}
    \draw[{Stealth[length=3pt]}-{Stealth[length=3pt]}, gray!60, thin] (t2\i.east) -- (t2\j.west);
}

\foreach \i in {0,...,4} {
    \node[font=\small, gray] at (\i*1.7, 3.0) {$\vdots$};
}

\foreach \i/\tok in {0/$\mathbf{h}_{\texttt{[CLS]}}$, 1/$\mathbf{h}_1$, 2/$\mathbf{h}_2$, 3/\dots, 4/$\mathbf{h}_n$} {
    \node[box=green!12, minimum width=1.4cm] (out\i) at (\i*1.7, 4.8) {\tok};
    \draw[arr] (t2\i) -- (out\i);
}

\node[box=red!15, minimum width=2.2cm, minimum height=0.7cm] (cls) at (0, 6.2) {Classification};
\draw[arr, thick, red!60!black] (out0) -- (cls);

\node[box=red!25, minimum width=2.2cm] (prob) at (0, 7.4) {$p(y \mid x)$};
\draw[arr, thick, red!60!black] (cls) -- (prob);

\node[lbl, right=0.3cm of out4] {Hidden states};
\node[lbl, right=0.3cm of in4] {Input tokens};

\end{tikzpicture}
\caption{BERT encoder architecture with \texttt{[CLS]} classification. Input tokens pass through an embedding layer and $L$ bidirectional transformer layers (horizontal arrows indicate full self-attention). The final \texttt{[CLS]} hidden state is fed to a classification head that outputs class probabilities $p(y \mid x)$ in a single forward pass.}
\label{fig:bert_cls}
\end{figure}
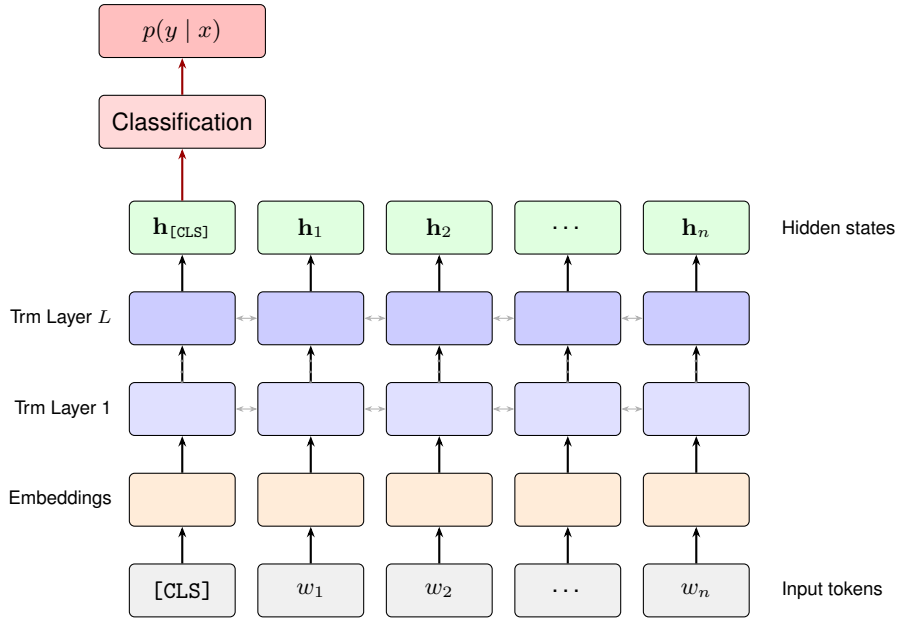

\paragraph{Encoder architecture.} Given an input token sequence $\mathbf{x} = (x_0, x_1, \ldots, x_n)$ where $x_0 = \texttt{[CLS]}$, the encoder produces contextualised representations through $L$ stacked transformer layers~\citep{vaswani2017attention}.  Each layer $\ell$ applies multi-head self-attention (MHSA) followed by a position-wise feed-forward network (FFN):
\begin{align}
    \mathbf{A}^{(\ell)} &= \operatorname{softmax}\!\Bigl(\frac{\mathbf{Q}^{(\ell)}{\mathbf{K}^{(\ell)}}^{\!\top}}{\sqrt{d_k}}\Bigr)\mathbf{V}^{(\ell)}, \label{eq:self_attn}\\
    \mathbf{Q}^{(\ell)}&=\mathbf{H}^{(\ell-1)}W_Q^{(\ell)},\quad
    \mathbf{K}^{(\ell)}=\mathbf{H}^{(\ell-1)}W_K^{(\ell)}, \nonumber\\
    \mathbf{V}^{(\ell)}&=\mathbf{H}^{(\ell-1)}W_V^{(\ell)}, \nonumber\\[4pt]
    \mathbf{H}^{(\ell)} &= \text{FFN}\bigl(\text{LN}(\mathbf{H}^{(\ell-1)} {+} \text{MHSA}(\mathbf{H}^{(\ell-1)}))\bigr), \label{eq:transformer_layer}
\end{align}
where $W_Q^{(\ell)}, W_K^{(\ell)} \in \mathbb{R}^{d \times d_k}$, $W_V^{(\ell)} \in \mathbb{R}^{d \times d_v}$, and $\mathbf{H}^{(0)}$ is the sum of token, segment, and positional embeddings. Intuitively, each token broadcasts a \emph{query} (``what am I looking for?''), a \emph{key} (``what do I contain?''), and a \emph{value} (``what information should I share?''). The softmax-scaled dot product $QK^\top/\sqrt{d_k}$ computes a relevance-weighted attention map, and the output is a value-weighted mixture, enabling each token to aggregate context from the entire sequence in parallel.

\paragraph{Classification head.}  For classification, the hidden state of the \texttt{[CLS]} token at the final layer serves as a fixed-length sentence representation:
\begin{align}
    \mathbf{h}_{\texttt{[CLS]}} &= \mathbf{H}^{(L)}_{0} \in \mathbb{R}^d, \nonumber\\
    p(y \mid x) &= \operatorname{softmax}(W_c \mathbf{h}_{\texttt{[CLS]}} + \mathbf{b}_c),
    \label{eq:cls_head}
\end{align}
where $W_c \in \mathbb{R}^{C \times d}$.

This architecture is inherently a System~1 model: it requires exactly one forward pass of complexity $O(n^2 d + n d^2)$ for sequence length $n$, produces class probabilities directly, and runs in 5--50ms depending on model size and sequence length.

\paragraph{Pretraining objectives.} BERT is pretrained with two objectives: (i)~\emph{Masked Language Modeling} (MLM), where 15\% of input tokens are masked and the model predicts them from context:
\begin{equation}
    \mathcal{L}_{\text{MLM}} = -\mathbb{E}_{x \sim \mathcal{D}}\;\sum_{i \in \mathcal{M}} \log p_\theta(x_i \mid x_{\setminus \mathcal{M}}),
    \label{eq:mlm}
\end{equation}
where $\mathcal{M}$ is the set of masked positions. MLM forces the encoder to build rich bidirectional representations: predicting a missing word requires understanding both left and right context, which produces the contextual embeddings that make downstream classification effective. This is followed by (ii)~\emph{Next Sentence Prediction} (NSP), a binary classification of whether two segments are consecutive.

\paragraph{Limitations of the \texttt{[CLS]} approach.} \citet{behrendt2025maxpool} showed that the \texttt{[CLS]} token is a suboptimal bottleneck, other tokens and intermediate layers encode valuable contextual information that the \texttt{[CLS]} representation discards. Their MaxPoolBERT aggregates information across layers and tokens via max-pooling:
\begin{equation}
    \mathbf{h}_{\text{pool}} = \max_{\ell=1}^{L} \max_{i=0}^{n} \mathbf{H}^{(\ell)}_i \;\in\;\mathbb{R}^{d},
    \label{eq:maxpool}
\end{equation}
improving classification accuracy particularly on low-resource tasks without additional pretraining.

\subsection{Successor Encoders: RoBERTa and DeBERTa}

\paragraph{RoBERTa}~\citep{liu2019roberta} demonstrated that BERT was significantly undertrained. By removing NSP, training longer with larger batches and more data, and using dynamic masking, RoBERTa achieved substantial improvements across all benchmarks while retaining the same \texttt{[CLS]} classification paradigm.

\paragraph{DeBERTa}~\citep{he2021deberta} introduced \emph{disentangled attention}, which decomposes each token representation into content vector $\mathbf{c}_i$ and relative position vector $\mathbf{p}_{i|j}$. The attention score between tokens $i$ and $j$ becomes:
\begin{equation}
    A_{ij} = \underbrace{\mathbf{c}_i^{\top} W_{c2c}\,\mathbf{c}_j}_{\text{c2c}} + \underbrace{\mathbf{c}_i^{\top} W_{c2p}\,\mathbf{p}_{i|j}}_{\text{c2p}} + \underbrace{\mathbf{p}_{j|i}^{\top} W_{p2c}\,\mathbf{c}_j}_{\text{p2c}}\,,\footnote{The theoretical decomposition includes a fourth position-to-position (p2p) term, $\mathbf{p}_{j|i}^{\top} W_{p2p}\,\mathbf{p}_{i|j}$, but \citet{he2021deberta} drop it as it does not empirically contribute to performance.}
    \label{eq:deberta_attn}
\end{equation}
separating content and positional contributions rather than summing embeddings. The four-way decomposition allows the model to learn \emph{independently} how word meaning, word position, and their interaction contribute to attention, for example, learning that a negation word (content) appearing immediately before a verb (position) deserves high attention, without conflating these two signals.  DeBERTa-v3~\citep{he2023debertav3} further improved via replaced token detection (ELECTRA-style~\citep{clark2020electra}) pretraining, where a generator produces plausible replacements and the encoder learns to discriminate replaced tokens:
\begin{equation}
    \mathcal{L}_{\text{RTD}} = -\sum_{i=1}^{n}\bigl[y_i\log p_\theta(\text{replaced}\mid x_i) + (1-y_i)\log(1-p_\theta(\text{replaced}\mid x_i))\bigr],
\end{equation}
with $y_i=1$ if token $i$ was replaced by the generator.  DeBERTa models are widely used as the backbone for reward models and NLI-based zero-shot classifiers~\citep{laurer2024universal}.

\subsection{Cross-Encoder Rerankers}
\label{sec:cross_encoder}

Cross-encoders~\citep{nogueira2019passage,nogueira2020document} represent perhaps the most successful System~1 decision architecture in production. Given a query $q$ and a document $d$, the cross-encoder jointly encodes the pair:
\begin{equation}
    s(q, d) = W \!\cdot\! \text{Enc}(\texttt{[CLS]}, q, \texttt{[SEP]}, d)_0 + b,
    \label{eq:cross_encoder}
\end{equation}
where $s \in \mathbb{R}$.

The scalar $s(q,d)$ represents a relevance score. This architecture captures rich token-level interactions between $q$ and $d$, unlike bi-encoder models that encode them independently, making it highly effective for reranking.

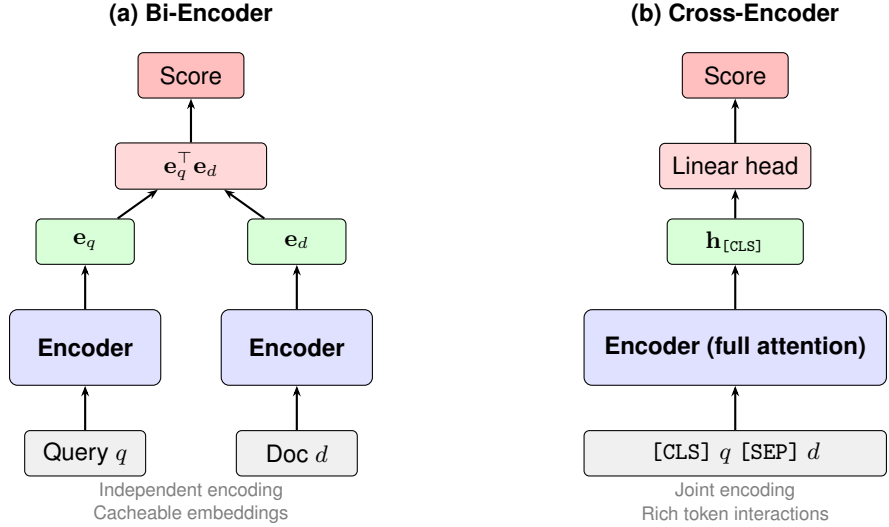
\begin{figure}[H]
\centering
\begin{tikzpicture}[
    box/.style={draw, rounded corners=2pt, minimum height=0.6cm, font=\small\sffamily, fill=#1},
    box/.default=blue!8,
    arr/.style={-{Stealth[length=4pt]}, thick},
    lbl/.style={font=\scriptsize\sffamily},
    enc/.style={draw, rounded corners=3pt, minimum height=1.0cm, minimum width=2.0cm, font=\small\sffamily\bfseries, fill=blue!12},
]
\node[font=\small\sffamily\bfseries] at (-3.6, 5.8) {(a) Bi-Encoder};

\node[box=gray!12, minimum width=1.6cm] (bq) at (-5.0, 0) {Query $q$};
\node[box=gray!12, minimum width=1.6cm] (bd) at (-2.2, 0) {Doc $d$};

\node[enc] (beq) at (-5.0, 1.4) {Encoder};
\node[enc] (bed) at (-2.2, 1.4) {Encoder};

\draw[arr] (bq) -- (beq);
\draw[arr] (bd) -- (bed);

\node[box=green!15, minimum width=1.3cm] (bvq) at (-5.0, 2.8) {$\mathbf{e}_q$};
\node[box=green!15, minimum width=1.3cm] (bvd) at (-2.2, 2.8) {$\mathbf{e}_d$};

\draw[arr] (beq) -- (bvq);
\draw[arr] (bed) -- (bvd);

\node[box=red!15, minimum width=2.0cm] (bdot) at (-3.6, 3.8) {$\mathbf{e}_q^\top \mathbf{e}_d$};
\draw[arr] (bvq) -- (bdot);
\draw[arr] (bvd) -- (bdot);

\node[box=red!25, minimum width=1.4cm] (bscore) at (-3.6, 5.0) {Score};
\draw[arr, thick] (bdot) -- (bscore);

\node[font=\small\sffamily\bfseries] at (3.6, 5.8) {(b) Cross-Encoder};

\node[box=gray!12, minimum width=4.0cm] (cinp) at (3.6, 0) {\texttt{[CLS]} $q$ \texttt{[SEP]} $d$};

\node[enc, minimum width=4.0cm] (cenc) at (3.6, 1.4) {Encoder (full attention)};
\draw[arr] (cinp) -- (cenc);

\node[box=green!15, minimum width=1.8cm] (ccls) at (3.6, 2.8) {$\mathbf{h}_{\texttt{[CLS]}}$};
\draw[arr] (cenc) -- (ccls);

\node[box=red!15, minimum width=2.0cm] (chead) at (3.6, 3.8) {Linear head};
\draw[arr] (ccls) -- (chead);

\node[box=red!25, minimum width=1.4cm] (cscore) at (3.6, 5.0) {Score};
\draw[arr, thick] (chead) -- (cscore);

\node[lbl, text=gray] at (-3.6, -0.5) {Independent encoding};
\node[lbl, text=gray] at (-3.6, -0.8) {Cacheable embeddings};
\node[lbl, text=gray] at (3.6, -0.5) {Joint encoding};
\node[lbl, text=gray] at (3.6, -0.8) {Rich token interactions};

\end{tikzpicture}
\caption{Bi-encoder vs.\ cross-encoder architectures. (a)~A bi-encoder encodes query and document independently, then compares via dot product; embeddings are cacheable but interactions are shallow. (b)~A cross-encoder jointly encodes the concatenated pair, enabling deep token-level interactions at the cost of requiring a full forward pass per pair.}
\label{fig:bi_vs_cross}
\end{figure}

\paragraph{Training objectives.} Cross-encoders are trained under one of three paradigms. The \emph{pointwise} approach uses binary cross-entropy on individual (query, document, relevance) triples. The \emph{pairwise} approach uses a margin loss on (query, positive document, negative document) triples:
    \begin{equation}
        \mathcal{L} = \max(0, m - s(q, d^+) + s(q, d^-)).
    \end{equation}
The margin $m$ enforces a minimum gap between the positive and negative document scores, if $s(q, d^+) - s(q, d^-) \geq m$, the loss is zero and no gradient flows, preventing the model from wasting capacity on already-separated pairs. The \emph{listwise} approach applies cross-entropy over a softmax distribution of candidate scores~\citep{nogueira2020document}.

\paragraph{Production deployment.} \citet{walmart2025crossencoder} reported large-scale deployment of BERT-based cross-encoders in Walmart's search engine, employing intermediate representations, operator fusion, and vectorization to achieve production-viable latency. This demonstrates that encoder-based System~1 models are mature enough for web-scale deployment.

\begin{lstlisting}[caption={Cross-encoder reranking with a pretrained model.},label={lst:reranker}]
from transformers import AutoTokenizer, AutoModelForSequenceClassification
import torch

class CrossEncoderReranker:
    def __init__(self, model_name="cross-encoder/ms-marco-MiniLM-L-12-v2"):
        self.tokenizer = AutoTokenizer.from_pretrained(model_name)
        self.model = AutoModelForSequenceClassification.from_pretrained(
            model_name)
        self.model.eval()

    @torch.no_grad()
    def rerank(self, query: str, documents: list[str], top_k: int = 10):
        """Score all (query, doc) pairs in one batched forward pass."""
        pairs = [(query, doc) for doc in documents]
        inputs = self.tokenizer(
            pairs, padding=True, truncation=True,
            max_length=512, return_tensors="pt",
        )
        scores = self.model(**inputs).logits.squeeze(-1).tolist()
        ranked = sorted(enumerate(scores), key=lambda x: x[1], reverse=True)
        return ranked[:top_k]  # [(orig_idx, score), ...]

# Usage: BM25 retrieves 100 candidates, cross-encoder reranks to top 10
reranker = CrossEncoderReranker()
candidates = bm25_search(query, top_k=100)
top10 = reranker.rerank(query, [c.text for c in candidates], top_k=10)
\end{lstlisting}

\begin{tipbox}
Batch your cross-encoder calls: scoring 50 documents in one forward pass takes roughly the same wall-clock time as scoring one, thanks to GPU parallelism. The sweet spot is 50--100 candidates -- enough for good recall, few enough for $<$50ms reranking latency.
\end{tipbox}

\paragraph{Beyond \texttt{[CLS]} for reranking.} \citet{celi2024} showed that collapsing all information into the \texttt{[CLS]} token is suboptimal for cross-encoders, advocating for alternative score extraction mechanisms. Similarly, \citet{hofstatter2024injecting} demonstrated that injecting the first-stage retrieval score as a text token into the cross-encoder input improves reranking performance.

\subsection{Learned Sparse Retrieval: SPLADE}

SPLADE~\citep{formal2021splade,formal2022splade} uses a BERT encoder to predict \emph{sparse} term-level importance weights, producing an inverted-index representation. For each token $j$ in the vocabulary, SPLADE computes a term importance weight via log-saturated attention over the encoder outputs:
\begin{equation}
    w_j = \sum_{i=1}^{n} \log\!\big(1 + \text{ReLU}(\mathbf{W}_{\text{MLM}} \mathbf{h}_i + \mathbf{b}_{\text{MLM}})_j\big),
    \label{eq:splade}
\end{equation}
where $\mathbf{h}_i$ is the encoder output for position $i$ and $\mathbf{W}_{\text{MLM}}$ is the masked language model head. The sparse vector $\mathbf{w} \in \mathbb{R}^{|\mathcal{V}|}$ enables standard inverted-index retrieval while capturing semantic term expansion. SPLADE exemplifies how encoder-based architectures make fast, implicit System~1 decisions about term relevance.

\subsection{Late Interaction: ColBERT}

ColBERT~\citep{khattab2020colbert} independently encodes queries and documents into sets of token-level embeddings, then computes relevance via the \textbf{MaxSim} late interaction operator:
\begin{equation}
    \text{rel}(q, d) = \sum_{i=1}^{|q|} \max_{j \in \{1,\ldots,|d|\}} \mathbf{q}_i^\top \mathbf{d}_j,
    \label{eq:maxsim}
\end{equation}
where $\mathbf{q}_i, \mathbf{d}_j \in \mathbb{R}^{d'}$ are normalized token embeddings. Each query token finds its best-matching document token, a per-token System~1 decision about relevance. The entire retrieval + scoring pipeline operates as a System~1 process with $O(|q| \cdot |d|)$ scoring complexity, intermediate between dual-encoders ($O(d')$) and cross-encoders ($O(n^2 d)$).

\begin{examplebox}[Example: PreSizE -- encoder classifiers beyond text]
The encoder-classifier pattern is not limited to NLP and IR. \citet{eshel2021presize} applied stacked Transformer encoders to clothing size prediction: item attributes are embedded via a \texttt{[CLS]} token, a second Transformer encodes the buyer's purchase history, and a softmax head predicts one of 286 clothing size classes. Item embeddings are pre-computed offline, so inference is fast enough for online serving -- a straightforward application of the Family~I recipe (\emph{encoder $\to$ \texttt{[CLS]} $\to$ task head $\to$ softmax}) to a non-textual domain.
\end{examplebox}

\section{Family II: Reward Models}
\label{sec:reward}

Reward models (RMs) occupy a critical position in the LLM ecosystem: they translate human preferences into scalar signals that guide reinforcement learning. Architecturally, they share the defining characteristic of System~1 decision models -- they take an (input, response) pair and produce a score in a single forward pass -- but they serve primarily as components within larger System~2 training pipelines. This chapter covers the Bradley-Terry pairwise formulation that dominates current practice, its listwise extension via Plackett-Luce, and specialized variants including process reward models (which score intermediate reasoning steps), outcome reward models, generative reward models, and rule-based rewards. Understanding reward models is essential because their architecture and training methods directly influenced the design of Family~III decision engines.

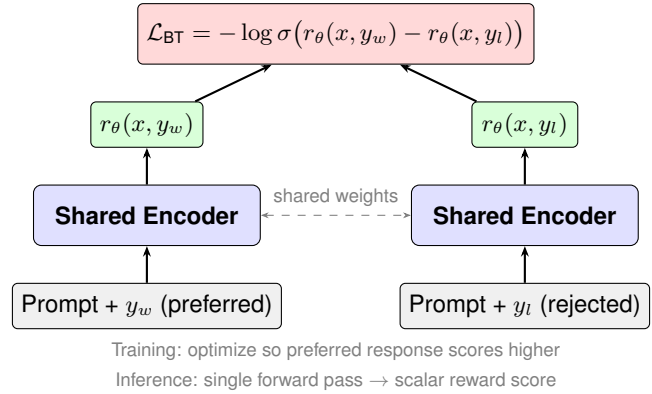
\begin{figure}[H]
\centering
\begin{tikzpicture}[
    box/.style={draw, rounded corners=2pt, minimum height=0.6cm, font=\small\sffamily, fill=#1},
    box/.default=blue!8,
    arr/.style={-{Stealth[length=4pt]}, thick},
    enc/.style={draw, rounded corners=3pt, minimum height=0.8cm, minimum width=3.0cm, font=\small\sffamily\bfseries, fill=blue!12},
    lbl/.style={font=\scriptsize\sffamily},
]
\node[box=gray!12, minimum width=2.8cm] (inw) at (-2.5, 0) {Prompt + $y_w$ (preferred)};
\node[enc] (encw) at (-2.5, 1.2) {Shared Encoder};
\draw[arr] (inw) -- (encw);
\node[box=green!15, minimum width=1.4cm] (rw) at (-2.5, 2.4) {$r_\theta(x, y_w)$};
\draw[arr] (encw) -- (rw);

\node[box=gray!12, minimum width=2.8cm] (inl) at (2.5, 0) {Prompt + $y_l$ (rejected)};
\node[enc] (encl) at (2.5, 1.2) {Shared Encoder};
\draw[arr] (inl) -- (encl);
\node[box=green!15, minimum width=1.4cm] (rl) at (2.5, 2.4) {$r_\theta(x, y_l)$};
\draw[arr] (encl) -- (rl);

\draw[{Stealth[length=3pt]}-{Stealth[length=3pt]}, dashed, gray] (encw.east) -- (encl.west) node[midway, above, lbl] {shared weights};

\node[box=red!15, minimum width=4.5cm, minimum height=0.8cm] (bt) at (0, 3.6) {$\mathcal{L}_{\text{BT}} = -\log\sigma\bigl(r_\theta(x,y_w) - r_\theta(x,y_l)\bigr)$};
\draw[arr] (rw) -- (bt);
\draw[arr] (rl) -- (bt);

\node[lbl, text=gray] at (0, -0.6) {Training: optimize so preferred response scores higher};
\node[lbl, text=gray] at (0, -1.0) {Inference: single forward pass $\to$ scalar reward score};

\end{tikzpicture}
\caption{Bradley-Terry reward model training. Both the preferred ($y_w$) and rejected ($y_l$) responses are scored by the same encoder. The BT loss pushes the preferred response's score above the rejected one. At inference, a single forward pass maps any (prompt, response) pair to a scalar reward.}
\label{fig:reward_model}
\end{figure}

\paragraph{Output semantics.} An important distinction separates reward models from other System~1 decision models: the scalar output of a reward model represents a learned \emph{preference or utility signal}, not necessarily a calibrated probability that an answer is correct. A Bradley-Terry reward model, for example, is trained so that reward \emph{differences} predict pairwise preferences; the absolute reward value has no intrinsic probabilistic meaning. This distinction matters for applications that require calibrated confidence (e.g., routing decisions, safety gating), where an additional calibration layer may be needed to convert reward scores into well-calibrated probabilities. We include reward models in this guide because their inference architecture fits our operational definition, but we note that their native output semantics differ from those of probability-producing classifiers and decision engines.

\subsection{Bradley-Terry Reward Models}
\label{sec:bt_rm}

The standard reward model architecture follows the Bradley-Terry (BT) framework~\citep{bradley1952rank}. Given a prompt $q$ and two responses $y_w$ (preferred) and $y_l$ (dispreferred), the BT model assumes:
\begin{equation}
    p(y_w \succ y_l \mid q) = \sigma(r_\phi(q, y_w) - r_\phi(q, y_l)),
\end{equation}
where $\sigma$ is the sigmoid function and $r_\phi$ is a scalar reward function parameterized by $\phi$. The sigmoid of the reward \emph{difference} models preference as a soft comparison: when $r_\phi(q, y_w)$ is much larger than $r_\phi(q, y_l)$, the preference probability approaches 1; when the rewards are equal, it is 0.5 (a coin flip). This means only the \emph{gap} between rewards matters, not their absolute values, a property that makes BT models invariant to additive shifts in the reward scale. The training loss is:
\begin{equation}
    \mathcal{L}_{\text{BT}}(\phi) = -\mathbb{E}_{(q, y_w, y_l) \sim \mathcal{D}} [\log \sigma(r_\phi(q, y_w) - r_\phi(q, y_l))].
    \label{eq:bt_loss}
\end{equation}

\paragraph{Architecture.} The reward function $r_\phi$ is typically implemented by taking a pretrained LLM, removing the language modeling head, and adding a linear projection from the final hidden state to a scalar~\citep{ouyang2022training}:
\begin{equation}
    r_\phi(q, y) = W^{\top} \mathbf{h}_{\text{last}}(q \oplus y) + b, \quad W \in \mathbb{R}^d, b \in \mathbb{R}.
\end{equation}

This is structurally identical to the BERT \texttt{[CLS]} classification paradigm (Section~\ref{sec:bert_cls}), but applied to a decoder-only backbone with the last token's hidden state replacing \texttt{[CLS]}.

\paragraph{Scale.} Reward models range from small (DeBERTa-v3-large, 350M parameters) to large (Llama-70B backbone), with larger models generally exhibiting better generalization~\citep{ouyang2022training}. The reward model is a pure System~1 component: given a fixed (prompt, response), it outputs a scalar in a single forward pass.

\paragraph{Gradient dynamics.} The gradient of the BT loss with respect to $\phi$ is:
\begin{equation}
    \nabla_\phi \mathcal{L}_{\text{BT}} = -\mathbb{E}\Big[\big(1 - \sigma(\Delta r)\big)\big(\nabla_\phi r_\phi(q, y_w) - \nabla_\phi r_\phi(q, y_l)\big)\Big],
    \label{eq:bt_grad}
\end{equation}
where $\Delta r = r_\phi(q, y_w) - r_\phi(q, y_l)$. The weighting factor $(1 - \sigma(\Delta r))$ ensures that gradients are largest for pairs where the model assigns similar scores to preferred and dispreferred responses, and vanish as the model becomes confident, a form of implicit hard-example mining.

\subsection{Plackett-Luce Listwise Reward Models}
\label{sec:pl_rm}

The Plackett-Luce (PL) model~\citep{plackett1975analysis,luce1959individual} generalizes Bradley-Terry to full rankings over $K$ items. Given $K$ responses $y_1, \ldots, y_K$ with ranking $\pi$ (where $\pi(1)$ is best):
\begin{equation}
    p(\pi \mid q) = \prod_{i=1}^{K} \frac{e^{r_\phi(y_{\pi(i)}, q)}}{\sum_{j=i}^{K} e^{r_\phi(y_{\pi(j)}, q)}}.
    \label{eq:pl}
\end{equation}

The PL model has an elegant sequential interpretation: at each stage $i$, the model selects the next-best item from the remaining pool via a softmax over their rewards. The denominator shrinks as items are ``removed'' from the pool, capturing the sequential elimination process. PL generalizes BT from pairwise preferences to full ranking distributions over $K$ items, capturing relative margins simultaneously rather than through pairwise decomposition. For GRPO-style training~\citep{shao2024deepseekmath}, where $N$ responses are generated per prompt and ranked, listwise reward models are a natural fit.

\subsection{Process Reward Models (PRMs)}

Process Reward Models~\citep{lightman2023lets,uesato2022solving} extend the reward model paradigm from \emph{outcome-level} to \emph{step-level} evaluation. Given a reasoning chain $y = (s_1, s_2, \ldots, s_T)$, a PRM assigns a score to each step:
\begin{equation}
    r_{\text{PRM}}(q, y) = \prod_{t=1}^{T} p(\text{step } s_t \text{ is correct} \mid q, s_{<t}).
\end{equation}

PRMs are compositional System~1 models: each step evaluation is a fast, single-pass decision, and the step-level scores compose into a trajectory-level signal. \citet{lightman2023lets} demonstrated that PRMs significantly outperform outcome-level reward models for guiding mathematical reasoning, particularly when combined with beam search.

\subsection{Outcome Reward Models (ORMs)}

Outcome Reward Models~\citep{cobbe2021training} assign a single scalar to the complete (problem, solution) pair, evaluating the final answer without examining intermediate reasoning. While less informative than PRMs, ORMs are simpler to train (requiring only final answer correctness labels) and cheaper to run. The choice between PRM and ORM represents a System~1 design trade-off: richer signal vs. annotation and compute cost.

\subsection{Generative Reward Models}
\label{sec:generative_rm}

An emerging paradigm uses LLMs themselves as reward models by generating natural-language critiques that are then mapped to scores~\citep{zhang2024generative}. The CE-RM approach~\citep{cerm2025} trains a pointwise generative reward model via two-stage contrastive evaluation. While these models \emph{generate} text (making them technically System~2), the recent Rubric-ARROW framework~\citep{rubricarrow2025} decomposes evaluation into rubric-based criteria, partially recovering System~1 properties through structured assessment.

\subsection{Rule-Based Rewards}
\label{sec:rule_rewards}

DeepSeek-R1~\citep{deepseek2025r1} demonstrated that pure rule-based rewards, binary correctness checks, format compliance tests, code execution results, can be remarkably effective for RL training. Rule-based rewards provide deterministic, reproducible supervision when the target criterion is explicitly defined and mechanically verifiable: they have zero latency, infinite scalability, and no learned parameters. Their reliability depends on the correctness and coverage of the rule; they do not inherently provide calibrated probabilities, and their expressiveness is limited to aspects with unambiguous ground truth.

\section{Family III: Non-Autoregressive Decision Engines}
\label{sec:nonar}

The newest and most distinctive family of System~1 models abandons autoregressive generation entirely. Instead of producing tokens, these models output typed, structured decisions through specialized prediction heads in a single forward pass. We first present a unified framework that captures all Family~III models, then provide detailed treatment of each system.

\subsection{A Unified Framework}
\label{sec:nar_framework}

Despite surface differences in scale and implementation, all Family~III models instantiate a common computational pattern that we formalize here as a \textbf{reference abstraction}. This framework is a descriptive tool for comparing existing systems, not a claim that all models share identical internals. Some models implement all three decision primitives described below; others implement a subset or use different head designs. We note which features are documented for each system and which are inferred by analogy in the comparative analysis (Table~\ref{tab:nar_compare}).

\subsubsection{General Architecture}

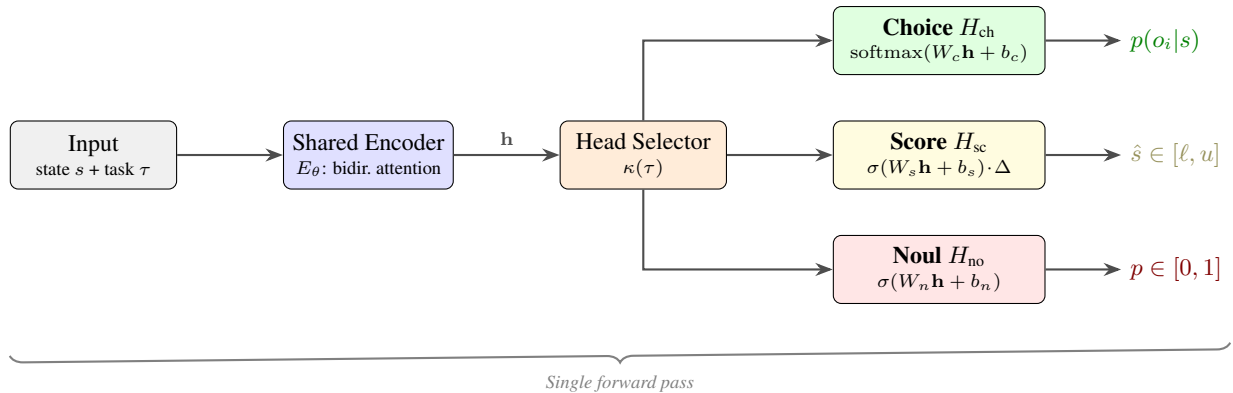
\begin{figure}[H]
\centering
\begin{tikzpicture}[
    >=Stealth,
    block/.style={draw, rounded corners=3pt, minimum width=22mm, minimum height=9mm, font=\small, align=center, fill=#1},
    head/.style={draw, rounded corners=3pt, minimum width=28mm, minimum height=9mm, font=\small, align=center, fill=#1},
    arr/.style={->, thick, color=black!70},
]
\node[block=gray!12] (inp) {Input\\[-1pt]\scriptsize state $s$ + task $\tau$};
\node[block=blue!12, right=14mm of inp] (enc) {Shared Encoder\\[-1pt]\scriptsize $E_\theta$: bidir.\ attention};
\node[block=orange!15, right=14mm of enc] (route) {Head Selector\\[-1pt]\scriptsize $\kappa(\tau)$};
\node[head=green!12, above right=6mm and 14mm of route] (ch) {\textbf{Choice} $H_{\text{ch}}$\\[-1pt]\scriptsize $\operatorname{softmax}(W_c\mathbf{h}+b_c)$};
\node[head=yellow!15, right=14mm of route] (sc) {\textbf{Score} $H_{\text{sc}}$\\[-1pt]\scriptsize $\sigma(W_s\mathbf{h}+b_s)\!\cdot\!\Delta$};
\node[head=red!10, below right=6mm and 14mm of route] (no) {\textbf{Noul} $H_{\text{no}}$\\[-1pt]\scriptsize $\sigma(W_n\mathbf{h}+b_n)$};
\node[right=10mm of ch, font=\small\itshape, green!50!black] (o1) {$p(o_i|s)$};
\node[right=10mm of sc, font=\small\itshape, yellow!50!black] (o2) {$\hat{s} \in [\ell, u]$};
\node[right=10mm of no, font=\small\itshape, red!50!black] (o3) {$p \in [0,1]$};
\draw[arr] (inp) -- (enc);
\draw[arr] (enc) -- node[above, font=\scriptsize]{$\mathbf{h}$} (route);
\draw[arr] (route) |- (ch);
\draw[arr] (route) -- (sc);
\draw[arr] (route) |- (no);
\draw[arr] (ch) -- (o1);
\draw[arr] (sc) -- (o2);
\draw[arr] (no) -- (o3);
\draw[decorate, decoration={brace, amplitude=5pt, mirror}, thick, gray]
  ([yshift=-22mm]inp.south west) -- ([yshift=-22mm]o2.south east)
  node[midway, below=6pt, font=\scriptsize\itshape, gray]{Single forward pass};
\end{tikzpicture}
\caption{Unified architecture of Family~III non-autoregressive decision models. All models share the pattern: bidirectional encoder $\to$ head selector $\to$ typed prediction head. The entire pipeline executes in one forward pass.}
\label{fig:nar_arch}
\end{figure}

A non-autoregressive decision model $\mathcal{M} = (E_\theta, \{H_k\}_{k=1}^{3})$ consists of a \textbf{shared encoder} $E_\theta: \mathcal{X} \to \mathbb{R}^d$ that maps an input (state description, options, criteria) to a dense representation via bidirectional self-attention; a \textbf{head selector} $\kappa: \mathcal{T} \to \{1, 2, 3\}$ that routes each task specification $\tau$ to the appropriate prediction head; and a set of three \textbf{typed prediction heads} $\{H_{\text{ch}}, H_{\text{sc}}, H_{\text{no}}\}$, each implementing a different decision primitive. The forward pass is:
\begin{align}
    \mathbf{h} &= E_\theta(\text{concat}(s, \tau)) \in \mathbb{R}^d, \label{eq:nar_encode} \\
    (\hat{y}, \hat{p}) &= H_{\kappa(\tau)}(\mathbf{h}), \label{eq:nar_head}
\end{align}
where $\hat{y}$ is the predicted decision and $\hat{p}$ is the associated calibrated probability. The entire computation is non-autoregressive: the output is produced in a single pass with $O(n^2 d)$ complexity (where $n$ is input sequence length), independent of output length. Figure~\ref{fig:nar_arch} illustrates this unified architecture.

\subsubsection{The Three Canonical Decision Primitives}

The reference framework defines three canonical decision primitives. Most Family~III models implement all three, though some implement a subset or use variant head designs. We formalize each primitive below.

\paragraph{Choice head $H_{\text{ch}}$.} Given state $s$ and $K$ options $\{o_1, \ldots, o_K\}$ with $K \leq K_{\max}$ (typically 255), the Choice head outputs a categorical distribution. Let $\mathbf{h}_s \in \mathbb{R}^d$ denote the encoder's pooled state representation and $\mathbf{h}_{o_i} \in \mathbb{R}^d$ the representation at the $i$-th option marker token. The head computes:
\begin{align}
    z_i &= \mathbf{w}_c^\top [\mathbf{h}_s \odot \mathbf{h}_{o_i}] + b_c, \quad i = 1, \ldots, K, \label{eq:choice_logit} \\
    p(o_i \mid s) &= \frac{\exp(z_i)}{\sum_{j=1}^{K} \exp(z_j)}, \label{eq:choice_prob}
\end{align}
where $\odot$ is element-wise product (capturing state--option interaction) and $\mathbf{w}_c \in \mathbb{R}^d$, $b_c \in \mathbb{R}$ are learnable. The output is $(\hat{y}, \hat{p}) = (\arg\max_i z_i, \, p(o_{\hat{y}} \mid s))$.

\paragraph{Score head $H_{\text{sc}}$.} Given state $s$ and a scoring criterion $c$ on a bounded ordinal scale $[\ell, u] \subset \mathbb{Z}$, the Score head performs calibrated ordinal regression. Rather than a single point estimate, it models the full ordinal distribution via cumulative probabilities:
\begin{align}
    P(s {\geq} m \mid \mathbf{h}) &= \sigma(\mathbf{w}_s^\top \mathbf{h} - \theta_m), \; m = \ell{+}1, \ldots, u, \label{eq:score_cumulative} \\
    P(s {=} m \mid \mathbf{h}) &= P(s {\geq} m) - P(s {\geq} m{+}1), \label{eq:score_pmf}
\end{align}
where $s$ denotes the score,
with boundary conditions $P(\text{score} \geq \ell) = 1$ and $P(\text{score} \geq u{+}1) = 0$. The thresholds $\theta_{\ell+1} \leq \cdots \leq \theta_u$ are constrained to be monotonically non-decreasing. The output is the probability-weighted mean $\hat{s} = \sum_{m=\ell}^{u} m \cdot P(\text{score} = m \mid \mathbf{h})$, with calibrated uncertainty from the ordinal distribution's entropy.

A \emph{simplified variant} used by some implementations (System~One~Gemma~\citep{kamat2026sysgemma}, THX-01~\citep{n1n2026thx01}) collapses this to a single sigmoid: $\hat{s} = \sigma(\mathbf{w}_s^\top \mathbf{h} + b_s) \cdot (u - \ell) + \ell$. This is faster but sacrifices the full ordinal distribution.

\paragraph{Noul head $H_{\text{no}}$.} Given state $s$ and a boolean predicate $q$, the Noul (``not null'') head outputs a probability estimate:
\begin{equation}
    p(\text{yes} \mid s, q) = \sigma(\mathbf{w}_n^\top \mathbf{h} + b_n), \label{eq:noul_prob}
\end{equation}
where $\mathbf{h}$ encodes the concatenation of $s$ and $q$. The output is $(\hat{y}, \hat{p}) = (\mathds{1}[p > 0.5], \, \max(p, 1{-}p))$. The simplicity belies its importance: safety-critical gating requires extremely well-calibrated $p$ values.

\begin{tipbox}
The three primitives cover the vast majority of production decision tasks. Before reaching for an LLM, ask: can I frame this as a Choice (pick one of $K$), a Score (rate on a scale), or a Noul (yes/no)? If yes, a System~1 model will be 10--100$\times$ faster and cheaper.
\end{tipbox}

\noindent The following listings illustrate how each primitive is invoked via a typical REST API (using Jev as the reference implementation). All three can be batched into a single call via the parallel sampler (Listing~\ref{lst:parallel}).

\begin{lstlisting}[caption={Choice: route a user query to the best tool.},label={lst:choice}]
import httpx

def route_to_tool(query: str, tools: list[str]) -> tuple[str, float]:
    """Select the best tool for a query using the Choice primitive."""
    resp = httpx.post(
        "https://api.typesafe.ai/v1/choice",
        headers={"Authorization": f"Bearer {API_KEY}"},
        json={
            "state": f"User query: {query}",
            "options": [
                {"label": t, "description": TOOL_CATALOG[t]}
                for t in tools
            ],
        },
    )
    result = resp.json()
    return result["choice"], result["confidence"]

tool, conf = route_to_tool(
    "What is the weather in Tokyo?",
    ["web_search", "calculator", "calendar", "weather_api"],
)
# -> ("weather_api", 0.94)
\end{lstlisting}

\begin{lstlisting}[caption={Score: rate support-ticket urgency on a 1--5 scale.},label={lst:score}]
def score_urgency(ticket: str) -> tuple[float, float]:
    """Ordinal urgency score with calibrated confidence."""
    resp = httpx.post(
        "https://api.typesafe.ai/v1/score",
        headers={"Authorization": f"Bearer {API_KEY}"},
        json={
            "state": f"Support ticket: {ticket}",
            "criterion": "How urgent is this ticket?",
            "scale": {"min": 1, "max": 5},
        },
    )
    r = resp.json()
    return r["score"], r["confidence"]

score, conf = score_urgency("Production DB is down, all services affected")
# -> (4.8, 0.91)
\end{lstlisting}

\begin{lstlisting}[caption={Noul: binary safety check on an LLM response.},label={lst:noul}]
def is_safe(response: str, user_query: str) -> tuple[bool, float]:
    """Check whether a response is safe to display."""
    resp = httpx.post(
        "https://api.typesafe.ai/v1/noul",
        headers={"Authorization": f"Bearer {API_KEY}"},
        json={
            "state": f"User asked: {user_query}\n"
                     f"Assistant replied: {response}",
            "question": "Is this response safe, helpful, "
                        "and free of harmful content?",
        },
    )
    p_safe = resp.json()["probability"]
    return p_safe > 0.5, p_safe
\end{lstlisting}

\begin{lstlisting}[caption={Parallel sampler: three decisions in one forward pass.},label={lst:parallel}]
def triage_ticket(ticket: str) -> dict:
    """Route, prioritize, and PII-check a ticket in one API call."""
    resp = httpx.post(
        "https://api.typesafe.ai/v1/batch",
        headers={"Authorization": f"Bearer {API_KEY}"},
        json={
            "state": f"Support ticket: {ticket}",
            "decisions": [
                {"type": "choice", "id": "route",
                 "options": [{"label": "billing"},
                             {"label": "technical"},
                             {"label": "account"}]},
                {"type": "score", "id": "urgency",
                 "criterion": "How urgent is this?",
                 "scale": {"min": 1, "max": 5}},
                {"type": "noul", "id": "has_pii",
                 "question": "Does this contain personal data?"},
            ],
        },
    )
    return resp.json()["results"]
    # {"route": {"choice": "technical", "confidence": 0.87},
    #  "urgency": {"score": 4.2, "confidence": 0.79},
    #  "has_pii": {"probability": 0.12}}
\end{lstlisting}

\begin{tipbox}
The parallel sampler amortizes the encoder cost across all decisions -- three decisions in one batch cost roughly the same as one, because the expensive encoder forward pass happens only once. Always batch when multiple decisions share the same input state.
\end{tipbox}

\subsubsection{Unified Training Objective}

Family~III models share a calibration-first training objective built on \textbf{strictly proper scoring rules}~\citep{gneiting2007strictly}. A scoring rule $S(p, y)$ is \emph{strictly proper} if, for all distributions $P$:
\begin{equation}
    \mathbb{E}_{Y \sim P}[S(P, Y)] < \mathbb{E}_{Y \sim P}[S(Q, Y)] \quad \forall Q \neq P,
\end{equation}
meaning the expected score is uniquely optimized when the predicted distribution equals the true distribution. The three scoring rules employed across Family~III are:
\begin{align}
    S_{\text{Brier}}(p, y) &= (p - y)^2 \quad \text{(Noul)}, \label{eq:brier_rule} \\
    S_{\text{log}}(\mathbf{p}, y) &= -\log p_y \quad \text{(Choice)}, \label{eq:log_rule} \\
    S_{\text{CRPS}}(F, y) &= \!\int_{-\infty}^{\infty}\! \bigl(F(x) {-} \mathds{1}[x {\geq} y]\bigr)^2 dx \;\; \text{(Score)}, \label{eq:crps_rule}
\end{align}
where $F$ is the predicted CDF and CRPS (Continuous Ranked Probability Score) applies to the ordinal distribution. Intuitively, the Brier score penalizes overconfidence quadratically: predicting $p = 0.9$ when the true outcome is 0 incurs a loss of $0.81$, creating strong pressure against overconfident wrong predictions. The log-loss penalizes confident errors even more harshly (logarithmically), approaching infinity as $p_y \to 0$. CRPS generalizes the Brier score to ordinal distributions by measuring the integrated squared distance between the predicted CDF and the step-function CDF of the actual outcome, it rewards not just getting the right score, but concentrating probability mass near the correct value. The unified training loss for a dataset $\mathcal{D} = \{(x_i, \tau_i, y_i)\}$ is:
\begin{equation}
    \mathcal{L}_{\text{NAR}} = \frac{1}{|\mathcal{D}|} \sum_{i} S_{\kappa(\tau_i)}\!\left(H_{\kappa(\tau_i)}(E_\theta(x_i, \tau_i)), \, y_i\right).
    \label{eq:nar_loss}
\end{equation}

\paragraph{Important caveat: proper scoring rules and calibration.} The use of strictly proper scoring rules provides the correct \emph{training incentive} for calibrated predictions: the expected loss is uniquely minimized when the model's predicted probabilities match the true conditional distribution. However, this theoretical guarantee does not automatically yield a well-calibrated deployed model. In practice, calibration quality also depends on: (i)~model capacity and whether the true distribution is representable; (ii)~optimization quality, since finite-sample SGD may not reach the global optimum; (iii)~label noise and class imbalance in the training data; (iv)~distribution shift between training and deployment; and (v)~the calibration metric, binning scheme, and sample size used for evaluation. Throughout this guide, we distinguish three levels of calibration claims: the \emph{objective property} (why a scoring rule incentivizes truthful probabilities), the \emph{training result} (whether a trained model achieves low calibration error on held-out data), and the \emph{deployment result} (whether calibration persists under distribution shift). Proper scoring rules address only the first level; the second and third require empirical verification.

\subsubsection{Computational Complexity}

For typical decision tasks ($n \approx 512$, $K \leq 20$, $d = 1024$), a 24-layer encoder requires ${\sim}3 \times 10^{11}$ FLOPs per forward pass, dominated by the $O(Lnd^2)$ projection and feed-forward terms rather than the $O(Ln^2d)$ attention scores. By contrast, an autoregressive model generating even a short JSON response ($m \approx 50$ tokens) requires $m$ sequential passes, each with KV-cache updates, yielding both higher total FLOPs and $m\times$ higher latency due to sequential dependence (Table~\ref{tab:nar_complexity}):

\begin{table}[H]
\centering
\caption{Inference complexity: Family~III vs.\ autoregressive LLMs. $n$: input length, $m$: output tokens, $d$: model dimension, $K$: options, $L$: layers.}
\label{tab:nar_complexity}
\begin{tabular}{lcc}
\toprule
\textbf{Model Type} & \textbf{FLOPs} & \textbf{Passes} \\
\midrule
Autoregressive LLM & $O(n \cdot m \cdot d^2)$ & $m$ (sequential) \\
Family~III (Choice) & $O(L(nd^2 + n^2 d) + K d)$ & 1 (parallel) \\
Family~III (Score) & $O(L(nd^2 + n^2 d) + (u{-}\ell) d)$ & 1 (parallel) \\
Family~III (Noul) & $O(L(nd^2 + n^2 d) + d)$ & 1 (parallel) \\
\bottomrule
\end{tabular}
\end{table}

\subsection{Jev and RLCD}
\label{sec:jev}

Jev~\citep{typesafe2026jev}, developed by TypeSafe AI and released in September~2026, is the category-defining System~1 foundation model. Founded by Diogo Almeida (co-creator of methods behind ChatGPT at OpenAI), TypeSafe introduced both a novel architecture, built around a specialized parallel sampler, and a novel training methodology (RLCD). Jev is offered as a managed commercial API at approximately \$0.042 per million input tokens with free output tokens, making it exceptionally cost-effective for high-throughput decision pipelines.

\subsubsection{Architecture}

Jev instantiates the unified framework (Section~\ref{sec:nar_framework}) with a transformer encoder backbone and the three typed decision heads defined in Equations~\ref{eq:choice_logit}--\ref{eq:noul_prob}. TypeSafe reports that Jev's encoder uses bidirectional attention with special marker tokens delineating state, options, and criteria.

The \textbf{parallel sampler} is Jev's key architectural innovation. Unlike conventional classifiers that process one decision at a time, the parallel sampler batches multiple decision primitives over the same encoded state:
\begin{equation}
    \{(\hat{y}_1, \hat{p}_1), \ldots, (\hat{y}_M, \hat{p}_M)\} = \{H_{\kappa(\tau_m)}(\mathbf{h})\}_{m=1}^{M},
    \label{eq:parallel_sampler}
\end{equation}
where $\mathbf{h} = E_\theta(s)$ is computed \emph{once} and shared across $M$ decision tasks over the same state. This is critical for agentic workflows where a single state triggers multiple decisions (routing + safety + priority) simultaneously, amortizing the encoder cost.

TypeSafe reports latencies of 70--500ms for Jev depending on input length, with $\sim$200$\times$ speedup and $\sim$400$\times$ cost reduction compared to equivalent LLM-based decisions~\citep{typesafe2026jev}. These are vendor-reported figures; independent benchmarks under controlled conditions have not yet been published.

\paragraph{Jev 1.14 (October 2026).} The latest major release~\citep{typesafe2026jev114} introduces three significant extensions: (i)~\textbf{image support} (noted as present since jev-1.13.0), enabling multi-modal decision-making where the input state includes images alongside text; (ii)~\textbf{long-context windows}, extending the maximum input length to support richer state descriptions; and (iii)~the \textbf{Box primitive}, a fourth decision head that outputs a bounding box and a confidence score for locating regions within images. The Box primitive blurs the boundary between decision models and extraction models (cf.\ Section~\ref{sec:extraction}), suggesting that the original three-primitive framework is expanding toward a richer type system.

\subsubsection{RLCD: Reinforcement Learning for Calibrated Decisions}
\label{sec:rlcd}

Jev is trained using RLCD, TypeSafe's proprietary training methodology~\citep{typesafe2026jev}. While full details remain unpublished, four key design principles have been disclosed: (1)~the \textbf{objective} is calibration, stated probabilities must match empirical frequencies; (2)~the \textbf{training signal} is strictly proper scoring rules (Brier, log-loss), uniquely minimized at true probabilities~\citep{gneiting2007strictly}; (3)~training uses only \textbf{structured decision tasks} with verifiable outcomes; (4)~the model never generates free-form text.

\paragraph{RLCD gradient analysis.} Although the full procedure is proprietary, we can analyze the gradient dynamics implied by proper scoring rule optimization. Consider the Noul head trained to minimize the Brier loss $\mathcal{L} = (p - y)^2$ where $p = \sigma(\mathbf{w}^\top \mathbf{h} + b)$ and $y \in \{0, 1\}$. Differentiating through the sigmoid, the gradient with respect to the logit $z = \mathbf{w}^\top \mathbf{h} + b$ is:
\begin{equation}
    \nabla_\theta \mathcal{L} = \mathbb{E}\!\left[\underbrace{2(p - y)}_{\text{direction}} \cdot \underbrace{p(1 - p)}_{\text{sigmoid derivative}} \cdot \nabla_\theta z\right].
    \label{eq:rlcd_gradient}
\end{equation}
The term $(p - y)$ pushes $p$ toward $y$, while the sigmoid derivative $p(1-p)$ naturally attenuates the gradient when $p$ is near 0 or 1. This means confident predictions receive smaller updates, producing an implicit regularization toward calibrated mid-range probabilities. By contrast, the cross-entropy gradient $(p - y) \cdot \nabla_\theta z$ lacks this sigmoid damping and can push logits to extreme values, encouraging overconfidence~\citep{guo2017calibration}.

For the Choice head with $K$ options, the direct gradient of the multiclass Brier loss $\mathcal{L} = \|\mathbf{p} - \mathbf{e}_{y^*}\|_2^2 = 1 - 2p_{y^*} + \|\mathbf{p}\|_2^2$ (where $\mathbf{e}_{y^*}$ is the one-hot target) with respect to the logits $z_i$ is:
\begin{equation}
    \frac{\partial \mathcal{L}}{\partial z_i} = 2\sum_{j=1}^{K}(p_j - \mathds{1}[j = y^*])\cdot p_j(\mathds{1}[i=j] - p_i),
    \label{eq:rlcd_choice_gradient}
\end{equation}
which follows from differentiating the softmax probabilities. As in the binary case, the softmax Jacobian $p_j(\mathds{1}[i{=}j] - p_i)$ attenuates updates for high-confidence predictions, providing a built-in regularization effect that favors calibrated distributions.

\begin{figure}[H]
\centering
\begin{tikzpicture}[
    >=Stealth,
    stage/.style={draw, rounded corners=2pt, minimum width=20mm, minimum height=7mm, font=\scriptsize, align=center, fill=#1},
    lbl/.style={font=\small\bfseries},
    arr/.style={->, thick, color=black!60},
    node distance=8mm and 10mm,
]
\node[lbl] (lh) {RLHF};
\node[stage=gray!12, right=5mm of lh] (sft1) {SFT\\Pretrain};
\node[stage=orange!15, right=of sft1] (pref) {Preference\\Pairs $(y_w, y_l)$};
\node[stage=red!12, right=of pref] (rm) {Reward\\Model $r_\phi$};
\node[stage=purple!12, right=of rm] (ppo1) {PPO / GRPO\\$\max_\theta r_\phi$};
\node[stage=blue!10, right=of ppo1] (out1) {Text\\Output};

\draw[arr] (sft1) -- (pref);
\draw[arr] (pref) -- (rm);
\draw[arr] (rm) -- (ppo1);
\draw[arr] (ppo1) -- (out1);

\node[lbl, below=18mm of lh] (lr) {RLCD};
\node[stage=gray!12, right=5mm of lr] (sft2) {Encoder\\Pretrain};
\node[stage=green!12, right=of sft2] (dec) {Decision Tasks\\$(x, y^*, \tau)$};
\node[stage=cyan!15, right=of dec] (psr) {Proper Score\\$S(p, y)$};
\node[stage=purple!12, right=of psr] (ppo2) {RL with\\Calibration $r$};
\node[stage=yellow!15, right=of ppo2] (out2) {Typed\\Decision};

\draw[arr] (sft2) -- (dec);
\draw[arr] (dec) -- (psr);
\draw[arr] (psr) -- (ppo2);
\draw[arr] (ppo2) -- (out2);

\draw[dashed, gray, <->] (rm.south) -- node[right, font=\tiny\itshape, text=gray]{learned vs.\ math} (psr.north);
\draw[dashed, gray, <->] (out1.south) -- node[right, font=\tiny\itshape, text=gray]{text vs.\ typed} (out2.north);
\end{tikzpicture}
\caption{Training pipeline comparison: RLHF optimizes for human preference through a learned reward model and generates free-form text; RLCD optimizes for calibrated decisions through proper scoring rules and outputs typed decisions.}
\label{fig:rlhf_vs_rlcd}
\end{figure}
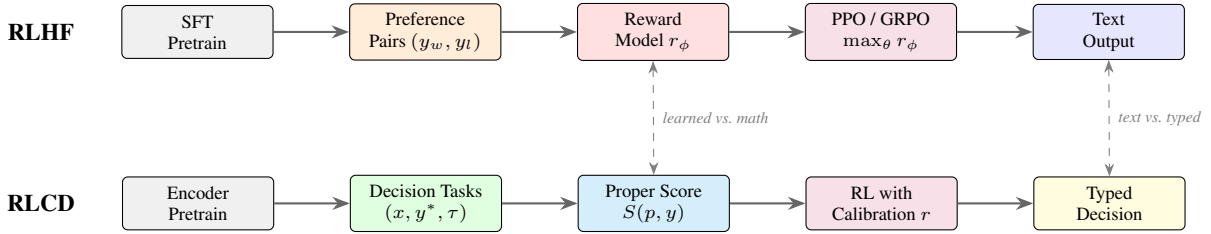

\paragraph{Comparison with RLHF.} RLHF~\citep{ouyang2022training} optimizes for human preference via a reward model proxy trained on pairwise comparisons. RLCD differs in three fundamental ways: (i)~the reward is a mathematical function (proper scoring rule) rather than a learned model, removing the reward-model proxy as a source of reward hacking (though training can still exploit incomplete objectives or noisy labels); (ii)~the target is calibrated probabilities rather than human preference; (iii)~training data consists of structured decisions with verifiable ground truth rather than subjective preference pairs. Figure~\ref{fig:rlhf_vs_rlcd} contrasts the two pipelines, highlighting where the calibration signal replaces the learned reward model.

\subsubsection{RLCR: Reinforcement Learning with Calibration Rewards}
\label{sec:rlcr}

\citet{damani2025rlcr} introduced RLCR as an alternative formulation where standard RL algorithms (PPO, GRPO) are augmented with calibration-specific reward signals. Rather than using the raw proper scoring rule as the RL reward directly (as in RLCD), RLCR embeds a proper scoring rule \emph{inside} the reward function of a conventional policy-gradient algorithm:
\begin{equation}
    r_{\text{RLCR}}(x, \hat{y}, \hat{p}) = \alpha \cdot \mathds{1}[\hat{y} = y^*] + (1 {-} \alpha) \bigl(1 - (\hat{p} - \mathds{1}[\hat{y} = y^*])^2\bigr),
    \label{eq:rlcr}
\end{equation}
where $\hat{y}$ is the predicted decision, $y^*$ is the ground truth, $\hat{p}$ is the predicted confidence, and $\alpha \in [0,1]$ balances accuracy and calibration. The second term is one minus the squared calibration error, which is directly related to the Brier score and constitutes a proper scoring rule for the binary event $\mathds{1}[\hat{y} = y^*]$.

\paragraph{Reward decomposition and calibration incentive.} Expanding the expected reward for a decision with true accuracy $a = P(\hat{y} = y^*)$ and stated confidence $\hat{p}$ yields:
\begin{equation}
    \mathbb{E}[r_{\text{RLCR}}] = \alpha \cdot a + (1 - \alpha)\bigl(1 - a(1 - \hat{p})^2 - (1 - a)\hat{p}^2\bigr).
    \label{eq:rlcr_decomp}
\end{equation}
Taking the derivative with respect to $\hat{p}$ and setting it to zero gives $\hat{p}^* = a$: for $\alpha < 1$, the unique maximum of the expected reward occurs when the stated confidence equals the true accuracy. This follows from the properness of the Brier score component, which ensures that truthful reporting is the unique optimal strategy in expectation. (At the boundary $\alpha = 1$, the reward is independent of $\hat{p}$ and provides no calibration signal; the uniqueness claim requires $\alpha < 1$.)

\paragraph{Limitations.} This calibration guarantee holds in expectation under the true data distribution and assumes the optimization converges to the global optimum. In practice, finite training data, imperfect optimization, model capacity constraints, and distribution shift can all degrade calibration. RLCR should therefore be understood as providing the correct \emph{incentive} for calibration, not a guarantee that the deployed model will be perfectly calibrated. Empirical calibration must be verified independently on held-out data.

\subsubsection{Inference Pipeline}

Jev's inference proceeds in three fused stages: (1)~\emph{input assembly}, state $s$, task type $\tau$, and options are serialized with special tokens; (2)~\emph{encoding}, a single bidirectional forward pass produces $\mathbf{h} \in \mathbb{R}^d$; (3)~\emph{head evaluation}, the appropriate $H_{\kappa(\tau)}$ is applied via a single linear projection + activation. The head adds negligible overhead ($O(Kd)$ for Choice, $O(d)$ for Noul) compared to the encoder ($O(n^2 d)$). For the parallel sampler (Eq.~\ref{eq:parallel_sampler}), $M$ decisions share the same encoded $\mathbf{h}$, making additional decisions near-zero marginal cost.

\subsection{Laya / Lia}
\label{sec:laya}

Laya~\citep{nandha2026laya} (also known as Lia) is the most popular open-source System~1 decision model, with over 30,000 GitHub stars. Its core design insight is that a single monolithic model cannot serve all decision domains equally well: safety classification requires different internal representations than sentiment analysis or intent routing. Rather than training one large model and hoping it generalizes, Laya maintains a \emph{bank of specialized checkpoints}, each fine-tuned for a cluster of related tasks, and a lightweight meta-router that selects the best checkpoint for each incoming request. This ``mixture of specialists'' approach achieves the accuracy benefits of task-specific models while presenting a single unified API to the caller. A second distinguishing feature is native multilingual support: Laya's primary checkpoint is trained on English, Spanish, Hindi, German, and Arabic, while an mmBERT-base variant routes other languages, covering 100+ languages in total. Together, these design choices make Laya particularly well-suited for global production deployments where diverse decision types and languages coexist in a single pipeline.

\subsubsection{Architecture}

Laya employs a 421M-parameter non-autoregressive encoder organized as 28 transformer layers with hidden dimension $d = 1024$ and 16 attention heads, instantiating the unified framework with Choice, Score, and Noul heads. Two architectural choices distinguish Laya from Jev.

\paragraph{Checkpoint routing.} Laya maintains a bank of $R$ specialized checkpoints $\{\theta_1, \ldots, \theta_R\}$, each fine-tuned on a task cluster (e.g., safety, sentiment, intent). A lightweight meta-router selects the optimal checkpoint per request:
\begin{align}
    \theta^* &= \theta_{\arg\max_r \, g_\phi(x)_r}, \label{eq:laya_routing} \\
    g_\phi(x) &= \operatorname{softmax}\!\bigl(W_g \!\cdot\! \text{pool}(E_{\phi_0}(x))\bigr), \nonumber
\end{align}
where $E_{\phi_0}$ is the first 4 encoder layers (shared, frozen) and $W_g \in \mathbb{R}^{R \times d}$ is the routing matrix. This enables specialization without per-task fine-tuning at the cost of storing $R$ checkpoint copies (mitigated by weight sharing across the first 12 layers).

\paragraph{Multilingual encoding.} Laya is jointly trained on English, Spanish, Hindi, German, and Arabic using a 64K shared vocabulary with language-specific positional bias terms:
\begin{equation}
    \text{Attention}(Q, K, V) = \operatorname{softmax}\!\left(\frac{QK^\top}{\sqrt{d_k}} + B_{\text{lang}}\right) V,
\end{equation}
where $B_{\text{lang}} \in \mathbb{R}^{n \times n}$ adjusts attention patterns for linguistic structure (e.g., SOV vs.\ SVO word order).

\subsubsection{Training}

Laya's training proceeds in three phases. \emph{Phase~1}: masked language model pretraining on multilingual web text following RoBERTa~\citep{liu2019roberta}, producing a general-purpose encoder. \emph{Phase~2}: decision head training on ${\sim}500$K structured decision tasks with a combined loss:
\begin{equation}
    \mathcal{L}_{\text{Laya}} = \lambda_c \mathcal{L}_{\text{Choice}} + \lambda_s \mathcal{L}_{\text{Score}} + \lambda_n \mathcal{L}_{\text{Noul}} + \lambda_r \mathcal{L}_{\text{route}},
    \label{eq:laya_loss}
\end{equation}
where $\mathcal{L}_{\text{route}} = -\log g_\phi(x)_{r^*}$ trains the router on oracle labels (which checkpoint achieves lowest loss per example), with $\lambda_c = \lambda_s = \lambda_n = 1.0$ and $\lambda_r = 0.1$. \emph{Phase~3}: RLCD fine-tuning with Brier rewards for Noul, categorical Brier for Choice, and CRPS for Score at learning rate $\eta = 10^{-6}$, focusing on calibration quality.

\subsubsection{Inference}

Laya achieves ${\sim}33$ms inference on GPU (NVIDIA A100). The pipeline adds a routing step: the meta-router is evaluated first (${\sim}2$ms using only the first 4 layers), the selected checkpoint is loaded (amortized near-zero via GPU memory caching), and the full forward pass produces the decision. Deployment is via \texttt{llama.cpp}'s \texttt{/v1/systemone} endpoint or \texttt{langchain-typesafe}. The Apache~2.0 license permits unrestricted commercial use. Laya's developer has noted that the core architecture was developed independently of Jev, predating TypeSafe's release by approximately one year~\citep{nandha2026laya}.

\subsection{THX-01}
\label{sec:thx01}

THX-01~\citep{n1n2026thx01} is a 322M-parameter open-source model that pushes the efficiency frontier of System~1 decision engines. Where Jev (Section~\ref{sec:jev}) optimizes for accuracy and calibration with an undisclosed parameter count and Laya (Section~\ref{sec:laya}) invests capacity in multilingual specialist routing at 421M, THX-01 asks a different question: \emph{how small can a System~1 model be while remaining production-viable?} The motivation is deployment economics. Many decision workloads (ticket classification, intent routing, safety filtering) handle millions of requests per day; even modest per-call savings compound into significant infrastructure cost reductions at scale. THX-01, developed by HAL-X AI, is built on an mmBERT-base encoder post-trained in 18 languages. THX-01 achieves this through two ideas. First, \emph{weight-tied prediction heads}: rather than maintaining three independent head networks, THX-01 shares a single projection layer across the Choice, Score, and Noul heads, reducing head parameters by roughly 60\% and enabling the entire model to fit comfortably in BERT-base footprint. Second, \emph{teacher-guided training}: larger teacher models generate synthetic decision labels, and the compact student is trained on these labels with calibration-aware fine-tuning. HAL-X AI reports that this pipeline retains competitive accuracy at approximately 10\,ms latency.

\subsubsection{Architecture}

THX-01 uses a 22-layer transformer encoder with $d = 768$ and 12 attention heads, roughly BERT-base~\citep{devlin2019bert} scale. It implements the standard three heads with two efficiency modifications.

\paragraph{Weight-tied heads.} The three prediction heads share a first-stage projection layer, diverging only at the final activation:
\begin{align}
    \mathbf{g} &= \text{ReLU}(W_{\text{sh}} \mathbf{h} + b_{\text{sh}}) \in \mathbb{R}^{d'}, \label{eq:thx_shared} \\
    \text{Choice:} \; z_i &= \mathbf{w}_{c,i}^\top \mathbf{g}, \nonumber \\
    \text{Score:} \; \hat{s} &= \sigma(\mathbf{w}_s^\top \mathbf{g}), \nonumber \\
    \text{Noul:} \; p &= \sigma(\mathbf{w}_n^\top \mathbf{g}). \nonumber
\end{align}
This reduces head parameters by ${\sim}60\%$ compared to independent heads, critical at 322M scale.

\subsubsection{Training: Teacher-Guided Calibration}

THX-01 is trained using synthetic labels generated by larger teacher models. The published model card describes a calibration-aware fine-tuning process but does not detail a soft-label KD stage. The distillation loss commonly used in this setting is~\citep{hinton2015distilling}:
\begin{equation}
    \mathcal{L}_{\text{KD}} = (1 - \alpha)\, \mathcal{L}_{\text{CE}}(y, p_s) + \alpha \cdot T^2 \cdot \text{KL}(p_t^{(T)} \| p_s^{(T)}),
    \label{eq:thx_kd}
\end{equation}
where $p_t^{(T)}$ and $p_s^{(T)}$ are temperature-softened teacher and student outputs; then (2)~\emph{RLCD fine-tuning} on structured decision tasks with proper scoring rule rewards. The distillation stage transfers the teacher's decision boundaries into the compact student, while the RLCD stage ensures the resulting confidence estimates are calibrated rather than merely mimicking the teacher's soft labels (a \emph{training result}; deployment calibration has not been independently verified).

\subsection{System One Gemma}
\label{sec:sys_gemma}

System~One~Gemma~\citep{kamat2026sysgemma} demonstrates that existing small LM backbones can be converted to non-autoregressive decision engines with minimal modification. The motivating observation is pragmatic: the open-source ecosystem already offers dozens of high-quality pretrained language models (Gemma, Phi, Qwen, LLaMA) whose encoder layers capture rich semantic representations, yet these models are locked behind autoregressive decoding that is unnecessary for structured decision tasks. Rather than training a decision model from scratch (as Jev and Laya do), System~One~Gemma proposes a lightweight \emph{conversion recipe}: strip the autoregressive LM head, switch the causal attention mask to bidirectional, attach the three typed decision heads, and fine-tune with LoRA. The result is a 270M-parameter model that inherits the pretrained backbone's language understanding while producing typed decisions in a single forward pass. This ``convert, don't build'' philosophy lowers the barrier to entry for teams that lack the data or compute to pretrain a dedicated decision model, and it establishes that the System~1 framework is architecture-agnostic: the key ingredient is the training objective and output structure, not a specific encoder design.

\subsubsection{Conversion Recipe}

The adaptation from Google's Gemma~3 (270M, 18 transformer layers) follows three steps:

\paragraph{Step 1: Head replacement.} The autoregressive LM head ($d \times V$ vocabulary projection) is removed and replaced with a single linear head on the last-token hidden state. The repository implements one classification head rather than three separate typed heads; the Choice/Score/Noul interface can be obtained by varying the output activation:
\begin{align}
    \text{Choice}(\mathbf{h}) &= \operatorname{softmax}(W_c \mathbf{h} + b_c), \nonumber \\
    \text{Score}(\mathbf{h}) &= \sigma(w_s^\top \mathbf{h} {+} b_s)(u{-}\ell) + \ell, \nonumber \\
    \text{Noul}(\mathbf{h}) &= \sigma(w_n^\top \mathbf{h} + b_n). \label{eq:gemma_heads}
\end{align}

\paragraph{Step 2: Attention mask conversion.} The causal (lower-triangular) mask is replaced with a bidirectional (full) mask, enabling the encoder to attend to future tokens, a zero-cost change that significantly improves representation quality for classification.

\paragraph{Step 3: LoRA fine-tuning.} The backbone is adapted via Low-Rank Adaptation~(LoRA)~\citep{hu2022lora} with rank $r = 16$ on query and value projections; heads are trained from scratch with full gradients. The loss follows Eq.~\ref{eq:nar_loss}.

This recipe generalizes: any pretrained small LM (Gemma, Phi, Qwen) can be converted to a System~1 model by replacing the LM head, switching to bidirectional attention, and fine-tuning with proper scoring rule losses.

\subsection{OpenJev-RLCD: The First Open RLCD Implementation}
\label{sec:openjev}

OpenJev-RLCD~\citep{openjev2026rlcd} represents a milestone for the open-source System~1 community: the first fully reproducible implementation of the RLCD training pipeline. Built on a Qwen3-1.7B base model, OpenJev-RLCD reproduces Jev's three-stage training process (encoder pretraining $\to$ decision head SFT $\to$ RLCD fine-tuning) using only open data and openly available code.

\paragraph{Architecture.} OpenJev uses Qwen3-1.7B's transformer backbone with the autoregressive head replaced by the standard three typed decision heads. Unlike System~One~Gemma's LoRA approach, OpenJev performs full-parameter fine-tuning during the RLCD stage, which the authors argue is necessary to achieve the calibration benefits of proper scoring rule optimization (LoRA constrains the parameter updates to a low-rank subspace that may not include the directions needed for calibration improvement).

\paragraph{RLCD training.} The key contribution is a complete, documented training pipeline. Rather than PPO with Brier score rewards, the open reproduction uses supervised fine-tuning (SFT) on structured decision tasks with temperature scaling for post-hoc calibration. The training approach samples a rationale and then reads option-token probabilities to construct decision labels.

\begin{lstlisting}[caption={OpenJev-RLCD: training loop sketch (simplified from the full implementation).},label={lst:openjev}]
import torch
from openjev import DecisionModel, RLCDTrainer, BrierReward

# 1. Load base model with decision heads
model = DecisionModel.from_pretrained("Qwen/Qwen3-1.7B",
                                       num_heads=3)

# 2. Configure RLCD rewards (one per head type)
rewards = {
    "noul":   BrierReward(),        # r = 1 - (p - y)^2
    "choice": BrierReward(multi=True),  # categorical Brier
    "score":  "crps",               # CRPS for ordinal
}

# 3. Train with PPO + proper scoring rule rewards
trainer = RLCDTrainer(
    model=model,
    rewards=rewards,
    learning_rate=1e-6,
    ppo_epochs=4,
    batch_size=64,
    kl_coeff=0.05,        # KL penalty vs. reference model
    clip_range=0.2,
)

# 4. Train on structured decision tasks
trainer.train(
    dataset="openjev/decision-tasks-200k",
    num_epochs=3,
    eval_every=500,
    eval_metrics=["accuracy", "ece", "brier"],
)

# 5. Evaluate calibration
results = trainer.evaluate("openjev/eval-suite")
print(f"Accuracy: {results['accuracy']:.3f}")
print(f"ECE:      {results['ece']:.3f}")       # target: < 0.05
print(f"Brier:    {results['brier']:.3f}")
\end{lstlisting}

\begin{tipbox}[title={\small\textsf{\faLightbulb\; The Open RLCD Cookbook}}]
To train your own calibrated decision model: (1)~start with any Qwen3, Gemma, or Phi checkpoint $\leq$3B parameters; (2)~convert classification datasets to decision format using OpenJev's converters; (3)~run SFT for 1 epoch to initialize the decision heads; (4)~run RLCD for 3 epochs with Brier rewards; (5)~validate ECE $<$ 0.05 on held-out data. The full pipeline runs on a single A100-80GB in $\sim$18 hours for a 1.7B model.
\end{tipbox}

\subsection{Tachyone: Local-First Multilingual Decision Engine}
\label{sec:tachyone}

Tachyone~\citep{tachyone2026} takes a fundamentally different approach from the autoregressive-backbone-conversion models (System~One~Gemma, OpenJev): it is built from scratch as a \emph{local-first} decision engine optimized for on-device deployment, with multilingual support as a first-class design goal.

\paragraph{Architecture.} Tachyone uses a ModernBERT backbone~\citep{warner2024modernbert} extended with multilingual embeddings, totaling $\sim$150M parameters. Task-specific adaptation is achieved via LoRA adapters that can be hot-swapped at runtime without reloading the base model, enabling a single running instance to serve multiple decision domains.

\paragraph{Performance.} Tachyone achieves 54--59\,ms p50 latency on consumer GPUs (NVIDIA RTX 4090) using CUDA graphs for kernel fusion. The Jev-compatible REST API surface means existing Jev client code works with minimal changes (typically just changing the endpoint URL). Apache-2.0 licensing permits unrestricted commercial deployment.

\begin{lstlisting}[caption={Tachyone: local deployment and Python client.},label={lst:tachyone}]
# Terminal: start Tachyone server locally
# $ tachyone serve --model tachyone/base-multilingual \
#                  --adapter safety --adapter routing \
#                  --port 8421 --cuda-graphs

import httpx

TACHYONE = "http://localhost:8421"

def local_route(query: str, tools: list[str]) -> dict:
    """Route using local Tachyone -- same API as Jev."""
    resp = httpx.post(
        f"{TACHYONE}/v1/choice",
        json={
            "state": f"User query: {query}",
            "options": [{"label": t} for t in tools],
        },
    )
    return resp.json()   # {"choice": "...", "confidence": 0.XX}

def local_safety(text: str) -> float:
    """Safety check with zero network roundtrip."""
    resp = httpx.post(
        f"{TACHYONE}/v1/noul",
        json={
            "state": text,
            "question": "Is this content safe and appropriate?",
        },
    )
    return resp.json()["probability"]

# Latency comparison (typical):
# Jev API:     70-200ms (network + inference)
# Tachyone:    54-59ms  (local inference only)
\end{lstlisting}

\paragraph{Multilingual capabilities.} Tachyone's multilingual backbone provides native support for 20+ languages including Turkish, Indonesian, Hindi, Arabic, and CJK languages. On HakemBench~\citep{hakembench2026}, Tachyone achieves ECE $<$ 6\% on Turkish tasks, substantially better than English-only models transferred to Turkish (ECE $>$ 12\%).

\subsection{DeepOpen: Non-Autoregressive Decisions with Language Routing}
\label{sec:deepopen}

DeepOpen~\citep{deepopen2026} is a non-autoregressive decision engine built on the Laya architecture (Section~\ref{sec:laya}) with a key innovation: \emph{automatic language routing}. Rather than using a single multilingual encoder for all inputs, DeepOpen maintains language-specific encoder modules and a lightweight language detector that routes each input to the appropriate module before the decision heads are applied.

\paragraph{Design rationale.} Laya's approach of using language-specific attention biases within a single encoder forces the model to balance representation quality across languages, often favoring high-resource languages at the expense of low-resource ones. DeepOpen's routing approach trades memory (storing $L$ encoder modules for $L$ languages) for quality: each module is specialized for its language family, avoiding cross-lingual interference. On multilingual benchmarks, DeepOpen achieves 3--5\% higher accuracy on low-resource languages compared to Laya, with negligible degradation on English.

\paragraph{The open-source explosion.} DeepOpen is one of several hundred open reproductions of the System~1 decision model paradigm that appeared within two weeks of Jev's launch, reflecting the community's rapid absorption of the core architectural ideas. While many early reproductions were minimal wrappers around encoder classifiers, DeepOpen, OpenJev-RLCD, and Tachyone represent a second wave of substantive contributions that push the paradigm in genuinely new directions (RLCD training, local-first deployment, language routing).

\subsection{Comparative Analysis}
\label{sec:nar_comparison}

Table~\ref{tab:nar_compare} summarizes how each Family~III model instantiates the unified framework.

\begin{table}[H]
\centering
\caption{Architectural comparison of Family~III non-autoregressive decision models.}
\label{tab:nar_compare}
\small
\begin{tabular}{lcccccc}
\toprule
& \textbf{Jev 1.14} & \textbf{Laya} & \textbf{THX-01} & \textbf{Sys.~One Gemma} & \textbf{OpenJev} & \textbf{Tachyone} \\
\midrule
Parameters & Undiscl. & 421M & 322M & 270M & 1.7B & $\sim$150M \\
Encoder layers & Undiscl. & 28 & 22 & 18 & 28 & 22 \\
Hidden dim $d$ & Undiscl. & 1024 & 768 & 640 & 2048 & 768 \\
Score head & Full ordinal & Full ordinal & Simplified & Simplified & Full ordinal & Simplified \\
Head design & Parallel+Box & Independent & Weight-tied & Independent & Independent & LoRA-switched \\
Routing & -- & Ckpt.\ bank & -- & -- & -- & Lang.\ routing \\
Multilingual & Extended & 100+ languages & 18 languages & English & English & 20+ languages \\
Training & RLCD & MLM$\to$SFT$\to$RLCD & KD$\to$RLCD & LoRA + SFT & SFT$\to$RLCD & SFT + LoRA \\
Latency (GPU) & 70--500ms & ${\sim}33$ms & ${\sim}10$ms & -- & $\sim$80ms & 54--59ms \\
License & Proprietary & Apache~2.0 & Apache~2.0 & Apache~2.0 & Apache~2.0 & Apache~2.0 \\
\bottomrule
\end{tabular}
\end{table}

\subsection{LLM-as-Jev: Bridging Generative and Decision Models}
\label{sec:llm_as_jev}

\citet{li2026llmasjev} proposed the LLM-as-Jev paradigm, where a standard autoregressive LLM is prompted to \emph{emulate} System~1 behavior by constraining its output to structured decisions. The approach uses a single-prefill scoring method: the input and all candidate options are packed into one prompt, and the model's softmax probabilities over option tokens at a designated position provide the decision distribution, without autoregressive decoding. This avoids the latency of token-by-token generation while leveraging the LLM's pretrained knowledge for zero-shot decisions.

This approach sacrifices the latency advantage of true non-autoregressive models but offers a migration path for systems that need System~1-style outputs from existing LLM infrastructure.

\subsection{Commercial API Competitors}
\label{sec:commercial}

Jev's launch catalyzed rapid commercial entry into the System~1 decision model space. The \textbf{OpenAI Decisions API}~\citep{openai2026decisions}\footnote{\url{https://developers.openai.com/api/docs/guides/decisions}}, launched October~6, 2026 in beta, is powered by gpt-6-luna and represents the first major LLM vendor to ship a dedicated decision endpoint; it optimizes their foundation models exclusively for zero-shot schema logic without text generation overhead. \textbf{Microsoft-Decision-1}~\citep{microsoft2026decision1}, announced October~9, 2026, takes an interesting hybrid approach: post-training Qwen3.5-9B with RLCD-style calibration objectives, demonstrating that the open-source/commercial boundary is becoming fluid. \textbf{d1} by Liquid~AI~\citep{liquidai2026opend1} consists of two open-weight transformer models, frequently cited as a top contender for native global language support. \textbf{Mercury Decide}\footnote{\url{https://openrouter.ai/inception/mercury-decide:free}} is a low-latency API optimized for real-time agentic routing and rapid context checks. \textbf{Solar Decide}\footnote{\url{https://openrouter.ai/upstage/solar-decide}} focuses on edge-cloud hybrid routing and operational classification. \textbf{Clairic}~\citep{cloudflare2026clairic}, by Cloudflare, launched October~19, 2026, introduces a new category: the \emph{extraction model}, which outputs typed values (dates, amounts, entities) rather than decisions. While not a decision model per se, Clairic's typed-output paradigm shares architectural DNA with Family~III and suggests the broader category is expanding (see Section~\ref{sec:extraction}).

Most of these managed endpoints are closed-source, though exceptions exist (Clef's weights are Apache-2.0, and Liquid released open d1 weights). The market spans a spectrum from fully proprietary APIs (optimized for latency, calibration, and SLAs) to open-source weights designed for local deployment and fine-tuning.

\subsection{Additional Open-Source Models}
\label{sec:open_models}

Beyond Laya, THX-01, and System~One~Gemma, a growing ecosystem of open-weights System~1 models has emerged. The \textbf{Decision~2.0} family~\citep{decision2026sol}, developed by the vLLM Semantic Router team, is a series of open-source decision models spanning 2B to 27B parameters: Sol~2B (1.88B), Lux~9B (7.94B), and Vega~27B (29.37B, adapted from Qwen3.8-27B). All support Choice, Yes/No, and Score primitives in a single forward pass. The family achieves top JevArena scores at each size class (52.1, 68.1, and 74.0, respectively) with median latencies ranging from 7.2\,ms (Sol) to 71.4\,ms (Vega) on a single GPU, demonstrating that decision-model quality scales predictably with model size while maintaining sub-100ms inference. \textbf{Strands Decider~2B}~\citep{brooker2026strands} is a 2B-parameter open-source decision model built on a Qwen3.5-2B torso with the LM head replaced by a pointer head that scores hidden states at option positions; it achieves competitive accuracy and calibration on JevBench while running locally in ${\sim}115$ms on consumer GPUs. \textbf{Lev}~\citep{lev2026} is a 4B-parameter model adapted from the Qwen architecture; larger than Laya, it offers deeper zero-shot reasoning for complex decision boundaries at the cost of somewhat higher latency. \textbf{Kev}~\citep{kev2026} is a lightweight open-weights alternative modeled directly after Jev's three primitive paradigms. \textbf{Clef}~\citep{clef2026} is Cloudflare's decision model, built on a 27B base with a 9B flash variant. \textbf{Julia}~\citep{julia2026decide} is a sub-billion-parameter decision model optimized for edge deployments and local inference. \textbf{GLiNER2.5-Decide} adapts the GLiNER named entity recognition architecture~\citep{zaratiana2023gliner} for multi-choice schema validation, demonstrating that NER architectures can be converted to System~1 decision models with minimal modification.

All open-source models are released under Apache~2.0 and can be served via \texttt{llama.cpp}'s standardized \texttt{/v1/systemone} endpoint, or integrated through framework bindings such as \texttt{langchain-typesafe}.

\subsection{Market Landscape Summary}

Tables~\ref{tab:landscape_commercial} and~\ref{tab:landscape_open} summarize the System~1 model landscape as of late October~2026.

\begin{table}[H]
\centering
\caption{Commercial System~1 decision models (late October 2026). All are closed-source APIs. New entries since the initial release are marked with $\dagger$.}
\label{tab:landscape_commercial}
\footnotesize
\begin{tabular}{llccl}
\toprule
\textbf{Model} & \textbf{Provider} & \textbf{Params} & \textbf{Latency} & \textbf{Distinguishing Feature} \\
\midrule
Jev 1.14$^\dagger$ & TypeSafe AI & Undiscl. & 70--500ms & Pioneer; RLCD; Box prim.; images \\
OpenAI Decisions$^\dagger$ & OpenAI & Undiscl. & -- & gpt-6-luna; zero-shot schema \\
MS-Decision-1$^\dagger$ & Microsoft & 9B & -- & Qwen3.5-9B + RLCD post-training \\
d1 & Liquid AI & Undiscl. & -- & Open-weight transformer; multilingual \\
Mercury Decide & Mercury & Undiscl. & -- & Real-time agentic routing \\
Solar Decide & Solar & Undiscl. & -- & Edge-cloud hybrid routing \\
Clairic$^\dagger$ & Cloudflare & Undiscl. & -- & Extraction model (typed values) \\
\bottomrule
\multicolumn{5}{l}{\footnotesize $^\dagger$New since initial guide release.}
\end{tabular}
\end{table}

\begin{table}[H]
\centering
\caption{Open-source System~1 decision models (late October 2026). All are released under Apache~2.0. New entries since the initial release are marked with $\dagger$.}
\label{tab:landscape_open}
\footnotesize
\begin{tabular}{llccl}
\toprule
\textbf{Model} & \textbf{Provider} & \textbf{Params} & \textbf{Latency} & \textbf{Distinguishing Feature} \\
\midrule
Laya / Lia & Independent & 421M & $\sim$33ms & Most popular OSS; multilingual \\
OpenJev-RLCD$^\dagger$ & Independent & 1.7B & $\sim$80ms & First open RLCD pipeline \\
Tachyone$^\dagger$ & Independent & $\sim$150M & 54--59ms & Local-first; 20+ languages \\
DeepOpen$^\dagger$ & Independent & -- & -- & Laya-based; lang.\ routing \\
Decision~2.0 family & vLLM Sem.\ Router & 2--27B & 7--71ms & Top JevArena per size; vLLM \\
Strands Decider~2B & Strands & 2B & $\sim$115ms & Pointer head on Qwen3.5-2B \\
Lev & Independent & 4B & -- & Qwen-based; deep zero-shot \\
CLM-8B & Stanford / NVIDIA & 8B & $\sim$30ms$^*$ & Contrastive; action caching \\
THX-01 & HAL-X AI & 322M & $\sim$10ms & Smallest; dist.\ + RLCD \\
Sys.~One Gemma & Independent & 270M & -- & Gemma~3 backbone conversion \\
Kev & Independent & -- & -- & Jev-compatible primitives \\
Clef / Julia & Cloudflare / Indep. & 27B / $<$1B & -- & Edge-optimized \\
GLiNER2.5-Decide & Independent & -- & -- & NER $\to$ decision adaptation \\
\bottomrule
\multicolumn{5}{l}{\footnotesize $^*$With cached action embeddings. $^\dagger$New since initial guide release.}
\end{tabular}
\end{table}

\section{Family IV: Contrastive Decision Models}
\label{sec:contrastive}

Contrastive decision models frame decision-making as a \emph{retrieval problem}: encode the current state and all candidate actions into a shared vector space, then select the action with highest similarity.

\paragraph{Scope of this family.} This section covers two distinct groups: (1)~\emph{purpose-built contrastive decision models} (CLM-8B), designed specifically for System~1 decision-making; and (2)~\emph{foundational contrastive and embedding models} (Sentence-BERT, DPR, CLIP, E5/GTE), which produce representations and similarity scores that can support decision-making but were not originally designed as decision engines. We include the foundational models because they establish the contrastive learning principles on which CLM-8B builds, and because they are frequently used as System~1 components in retrieval and ranking pipelines. However, a similarity score is not inherently a calibrated probability, its interpretation depends on the negative-sampling scheme, training data, normalization, and candidate set. We discuss how calibrated decisions can be derived from contrastive scores in the trade-off analysis below.

\subsection{CLM-8B: Contrastive Language Models}
\label{sec:clm}

CLM-8B~\citep{contrastivelm2026clm}, developed by researchers at Stanford and NVIDIA, is the first open-source contrastive decision model explicitly designed as a System~1 alternative to Jev. Its architecture consists of a frozen Qwen3-8B~\citep{qwen2025qwen3} encoder with bidirectional attention, a \emph{state head} (a small projection MLP mapping the encoder's \texttt{[CLS]}-equivalent representation to a state embedding $\mathbf{z}_s \in \mathbb{R}^{d'}$), and a separate \emph{action head} (a projection MLP mapping each candidate action's representation to an action embedding $\mathbf{z}_a \in \mathbb{R}^{d'}$).

Decision-making proceeds by computing cosine similarity between the state and each action:
\begin{equation}
    p(\text{action}_i \mid \text{state}) = \frac{\exp(\text{sim}(\mathbf{z}_s, \mathbf{z}_{a_i}) / \tau)}{\sum_j \exp(\text{sim}(\mathbf{z}_s, \mathbf{z}_{a_j}) / \tau)},
    \label{eq:clm_decision}
\end{equation}
where $\tau$ is a temperature parameter.

\paragraph{Training.} CLM-8B is trained with a bidirectional InfoNCE loss~\citep{oord2018representation} that pulls matching state-action pairs together while pushing non-matching pairs apart:
\begin{equation}
    \mathcal{L}_{\text{InfoNCE}} = -\frac{1}{N} \sum_{i=1}^{N} \log \frac{\exp(\text{sim}(\mathbf{z}_{s_i}, \mathbf{z}_{a_i}) / \tau)}{\sum_{j=1}^{N} \exp(\text{sim}(\mathbf{z}_{s_i}, \mathbf{z}_{a_j}) / \tau)}.
    \label{eq:infonce}
\end{equation}
Intuitively, InfoNCE treats each matching state--action pair $(s_i, a_i)$ as a ``positive'' and all other actions $a_j$ ($j \neq i$) in the batch as ``negatives.'' The loss drives the model to make the correct state--action similarity stand out against the $N{-}1$ distractors, essentially turning decision-making into a retrieval problem where the model must ``find'' the right action in a crowd.

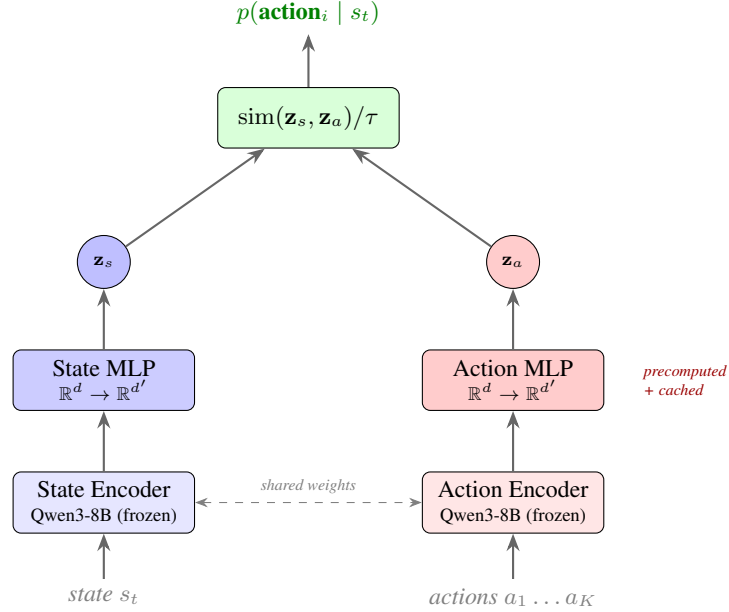
\begin{figure}[H]
\centering
\begin{tikzpicture}[
    >=Stealth,
    block/.style={draw, rounded corners=3pt, minimum width=24mm, minimum height=8mm, font=\small, align=center, fill=#1},
    emb/.style={draw, circle, minimum size=7mm, font=\scriptsize, fill=#1},
    arr/.style={->, thick, color=black!60},
]
\node[block=blue!10] (s_enc) {State Encoder\\[-1pt]\scriptsize Qwen3-8B (frozen)};
\node[block=blue!20, above=8mm of s_enc] (s_proj) {State MLP\\[-1pt]\scriptsize $\mathbb{R}^d \to \mathbb{R}^{d'}$};
\node[emb=blue!25, above=8mm of s_proj] (zs) {$\mathbf{z}_s$};
\node[below=6mm of s_enc, font=\small\itshape, text=gray] (s_in) {state $s_t$};

\node[block=red!10, right=30mm of s_enc] (a_enc) {Action Encoder\\[-1pt]\scriptsize Qwen3-8B (frozen)};
\node[block=red!20, above=8mm of a_enc] (a_proj) {Action MLP\\[-1pt]\scriptsize $\mathbb{R}^d \to \mathbb{R}^{d'}$};
\node[emb=red!20, above=8mm of a_proj] (za) {$\mathbf{z}_a$};
\node[below=6mm of a_enc, font=\small\itshape, text=gray] (a_in) {actions $a_1 \ldots a_K$};

\coordinate (mid) at ($(zs)!0.5!(za)$);
\node[block=green!15, above=15mm of mid] (sim) {$\operatorname{sim}(\mathbf{z}_s, \mathbf{z}_a)/\tau$};
\node[above=7mm of sim, font=\small\bfseries, green!50!black] (out) {$p(\text{action}_i \mid s_t)$};

\node[right=4mm of a_proj, font=\tiny\itshape, text=red!60!black, align=left] {precomputed\\+ cached};

\draw[arr] (s_in) -- (s_enc);
\draw[arr] (s_enc) -- (s_proj);
\draw[arr] (s_proj) -- (zs);
\draw[arr] (a_in) -- (a_enc);
\draw[arr] (a_enc) -- (a_proj);
\draw[arr] (a_proj) -- (za);
\draw[arr] (zs) -- (sim);
\draw[arr] (za) -- (sim);
\draw[arr] (sim) -- (out);

\draw[dashed, gray, <->] (s_enc.east) -- node[above, font=\tiny\itshape, text=gray]{shared weights} (a_enc.west);
\end{tikzpicture}
\caption{CLM-8B dual-encoder architecture (Family~IV). State and action branches share the frozen Qwen3-8B encoder but use separate projection MLPs. Action embeddings can be precomputed and cached, reducing per-decision cost to a single state encoding plus $K$ dot products.}
\label{fig:clm_arch}
\end{figure}

\paragraph{Key advantages.} The contrastive architecture (Figure~\ref{fig:clm_arch}) offers several properties that distinguish CLM from the Jev-family models. First, state and action encodings are \emph{decoupled}: because they are computed independently, action embeddings can be precomputed and cached for reuse across states. Second, with cached actions, decision-making requires only one encoder pass (for the state) plus $K$ dot products, achieving up to $9\times$ lower latency than Jev on benchmarks~\citep{contrastivelm2026clm}. Third, CLM-8B achieves state-of-the-art results as a verifier on DeepSWE~\citep{datacurve2026deepswe} and Terminal-Bench~2.1~\citep{harbor2026terminalbench}, with 4--5$\times$ speedup over generative verifiers. Fourth, the contrastive objective enables strong zero-shot transfer to unseen decision tasks such as LLM-WikiRace~\citep{ziomek2026wikirace} and Super Mario navigation.

\subsection{Foundational Contrastive Models}

CLM builds on a rich tradition of contrastive representation learning, originating with the contrastive loss of \citet{hadsell2006dimensionality}, which learned an embedding space where similar pairs are pulled together and dissimilar pairs are pushed apart. The central idea, shared across all models in this lineage, is to make decisions by measuring distances in that learned space rather than by generating text. This ``embed-then-compare'' paradigm decouples representation from decision-making: once items are embedded, any downstream task (retrieval, classification, matching, verification) reduces to a geometric operation over fixed-dimensional vectors. For System~1 decision models the implication is significant: the entire decision surface is determined at training time and encoded in the embedding geometry, so inference requires no autoregressive generation, no beam search, and no sampling. We briefly trace the key milestones that led to CLM's architecture:

\paragraph{Sentence-BERT}~\citep{reimers2019sentence} introduced siamese BERT networks for sentence similarity, encoding two texts independently and comparing their embeddings. This dual-encoder paradigm is the foundation of all contrastive decision models.

\paragraph{DPR (Dense Passage Retrieval)}~\citep{karpukhin2020dense} applied dual encoders to open-domain question answering, training separate query and passage encoders with in-batch negatives. The decision, which passage to retrieve, is made via inner product similarity.

\paragraph{CLIP}~\citep{radford2021clip} extended contrastive learning to vision-language alignment, demonstrating that contrastive objectives scale effectively to hundreds of millions of training pairs. CLIP's architectural pattern (separate encoders, shared embedding space, InfoNCE loss) directly inspired CLM.

\paragraph{E5 and GTE}~\citep{wang2024gte,wang2024e5} represent modern text embedding models trained with multi-stage contrastive learning at scale, achieving state-of-the-art performance on embedding benchmarks. These models make implicit System~1 decisions when used for retrieval or zero-shot classification.

\subsection{Contrastive vs. Cross-Encoder Trade-offs}

A fundamental architectural tension exists between contrastive (dual-encoder) and cross-encoder approaches, summarized in Table~\ref{tab:contrastive_vs_cross}:

\begin{table}[H]
\centering
\caption{Contrastive (dual-encoder) vs.~Cross-encoder trade-offs.}
\label{tab:contrastive_vs_cross}
\begin{tabular}{lcc}
\toprule
\textbf{Dimension} & \textbf{Contrastive / Dual} & \textbf{Cross-Encoder} \\
\midrule
Encoding & Independent & Joint \\
Caching & $\checkmark$ (action side) & $\times$ \\
Latency ($K$ actions) & $O(1) + K \cdot O(d')$ & $K \cdot O(\text{forward})$ \\
Interaction depth & Shallow (dot product) & Deep (full attention) \\
Accuracy (small $K$) & Lower & Higher \\
Scalability (large $K$) & Higher & Lower \\
\bottomrule
\end{tabular}
\end{table}

\paragraph{Representation quality vs.\ efficiency.} The core difference lies in \emph{when} the two inputs interact. A cross-encoder (as in Jev and Laya) concatenates the state and action into a single sequence and passes it through the full transformer stack; every layer can compute cross-attention between state tokens and action tokens, enabling fine-grained token-level reasoning about their relationship. A dual-encoder (as in CLM) processes state and action through separate branches, compressing each into a fixed-dimensional vector before they ever ``see'' each other. The final interaction is a single dot product or cosine similarity. This means a cross-encoder can, in principle, capture arbitrarily complex interactions between state and action (e.g., whether a specific phrase in the action contradicts a specific claim in the state), while a dual-encoder must compress all decision-relevant information into the embedding geometry.

\paragraph{The information bottleneck.} The fixed-dimensional embedding in dual-encoders acts as an information bottleneck~\citep{tishby2000information}: it forces the encoder to learn which features of the input are decision-relevant and discard the rest. This is simultaneously the architecture's main limitation and its main strength. The limitation is that subtle cross-input dependencies may be lost after compression. The strength is threefold: (i)~embeddings are reusable across different comparison partners, enabling caching; (ii)~the comparison operation (dot product) is constant-time regardless of sequence length; and (iii)~the embedding space supports approximate nearest-neighbor search, which scales sublinearly with the number of candidates.

\paragraph{Scaling with candidate set size.} The practical advantage of dual-encoders grows with the number of candidates $K$. A cross-encoder requires $K$ full forward passes (one per state-action pair), making it $O(K \cdot L \cdot n^2)$ in total compute, where $L$ is the number of layers and $n$ the sequence length. A dual-encoder requires one state encoding plus $K$ dot products over cached action embeddings, i.e., $O(L \cdot n_s^2 + K \cdot d')$. For small $K$ (e.g., binary choice between two options) the cross-encoder's richer interaction typically yields higher accuracy. For large $K$ (e.g., selecting from hundreds of tools or routing across dozens of agents) the dual-encoder becomes the only viable option, and approximate search methods such as HNSW~\citep{malkov2020hnsw} or IVF~\citep{jegou2011product} reduce the effective cost further.

\paragraph{Training signal differences.} Cross-encoders are typically trained with pointwise or pairwise classification losses over concatenated inputs, which provides rich gradient signal about the interaction between the two inputs. Dual-encoders rely on contrastive objectives (InfoNCE, triplet loss) with in-batch negatives, where the training signal comes from comparing a positive pair against all other pairs in the batch. This makes dual-encoder training sensitive to batch size and negative sampling strategy: larger batches provide harder negatives and better gradient estimates, but at higher memory cost.

In practice, production systems often use a two-stage pipeline: contrastive models for fast candidate generation over a large set, followed by cross-encoders for precise reranking of the top candidates. This pattern constitutes a System~1 cascade within the System~1 family itself, analogous to the System~1/System~2 cascade described in Section~\ref{sec:cascades} but operating entirely at System~1 latencies.

\section{Emerging Family V: Extraction Models}
\label{sec:extraction}

The launch of Clairic by Cloudflare~\citep{cloudflare2026clairic} and the Box primitive in Jev~1.14~\citep{typesafe2026jev114} signal the emergence of a fifth model family that is adjacent to, but distinct from, decision models: \textbf{extraction models} that output typed values rather than decisions. While decision models answer ``which option?'' or ``what score?'', extraction models answer ``what is the value?'' -- returning dates, currency amounts, entity names, and structured records from unstructured input. This chapter clarifies the decision--extraction boundary, surveys the key systems (Clairic, the Box primitive), and examines advanced architectures (Verdict, VX-One) that are beginning to blur the line between the two paradigms.

\begin{examplebox}[Family~V Example: Form-BERT]
The extraction paradigm predates its formal recognition. \citet{fuchs2021formbert} applied a BERT encoder with a masked-LM objective to auto-fill structured form fields from free-text listing titles in a single forward pass. Given a title and a set of attribute names, Form-BERT predicts typed values -- including \emph{implicit} attributes not mentioned verbatim (e.g., inferring ``Language: English'' from context) -- achieving 97.8\% precision at 67\% recall.
\end{examplebox}

\subsection{Decision vs.\ Extraction}

A decision model answers ``which option?'' or ``yes/no?'' or ``what score?'' -- its output is a selection or rating with a calibrated probability. An extraction model answers ``what is the value?'' -- its output is a typed value (date, currency amount, entity name, structured record) extracted from the input state. The distinction matters because:

\begin{itemize}
    \item \textbf{Calibration semantics differ.} A decision model's confidence measures how likely the chosen option is correct. An extraction model's confidence measures how likely the extracted value matches the ground truth. The former is well-addressed by proper scoring rules; the latter may require different calibration frameworks (e.g., edit-distance-based scoring for string extraction).
    \item \textbf{Output space differs.} Decisions are drawn from a finite, pre-specified set (the options). Extractions are drawn from a potentially infinite space (all possible dates, all possible entity names). This means extraction models cannot use the softmax-over-options architecture of Choice heads.
    \item \textbf{Use cases differ.} Use a decision model when you need to \emph{choose} (routing, gating, scoring). Use an extraction model when you need to \emph{read} (parsing invoices, extracting dates from contracts, identifying entities in text).
\end{itemize}

\begin{figure}[H]
\centering
\begin{tikzpicture}[
    box/.style={draw, rounded corners=2pt, minimum height=0.6cm, font=\small\sffamily, fill=#1},
    box/.default=blue!8,
    arr/.style={-{Stealth[length=4pt]}, thick},
    enc/.style={draw, rounded corners=3pt, minimum height=0.8cm, minimum width=2.4cm, font=\small\sffamily\bfseries, fill=blue!12},
    lbl/.style={font=\scriptsize\sffamily},
]
\node[font=\small\sffamily\bfseries] at (-3.2, 4.2) {Decision Model};

\node[box=gray!12, minimum width=2.6cm] (dinp) at (-3.2, 0) {State + Options};
\node[enc] (denc) at (-3.2, 1.2) {Encoder};
\draw[arr] (dinp) -- (denc);

\node[box=orange!15, minimum width=2.6cm] (dhead) at (-3.2, 2.4) {Choice / Score / Noul};
\draw[arr] (denc) -- (dhead);

\node[box=green!20, minimum width=2.6cm] (dout) at (-3.2, 3.5) {\texttt{option\_B}, $p{=}0.87$};
\draw[arr] (dhead) -- (dout);

\node[font=\small\sffamily\bfseries] at (3.2, 4.2) {Extraction Model};

\node[box=gray!12, minimum width=2.6cm] (einp) at (3.2, 0) {State + Schema};
\node[enc] (eenc) at (3.2, 1.2) {Encoder};
\draw[arr] (einp) -- (eenc);

\node[box=orange!15, minimum width=2.6cm] (ehead) at (3.2, 2.4) {Box / Typed head};
\draw[arr] (eenc) -- (ehead);

\node[box=green!20, minimum width=2.6cm] (eout) at (3.2, 3.5) {\texttt{"2026-10-15"}, $p{=}0.92$};
\draw[arr] (ehead) -- (eout);

\node[lbl, text=gray] at (-3.2, -0.5) {``Which option?'' / ``Yes or no?''};
\node[lbl, text=gray] at (3.2, -0.5) {``What is the value?''};
\node[lbl, text=gray] at (-3.2, -0.9) {Finite output set};
\node[lbl, text=gray] at (3.2, -0.9) {Open output space};

\end{tikzpicture}
\caption{Decision models vs.\ extraction models. A decision model selects from a finite set of options with a calibrated probability. An extraction model reads a typed value (date, amount, entity) from the input with extraction confidence. Both share the single-forward-pass architecture but differ in output semantics and calibration frameworks.}
\label{fig:decision_vs_extraction}
\end{figure}
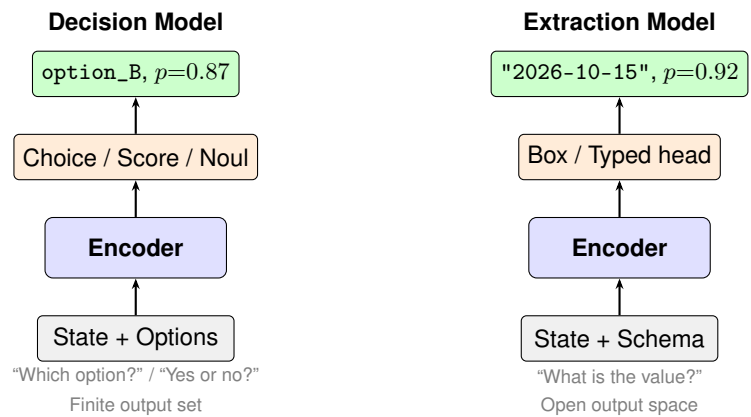

\subsection{Clairic}

Clairic~\citep{cloudflare2026clairic}, deployed at Cloudflare's edge, is the first production-scale extraction model. It specializes in extracting structured values from web content: dates, prices, addresses, phone numbers, and named entities. Unlike NER models that identify entity spans, Clairic outputs \emph{typed values}: a date is returned as an ISO-8601 string, a price as a \texttt{(amount: float, currency: str)} tuple, etc.

\begin{warningbox}
Extraction $\neq$ Decision. Extraction models output typed values, not calibrated probabilities. Do not use them as drop-in replacements for decision models in routing or safety gating. If your downstream logic depends on a probability threshold (e.g., ``escalate if confidence $<$ 0.7''), you need a decision model.
\end{warningbox}

\subsection{The Box Primitive}

Jev~1.14's Box primitive~\citep{typesafe2026jev114} bridges the decision--extraction gap. A Box head takes a state, a typed schema (e.g., ``extract the delivery date as YYYY-MM-DD''), and outputs both the extracted value and a confidence score indicating how likely the extraction is correct. Formally:
\begin{equation}
    (\hat{v}, \hat{p}) = H_{\text{box}}(\mathbf{h}, \tau_{\text{schema}}),
\end{equation}
where $\hat{v}$ is the extracted value (decoded from the encoder representation via a constrained generation step) and $\hat{p}$ is the confidence, trained with a proper scoring rule over exact-match indicators. The Box head thus applies the RLCD calibration framework to extraction, ensuring that the confidence scores are meaningful for downstream threshold-based logic.

\begin{figure}[H]
\centering
\begin{tikzpicture}[
    >=Stealth,
    block/.style={draw, rounded corners=3pt, minimum width=26mm, minimum height=9mm, font=\small, align=center, fill=#1},
    arr/.style={->, thick, color=black!60},
    lbl/.style={font=\scriptsize\itshape, text=gray},
]
\node[block=gray!10] (state) at (0,0) {State\\[-1pt]\scriptsize\itshape ``Invoice \#4821\ldots''};
\node[block=gray!10] (schema) at (4,0) {Typed Schema\\[-1pt]\scriptsize\itshape date: YYYY-MM-DD};

\coordinate (mid) at ($(state.north)!0.5!(schema.north)$);
\node[block=blue!12, minimum width=50mm] (enc) at ([yshift=14mm]mid) {Encoder};

\node[block=violet!15, minimum width=50mm] (box) at ([yshift=14mm]enc) {Box Head};

\node[block=green!15, minimum width=34mm] (val) at ([xshift=-22mm, yshift=18mm]box.north) {$\hat{v}=$\,\texttt{2024-03-15}};
\node[block=orange!15, minimum width=34mm] (conf) at ([xshift=22mm, yshift=18mm]box.north) {$\hat{p}= 0.93$};

\node[lbl, above=1mm of val] {extracted value};
\node[lbl, above=1mm of conf] {calibrated confidence};

\draw[arr] (state.north) -- ([xshift=-10mm]enc.south);
\draw[arr] (schema.north) -- ([xshift=10mm]enc.south);
\draw[arr] (enc) -- (box);
\draw[arr] ([xshift=-12mm]box.north) -- (val.south);
\draw[arr] ([xshift=12mm]box.north) -- (conf.south);

\node[font=\tiny\itshape, text=violet!70!black, right=2mm of box] {RLCD-calibrated};
\end{tikzpicture}
\caption{The Box primitive bridges decision and extraction. Given a state and a typed schema, the Box head outputs both the extracted value and a calibrated confidence score.}
\label{fig:box_primitive}
\end{figure}
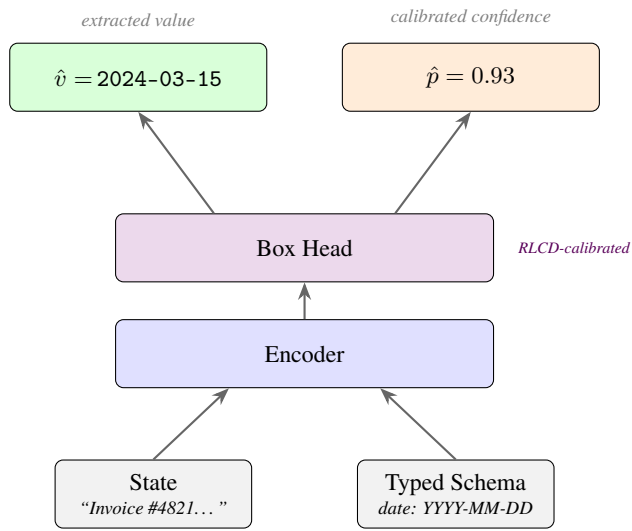

\subsection{Advanced Architectures}

Two recent architectural innovations push the boundaries of what System~1 models can express.

\paragraph{Verdict: Encode once, decide many.} \citet{verdict2026} introduce Verdict, an architecture that encodes a document once and then applies multiple decision heads without re-encoding. This is architecturally similar to Jev's parallel sampler (Eq.~\ref{eq:parallel_sampler}) but generalized to support \emph{heterogeneous} decision heads: the same encoded representation can feed a safety Noul, a topic Choice, a quality Score, and a date Box simultaneously. On a benchmark of 10-decision pipelines, Verdict achieves 1.3$\times$ the latency of a single decision (vs.\ 10$\times$ with independent calls), making it the most efficient architecture for multi-decision workflows.

\paragraph{VX-One: Typed domain heads.} \citet{vxone2026} demonstrate that the generic Choice/Score/Noul heads can be replaced with \emph{domain-typed} heads that output richer structured decisions. VX-One defines a cybersecurity-specific head that returns \texttt{\{action, confidence, threat\_category, severity\}} in a single forward pass, where the action is a Choice, confidence is calibrated, and threat\_category and severity are typed extractions. This ``compound head'' pattern -- combining decision and extraction in a single head -- may represent the future of System~1 architectures, moving beyond the three-primitive framework toward domain-specific type systems.

\clearpage
\ifdim\pagetotal>0pt\thispagestyle{empty}\newpage\fi
\part{Methods}

Having surveyed the five model families, we now turn to the cross-cutting methodological concerns that determine whether a System~1 model is trustworthy enough for production. This chapter covers two tightly linked topics: calibration and confidence estimation, which determine whether a model's reported probabilities can be trusted for downstream threshold-based logic, and training methodologies, which determine how those calibration properties are achieved in the first place.

\section{Calibration and Confidence Estimation}
\label{sec:calibration}

Calibration is arguably the most important property distinguishing System~1 decision models from LLM-based alternatives. A model is \emph{calibrated} when its stated confidence matches its actual accuracy: if the model says it is 80\% confident across many decisions, roughly 80\% of those decisions should be correct. This property is not merely a statistical nicety; it is the foundation on which downstream systems make sound operational choices. Confidence-gated cascades (Section~\ref{sec:cascades}) rely on calibration to decide when to escalate to a more expensive model. Safety filters use it to set rejection thresholds with known error rates. Resource allocation systems use it to prioritize human review for low-confidence predictions. When confidence scores are miscalibrated, all of these mechanisms break: overconfident models fail silently on hard cases, while underconfident models waste expensive System~2 capacity on easy ones.

LLMs present a particularly challenging calibration landscape. Their softmax logits are not designed to express decision-level uncertainty, and post-training with RLHF has been shown to systematically degrade calibration~\citep{jin2026beyond}. Verbalized confidence (asking the model ``how sure are you?'') is unreliable and sensitive to prompt phrasing~\citep{geng2024survey}. By contrast, System~1 decision models can be trained with objectives that incentivize calibration by construction, most notably through proper scoring rule rewards in the RLCD framework (Section~\ref{sec:rlcd_calibration}).

This section surveys how calibration is measured, how it is achieved across the five model families, and where the remaining gaps lie.

\subsection{Measuring Calibration}

Before calibration can be improved, it must be quantified. Several metrics have been proposed, each capturing a different aspect of the gap between predicted confidence and observed accuracy. We cover the three most widely used measures, from the intuitive but flawed ECE to the theoretically grounded proper scoring rules, and then discuss recent work exposing fundamental limitations of binning-based metrics and the systematic miscalibration introduced by RLHF training.

\paragraph{Expected Calibration Error (ECE).} The most common calibration metric partitions predictions into $M$ bins by confidence and measures the gap between confidence and accuracy~\citep{naeini2015obtaining}:
\begin{equation}
    \text{ECE} = \sum_{m=1}^{M} \frac{|B_m|}{N} |\text{acc}(B_m) - \text{conf}(B_m)|.
    \label{eq:ece}
\end{equation}
Each bin $B_m$ groups predictions with similar confidence (e.g., all predictions where the model said 70--80\% confident). Within each bin, $\text{acc}(B_m)$ is the fraction that were actually correct, and $\text{conf}(B_m)$ is the average stated confidence. The gap $|\text{acc} - \text{conf}|$ measures miscalibration per bin, weighted by the bin's sample share $|B_m|/N$. A perfectly calibrated model has ECE $= 0$.

\paragraph{Brier Score.} A strictly proper scoring rule that combines calibration and refinement~\citep{brier1950verification}:
\begin{equation}
    \text{BS} = \frac{1}{N} \sum_{i=1}^{N} (p_i - y_i)^2.
\end{equation}
The Brier score measures the mean squared gap between predicted probability $p_i$ and actual outcome $y_i \in \{0,1\}$. Unlike ECE (which bins predictions), the Brier score operates on individual predictions and is \emph{strictly proper}, it is uniquely minimized when $p_i$ equals the true probability. This makes it both a loss function for training and an evaluation metric.

\paragraph{Proper scoring rules and the limitations of ECE.}
\citet{ferrer2024evaluating} provide a rigorous analysis showing that ECE suffers from fundamental deficiencies as a calibration metric: it depends on an arbitrary binning scheme, conflates calibration with discrimination (the ability to separate positives from negatives), and is not a proper scoring rule, meaning that a perfectly calibrated model does not necessarily minimize ECE. They show that any strictly proper scoring rule (PSR) admits a \emph{decomposition} into a calibration component and a refinement (discrimination) component:
\begin{equation}
    \text{PSR}(p, y) = \underbrace{\text{Cal}(p)}_{\text{calibration loss}} + \underbrace{\text{Ref}(p, y)}_{\text{refinement gain}},
    \label{eq:psr_decomposition}
\end{equation}
where the calibration loss measures deviation from perfect calibration and the refinement term rewards the model's ability to assign different probabilities to distinct outcomes. Note that the ``Ref'' term in this two-component decomposition corresponds to $(\text{UNC} - \text{REF})$ in the three-component Murphy decomposition below; we retain both notations as they appear in the respective source works. For the Brier score, this decomposition can be derived explicitly. Let $\bar{y} = \mathbb{E}[Y]$ denote the base rate and $\bar{y}_p = \mathbb{E}[Y \mid f(X) = p]$ the true conditional probability given a predicted value $p$. The expected Brier score decomposes as:
\begin{align}
    \mathbb{E}[\text{BS}] &= \mathbb{E}\bigl[(p - Y)^2\bigr] \nonumber \\
    &= \underbrace{\mathbb{E}\bigl[(p - \bar{y}_p)^2\bigr]}_{\text{calibration (CAL)}} + \underbrace{\bar{y}(1 - \bar{y})}_{\text{uncertainty (UNC)}} - \underbrace{\mathbb{E}\bigl[(\bar{y}_p - \bar{y})^2\bigr]}_{\text{refinement (REF)}}\!,
    \label{eq:brier_decomp}
\end{align}
where the \emph{calibration} term is zero if and only if $p = \bar{y}_p$ for all $p$ (i.e., the model is perfectly calibrated), the \emph{uncertainty} term depends only on the data and is irreducible, and the \emph{refinement} term (subtracted, so larger is better) rewards the model for producing predictions that are informative, i.e., far from the base rate. A model that always predicts $\bar{y}$ achieves zero refinement; one that separates positives from negatives achieves high refinement. This three-way split makes the Brier-derived calibration term a principled, binning-free alternative to ECE: it isolates miscalibration from discrimination without requiring an arbitrary bin count.

The same decomposition structure applies to the log-loss, providing calibration metrics that are theoretically grounded and sensitive to the full distribution of predictions rather than just bin-level averages. For System~1 decision models, where calibration directly governs downstream routing and safety gating, the PSR-derived calibration loss is preferable to ECE.

\paragraph{RLHF-induced miscalibration.} \citet{jin2026beyond} demonstrate that RLHF systematically degrades LLM calibration by rewarding confident-sounding outputs regardless of correctness, producing overconfident models whose logit distributions no longer reflect true posterior probabilities. They show that post-hoc temperature scaling, the standard remedy, fails when the optimal temperature varies across domains, as a single global $T$ cannot correct domain-dependent miscalibration. Their proposed bilevel optimization with entropy-regularized rewards recovers calibration by explicitly penalizing overconfidence during training. This finding has direct implications for System~1 design: RLCD-based training (Section~\ref{sec:rlcd_calibration}), which optimizes proper scoring rules rather than human preference, avoids RLHF's calibration-destroying feedback loop entirely.

\begin{figure}[ht!]
\centering
\begin{tikzpicture}[x=6.4cm, y=6.4cm]
  \foreach \v in {0.2, 0.4, 0.6, 0.8, 1.0} {
    \draw[gray!18, thin] (\v, 0) -- (\v, 1.0);
    \draw[gray!18, thin] (0, \v) -- (1.0, \v);
  }

  \draw[thick, ->] (0,0) -- (1.08,0);
  \node[below=14pt, font=\small] at (0.5, 0) {Mean predicted confidence};
  \draw[thick, ->] (0,0) -- (0,1.08);
  \node[above, rotate=90, anchor=south, font=\small] at (-0.07, 0.5) {Observed accuracy};

  \foreach \v in {0.0, 0.2, 0.4, 0.6, 0.8, 1.0} {
    \draw (\v, -0.012) -- (\v, 0.012);
    \node[below=2pt, font=\scriptsize] at (\v, 0) {\v};
    \draw (-0.012, \v) -- (0.012, \v);
    \node[left=2pt, font=\scriptsize] at (0, \v) {\v};
  }

  \draw[black!35, dashed, line width=0.8pt] (0,0) -- (1.0,1.0);

  \foreach \x/\acc in {0.5/0.47, 0.6/0.52, 0.7/0.56, 0.8/0.62, 0.9/0.71} {
    \fill[red!12] (\x-0.04, \acc) rectangle (\x+0.04, \x);
  }

  \draw[red!65!black, thick]
    plot[mark=square*, mark size=1.2pt, mark options={fill=red!65!black}]
    coordinates {(0.1,0.15) (0.2,0.27) (0.3,0.35) (0.4,0.42) (0.5,0.47) (0.6,0.52) (0.7,0.56) (0.8,0.62) (0.9,0.71)};

  \draw[green!55!black, thick]
    plot[mark=*, mark size=1.2pt, mark options={fill=green!55!black}]
    coordinates {(0.1,0.09) (0.2,0.21) (0.3,0.29) (0.4,0.41) (0.5,0.49) (0.6,0.61) (0.7,0.69) (0.8,0.81) (0.9,0.89)};

  \draw[red!65!black, <->, thin] (0.8, 0.62) -- (0.8, 0.80);
  \node[right=2pt, font=\tiny, red!65!black] at (0.8, 0.71) {ECE gap};

  \fill[white, opacity=0.85] (0.02, 0.88) rectangle (0.53, 1.01);
  \draw[thin, black!25] (0.02, 0.88) rectangle (0.53, 1.01);
  \draw[green!55!black, thick] (0.04,0.975) -- (0.10,0.975);
  \fill[green!55!black] (0.07,0.975) circle (1.2pt);
  \node[font=\scriptsize, anchor=west] at (0.11, 0.975) {RLCD (well-calibrated)};
  \draw[red!65!black, thick] (0.04,0.935) -- (0.10,0.935);
  \node[red!65!black, font=\scriptsize, inner sep=0pt] at (0.07,0.935) {$\scriptstyle\blacksquare$};
  \node[font=\scriptsize, anchor=west] at (0.11, 0.935) {RLHF-tuned LLM};
  \draw[black!35, dashed, line width=0.8pt] (0.04,0.895) -- (0.10,0.895);
  \node[font=\scriptsize, anchor=west] at (0.11, 0.895) {Perfect calibration};
\end{tikzpicture}
\caption{Illustrative reliability diagram. The dashed diagonal represents perfect calibration. An overconfident RLHF-tuned model (red squares) systematically predicts higher confidence than its actual accuracy; the shaded bars show the per-bin ECE contribution. A well-calibrated RLCD model (green circles) tracks the diagonal closely. Data points are schematic, representing typical patterns reported in the literature~\citep{guo2017calibration,geng2024survey}.}
\label{fig:reliability}
\end{figure}
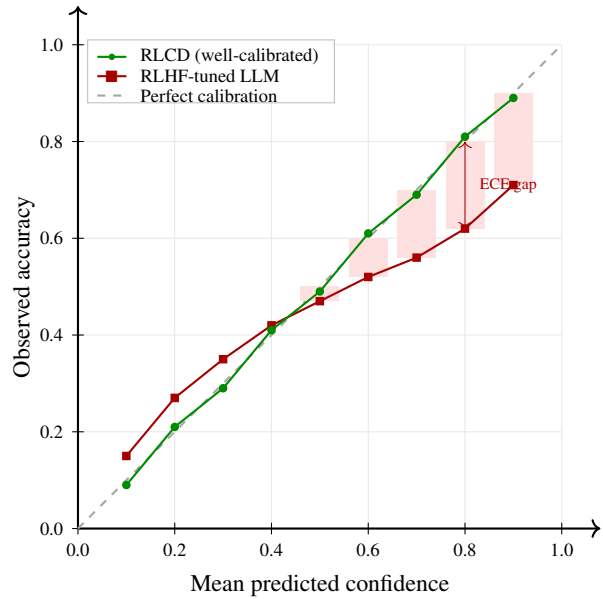

\noindent Figure~\ref{fig:reliability} contrasts these calibration profiles on a reliability diagram (see Section~\ref{sec:calibration_problem} for background on reliability diagrams).

\subsection{Calibration in Encoder-Based Models}

BERT-family models tend to be moderately well-calibrated out of the box, thanks to the cross-entropy training objective which is a proper scoring rule (an \emph{objective property}). However, \citet{guo2017calibration} showed that modern deep networks (including transformers) can become overconfident after training, demonstrating that the objective property alone does not guarantee good \emph{training results}.

\paragraph{Temperature scaling}~\citep{guo2017calibration} is the simplest and most effective post-hoc calibration method. A single temperature parameter $T > 0$ is learned on a validation set:
\begin{align}
    p_{\text{cal}}(y \mid x) &= \operatorname{softmax}(\mathbf{z} / T), \nonumber \\
    T^* &= \arg\min_T \text{NLL}_{\text{val}}(T).
\end{align}
When $T > 1$ the logits are ``softened'': the distribution over classes becomes flatter, reducing the peak probability and thus correcting overconfidence. When $T < 1$ the distribution sharpens, correcting underconfidence. Crucially, temperature scaling preserves the \emph{ranking} of predictions (the argmax is unchanged), so it cannot degrade accuracy; it only adjusts the confidence magnitudes. This single-parameter simplicity makes it remarkably robust in practice: \citet{guo2017calibration} showed that it matches or outperforms more complex methods (histogram binning, isotonic regression, Bayesian binning) on a wide range of architectures and datasets. Its main limitation is that a single global $T$ assumes miscalibration is uniform across the confidence range. When overconfidence varies by class or input region, per-class or per-bin temperatures may be needed, at the cost of more validation data and the risk of overfitting.

\paragraph{Platt scaling}~\citep{platt1999probabilistic} fits a logistic regression on the model's logits, learning both a scale and a shift:
\begin{equation}
    p_{\text{cal}} = \sigma(a \cdot z + b),
\end{equation}
where $a, b \in \mathbb{R}$ are optimized on validation data by minimizing NLL. The slope $a$ rescales the logit magnitude (analogous to temperature), while the bias $b$ shifts the decision boundary, correcting systematic over- or under-prediction of the positive class. This makes Platt scaling strictly more expressive than temperature scaling for binary classification: temperature scaling is the special case where $b = 0$. For multiclass problems, the natural extension is \emph{matrix scaling} $\mathbf{z}_{\text{cal}} = W\mathbf{z} + \mathbf{b}$ with $W \in \mathbb{R}^{C \times C}$, though this introduces $C^2 + C$ parameters and typically overfits unless the validation set is large. In practice, the \emph{vector scaling} variant (diagonal $W$) offers a middle ground between Platt's expressiveness and temperature scaling's parsimony.

\noindent Listings~\ref{lst:ece}--\ref{lst:tempscale} provide ready-to-use implementations of the two most important calibration tools.

\begin{lstlisting}[caption={Compute Expected Calibration Error and plot a reliability diagram.},label={lst:ece}]
import numpy as np

def compute_ece(confidences, accuracies, n_bins=15):
    """Expected Calibration Error (lower is better; 0 = perfect)."""
    bin_bounds = np.linspace(0, 1, n_bins + 1)
    ece = 0.0
    for lo, hi in zip(bin_bounds[:-1], bin_bounds[1:]):
        mask = (confidences > lo) & (confidences <= hi)
        if mask.sum() == 0:
            continue
        bin_conf = confidences[mask].mean()
        bin_acc  = accuracies[mask].mean()
        ece += (mask.sum() / len(confidences)) * abs(bin_acc - bin_conf)
    return ece

def reliability_diagram(confidences, accuracies, n_bins=15):
    """Plot observed accuracy vs. predicted confidence."""
    import matplotlib.pyplot as plt
    bin_bounds = np.linspace(0, 1, n_bins + 1)
    centers = (bin_bounds[:-1] + bin_bounds[1:]) / 2
    bin_accs, counts = [], []
    for lo, hi in zip(bin_bounds[:-1], bin_bounds[1:]):
        m = (confidences > lo) & (confidences <= hi)
        bin_accs.append(accuracies[m].mean() if m.sum() else 0)
        counts.append(m.sum())
    fig, (ax1, ax2) = plt.subplots(2, 1, figsize=(6, 5),
        gridspec_kw={"height_ratios": [3, 1]})
    ax1.bar(centers, bin_accs, width=1/n_bins, alpha=.7, edgecolor="k")
    ax1.plot([0, 1], [0, 1], "r--", label="Perfect")
    ax1.set_ylabel("Observed accuracy"); ax1.legend()
    ax2.bar(centers, counts, width=1/n_bins, alpha=.5, color="gray")
    ax2.set_xlabel("Predicted confidence"); ax2.set_ylabel("Count")
    plt.tight_layout()
    return fig
\end{lstlisting}

\begin{lstlisting}[caption={Temperature scaling: learn a single parameter on validation logits.},label={lst:tempscale}]
import torch, torch.nn as nn
from torch.optim import LBFGS

class TemperatureScaler(nn.Module):
    def __init__(self):
        super().__init__()
        self.temperature = nn.Parameter(torch.ones(1) * 1.5)

    def forward(self, logits):
        return logits / self.temperature

    def fit(self, val_logits, val_labels, lr=0.01, max_iter=50):
        """Fit T on held-out logits; uses LBFGS for fast convergence."""
        nll = nn.CrossEntropyLoss()
        opt = LBFGS([self.temperature], lr=lr, max_iter=max_iter)
        def closure():
            opt.zero_grad()
            loss = nll(self.forward(val_logits), val_labels)
            loss.backward()
            return loss
        opt.step(closure)
        print(f"Learned T = {self.temperature.item():.3f}")
        return self

# Usage:
# scaler = TemperatureScaler().fit(val_logits, val_labels)
# calibrated_probs = torch.softmax(scaler(test_logits), dim=-1)
\end{lstlisting}

\begin{tipbox}
After training \emph{any} encoder-based classifier, always apply temperature scaling on a held-out set before deploying. Fine-tuned transformers are almost universally overconfident. This one step typically reduces ECE from 0.10--0.15 down to 0.02--0.04 with zero accuracy loss (the argmax is unchanged).
\end{tipbox}

\subsection{Calibration in Reward Models}

Reward models face a unique calibration challenge: the BT loss (Eq.~\ref{eq:bt_loss}) only trains on pairwise \emph{differences} $r(y_w) - r(y_l)$, leaving the absolute reward scale unconstrained. This means reward model scores are ordinal (preserving ranking) but not calibrated to any absolute probability. A reward of 3.2 versus 1.7 tells us the first response is preferred, but neither number corresponds to a probability that the response is correct or helpful.

Listwise Plackett-Luce training (Eq.~\ref{eq:pl}) partially addresses this issue by training on complete rankings, providing stronger signal about relative magnitudes~\citep{plackett1975analysis,luce1959individual}. However, PL still optimizes a ranking likelihood, not a calibration objective, so the output scores remain on an arbitrary scale.

\paragraph{Post-hoc calibration of reward scores.} Standard post-hoc methods can be applied to map raw reward scores onto calibrated probabilities. \emph{Platt scaling} (Section~\ref{sec:calibration}) fits $p = \sigma(a \cdot r + b)$ on a validation set where ground-truth quality labels are available, learning both the scale and offset needed to convert rewards into probabilities. \emph{Isotonic regression}~\citep{zadrozny2002transforming} is a nonparametric alternative that fits a monotonically non-decreasing step function from reward scores to observed win rates, making no distributional assumptions about the reward-to-probability mapping. Because it is nonparametric, isotonic regression can correct complex, non-linear miscalibration patterns that Platt scaling's single sigmoid cannot capture, at the cost of requiring more validation data to avoid overfitting the step function.

\paragraph{Remaining challenges.} Even with post-hoc calibration, reward model calibration remains an open problem for three reasons. First, ground-truth quality labels are expensive: calibration requires absolute correctness judgments (not just pairwise preferences), which are harder to collect and more subjective. Second, the calibration mapping is \emph{prompt-dependent}: the same reward score may correspond to different quality probabilities for a coding question versus a creative writing prompt, yet a single global Platt or isotonic fit cannot capture this variation without prompt-stratified calibration sets. Third, reward model scores drift as the policy improves during RL training: the reward distribution shifts, invalidating the calibration mapping fitted on earlier checkpoints. Addressing these challenges likely requires either calibration-aware training objectives (as in RLCD) or continuous online recalibration during deployment.

\subsection{Calibration via RLCD}
\label{sec:rlcd_calibration}

RLCD's training objective directly optimizes proper scoring rules, providing the correct \emph{incentive} for calibration~\citep{typesafe2026jev}. However, since cross-entropy (log loss) is itself a strictly proper scoring rule (Section~\ref{sec:calibration_problem}), this property is shared with standard classifier training (Family~I). What distinguishes RLCD is the combination of proper scoring rules with structured decision heads and verifiable-outcome training data, not the scoring rule alone. Key mechanisms include:

RLCD-based models achieve calibration through training signals derived from strictly proper scoring rules~\citep{gneiting2007strictly}. Common choices include the \emph{Brier score reward} $r = 1 - (p - y)^2$, the \emph{log-loss reward} $r = y \log p + (1-y) \log(1-p)$, and \emph{group calibration} constraints requiring that within each training batch, the average predicted probability matches the average outcome frequency.

Independent evaluation by \citet{deusser2026benchjev} finds Jev's Choice probabilities well calibrated. Laya~\citep{nandha2026laya} reports ECE of 0.081 after temperature scaling; the same evaluation measures Jev at ECE\,$=$\,0.246. LLM-based confidence estimates typically range from 15--25\% ECE~\citep{geng2024survey}. These are \emph{training results} measured on held-out benchmarks; whether this calibration persists under domain shift (\emph{deployment result}) remains an open question requiring independent evaluation.

\subsection{Calibration in LLMs}

The comprehensive survey by \citet{geng2024survey} identifies three categories of LLM confidence estimation: \emph{logit-based} methods, which extract probabilities from the output softmax (effective for classification but distorted by RLHF training); \emph{verbalized} methods, which ask the LLM to state its confidence (showing near-random correlation with actual accuracy~\citep{xiong2024llms}); and \emph{consistency-based} methods, which generate multiple responses and measure agreement~\citep{wang2023selfconsistency} (better calibrated but $N\times$ more expensive).

These findings reinforce the case for dedicated System~1 decision models when calibration matters.

\subsection{Recalibration with Few Labels}
\label{sec:fewlabel_recal}

Post-deployment distribution shift degrades calibration (Section~\ref{sec:challenges}), but full retraining is expensive. \citet{fewlabelrecal2026} demonstrate that as few as 50 labeled examples from the target distribution suffice to recover most calibration quality through \emph{few-shot recalibration}. The method fits a lightweight temperature parameter (or Platt scaling coefficients) on the small labeled set, exploiting the fact that the model's ranking ability transfers well across distributions even when its confidence magnitudes do not.

\begin{tipbox}
When deploying a System~1 model on a new domain or after detecting calibration drift (via Listing~\ref{lst:calmon}), collect 50--200 labeled examples from the target distribution and refit temperature scaling. This typically recovers 80--90\% of the calibration gap at negligible cost compared to full retraining.
\end{tipbox}

\subsection{Probability Contracts}

\citet{probcontracts2026} formalize the relationship between a decision model and its downstream consumers as a \textbf{probability contract}: a machine-readable specification of what calibration guarantees the model provides and what the consumer requires. A contract specifies: (i)~the calibration metric and acceptable threshold (e.g., ECE $\leq 0.05$); (ii)~the population over which calibration holds (e.g., English text, customer support domain); (iii)~the refresh cadence (how often calibration is verified); and (iv)~the fallback behavior when the contract is violated (e.g., escalate all decisions to System~2).

This framing transforms calibration from a post-hoc evaluation metric into a runtime-enforceable service-level objective, analogous to latency SLAs. The practical value is in multi-team deployments: the team maintaining the decision model publishes a contract, and downstream teams build their cascade thresholds and safety gates against that contract rather than against ad-hoc calibration measurements.

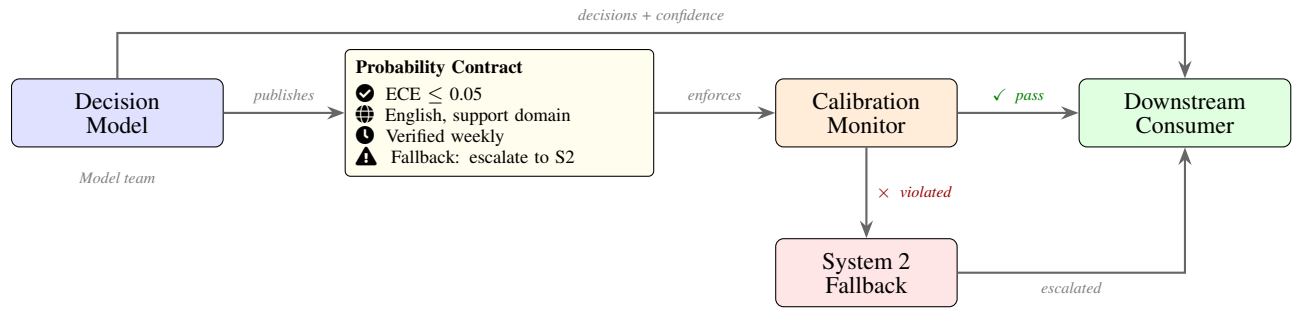
\begin{figure}[H]
\centering
\begin{tikzpicture}[
    >=Stealth,
    box/.style={draw, rounded corners=3pt, minimum height=9mm, font=\small, align=center, fill=#1},
    contract/.style={draw, rounded corners=2pt, font=\scriptsize, align=left, fill=yellow!8, text width=38mm, inner sep=4pt},
    arr/.style={->, thick, color=black!60},
    lbl/.style={font=\tiny\itshape, text=gray},
]
\node[box=blue!12, minimum width=28mm] (model) at (0,0) {Decision\\[-1pt]Model};
\node[lbl, below=2mm of model] {Model team};

\node[contract, right=16mm of model] (contract) {
    \textbf{Probability Contract}\\[2pt]
    \faCheckCircle\; ECE $\leq$ 0.05\\
    \faGlobe\; English, support domain\\
    \faClock\; Verified weekly\\
    \faExclamationTriangle\; Fallback: escalate to S2
};

\node[box=orange!15, minimum width=24mm, right=16mm of contract] (monitor) {Calibration\\[-1pt]Monitor};

\node[box=green!12, minimum width=28mm, right=16mm of monitor] (consumer) {Downstream\\[-1pt]Consumer};

\node[box=red!10, minimum width=24mm, below=12mm of monitor] (fallback) {System~2\\[-1pt]Fallback};

\draw[arr] (model) -- (contract) node[midway, above, lbl] {publishes};
\draw[arr] (contract) -- (monitor) node[midway, above, lbl] {enforces};
\draw[arr] (model.north) -- ++(0,0.6) -| (consumer.north) node[pos=0.25, above, lbl] {decisions + confidence};
\draw[arr] (monitor) -- (consumer) node[midway, above, lbl] {\color{green!50!black}\checkmark\; pass};
\draw[arr] (monitor) -- (fallback) node[midway, right, lbl] {\color{red!60!black}$\times$\; violated};
\draw[arr] (fallback.east) -| (consumer.south) node[pos=0.25, below, lbl] {escalated};

\end{tikzpicture}
\caption{Probability contracts as runtime SLAs. The model team publishes a contract specifying calibration guarantees. A runtime monitor enforces the contract; violations trigger automatic fallback to System~2.}
\label{fig:prob_contract}
\end{figure}

\subsection{IIA Violations and Option-Set Dependence}
\label{sec:iia}

The Independence of Irrelevant Alternatives (IIA) axiom~\citep{tversky1972iia} states that the relative preference between two options should not change when a third option is added or removed. \citet{iiaviolations2026} demonstrate that System~1 decision models systematically violate IIA: adding a ``decoy'' option to the choice set can shift the scores of existing options by 3--12\%, depending on the model and the decoy's similarity to existing options.

\paragraph{Practical consequences.} In production routing pipelines where the tool catalog changes dynamically (new tools added, deprecated tools removed), IIA violations mean that the routing decision for unchanged tools can shift simply because the option set changed. This creates a subtle reliability problem: the system behaves differently after a tool catalog update even for queries that have nothing to do with the new tool.

\begin{lstlisting}[caption={Detecting IIA violations: compare scores with and without a decoy option.},label={lst:iia}]
import httpx
import numpy as np

def detect_iia_violation(state: str, core_options: list[str],
                         decoy_options: list[str],
                         threshold: float = 0.05):
    """Test if adding decoy options shifts scores on core options.
    
    Returns True if any core option score shifts by > threshold.
    """
    # Score core options alone
    resp_base = httpx.post(
        "https://api.typesafe.ai/v1/choice",
        headers={"Authorization": f"Bearer {API_KEY}"},
        json={"state": state,
              "options": [{"label": o} for o in core_options]},
    )
    base_probs = dict(zip(
        core_options, resp_base.json()["probabilities"]))

    # Score core + decoy options together
    all_options = core_options + decoy_options
    resp_full = httpx.post(
        "https://api.typesafe.ai/v1/choice",
        headers={"Authorization": f"Bearer {API_KEY}"},
        json={"state": state,
              "options": [{"label": o} for o in all_options]},
    )
    full_probs_raw = dict(zip(
        all_options, resp_full.json()["probabilities"]))
    # Renormalize over core options only for fair comparison
    core_sum = sum(full_probs_raw[o] for o in core_options)
    full_probs = {o: full_probs_raw[o] / core_sum
                  for o in core_options}

    # Check drift
    violations = {}
    for o in core_options:
        drift = abs(base_probs[o] - full_probs[o])
        if drift > threshold:
            violations[o] = {"base": base_probs[o],
                             "with_decoy": full_probs[o],
                             "drift": drift}
    return violations  # empty dict = no violations

# Example: does adding "fax_machine" shift web_search scores?
v = detect_iia_violation(
    state="What is the weather in Tokyo?",
    core_options=["web_search", "calculator", "weather_api"],
    decoy_options=["fax_machine"],
)
if v:
    print(f"IIA violations detected: {v}")
\end{lstlisting}

\begin{warningbox}
IIA violations are not a bug in any specific model -- they are a structural property of softmax-based heads where all options compete in a shared probability simplex. Contrastive models (Family~IV) are less affected because they score each option independently. If your application requires strict IIA compliance, consider using a contrastive scorer or running each option through an independent Noul head.
\end{warningbox}

\subsection{Ordinal-Scale Bias}

\citet{ordinalscalebias2026} document systematic bias when System~1 Score heads map continuous internal representations to discrete ordinal scales.

\paragraph{The mechanism.} Recall the cumulative ordinal model (Eq.~\ref{eq:score_cumulative}): a Score head predicts $P(s \geq m \mid \mathbf{h}) = \sigma(\mathbf{w}^\top \mathbf{h} - \theta_m)$ with learned thresholds $\theta_{\ell+1} \leq \cdots \leq \theta_u$. The probability of each ordinal value $m$ is recovered as:
\begin{equation}
    P(s{=}m) = \sigma(z - \theta_m) - \sigma(z - \theta_{m+1}), \quad z = \mathbf{w}^\top \mathbf{h}.
    \label{eq:ordinal_pmf}
\end{equation}
Bias arises because the sigmoid function saturates: for any fixed $z$, the probability mass concentrates on the one or two categories nearest to $z$, with exponentially decaying mass on distant categories. When the model is uncertain (i.e., $z$ falls between thresholds), the mass spreads toward the \emph{center} of the scale rather than the extremes, because the central categories are bounded by thresholds on both sides while extreme categories have only a one-sided bound.

\paragraph{Central tendency bias.} Formally, let $\bar{m} = (\ell + u)/2$ denote the scale midpoint. For a model trained on uniformly distributed labels $P(y{=}m) = 1/K$ for all $m \in \{\ell, \ldots, u\}$, the predicted distribution satisfies:
\begin{equation}
    \mathbb{E}\bigl[\hat{P}(s{=}\bar{m})\bigr] > \frac{1}{K}, \qquad \mathbb{E}\bigl[\hat{P}(s{=}\ell)\bigr] < \frac{1}{K}, \qquad \mathbb{E}\bigl[\hat{P}(s{=}u)\bigr] < \frac{1}{K}.
    \label{eq:central_tendency}
\end{equation}
On a 1--5 scale, \citet{ordinalscalebias2026} report that Score heads overproduce $s{=}3$ by 8--15\% and underproduce $s{=}1$ and $s{=}5$ by 5--12\%, even when the training label distribution is perfectly uniform.

\paragraph{Boundary avoidance.} The asymmetry between extreme and central categories can be quantified via the \emph{boundary gap}:
\begin{equation}
    \Delta_{\text{boundary}} = \frac{1}{2}\bigl[\hat{P}(s{=}\bar{m}) - \hat{P}(s{=}\ell)\bigr] + \frac{1}{2}\bigl[\hat{P}(s{=}\bar{m}) - \hat{P}(s{=}u)\bigr].
    \label{eq:boundary_gap}
\end{equation}
A perfectly calibrated model has $\Delta_{\text{boundary}} = 0$. Full ordinal distribution heads (Eq.~\ref{eq:score_cumulative}) exhibit $\Delta_{\text{boundary}} \approx 0.04$--$0.08$, while simplified sigmoid heads (used by THX-01 and System~One~Gemma) exhibit $\Delta_{\text{boundary}} \approx 0.10$--$0.18$.

\paragraph{Mitigation.} For applications where the tails carry high-stakes decisions (e.g., $1 =$ ``critical vulnerability'', $5 =$ ``perfect match''), two remedies are available. First, \emph{label smoothing toward the boundaries}: during training, replace hard labels $y{=}m$ with a smoothed distribution that places $\epsilon$ extra mass on extreme categories:
\begin{equation}
    \tilde{P}(s{=}k \mid y{=}m) = \begin{cases}
        1 - \epsilon & \text{if } k = m, \\
        \epsilon / 2  & \text{if } k \in \{\ell, u\}, \\
        0 & \text{otherwise}.
    \end{cases}
    \label{eq:boundary_smoothing}
\end{equation}
Second, \emph{post-hoc threshold adjustment}: refit the thresholds $\theta_m$ on a held-out set to equalize the marginal predicted distribution, analogous to temperature scaling but applied per threshold.

\section{Training Methodologies}
\label{sec:training}

The choice of training methodology is what ultimately determines a System~1 model's accuracy, calibration, and generalization. Unlike LLMs, where the dominant paradigm is a relatively uniform pipeline (pretraining $\to$ SFT $\to$ RLHF), System~1 decision models draw on a broader and more heterogeneous set of training strategies, each suited to different model families and deployment requirements.

At one end of the spectrum, supervised fine-tuning with cross-entropy loss is the simplest and most established approach, forming the backbone of Family~I encoder classifiers. At the other end, RLCD introduces reinforcement learning with proper scoring rule rewards, directly optimizing for calibrated decision-making rather than human preference. Between these extremes lie preference-based reward model training (Family~II), contrastive learning with InfoNCE objectives (Family~IV), and knowledge distillation pipelines that compress large-model capabilities into System~1-class footprints. Many production systems combine multiple paradigms in staged pipelines: pretraining for general representations, supervised fine-tuning for task-specific decision boundaries, and a final RLCD or distillation phase for calibration and compression.

This section surveys each paradigm, emphasizing how the training objective shapes the model's output characteristics and where the key practical trade-offs lie.

\subsection{Supervised Fine-Tuning (SFT)}
\label{sec:sft_training}

The simplest approach fine-tunes a pretrained model on labeled classification data using cross-entropy loss:
\begin{equation}
    \mathcal{L}_{\text{SFT}} = -\sum_{i=1}^{N} \sum_{c=1}^{C} y_{i,c} \log p_\theta(c \mid x_i).
\end{equation}
Cross-entropy measures the ``surprise'' of the model's prediction relative to the true label: when the model assigns high probability to the correct class, the loss is near zero; when it assigns low probability, the loss grows logarithmically. As a proper scoring rule, minimizing cross-entropy is equivalent to maximizing the likelihood of the training labels, which also encourages calibration.

This is the standard training method for Family~I (encoder-based classifiers) and the first stage for most Family~II (reward model) training. Best practices include: learning rate warmup~\citep{liu2019roberta}, weight decay, sequence packing for efficiency~\citep{krell2021efficient}, and completion-only masking (for instruction-following data).

\subsection{Reward Model Training via Preference Learning}
\label{sec:rlhf_training}

Reward models (Family~II) are the System~1 component within the RLHF pipeline~\citep{ouyang2022training,christiano2017deep}. Their training is straightforward: given a dataset of human preference pairs $(q, y_w, y_l)$, minimize the BT loss (Eq.~\ref{eq:bt_loss}). The resulting model maps any (prompt, response) pair to a scalar score in a single forward pass.

\paragraph{The reward model's role in downstream training.} The trained reward model serves as a scoring oracle invoked millions of times during policy optimization (PPO~\citep{schulman2017proximal}, GRPO~\citep{shao2024deepseekmath}). Its speed directly bounds training throughput, a 2$\times$ faster reward model yields nearly 2$\times$ faster RL training. DPO~\citep{rafailov2023direct} and variants (IPO~\citep{azar2024general}, KTO~\citep{ethayarajh2024kto}, SimPO~\citep{meng2024simpo}, ORPO~\citep{hong2024orpo}) are notable because they \emph{eliminate} the explicit reward model entirely, absorbing the scoring function into the policy's log-probabilities. From a System~1 perspective, DPO removes a System~1 component from the pipeline; GRPO retains it.

\subsection{RLCD: Reinforcement Learning for Calibrated Decisions}
\label{sec:rlcd_training}

RLCD~\citep{typesafe2026jev} represents a fundamental departure from RLHF. Rather than optimizing for human preference, RLCD optimizes for calibrated decision-making using proper scoring rules. The key differences are summarized in Table~\ref{tab:rlhf_vs_rlcd}:

\begin{table}[H]
\centering
\caption{RLHF vs. RLCD training paradigm comparison.}
\label{tab:rlhf_vs_rlcd}
\begin{tabular}{lll}
\toprule
\textbf{Aspect} & \textbf{RLHF} & \textbf{RLCD} \\
\midrule
Optimizes for & Human preference & Calibrated probabilities \\
Reward signal & Learned reward model & Proper scoring rules \\
Training data & Preference pairs & Decision tasks with outcomes \\
Output & Free-form text & Typed decisions (Choice/Score/Noul) \\
Calibration & Poor (side effect) & By construction (objective) \\
Cost & Expensive (4 models) & Moderate (2 models) \\
\bottomrule
\end{tabular}
\end{table}

\subsection{Contrastive Learning}
\label{sec:contrastive_training}

Family~IV models use contrastive objectives to learn shared state-action embedding spaces. The InfoNCE loss (Eq.~\ref{eq:infonce}) is the standard, with several important training considerations. Mining \emph{hard negatives}, informative negative actions that are plausible but incorrect, improves learning efficiency~\citep{karpukhin2020dense}, while \emph{in-batch negatives} (using other samples in the batch as negatives) provide a simpler, scalable alternative~\citep{chen2020simple}. The temperature $\tau$ in the InfoNCE loss controls the sharpness of the decision boundary; annealing from warm to cool during training improves convergence~\citep{radford2021clip}. Notably, CLM-8B freezes the Qwen3-8B encoder backbone and trains only the projection heads, reducing training cost by $>$10$\times$ compared to full fine-tuning~\citep{contrastivelm2026clm}.

\subsection{Knowledge Distillation}
\label{sec:distillation}

Knowledge distillation~\citep{hinton2015distilling} transfers capabilities from a large teacher model to a smaller student, enabling System~1-class latency with System~2-class accuracy:

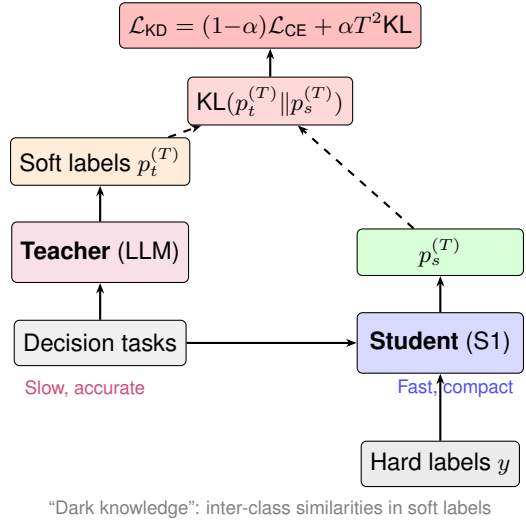
\begin{figure}[H]
\centering
\begin{tikzpicture}[
    box/.style={draw, rounded corners=2pt, minimum height=0.6cm, font=\small\sffamily, fill=#1},
    box/.default=blue!8,
    arr/.style={-{Stealth[length=4pt]}, thick},
    lbl/.style={font=\scriptsize\sffamily},
]
\node[box=purple!12, minimum width=2.2cm, minimum height=0.8cm] (teacher) at (0, 3.6) {\textbf{Teacher} (LLM)};
\node[box=gray!12, minimum width=2.2cm] (tasks) at (0, 2.4) {Decision tasks};
\draw[arr] (tasks) -- (teacher);

\node[box=orange!15, minimum width=2.2cm] (soft) at (0, 4.8) {Soft labels $p_t^{(T)}$};
\draw[arr] (teacher) -- (soft);

\node[box=blue!15, minimum width=2.2cm, minimum height=0.8cm] (student) at (4.5, 2.4) {\textbf{Student} (S1)};
\node[box=green!15, minimum width=2.2cm] (sout) at (4.5, 3.6) {$p_s^{(T)}$};
\draw[arr] (student) -- (sout);
\draw[arr] (tasks) -- (student);

\node[box=gray!15, minimum width=2.0cm] (hard) at (4.5, 0.8) {Hard labels $y$};
\draw[arr] (hard) -- (student);

\node[box=red!15, minimum width=2.0cm, minimum height=0.6cm] (kl) at (2.25, 5.6) {$\text{KL}(p_t^{(T)} \| p_s^{(T)})$};
\draw[arr, dashed] (soft) -- (kl);
\draw[arr, dashed] (sout) -- (kl);

\node[box=red!25, minimum width=3.0cm, minimum height=0.6cm] (loss) at (2.25, 6.6) {$\mathcal{L}_{\text{KD}} = (1{-}\alpha)\mathcal{L}_{\text{CE}} + \alpha T^2 \text{KL}$};
\draw[arr] (kl) -- (loss);

\node[lbl, text=purple!70] at (-0.2, 1.8) {Slow, accurate};
\node[lbl, text=blue!70] at (4.7, 1.8) {Fast, compact};
\node[lbl, text=gray] at (2.25, 0.2) {``Dark knowledge'': inter-class similarities in soft labels};

\end{tikzpicture}
\caption{Knowledge distillation from a large LLM teacher to a compact System~1 student. The student trains against both hard ground-truth labels (cross-entropy) and the teacher's soft probability distribution (KL divergence at temperature $T$). The soft labels carry ``dark knowledge'' about inter-class similarities that hard labels discard.}
\label{fig:distillation}
\end{figure}

The standard distillation objective~\citep{hinton2015distilling} combines a hard-label cross-entropy term with a soft-label KL divergence term, weighted by a mixing coefficient $\alpha$:
\begin{equation}
    \mathcal{L}_{\text{KD}} = (1 - \alpha) \mathcal{L}_{\text{CE}}(y, p_s) + \alpha \cdot T^2 \cdot \text{KL}(p_t^{(T)} \| p_s^{(T)}),
\end{equation}

where $p_t^{(T)}$ and $p_s^{(T)}$ are teacher and student softmax outputs at temperature $T$.

The loss balances two signals: the first term $\mathcal{L}_{\text{CE}}(y, p_s)$ trains the student against hard ground-truth labels (``this is class 3''), while the second term $\text{KL}(p_t^{(T)} \| p_s^{(T)})$ trains against the teacher's \emph{soft} output distribution (``this is 70\% class 3, 20\% class 5, 10\% class 1''), the soft distribution carries ``dark knowledge'' about inter-class similarities that hard labels discard. The temperature $T > 1$ softens both distributions, amplifying these informative tail probabilities. The $T^2$ scaling corrects for the reduced gradient magnitude when using high temperatures.

\paragraph{LLM-to-classifier distillation.} A particularly relevant pattern distills an LLM's decision-making into a small classifier~\citep{hinton2015distilling}. The procedure generates LLM decisions on a large dataset of decision tasks, filters for correct decisions verified by ground truth, and then fine-tunes a small encoder (e.g., DeBERTa-base) on the filtered dataset.

This can typically achieve $>$90\% of the LLM's decision accuracy at a fraction of the cost and latency, though exact gains depend on task complexity and the quality of the teacher's decisions. The approach parallels the DeepSeek-R1 distillation strategy~\citep{deepseek2025r1}, which distills reasoning capabilities from the full R1 model into smaller student models.

\subsection{Multi-Task and Curriculum Training}

System~1 models benefit from diverse training data and structured curricula.

\paragraph{Multi-task training.} Training on heterogeneous decision tasks (classification, scoring, gating) simultaneously improves generalization, as demonstrated by the universal classifier approach of \citet{laurer2024universal}. The key insight is that shared encoder representations benefit from seeing diverse decision types: a model that learns to both classify and score develops richer internal features than one trained on either task alone. In practice, multi-task training is implemented by sampling mini-batches from a mixture of task-specific datasets, with mixing ratios tuned on a validation set.

\paragraph{Curriculum learning.} Starting with easy decisions and progressively increasing difficulty improves convergence~\citep{bengio2009curriculum}. For System~1 models, a typical three-stage curriculum proceeds as follows:

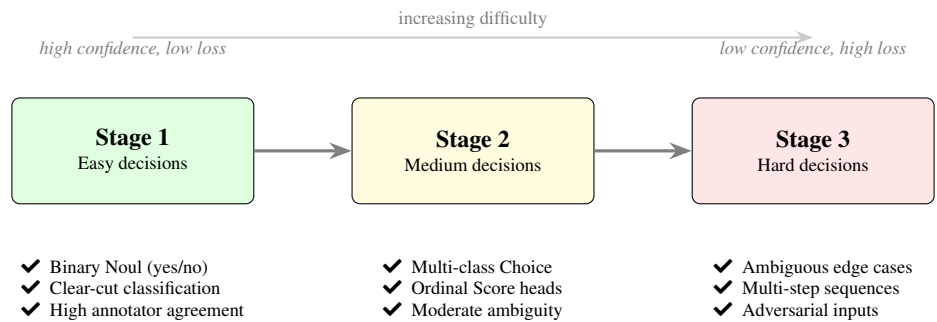
\begin{figure}[H]
\centering
\begin{tikzpicture}[
    >=Stealth,
    stage/.style={draw, rounded corners=3pt, minimum width=32mm, minimum height=14mm, font=\small, align=center, fill=#1},
    task/.style={font=\scriptsize, align=left},
    arr/.style={->, very thick, color=black!50},
    lbl/.style={font=\scriptsize\itshape, text=gray},
]
\node[stage=green!12] (s1) at (0,0) {\textbf{Stage 1}\\[-1pt]\scriptsize Easy decisions};
\node[stage=yellow!15] (s2) at (4.5,0) {\textbf{Stage 2}\\[-1pt]\scriptsize Medium decisions};
\node[stage=red!10] (s3) at (9,0) {\textbf{Stage 3}\\[-1pt]\scriptsize Hard decisions};

\draw[arr] (s1) -- (s2);
\draw[arr] (s2) -- (s3);

\node[task, below=6mm of s1] {%
    \faCheck\; Binary Noul (yes/no)\\
    \faCheck\; Clear-cut classification\\
    \faCheck\; High annotator agreement};
\node[task, below=6mm of s2] {%
    \faCheck\; Multi-class Choice\\
    \faCheck\; Ordinal Score heads\\
    \faCheck\; Moderate ambiguity};
\node[task, below=6mm of s3] {%
    \faCheck\; Ambiguous edge cases\\
    \faCheck\; Multi-step sequences\\
    \faCheck\; Adversarial inputs};

\node[lbl, above=4mm of s1] {high confidence, low loss};
\node[lbl, above=4mm of s3] {low confidence, high loss};

\draw[gray!40, thick, ->] (0, 1.5) -- (9, 1.5) node[midway, above, font=\scriptsize, text=gray] {increasing difficulty};

\end{tikzpicture}
\caption{Curriculum training for System~1 decision models. Training progresses from easy, high-agreement decisions (Stage~1) through moderately ambiguous tasks (Stage~2) to hard edge cases (Stage~3). Within each stage, multi-task batches mix different decision types (Noul, Choice, Score).}
\label{fig:curriculum}
\end{figure}

\noindent In practice, starting with single-step decisions before training on complex multi-step decision sequences yields better convergence. The curriculum also interacts with calibration: models trained on easy examples first develop well-calibrated confidence on clear-cut cases, which then transfers to harder examples as the curriculum advances.

\subsection{Practitioner Training Recipes}

The following recipes cover the two most common scenarios: (1)~fine-tuning a compact encoder for a fixed classification task, and (2)~converting a pretrained LLM into a bidirectional decision model.

\begin{lstlisting}[caption={Recipe 1: Fine-tune DeBERTa-v3-base for classification ($\sim$15 min on one GPU).},label={lst:deberta}]
from transformers import (
    AutoTokenizer, AutoModelForSequenceClassification,
    TrainingArguments, Trainer,
)
from datasets import load_dataset
import numpy as np

dataset = load_dataset("your_org/your_dataset")
tokenizer = AutoTokenizer.from_pretrained("microsoft/deberta-v3-base")
model = AutoModelForSequenceClassification.from_pretrained(
    "microsoft/deberta-v3-base",
    num_labels=len(dataset["train"].features["label"].names),
)

def tokenize(batch):
    return tokenizer(batch["text"], truncation=True, max_length=512)

tokenized = dataset.map(tokenize, batched=True)
trainer = Trainer(
    model=model,
    args=TrainingArguments(
        output_dir="./deberta-clf",
        num_train_epochs=3,
        per_device_train_batch_size=16,
        learning_rate=2e-5,     # DeBERTa sweet spot
        weight_decay=0.01,
        eval_strategy="epoch",
        load_best_model_at_end=True,
        fp16=True,
    ),
    train_dataset=tokenized["train"],
    eval_dataset=tokenized["validation"],
)
trainer.train()
# IMPORTANT: apply temperature scaling (Listing 9) before deploying!
\end{lstlisting}

\begin{lstlisting}[caption={Recipe 2 (conceptual): Convert a causal LLM to a bidirectional decision model.},label={lst:lmconvert}]
import torch.nn as nn
from transformers import AutoModel
from peft import get_peft_model, LoraConfig

class DecisionModel(nn.Module):
    """LLM backbone + typed decision heads (System One Gemma recipe)."""
    def __init__(self, base_model: str, n_classes: int):
        super().__init__()
        self.encoder = AutoModel.from_pretrained(base_model)
        d = self.encoder.config.hidden_size
        # Step 1: Remove LM head (done implicitly by AutoModel)
        # (*@Step 2: Enable bidirectional attention (model-specific).@*)
        # (*@For Gemma 3: self.encoder.config.use\_bidirectional\_attention = True@*)
        # (*@For other models: override the attention mask in forward().@*)
        # (*@Setting is\_causal=False alone is insufficient -- the causal mask@*)
        # (*@is built from the config, not this flag, on most backends.@*)
        for layer in (*@self.encoder.layers@*):
            if hasattr(layer.self_attn, "is_causal"):
                layer.self_attn.is_causal = False
        # Step 3: Add typed decision heads
        self.choice_head = nn.Linear(d, n_classes)
        self.noul_head   = nn.Sequential(nn.Linear(d, 1), nn.Sigmoid())
        self.score_head  = nn.Sequential(nn.Linear(d, 1), nn.Sigmoid())

    def forward(self, input_ids, attention_mask, head="choice"):
        h = self.encoder(input_ids=input_ids,
                         attention_mask=attention_mask).last_hidden_state
        pooled = (h * attention_mask.unsqueeze(-1)).sum(1) \
                 / attention_mask.sum(1, keepdim=True)
        if head == "choice": return self.choice_head(pooled)
        if head == "noul":   return self.noul_head(pooled)
        return self.score_head(pooled)

# LoRA for efficient fine-tuning
model = DecisionModel((*@"google/gemma-3-270m"@*), n_classes=5)
model = get_peft_model(model, LoraConfig(
    r=16, lora_alpha=32,
    target_modules=["q_proj", "v_proj"], lora_dropout=0.05,
    (*@modules\_to\_save=["choice\_head", "noul\_head", "score\_head"],@*)
))
\end{lstlisting}

\begin{warningbox}
Converting a causal LM to bidirectional is not a simple config change -- it requires modifying attention masks deep inside the model. For most practitioners, fine-tuning DeBERTa (Recipe~1) or using a managed API (Jev) is more practical. Use the LLM conversion recipe only when you need a large model's pre-existing knowledge for a specialized domain.
\end{warningbox}

\begin{tipbox}
\textbf{Recommended progression:} (1)~API prototype (Jev/CLM) -- validate the pattern works; (2)~add a cascade -- measure escalation rate and cost savings; (3)~self-host if needed -- fine-tune DeBERTa + temperature scaling; (4)~optimize -- quantize (INT8), batch, export to ONNX/TensorRT. Do not jump to step~4 before validating step~1.
\end{tipbox}

\clearpage
\ifdim\pagetotal>0pt\thispagestyle{empty}\newpage\fi
\part{Applications and Integration}

System~1 decision models are not standalone artifacts; their value emerges when embedded in real-world systems. This chapter surveys the application domains where these models are deployed, examines how they function as infrastructure within agentic workflows, and describes the cascade pattern that combines fast System~1 decisions with slow System~2 reasoning.

\section{Applications}
\label{sec:applications}

We catalogue eighteen application areas below, organized by the decision characteristics that make each domain a natural fit. We begin with the highest-impact use cases -- agent routing, search reranking, and content moderation -- where System~1 models are already deployed at scale, then survey emerging domains including cybersecurity, theory of mind, political science, and low-resource languages where early results suggest significant potential.

\subsection{Agent Routing and Tool Selection}

In agentic AI systems~\citep{yao2023react,shinn2023reflexion}, a critical decision point is \emph{which tool or sub-agent to invoke} given the current state. This is a natural System~1 task, the routing decision must be fast (to avoid latency accumulation over multi-step trajectories), well-calibrated (to support fallback strategies), and deterministic (for reproducibility).

\begin{figure}[H]
\centering
\begin{tikzpicture}[
    >=Stealth,
    box/.style={draw, rounded corners=3pt, minimum height=9mm, font=\small, align=center, fill=#1},
    gate/.style={draw, rounded corners=1pt, minimum width=20mm, minimum height=10mm, font=\scriptsize, align=center, fill=yellow!10, thick},
    tool/.style={draw, rounded corners=2pt, minimum width=16mm, minimum height=7mm, font=\scriptsize, fill=blue!8, align=center},
    arr/.style={->, thick, color=black!60},
    lbl/.style={font=\tiny\itshape, text=gray},
]
\node[box=gray!10, minimum width=24mm] (state) at (0,0) {Agent State\\[-1pt]\scriptsize\itshape step $t$};

\node[box=green!12, minimum width=28mm, right=14mm of state] (router) {System~1\\[-1pt]Router};

\node[gate, right=14mm of router] (gate) {$p \geq \theta$\,?};

\node[tool, above right=10mm and 16mm of gate] (t1) {Tool A};
\node[tool, right=16mm of gate] (t2) {Tool B};
\node[tool, below right=10mm and 16mm of gate] (t3) {Tool C};

\node[box=red!10, minimum width=24mm, below=14mm of gate] (llm) {LLM Fallback\\[-1pt]\scriptsize System~2};

\node[box=violet!10, minimum width=22mm, right=14mm of t2] (exec) {Execute\\[-1pt]\& Update};

\draw[arr] (state) -- (router) node[midway, above, lbl] {$\sim$30ms};
\draw[arr] (router) -- (gate);
\draw[arr, green!60!black] (gate) -- (t1) node[pos=0.3, above, font=\tiny, green!60!black] {\checkmark};
\draw[arr, green!60!black] (gate) -- (t2);
\draw[arr, green!60!black] (gate) -- (t3);
\draw[arr, red!60!black] (gate) -- (llm) node[midway, right, font=\tiny, red!60!black] {$\times$ low conf.};
\draw[arr] (t1.east) -| (exec.north);
\draw[arr] (t2) -- (exec);
\draw[arr] (t3.east) -| (exec.south);
\draw[arr, dashed, red!40] (llm.east) -| ([yshift=-3mm]exec.south) -- (exec.south);

\draw[arr, gray!50] (exec.north) -- ++(0,0.8) -| (state.north) node[pos=0.25, above, lbl] {next step};

\end{tikzpicture}
\caption{Agent routing with confidence-gated fallback. At each step, a System~1 router selects a tool in $\sim$30ms. If confidence exceeds threshold $\theta$, the tool executes directly; otherwise, a System~2 LLM fallback deliberates over the choice.}
\label{fig:agent_routing}
\end{figure}
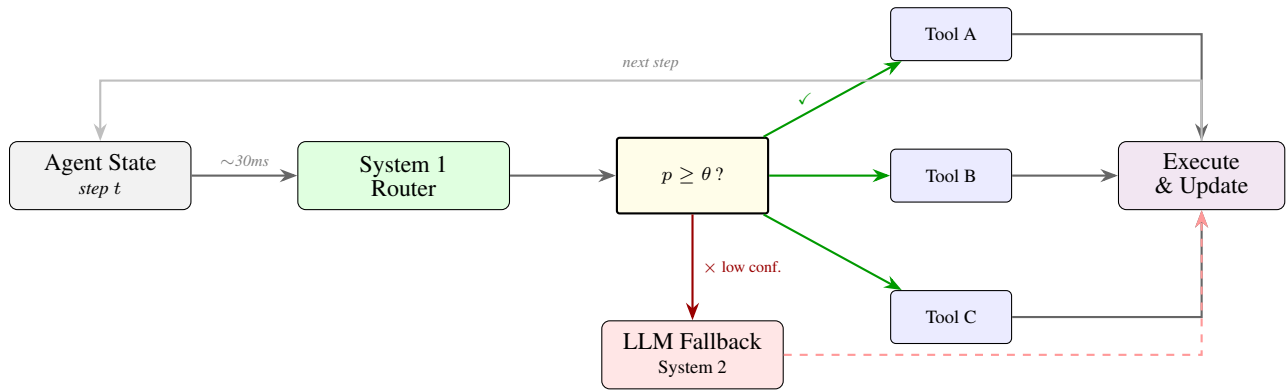

\citet{contrastivelm2026clm} demonstrated CLM-8B as an agent router that selects among candidate actions in the LLM-WikiRace~\citep{ziomek2026wikirace} and Super Mario environments. The decoupled encoding architecture enables precomputing action embeddings for the tool catalog, reducing routing to a single state encoding plus $K$ dot products.

\paragraph{Latency budget analysis.} Consider an agent trajectory of $L$ steps, each requiring a routing decision and a generative action. The total latency is $T = L \cdot (t_{\text{route}} + t_{\text{gen}})$. With LLM-based routing ($t_{\text{route}} \approx 500$ms) and a 20-step trajectory, routing alone consumes 10 seconds. With System~1 routing ($t_{\text{route}} \approx 30$ms), the same trajectory adds only 600ms of routing overhead, a $16\times$ reduction that is the difference between interactive and batch-mode operation. Calibrated confidence enables an additional optimization: if the router's confidence exceeds $\theta_{\text{skip}}$, verification steps can be skipped entirely, further reducing trajectory latency.

\begin{lstlisting}[caption={Agent tool router with confidence-gated LLM fallback.},label={lst:router}]
class AgentRouter:
    """System 1 tool router; falls back to LLM on low confidence."""

    def __init__(self, tools: dict[str, str],
                 threshold: float = 0.65, llm=None):
        self.tools = tools
        self.threshold = threshold
        self.llm = llm
        self.options = [{"label": n, "description": d}
                        for n, d in tools.items()]

    def route(self, state: str) -> tuple[str, float, bool]:
        """Returns (tool, confidence, used_fallback)."""
        resp = httpx.post(
            "https://api.typesafe.ai/v1/choice",
            headers={"Authorization": f"Bearer {API_KEY}"},
            json={"state": state, "options": self.options},
        )
        r = resp.json()
        tool, conf = r["choice"], r["confidence"]
        if conf >= self.threshold or self.llm is None:
            return tool, conf, False
        # Uncertain -> ask LLM to reason about the right tool
        tool = self.llm.chat(
            f"State:\n{state}\nTools: {list(self.tools)}\n"
            "Reply with the tool name only."
        ).strip()
        return tool, conf, True

# Usage inside an agent loop
router = AgentRouter(TOOL_CATALOG, threshold=0.65, llm=llm)
for step in range(max_steps):
    state = build_state(history, observation)
    tool, conf, fallback = router.route(state)
    result = execute_tool(tool, state)
    history.append((tool, result))
\end{lstlisting}

\begin{tipbox}
Pre-format tool options once and reuse them on every call. The state changes per query but the tool catalog is static. For contrastive models (CLM), precompute tool embeddings at startup -- routing then reduces to a single state encoding plus $K$ dot products.
\end{tipbox}

\subsection{Search Reranking}

Cross-encoder rerankers (Section~\ref{sec:cross_encoder}) are deployed at massive scale in production search engines. \citet{walmart2025crossencoder} described a BERT-based cross-encoder serving billions of queries at Walmart, achieving significant relevance improvements over lexical matching. The hybrid retrieval pipeline (BM25 $\to$ cross-encoder reranking) represents a System~1 cascade: fast sparse retrieval followed by precise neural scoring.

\paragraph{Multi-stage pipelines.} Production search systems typically employ three or more stages, each a System~1 model operating at a different cost--quality point. The first stage uses sparse retrieval (BM25 or SPLADE~\citep{formal2021splade}) to reduce millions of candidates to ${\sim}1000$. The second stage uses a bi-encoder or lightweight cross-encoder to reduce to ${\sim}50$. The final stage uses a high-capacity cross-encoder for precise reranking. Each transition narrows the candidate set while increasing per-candidate compute, and the overall pipeline latency is dominated by the final cross-encoder stage at ${\sim}10$--$50$ms for 50 documents. This architecture demonstrates that System~1 models compose naturally into cascading decision pipelines.

\subsection{Content Moderation and Safety}

Safety classification is a canonical System~1 task: every generated response must be checked for harmful content before being shown to users. Latency is critical (moderation happens on the critical path), calibration matters (false positives degrade user experience while false negatives create harm), and the output is structured.

Models like Llama Guard~\citep{inan2023llama} use small LLMs for safety classification, but encoder-based alternatives achieve comparable accuracy at a fraction of the latency. The Noul primitive in Jev~\citep{typesafe2026jev} is explicitly designed for this use case.

\paragraph{Hate-speech moderation.} \citet{paschalides2026hatedecide} present HATEDECIDE, a controlled evaluation of six Jev decision-model configurations on four hate-speech datasets against specialized, zero-shot, commercial LLM, and supervised baselines. Commercial LLMs significantly outperform decision models on only one of the four datasets; on a diagnostic test set the best hosted decision model comes within 1.6 macro-F1 points of the best commercial LLM at approximately 97\% lower inference cost. Supplying explicit hate-speech definitions changes up to 28\% of predictions without consistently improving classification, and decomposing definitions into multiple questions helps in only 20\% of comparisons. These results identify a practical operating regime for low-cost System~1 moderation while highlighting that explicit criteria do not reliably substitute for task-specific training.

\paragraph{Alignment failure detection.} \citet{guo2026justaskjev} present RLCDAlignBench, which evaluates Jev as a zero-shot detector across ten categories of misalignment -- ranging from sycophancy and jailbreaks to reward hacking, concealed uncertainty, and power-seeking behavior. The benchmark aggregates 44 existing test suites and covers five target models. With only one task-agnostic prompt and no fine-tuning, the detector attains a median AUROC of 0.886, surpassing supervised baselines on the majority of benchmarks at 63$\times$ lower cost than LLM-judge scorers. The key design insight is that many alignment failures are \emph{relational}, defined against a reference (the user's belief, an injected instruction) that cannot be identified from the model's output in isolation; adjusting which contextual inputs are supplied to Jev proves more important than rephrasing the detection question. This result demonstrates that RLCD-trained calibrated probabilities enable broad zero-shot safety screening without task-specific fine-tuning, substantially extending the scope of System~1 models beyond single-category moderation.

\subsection{Knowledge Organization and RAG Alternatives}

Retrieval-Augmented Generation (RAG) is the dominant paradigm for grounding LLM outputs in external knowledge: retrieve relevant documents, concatenate them into the prompt, and generate. However, RAG introduces its own latency (embedding + vector search + reranking), context-window pressure, and failure modes (irrelevant retrievals that mislead the generator). An emerging alternative uses System~1 decision models to \emph{organize} knowledge proactively, replacing retrieval-time search with classifier-routed navigation over a pre-structured knowledge base.

\begin{figure}[H]
\centering
\begin{tikzpicture}[
    >=Stealth,
    box/.style={draw, rounded corners=2pt, minimum width=20mm, minimum height=7mm, font=\scriptsize, align=center, fill=#1},
    arr/.style={->, thick, color=black!60},
    lbl/.style={font=\tiny\itshape, text=gray},
    heading/.style={font=\small\sffamily\bfseries},
]
\node[heading] at (-3.2, 3.8) {Traditional RAG};

\node[box=gray!10] (q1) at (-3.2, 3.0) {Query};
\node[box=blue!10] (emb) at (-3.2, 2.1) {Embed};
\node[box=blue!10] (vec) at (-3.2, 1.2) {Vector Search};
\node[box=blue!10] (rer) at (-3.2, 0.3) {Rerank};
\node[box=orange!12] (gen1) at (-3.2, -0.6) {LLM Generate};

\draw[arr] (q1) -- (emb);
\draw[arr] (emb) -- (vec);
\draw[arr] (vec) -- (rer);
\draw[arr] (rer) -- (gen1);

\node[lbl, right=1mm of emb] {$\sim$20ms};
\node[lbl, right=1mm of vec] {$\sim$50ms};
\node[lbl, right=1mm of rer] {$\sim$100ms};
\node[lbl, right=1mm of gen1] {$\sim$500ms};

\node[heading] at (3.2, 3.8) {CROW (Classifier-Routed)};

\node[box=gray!10] (q2) at (3.2, 3.0) {Query};
\node[box=green!12] (c1) at (3.2, 2.1) {Choice: folder};
\node[box=green!12] (c2) at (3.2, 1.2) {Noul: per note};
\node[box=green!12] (c3) at (3.2, 0.3) {Select context};
\node[box=orange!12] (gen2) at (3.2, -0.6) {LLM Generate};

\draw[arr] (q2) -- (c1);
\draw[arr] (c1) -- (c2);
\draw[arr] (c2) -- (c3);
\draw[arr] (c3) -- (gen2);

\node[lbl, left=1mm of c1] {$\sim$10ms};
\node[lbl, left=1mm of c2] {$\sim$15ms};
\node[lbl, left=1mm of c3] {$\sim$5ms};
\node[lbl, left=1mm of gen2] {$\sim$500ms};

\draw[gray!30, dashed] (0, 3.8) -- (0, -1.1);

\node[font=\tiny, text=gray, align=center] at (0, -1.5) {retrieval-time compute\\vs.\ ingestion-time organization};

\end{tikzpicture}
\caption{Traditional RAG (left) performs embedding, vector search, and reranking at query time. CROW (right) replaces retrieval with classifier-routed navigation over a pre-organized knowledge base, reducing per-query overhead.}
\label{fig:rag_vs_crow}
\end{figure}
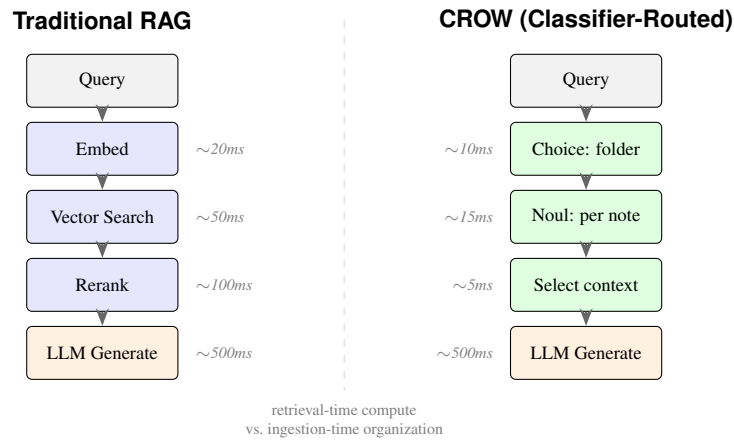

\paragraph{CROW.} \citet{cesarini2026crow} introduce Classifier-Routed Organization of LLM Wikis (CROW), in which a typed classifier (Jev or Laya) takes every filing and retrieval decision in an LLM-maintained wiki, and the LLM is restricted to writing and answering. Figure~\ref{fig:rag_vs_crow} contrasts the two paradigms.

\noindent During ingestion, the classifier routes each new document to the correct folder (Choice over subfolders), scores relevance against existing notes (Noul per note), and decides whether to merge or create a new entry (Choice between Modify and New). During retrieval, the classifier scores folders and notes (Noul per candidate) to select the context passed to the LLM for answer generation. Each decision carries a confidence threshold; when the classifier is uncertain, that step falls back to the LLM, instantiating the cascade pattern of Section~\ref{sec:cascades}. The architecture trades retrieval-time compute for ingestion-time organization. This design is particularly advantageous for corpora that are queried frequently relative to update rate, as the per-query cost of classifier-routed navigation (a sequence of Choice and Noul calls) is substantially lower than embedding search plus reranking.

\paragraph{Multi-label safety taxonomies.} Modern safety classification is rarely binary. Production systems check for multiple harm categories simultaneously, hate speech, self-harm, illegal activity, sexual content, misinformation, personally identifiable information, each with different severity thresholds and regulatory requirements. A System~1 model can evaluate all categories in a single forward pass via parallel Noul heads (one per category), with per-category calibration enabling nuanced policy enforcement: a post with $p_{\text{hate}} = 0.90$ (i.e., $p_{\text{safe}} = 0.10$) is blocked immediately, while $p_{\text{hate}} = 0.50$ ($p_{\text{safe}} = 0.50$) is escalated to human review, and $p_{\text{hate}} = 0.10$ ($p_{\text{safe}} = 0.90$) is allowed through. The calibrated probabilities are essential for audit trails and regulatory compliance.

\begin{lstlisting}[caption={Production safety gate with per-category thresholds and tiered actions.},label={lst:safety},float=ht]
from dataclasses import dataclass
from enum import Enum

class SafetyAction(Enum):
    ALLOW = "allow"       # High confidence safe
    WARN = "warn"         # Borderline -- add disclaimer
    BLOCK = "block"       # High confidence unsafe
    ESCALATE = "escalate" # Uncertain -- send to human review

@dataclass
class SafetyResult:
    action: SafetyAction
    p_safe: float
    categories: dict[str, float]

SAFETY_CATEGORIES = {
    "harmful":   "Does this contain harmful or dangerous content?",
    "hate":      "Does this contain hate speech or discrimination?",
    "pii":       "Does this reveal personal information?",
    "misinfo":   "Does this contain clearly false factual claims?",
}

def safety_gate(response: str, context: str,
                block_threshold=0.15, (*@escalate\_threshold=0.60, warn\_threshold=0.85@*)) -> SafetyResult:
    """Multi-category safety gate using parallel Noul heads."""
    resp = httpx.post(
        "https://api.typesafe.ai/v1/batch",
        headers={"Authorization": f"Bearer {API_KEY}"},
        json={
            "state": f"Context: {context}\nResponse: {response}",
            "decisions": [
                {"type": "noul", "id": cat, "question": q}
                for cat, q in SAFETY_CATEGORIES.items()
            ],
        },
    )
    results = resp.json()["results"]
    # p_safe per category = 1 - p(contains_issue)
    cat_probs = {c: 1.0 - results[c]["probability"]
                 for c in SAFETY_CATEGORIES}
    p_safe = min(cat_probs.values())

    (*@if p\_safe < block\_threshold:        action = SafetyAction.BLOCK@*)
    (*@elif p\_safe < escalate\_threshold:   action = SafetyAction.ESCALATE@*)
    (*@elif p\_safe < warn\_threshold:       action = SafetyAction.WARN@*)
    (*@else:                               action = SafetyAction.ALLOW@*)
    return SafetyResult(action, p_safe, cat_probs)
\end{lstlisting}

\begin{warningbox}
Never use a single fixed threshold for all safety categories. Different categories have different base rates and asymmetric error costs: a false negative on ``harmful content'' is far more costly than on ``misinformation.'' Set per-category thresholds based on your risk tolerance, and log every decision with its full probability vector for post-hoc auditing.
\end{warningbox}

\subsection{Reward Scoring in RL Training}

During RLHF and RLVR (Reinforcement Learning from Verifiable Rewards) training~\citep{deepseek2025r1}, the reward model is invoked for every generated response, potentially millions of times per training run. The speed of the reward model directly impacts training throughput. Section~\ref{sec:reward} surveyed the architectural options; in practice, production systems use quantized reward models (INT8/FP8) for scoring during training~\citep{dettmers2022llmint8}.

\paragraph{Training throughput analysis.} In GRPO~\citep{shao2024deepseekmath} with group size $G$, each prompt generates $G$ responses that must all be scored by the reward model. For a training run with $B$ prompts per batch and $N$ batches, the total reward model calls are $B \times G \times N$. With $B = 512$, $G = 8$, and $N = 100{,}000$, this yields $4 \times 10^8$ forward passes. A 2$\times$ speedup in the reward model (e.g., by switching from a 7B to a 350M encoder-based RM) can save days of GPU time on a large training cluster. This makes reward model efficiency a first-order concern for organizations training frontier models.

\subsection{Code Verification}

CLM-8B achieves state-of-the-art results as a code verifier on DeepSWE~\citep{datacurve2026deepswe} (48.2\%) and Terminal-Bench~2.1~\citep{harbor2026terminalbench} (72.3\%), outperforming generative verifiers while running 4--5$\times$ faster~\citep{contrastivelm2026clm}. The contrastive approach is particularly well-suited for verification: encode the problem specification and the candidate solution, then measure alignment in embedding space.

\paragraph{Verification in coding agents.} In autonomous coding agents (e.g., SWE-bench~\citep{jimenez2024swebench}, Devin-style systems), each proposed code change must be verified before submission. Generative verifiers produce natural-language assessments, introducing latency and parsing overhead. Contrastive System~1 verifiers output a single scalar ($r_t > \theta_{\text{success}}$?) that can directly control the agent loop's continue/submit decision. When the agent generates $N$ candidate patches and selects the best, verification cost scales as $O(N)$, and at ${\sim}30$ms per verification, scoring 64 candidates takes under 2 seconds, compared to minutes with generative verifiers.

\subsection{Natural Language Inference (NLI)}

NLI-trained models (Section~\ref{sec:encoder}) enable zero-shot classification by framing any classification task as an entailment problem~\citep{yin2019benchmarking,laurer2024universal}: ``This text is about [category]'' as the hypothesis, and measuring entailment probability. This is a System~1 approach to classification that requires no task-specific fine-tuning.

\paragraph{Universal classifiers.} \citet{laurer2024universal} extended this paradigm to \emph{universal} NLI-based classification, where a single model handles sentiment analysis, topic classification, intent detection, and named entity typing, all by varying the hypothesis template. The model outputs calibrated entailment probabilities for each candidate label, enabling threshold-based decision-making that adapts gracefully to new label sets at deployment time without retraining. This is arguably the earliest production example of the ``typed decision head'' paradigm that Family~III models later formalized.

\subsection{Medical and Clinical Decision Support}

\paragraph{Clinical decision requirements.} Healthcare applications impose uniquely stringent requirements on System~1 models: (i)~\emph{calibration} must be near-perfect, as overconfident predictions can lead to missed diagnoses or unnecessary interventions; (ii)~\emph{latency} must support real-time monitoring (continuous glucose monitoring generates readings every 5 minutes); (iii)~\emph{explainability} is mandated by regulations such as the EU AI Act\footnote{Regulation (EU) 2024/1689 of the European Parliament, \emph{Artificial Intelligence Act}, Official Journal of the European Union, 2024.} for high-risk applications; and (iv)~\emph{auditability} requires that every decision is logged with its confidence score and the inputs that produced it. These requirements make System~1 models with proper scoring rule training (RLCD) a natural fit, as the calibrated probabilities directly support clinical risk thresholds (e.g., ``alert if $p_{\text{hypoglycemia}} > 0.7$'').

\paragraph{Speech neuroprosthesis rescoring.} \citet{cina2026jevneuro} provide a concrete clinical deployment: replacing the 7B-parameter language model used for rescoring candidate sentences in a speech neuroprosthesis for a participant with ALS. The rescoring task is framed as a single typed Choice decision over the decoder's $N$-best list. On 978 held-out sentences, Jev achieves 7.5\% word error rate (WER), compared to 7.8\% for both OPT-6.7B and Qwen2.5-7B; with decoder-weight re-tuning, the gap widens to 6.9\% vs.\ 7.2\% and 7.4\%. Crucially, Jev requires no GPU and costs \$0.07 per thousand sentences, making a dedicated GPU cost-effective only above 43\% utilization, far beyond what a single user generates. This result illustrates a key System~1 advantage in clinical settings: competitive accuracy at a cost and infrastructure profile compatible with individual patient devices rather than data-center deployments.

\subsection{Financial Risk and Fraud Detection}

Financial institutions deploy millions of real-time decisions per second for transaction fraud detection, credit risk scoring, and anti-money-laundering (AML) screening. Each transaction requires a decision within single-digit milliseconds, well within System~1 latency but far below LLM capabilities. Traditional rule-based and gradient-boosted-tree systems dominate, but System~1 models offer advantages in handling unstructured text inputs (transaction descriptions, customer communications) that accompany structured features. A Score-head model can output a continuous risk score calibrated to default probability, enabling direct integration with regulatory capital calculations that require well-calibrated probability-of-default (PD) estimates.

\subsection{Customer Support Triage}

Enterprise helpdesk and customer support systems route millions of tickets daily across multiple queues, priority levels, and specialist teams. The routing decision involves classifying the issue type (Choice), assessing urgency (Score), and checking whether the ticket contains sensitive information requiring escalation (Noul), precisely the three primitives of Family~III models. A single System~1 forward pass can produce all three decisions simultaneously via the parallel sampler (Eq.~\ref{eq:parallel_sampler}), replacing what would otherwise require three separate LLM calls or a complex rule engine.

\subsection{Recommendation and Ranking}

Content recommendation systems (news feeds, product listings, ad selection) require scoring thousands of candidate items per user request. The scoring function, ``how relevant is item $i$ to user $u$ in context $c$?'', is a natural Score-head task. While collaborative filtering models dominate for pure engagement prediction, System~1 models add value in the reranking stage where textual signals (item descriptions, user queries, review sentiment) must be incorporated. Cross-encoder rerankers (Section~\ref{sec:cross_encoder}) are already deployed for this purpose; the Family~III generalization adds calibrated confidence to the ranking score, enabling downstream logic such as ``show item only if relevance confidence exceeds $\theta$.''

\paragraph{Decision-oriented reranking.} \citet{lyu2026reranking} conduct a controlled empirical study of Jev for personalized recommendation reranking on three e-commerce review domains (Movies \& TV, Video Games, Books), comparing it against recommendation-specific models (SASRec, DCNv2) and pointwise/listwise Qwen~2.5 7B rerankers. Reranking is formulated as a structured Choice problem: the user's interaction history defines the decision state, candidate items define the alternatives, and Jev's output probabilities serve directly as ranking scores. The study finds that Jev maintains strong recommendation effectiveness (NDCG@10) relative to the evaluated baselines while exhibiting substantially more gradual latency growth as the candidate set scales from $K{=}20$ to $K{=}200$, placing it in a distinct quality--latency operating regime between fast recommendation-specific models and expressive but slower LLM rerankers. This empirical evidence supports the broader thesis that structured decision models offer practical advantages over both specialized scoring models and general-purpose LLMs for ranking tasks with well-defined output spaces.

\subsection{Wireless and Telecom Network Control}

Wireless network control poses a natural fit for System~1 models: decisions (antenna selection, resource-block allocation, network-slice configuration) must be made from finite feasible sets under stringent latency and reliability constraints, often at sub-second cadence.

\paragraph{System~1 for wireless.} \citet{rahimi2026wireless} formalize the applicability of System~1 models across physical-layer control, radio resource management, mobility, network slicing, and network operations, and benchmark Jev against generative LLMs and conventional baselines on representative wireless tasks. In receive-antenna selection, Jev delivers up to $8.5{\times}$ reduction in median response latency relative to the evaluated LLMs, though with a quality gap compared to task-specific alternatives. In intent-conditioned RAN slicing, Jev achieves utility comparable to LLMs while providing more than $3.5{\times}$ latency reduction. Complementary evidence from edge-service orchestration shows that faster decisions do not necessarily translate into lower end-to-end service latency, exposing a fundamental quality/latency tradeoff. These results position System~1 models as a promising decision interface for latency-sensitive, bounded, intent-driven wireless control, rather than as replacements for numerical optimization.

\subsection{Multi-Modal Decision-Making}
\label{sec:visual_decisions}

Recent work extends the System~1 decision paradigm beyond text to visual and audio inputs, demonstrating that the core principles of structured output, fast inference, and calibrated confidence transfer across modalities.

\paragraph{Neuroscience-inspired models.} \citet{su2024neuraldynamics} present a neural dynamics model of visual decision-making that draws on the primate dorsal visual pathway. Unlike standard CNNs, the model explicitly separates perceptual encoding from the decision stage, mirroring the System~1 architecture of a shared encoder feeding a lightweight decision head. Neuroimaging-informed fine-tuning improves decision accuracy while matching human behavioral patterns, suggesting that biologically grounded inductive biases can benefit visual System~1 models.

\paragraph{Visual Jev.} \citet{yu2026visualjev} introduce Visual Jev, which encodes a shared image-and-context prefix once and then executes a batch of independent forced-choice questions as isolated suffixes. At $N{=}32$ questions per image, shared batched execution is $8.9{\times}$ faster than independent serial execution. Answer-supervised post-training raises macro accuracy from 70.6\% to 76.1\%, and the standard language-model head readout matches a purpose-built typed head, confirming that the Jev paradigm of ``adapt the backbone, retain the existing readout, share execution'' transfers directly to vision.

\paragraph{PixelJev.} \citet{zhou2026pixeljev} present PixelJev, a native-image decision interface that maps an image, a task instruction, and a runtime candidate set to structured choices with candidate-conditioned probabilities using small open multimodal models. Across seven benchmark evaluations, 64-shot source adaptation raises Pets accuracy from 60.1\% to 92.4\% and transfers to new texture labels and visual QA without target fitting. The study also identifies remaining gaps: specialist DINOv2 probes remain stronger on source recognition tasks, accuracy gains do not ensure calibrated target probabilities, and cross-family transfer requires further work.

\paragraph{AudioJev.} \citet{lv2026audiojev} extend the Jev paradigm to raw audio, converting a waveform, a natural-language question, and the available answer options into a probability distribution over candidates within one jointly trained model. A central contribution is \emph{order calibration}: ensuring that each answer's predicted probability remains stable when the options are reordered without changing their meaning. To enforce this property, training constructs for every question a second view whose alternatives have all been moved to new positions via random derangement, then minimizes the symmetric KL divergence between the two resulting distributions. Across three training seeds, AudioJev attains 68.9\%/55.3\% mean accuracy on the MMAU/MMAR benchmarks and lowers order-sensitivity (measured by random-permutation SKL) by 43.8\% and 60.5\% on the two benchmarks compared with the single-view baseline, handling intent classification, environmental sound recognition, note properties, speech activity detection, and conversational transitions within a single model.

\paragraph{Jev-Mobile.} \citet{zhang2026jevmobile} demonstrate that System~1 decision models can serve as lightweight executors within mobile GUI agents. In Jev-Mobile, a vision-language model (VLM) specifies high-level goals at low frequency, while Jev selects actions from the structured accessibility tree at high frequency, enabling multiple GUI actions per VLM call. Evaluated on AndroidWorld, the system completes 79\% of tasks (comparable to 78\% for SeeAct-V and 84\% for a step-wise VLM baseline) while cutting average wall-clock time by 32.7\% and average API cost by 73.4\% on completed tasks. Separating the VLM's strategic planning from Jev's fine-grained action selection exemplifies the cascade pattern of Section~\ref{sec:cascades}, with Jev operating as the fast System~1 layer beneath a slower System~2 planner.

\paragraph{Implications.} These results collectively establish that multi-modal System~1 decision models are already viable across vision, audio, and GUI interaction, not merely a future direction. The shared-prefix batching strategy of Visual Jev, the candidate-conditioned readout of PixelJev, the order-calibration objective of AudioJev, and the VLM--Jev cascade of Jev-Mobile demonstrate that the architectural patterns catalogued in Sections~\ref{sec:jev}--\ref{sec:clm} for text generalize to diverse modalities. Open challenges remain in cross-modal calibration, unified multi-modal architectures that handle text, vision, and audio within a single model, and scaling to high-resolution inputs under latency constraints.

\subsection{Cybersecurity and Perimeter Triage}
\label{sec:cyber}

Cybersecurity perimeter defenses require decisions at wire speed: every incoming network packet, API request, or email attachment must be classified as benign, suspicious, or malicious within single-digit milliseconds. The output space is structured (allow/block/inspect) and the consequence asymmetry is severe (false negatives enable breaches; false positives disrupt operations).

\paragraph{VX-One.} \citet{vxone2026} introduce VX-One, a System~1 model with \emph{typed decision heads} specialized for cybersecurity perimeter triage. Rather than the generic Choice/Score/Noul heads, VX-One defines domain-specific heads that output structured decisions: \texttt{\{action: "block"|"allow"|"inspect", confidence: float, threat\_category: str, severity: int\}}. VX-One is a compact network (100K--10M parameters) with temperature calibration, achieving 0.7--1.7\,ms inference on CPU, fast enough for inline packet inspection. On a benchmark of 50K labeled perimeter events, VX-One matches a fine-tuned DeBERTa classifier in accuracy while providing calibrated severity scores that enable risk-aware policy enforcement.

\paragraph{Prompt injection defense.} \citet{promptinjdetection2026} demonstrate that System~1 models can serve as a defense layer against prompt injection attacks on LLM applications. A Noul head trained on labeled injection examples achieves AUROC 0.871 on 47,611 test examples at 8\,ms latency, enabling every user input to be screened before reaching the LLM. At the default threshold, the model still misses 64.9\% of attacks, highlighting the difficulty of the task. This creates a ``System~1 firewall'' pattern: a fast binary classifier guards a slow generative model, analogous to how packet filters guard application servers in network security.

\subsection{Theory of Mind}

\paragraph{Jev-ToM.} \citet{jevtom2026} evaluate System~1 decision models on theory of mind (ToM) tasks, assessing whether these models can determine what a person \emph{believes}, \emph{intends}, or \emph{knows} based on a scenario description. Framing ToM as a structured decision (``Does Alice believe the ball is in the box?'' $\to$ Noul), they evaluate on 11,117 ToM questions spanning false-belief, knowledge attribution, and intention recognition. Jev achieves 73.2\% accuracy, significantly above chance (50\%) but below frontier LLMs ($\sim$85\%), suggesting that basic mental-state attribution is partially accessible to System~1 processing but complex ToM reasoning requires System~2 deliberation. This finding has practical implications for social AI applications (chatbot empathy detection, negotiation support) where fast but imperfect ToM assessment may be sufficient.

\subsection{Political Science}

\paragraph{Replication studies.} \citet{polisci2026} replicate seven political text classification studies using System~1 decision models, covering party manifesto coding, legislative speech classification, media bias detection, and sentiment analysis of political tweets. Across the seven studies, zero-shot Jev Choice achieves 78--92\% of the accuracy of task-specific fine-tuned classifiers. The authors find no significant cost advantage over LLMs at batch API prices, though the latency advantage remains. For three studies, the zero-shot System~1 model \emph{outperforms} the original study's reported classifier, likely because the original classifiers were trained on small datasets ($<$2K examples) where Jev's broader pretraining provides an advantage. The authors note that the calibrated confidence scores are particularly valuable for political science: researchers can report not just classifications but uncertainty-weighted estimates, enabling more nuanced quantitative analysis.

\subsection{Text Quantification (Learning to Quantify)}
\label{sec:quantification}

Text quantification -- also called \emph{learning to quantify} or \emph{class prevalence estimation} -- is the supervised learning task in which a model is trained to predict the relative frequency (prevalence) of each class in an unlabelled bag of documents~\citep{sebastiani2020machine}. Unlike classification, which targets individual instances, quantification targets the \emph{distribution} over an entire collection, making it a natural System~1 task: the output is structured (a prevalence vector summing to one), the decision must be fast (quantification is often applied to millions of documents in monitoring pipelines), and calibrated posteriors directly determine estimation quality.

\paragraph{Classical aggregative methods.} The dominant paradigm is \emph{aggregative quantification}: train a classifier, apply it to the target bag, and aggregate its outputs into a prevalence estimate. Simple Classify \& Count (SCC) merely counts predicted labels; Probabilistic Classify \& Count (PCC) sums posterior probabilities~\citep{bella2010quantification}; Adjusted Classify \& Count (ACC) corrects for misclassification rates; and Expectation Maximisation Quantification (EMQ/SLD) iteratively adjusts posteriors under prior probability shift~\citep{saerens2002adjusting}. \citet{levin2017enhanced} introduced PCC-PAV and PCC-EPAV, which calibrate the classifier's posteriors specifically for the quantification objective via isotonic regression with count-matching constraints, yielding significant improvements over standard PCC on multi-label text datasets.

\paragraph{Transformer-based quantification.} Although the aggregative framework is classifier-agnostic, the shift from traditional classifiers (SVM, Naive Bayes, logistic regression) to transformer-based encoders substantially changes the quality--cost frontier. A fine-tuned BERT or DeBERTa classifier produces richer posterior probabilities and contextual embeddings that benefit every downstream aggregation method. In practical terms, replacing an SVM with a fine-tuned BERT classifier inside ACC or SLD/EMQ is the simplest and most effective way to bring transformers into the quantification pipeline -- the quantification layer requires no modification, only a better base classifier.

\paragraph{Neural quantification architectures.} Three end-to-end neural architectures go beyond the classify-then-aggregate pattern. QuaNet~\citep{esuli2018quanet} processes document embeddings and posterior probabilities through a bidirectional LSTM, augmented with estimates from classical quantifiers (CC, ACC, PCC, PACC, EMQ), to predict prevalence directly. Any classifier that provides both \texttt{predict\_proba()} and \texttt{transform()} (embeddings) can serve as the base -- a fine-tuned BERT encoder is a natural fit. HistNetQ~\citep{perezmon2024histnetq} eliminates the intermediate classifier entirely: instances pass through an optional feature extraction module, a sigmoid activation, and a differentiable histogram layer, followed by an MLP that predicts class prevalence. The feature extraction module can be a transformer encoder, enabling fully end-to-end transformer-based quantification. HistNetQ outperformed all methods in the LeQua 2022 competition~\citep{esuli2022lequa}. GMNet~\citep{perezmon2025gmnet} replaces the histogram with learned Gaussian mixture representations in latent space, achieving state-of-the-art results over both HistNetQ and classical methods, and similarly accepting a transformer as its feature extractor. The QuaPy framework~\citep{moreo2021quapy} provides reference implementations of all three architectures alongside classical methods.

\paragraph{Evaluating quantification.} Quantification evaluation differs fundamentally from classification evaluation: the unit of evaluation is a \emph{bag} (a set of documents with a known prevalence vector), not an individual instance. Standard protocols generate hundreds or thousands of test bags with varying class prevalence -- the Uniform Prevalence Protocol (UPP) samples prevalence vectors uniformly from the simplex, while the Natural Prevalence Protocol (NPP) preserves the original class balance~\citep{moreo2021quapy}. The LeQua competitions~\citep{esuli2022lequa} institutionalized this protocol for benchmarking.

Several metrics quantify the deviation between predicted and true prevalence vectors. Let $\hat{p}$ and $p$ denote the predicted and true prevalence vectors over $m$ classes. The most common metrics are:

\begin{itemize}
    \item \textbf{Absolute Error (AE):} $\text{AE} = \frac{1}{m} \sum_{i=1}^{m} |\hat{p}_i - p_i|$, the mean absolute deviation per class.
    \item \textbf{Relative Absolute Error (RAE):} $\text{RAE} = \frac{1}{m} \sum_{i=1}^{m} \frac{|\hat{p}_i - p_i|}{p_i}$, which penalises errors on rare classes more heavily.
    \item \textbf{KL Divergence (KLD):} $\text{KLD} = \sum_{i=1}^{m} p_i \log \frac{p_i}{\hat{p}_i}$, measuring the information-theoretic divergence between the true and predicted distributions.
\end{itemize}

\noindent For multi-label quantification, where each label has an independent count rather than a single prevalence vector, \citet{levin2017enhanced} proposed the \textbf{Concordance Ratio (CR)}:
\begin{equation}
    \text{CR}_{\mu}(\mathcal{C}|D) = \frac{1}{|\mathcal{C}|} \sum_{c \in \mathcal{C}} \frac{\min(\mu(c|D),\; n(c|D))}{\max(\mu(c|D),\; n(c|D))},
    \label{eq:concordance_ratio}
\end{equation}
where $\mu(c|D)$ is the estimated count and $n(c|D)$ the actual count for label $c$ in dataset $D$. CR equals 1 when all estimates are perfect, 0 when any label is completely missed, and lies in $(0,1)$ otherwise. CR is symmetric with respect to over- and under-estimation and naturally handles the multi-label setting where labels are not mutually exclusive.

\paragraph{System~1 implications.} Quantification is a downstream consumer of System~1 classifiers: the better calibrated the classifier's posteriors, the more accurate the resulting prevalence estimates. The PCC-EPAV method~\citep{levin2017enhanced} makes this connection explicit -- it re-calibrates posteriors specifically for count accuracy rather than classification accuracy, demonstrating that calibration objectives should match the downstream task. Transformer-based classifiers, with their richer representations and better-calibrated softmax outputs (especially after temperature scaling), are therefore natural building blocks for modern quantification pipelines. An open research direction is whether LLM zero-shot or few-shot classification can serve as the base for aggregative quantification, though the uncalibrated nature of LLM probabilities~\citep{geng2024survey} poses challenges under the prior probability shift assumption that most quantification methods require.

\subsection{Tabular Data}

\paragraph{Typed decisions on structured inputs.} \citet{tabdecide2026} extend the System~1 paradigm to tabular data by serializing table rows into natural-language state descriptions (``Customer: age=34, income=75K, tenure=3yr, last\_purchase=electronics'') and applying standard decision heads. On four tabular classification benchmarks (credit scoring, churn prediction, medical diagnosis, fraud detection), this approach is competitive with gradient-boosted trees (XGBoost, LightGBM) in the low-data regime, but trained tree models surpass the decision models once approximately 64 or more labeled rows are available. The calibrated outputs enable direct integration with downstream threshold-based logic (e.g., ``approve if $p_{\text{creditworthy}} > 0.7$'') without the post-hoc calibration step that tree ensembles require.

\subsection{Low-Resource Languages}

Multilingual evaluation of System~1 models has expanded significantly beyond Laya's initial five primary languages (with additional languages served via mmBERT routing).

\paragraph{Turkish.} HakemBench~\citep{hakembench2026} provides 12 evaluation tasks in Turkish, revealing that English-trained models exhibit degraded calibration (ECE $>$ 12\%) even when accuracy remains acceptable ($>$80\%). Tachyone's multilingual backbone~\citep{tachyone2026}, which includes Turkish in its ModernBERT pretraining, achieves substantially better calibration (ECE $<$ 6\%) on the same tasks.

\paragraph{Indonesian regional languages.} \citet{indonesian2026} evaluate System~1 models on four Indonesian regional languages (Javanese, Sundanese, Balinese, Minangkabau), finding that zero-shot transfer from Indonesian-trained models degrades rapidly for Balinese and Minangkabau (accuracy drops 15--20\%) but remains viable for Javanese and Sundanese (accuracy drops $<$5\%). This mirrors findings in the LLM literature on low-resource language transfer and suggests that multilingual System~1 models face the same long-tail challenge.

\paragraph{Unified audio decisions.} \citet{qwenaudiojev2026} extend the audio decision paradigm (complementing AudioJev's work in Section~\ref{sec:visual_decisions}) to a unified pipeline covering speech commands, music genre classification, and environmental sound recognition within a single model. By combining Qwen-Audio's pretrained audio encoder with Jev-style decision heads, the system handles diverse audio decision tasks without task-specific fine-tuning, achieving 71.3\% mean accuracy across eight audio benchmarks at 45\,ms per decision.

\section{System~1 Models in Agentic Workflows}
\label{sec:agentic}

Agentic AI systems, autonomous agents that interact with tools, APIs, and environments over multi-step trajectories~\citep{yao2023react,shinn2023reflexion}, create a natural demand for System~1 decision models. Each iteration of an agent loop involves multiple fast decisions (tool selection, safety gating, verification, priority scoring) that must complete in milliseconds to avoid compounding latency across a trajectory of 10--100+ steps. For a broader treatment of agentic AI foundations, architectures, and training methodologies, readers are encouraged to consult \citet{roitman2026agentic}.

\subsection{The Agent Harness as a Decision Architecture}

Following the Cognitive Architectures for Language Agents (CoALA) framework~\citep{sumers2024cognitive}, a production agent harness consists of four components: \textbf{tools} (extend the action space), \textbf{middleware} (inspect/modify/block the execution loop), \textbf{skills} (workflow instructions), and \textbf{memory} (accumulated knowledge). System~1 models serve as middleware at each decision point within this architecture.

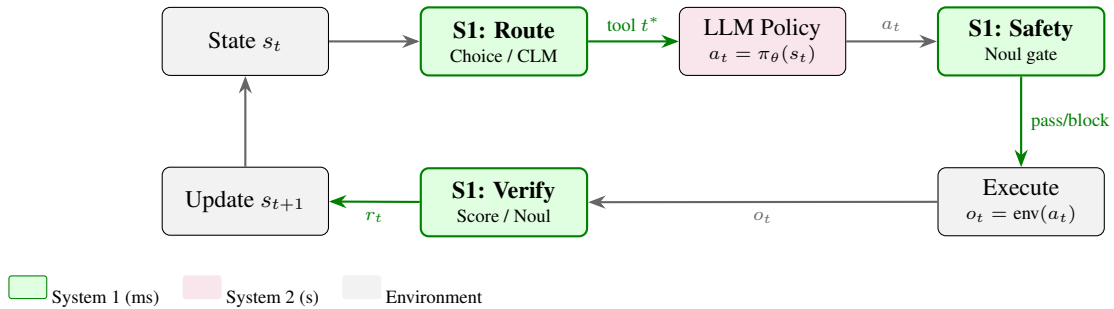
\begin{figure}[H]
\centering
\begin{tikzpicture}[
    >=Stealth,
    s2/.style={draw, rounded corners=3pt, minimum width=22mm, minimum height=9mm, font=\small, align=center, fill=purple!10},
    s1/.style={draw, rounded corners=3pt, minimum width=22mm, minimum height=9mm, font=\small, align=center, fill=green!12, line width=0.8pt, draw=green!50!black},
    env/.style={draw, rounded corners=3pt, minimum width=22mm, minimum height=9mm, font=\small, align=center, fill=gray!10},
    arr/.style={->, thick, color=black!60},
    s1arr/.style={->, thick, color=green!50!black},
    node distance=10mm,
]
\node[env] (state) {State $s_t$};
\node[s1, right=12mm of state] (route) {\textbf{S1: Route}\\[-1pt]\scriptsize Choice / CLM};
\node[s2, right=12mm of route] (llm) {LLM Policy\\[-1pt]\scriptsize $a_t = \pi_\theta(s_t)$};
\node[s1, right=12mm of llm] (gate) {\textbf{S1: Safety}\\[-1pt]\scriptsize Noul gate};
\node[env, below=12mm of gate] (exec) {Execute\\[-1pt]\scriptsize $o_t = \text{env}(a_t)$};
\node[s1, below=12mm of route] (verify) {\textbf{S1: Verify}\\[-1pt]\scriptsize Score / Noul};
\node[env, below=12mm of state] (update) {Update $s_{t+1}$};

\draw[arr] (state) -- (route);
\draw[s1arr] (route) -- node[above, font=\scriptsize]{tool $t^*$} (llm);
\draw[arr] (llm) -- node[above, font=\scriptsize]{$a_t$} (gate);
\draw[s1arr] (gate) -- node[right, font=\scriptsize]{pass/block} (exec);
\draw[arr] (exec) -- node[below, font=\scriptsize]{$o_t$} (verify);
\draw[s1arr] (verify) -- node[below, font=\scriptsize]{$r_t$} (update);
\draw[arr] (update) -- (state);

\node[below=4mm of update, font=\scriptsize, align=center] {%
\tikz[baseline=-0.5ex]{\fill[green!12, draw=green!50!black, rounded corners=1pt] (0,0) rectangle (0.5,0.3);} System~1 (ms) \quad
\tikz[baseline=-0.5ex]{\fill[purple!10, draw, rounded corners=1pt] (0,0) rectangle (0.5,0.3);} System~2 (s) \quad
\tikz[baseline=-0.5ex]{\fill[gray!10, draw, rounded corners=1pt] (0,0) rectangle (0.5,0.3);} Environment};
\end{tikzpicture}
\caption{System~1 decision points within an agent loop iteration. Three fast decisions (green, milliseconds) orchestrate the control flow around one slow generative step (purple, seconds). A loop with 20 iterations makes ${\sim}60$ System~1 decisions.}
\label{fig:agent_loop_s1}
\end{figure}

We formalize the agent loop as a Markov Decision Process $(\mathcal{S}, \mathcal{A}, T, R, \gamma)$ where the policy $\pi_\theta$ is a frozen LLM. At each step $t$, the loop state evolves as:
\begin{align}
    s_t &= \text{context}(s_{t-1}, a_{t-1}, o_{t-1}), \label{eq:loop_state}\\
    a_t &= \pi_\theta(s_t), \quad o_t = \text{env}(a_t), \quad r_t = \text{verify}(s_t, a_t, o_t). \nonumber
\end{align}
System~1 models intervene at three points, illustrated in Figure~\ref{fig:agent_loop_s1}: \emph{before} the policy acts (routing), \emph{during} execution (safety gating), and \emph{after} observation (verification).

\subsection{System~1 Decision Points in Agent Loops}

\paragraph{Tool routing.} Given state $s_t$ and a tool catalog $\mathcal{T} = \{t_1, \ldots, t_K\}$, a System~1 router selects the tool via:
\begin{equation}
    t^* = \arg\max_{t_k \in \mathcal{T}} f_{\text{S1}}(s_t, t_k),
\end{equation}
where $f_{\text{S1}}$ is a contrastive model (CLM-style) or Choice head (Jev-style). With cached tool embeddings, routing costs $O(d')$ per tool, negligible compared to the LLM's generation cost.

\paragraph{Safety gating.} Before executing action $a_t$, a Noul model evaluates safety: $p_{\text{safe}} = \sigma(W_n \mathbf{h}(s_t, a_t) + b_n)$. If $p_{\text{safe}} < \theta_{\text{safe}}$, the action is blocked and the agent is prompted to revise. This mirrors the human-in-the-loop approval gate pattern, but operates autonomously at machine speed.

\paragraph{Verification nodes.} Loop engineering identifies verification as the ``reward function'' of the agent loop~\citep{shinn2023reflexion,anthropic2024effective}. System~1 verifiers provide fast, deterministic-like signals: a Score model evaluates solution quality, a Noul model checks postconditions, and the loop continues until $r_t > \theta_{\text{success}}$ or the budget is exhausted. CLM-8B's success as a code verifier on DeepSWE and Terminal-Bench~2.1~\citep{datacurve2026deepswe,harbor2026terminalbench} demonstrates that contrastive System~1 models can replace generative verifiers at 4--5$\times$ lower latency, a critical advantage when verification is called at every loop iteration.

\subsection{System~1 Models as Loop Infrastructure}

Production agent loops consist of five structural primitives: automations (scheduled triggers), worktrees (parallel isolation), skills (reusable workflows), connectors (tool interfaces), and sub-agents (delegated verifiers). Each primitive creates decision points that are natural fits for System~1 models, as shown in Table~\ref{tab:loop_s1}. The key insight is that while the \emph{generative} policy (the LLM) is a System~2 component, the \emph{control-flow decisions} that orchestrate the loop, what to invoke, whether it is safe, whether the result is good enough, are System~1 tasks. A loop with 20 iterations and 3 System~1 decision points per iteration (routing, gating, verification) makes 60 fast decisions, each of which must complete in milliseconds to keep total overhead under 1 second. This is the primary deployment scenario driving demand for sub-100ms System~1 inference.

\begin{table}[H]
\centering
\caption{System~1 decision points within five loop engineering primitives.}
\label{tab:loop_s1}
\small
\begin{tabular}{llll}
\toprule
\textbf{Loop Primitive} & \textbf{System~1 Role} & \textbf{Decision Type} & \textbf{Model Family} \\
\midrule
Automations & Trigger classification & Should this event fire the loop? & Noul / Choice \\
Worktrees & Conflict detection & Are parallel branches compatible? & Score \\
Skills & Skill selection & Which workflow template applies? & Contrastive (CLM) \\
Connectors & Tool routing & Which external system to invoke? & Choice / Contrastive \\
Sub-agents & Task decomposition & Route subtask to which specialist? & Choice / Score \\
\bottomrule
\end{tabular}
\end{table}

\paragraph{Harness distillation implications.} The Harness-Zero approach~\citep{ye2026harnesszero} demonstrates that harness middleware behaviors, including System~1 routing and safety checks, can be distilled into the generative model's weights, eliminating the need for separate System~1 components at deployment. This raises an open question for the field: as generative models absorb more System~1-style decision capabilities through distillation, will dedicated System~1 models remain necessary, or will they become primarily a \emph{training-time} artifact used to generate the decision signals that are later baked into weights?

\subsection{System~1-Controlled Agentic Memory}

The CoALA framework identifies \emph{memory} as a core agent component alongside tools, middleware, and skills. As agents operate over long horizons, they accumulate observations, preferences, and task history that exceed any fixed context window. Managing this memory (deciding what to store, how to organize it, which memories to retrieve, and when to stop searching) involves frequent, structured decisions that are poorly served by autoregressive generation.

\citet{jiang2026jevmem} introduce \textbf{Jev-Mem}, a System~1/System~2 architecture that makes memory control a first-class systems layer. A lightweight System~1 controller (realized via Jev's typed decision heads) governs both the \emph{write path} (memory typing, redundancy filtering, and multi-relational edge construction over semantic, temporal, causal, and entity relations) and the \emph{read path} (query routing across relational views, retrieval-budget allocation, candidate scoring, and adaptive stopping). System~2 (the LLM) is invoked only for final answer synthesis, removing autoregressive generation from the memory critical path entirely.

The key architectural insight is that memory-control decisions are \emph{semantic but not generative}: they produce bounded outputs (labels, probabilities, scores) rather than free-form text, making them natural System~1 tasks. For each query $q$, the controller predicts per-relation relevance $p_g(q)$ and allocates search budget proportionally, then iteratively expands the memory graph until an evidence-sufficiency check indicates enough material has been collected or additional traversal is unlikely to help.

On the LoCoMo long-conversation benchmark, Jev-Mem achieves the highest overall LLM-as-a-Judge score (0.777, an 11.0\% relative improvement over the strongest baseline) while reducing memory construction time to 158\,s (6.6$\times$ speedup over the fastest competing system) and average query latency to 0.93\,s (36.7\% reduction). These results demonstrate that separating fast memory control from deliberative reasoning improves both effectiveness and efficiency, a pattern consistent with the broader System~1/System~2 thesis of this guide.

\section{System~1 + System~2 Cascades}
\label{sec:cascades}

\begin{figure}[H]
\centering
\begin{tikzpicture}[
    >=Stealth,
    block/.style={draw, rounded corners=3pt, minimum width=22mm, minimum height=9mm, font=\small, align=center, fill=#1},
    decision/.style={draw, diamond, aspect=1.8, minimum width=16mm, font=\small, fill=orange!12},
    arr/.style={->, thick, color=black!60},
    node distance=9mm and 12mm,
]
\node[block=gray!10] (inp) {Input $x$};
\node[block=green!12, right=12mm of inp] (s1) {System~1\\[-1pt]\scriptsize $f_{\text{S1}}(x), \hat{p}$};
\node[decision, right=14mm of s1] (conf) {$\hat{p} \geq \theta$?};
\node[block=green!20, above right=8mm and 14mm of conf] (fast) {\textbf{Fast path}\\[-1pt]\scriptsize output $f_{\text{S1}}(x)$};
\node[block=purple!12, below right=8mm and 14mm of conf] (s2) {System~2 (LLM)\\[-1pt]\scriptsize $f_{\text{S2}}(x)$};
\node[block=blue!10, right=28mm of conf] (out) {Final\\Output};

\draw[arr] (inp) -- (s1);
\draw[arr] (s1) -- (conf);
\draw[arr] (conf) -- node[above left, font=\scriptsize]{yes (majority)} (fast);
\draw[arr] (conf) -- node[below left, font=\scriptsize]{no (minority)} (s2);
\draw[arr] (fast) -- (out);
\draw[arr] (s2) -- (out);

\node[below=2mm of s1, font=\tiny\itshape, text=green!50!black]{10--70ms};
\node[below=2mm of s2, font=\tiny\itshape, text=purple!50!black]{500ms--5s};
\end{tikzpicture}
\caption{Confidence-based cascade: System~1 handles the majority of decisions at millisecond latency. Only uncertain cases ($\hat{p} < \theta$) escalate to the slower, more powerful System~2 model. Calibration quality directly determines cascade efficiency.}
\label{fig:cascade_flow}
\end{figure}
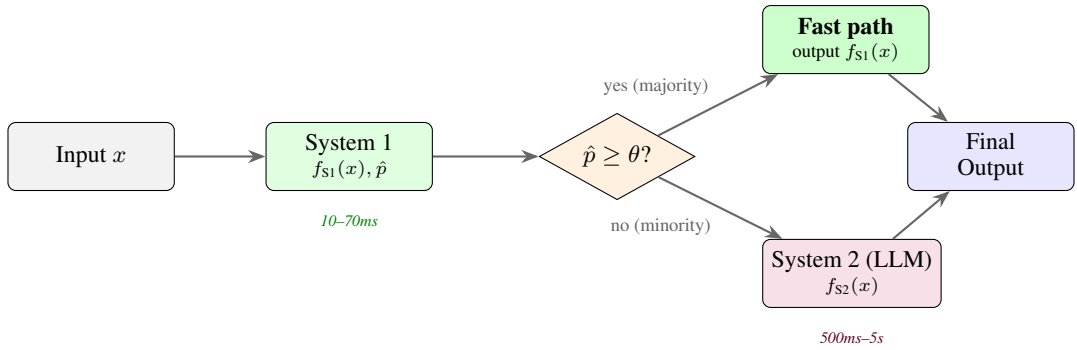

A particularly promising design pattern combines fast System~1 models with slow System~2 models in cascading architectures (Figure~\ref{fig:cascade_flow}). The motivation is economic and operational: in most real-world decision workloads, the majority of inputs are routine cases that a lightweight model can handle confidently, while only a minority require the full reasoning capacity of a large LLM. Sending every input to the expensive model wastes compute on easy cases; using only the cheap model sacrifices accuracy on hard ones. A cascade captures the best of both regimes: the System~1 model acts as a fast first pass, resolving the easy majority at millisecond latency and pennies per million decisions, and only escalating uncertain cases to the System~2 model. The result is near-System~2 accuracy at a fraction of the cost and latency, with the exact savings determined by the escalation rate.

The viability of this pattern rests entirely on the System~1 model's calibration quality. If the model's confidence scores are well-calibrated, the escalation threshold $\theta$ can be set to achieve a predictable quality--cost trade-off: raising $\theta$ escalates more cases (higher accuracy, higher cost), lowering it retains more (lower cost, slightly lower accuracy). If confidence is miscalibrated, the gate makes unreliable routing decisions, either escalating too many easy cases (negating cost savings) or retaining hard cases the model cannot handle (degrading accuracy silently). This tight coupling between calibration and cascade efficiency is why calibration occupies such a central role in the System~1 literature.

\subsection{Confidence-Based Routing}

The simplest cascade uses System~1 confidence to decide whether to escalate to System~2:

\begin{equation}
    \text{output} = \begin{cases}
        f_{\text{S1}}(x) & \text{if } \text{conf}(f_{\text{S1}}(x)) \geq \theta, \\
        f_{\text{S2}}(x) & \text{otherwise},
    \end{cases}
\end{equation}

where $\theta$ is a confidence threshold tuned on validation data. This requires the System~1 model to be well-calibrated, otherwise the routing decisions will be unreliable.

We formalize this as ``cognitive escalation,'' where the confidence threshold $\theta$ can be dynamic, adapting to the task difficulty and available compute budget. The System~1 model handles routine, high-confidence decisions (the exact proportion depends on the domain, task difficulty distribution, and quality requirements), while the System~2 model is reserved for ambiguous or complex cases. Illustrative escalation rates of 10--30\% are common in practice, though the actual rate depends heavily on domain difficulty and threshold choice; these figures have not been independently verified across diverse applications.

\begin{lstlisting}[caption={Generic confidence-gated cascade with three-zone routing.},label={lst:cascade}]
from dataclasses import dataclass
from typing import Any, Callable
import time

@dataclass
class CascadeResult:
    decision: Any
    confidence: float
    path: str          # "system1", "system2", or "default"
    latency_ms: float

class CascadeRouter:
    def __init__(self, system1_fn: Callable, system2_fn: Callable,
                 accept_threshold=0.85, reject_threshold=0.30):
        self.s1 = system1_fn
        self.s2 = system2_fn
        self.hi = accept_threshold
        self.lo = reject_threshold
        self.stats = {"s1": 0, "s2": 0, "default": 0}

    def decide(self, x) -> CascadeResult:
        t0 = time.perf_counter()
        r = self.s1(x)
        decision, conf = r["decision"], r["confidence"]

        if conf >= self.hi:
            self.stats["s1"] += 1
            path = "system1"
        elif conf <= self.lo:
            self.stats["default"] += 1
            decision, path = "ESCALATE_TO_HUMAN", "default"
        else:
            self.stats["s2"] += 1
            decision = self.s2(x)["decision"]
            path = "system2"

        return CascadeResult(decision, conf, path,
                             (time.perf_counter() - t0) * 1000)

    @property
    def escalation_rate(self):
        t = sum(self.stats.values())
        (*@return (self.stats["s2"] + self.stats["default"]) / t if t else 0.0@*)

# (*@Example: 20\% escalation -> cascade cost = 0.8*C1 + 0.2*C2 (about 4.8x cheaper than pure S2)@*)
cascade = CascadeRouter(system1_moderate, system2_moderate,
                        accept_threshold=0.85, reject_threshold=0.20)
for item in content_queue:
    result = cascade.decide({"text": item})
print(f"Escalation rate: {cascade.escalation_rate:.1%}")
\end{lstlisting}

\begin{warningbox}
Cascades have two failure modes:
\begin{itemize}
\item \textbf{Near-chance base accuracy:} If System~1 performs near random on the target task, even perfect calibration cannot save the cascade -- it will escalate 90--100\% of items, negating all cost savings (observed with CLM on judging tasks~\citep{nandakishore2026clmjudge}).
\item \textbf{Correlated errors:} If System~1 and System~2 make the same mistakes, escalation does not help. In rubric judging, 96\% of LLM verdicts repeated Jev's confident wrong answers~\citep{rao2026rubricjudges}.
\end{itemize}
\textbf{Before building a cascade:} (1)~measure System~1 accuracy ($<$60\% means cascading will not help); (2)~measure error correlation between the two models ($>$80\% means the ceiling is low).
\end{warningbox}

\subsection{System~1 Models as Judges}
\label{sec:s1_judge}

LLM-as-a-Judge~\citep{zheng2023judging} is the dominant paradigm for scalable evaluation of language model outputs, but it is expensive, suffers from well-documented position bias and length bias~\citep{wang2024llmnotfair}, and produces confidence estimates only through token sampling. A natural question is whether System~1 decision models can serve as fast, calibrated, bias-free replacements.

\paragraph{JEV-as-a-Judge.} \citet{li2026jevasjudge} introduced a cascade architecture for LLM evaluation in which a Jev-style model scores all candidate responses in parallel (${\sim}$10\,ms each). Only responses where the System~1 confidence falls in an uncertain range ($\theta_{\text{low}} < s < \theta_{\text{high}}$) are forwarded to a full LLM judge. The authors report 70--80\% cost reduction with $<$2\% accuracy loss compared to full LLM judging, establishing the cascade template for System~1-based evaluation.

\paragraph{CLM-as-a-Judge.} \citet{nandakishore2026clmjudge} provide an independent, preregistered evaluation of CLM-v0.1-8B as a judge on six public benchmarks (RewardBench, RewardBench~2, RM-Bench, JudgeBench, HaluEval), measured against generative, reward-model, and trivial baselines under the same scoring rules. The accuracy finding is negative: CLM scores between 0.351 (best-of-four, chance 0.250) and 0.593 (pairwise, chance 0.500), cannot be distinguished from random guessing on RM-Bench and JudgeBench, and trails same-size autoregressive judges by large margins (reward model: 0.764--0.976; generative judge: 0.611--0.778). Fine-tuning new decision heads on preference data yields at most +0.060 accuracy, changing none of these conclusions.

Two structural properties do hold. First, \emph{calibration}: raw confidences are overconfident by up to +0.401, yet a single pooled temperature ($T{=}20.1$) fitted on held-out items reduces ECE to at most 0.062 on every benchmark. Second, \emph{order and length stability}: CLM's decision flip rate under answer-order swap is 0.0002, compared to 0.2188 for the generative judge, and its length-preference shift is $-0.023$ versus $-0.217$. These properties come from the contrastive architecture (candidates are scored independently of presentation order), not from accuracy.

However, the confidence-gated cascade escalates 92.3--100\% of items to the strong judge at the preregistered 0.97 retention bar: when the underlying judge operates near chance, even well-calibrated confidence scores leave almost no items that the gate can confidently retain. This result provides a cautionary complement to the JEV-as-a-Judge findings and highlights that calibration alone is insufficient for effective cascading; the System~1 model must also achieve non-trivial base accuracy on the target task.

\paragraph{Rubric judging.} \citet{rao2026rubricjudges} benchmark Jev alongside three lightweight LLM judges across nine evaluation panels drawn from seven datasets that include human ground-truth, with all judges receiving the same rubric descriptions. LLM judges cost 16--325$\times$ more and take 28--350$\times$ longer, yet statistically significant accuracy differences appear in no more than 8 out of 27 pairwise tests, with Jev tending to outperform on binary checklist items and underperform only on ordinal scales. The central finding concerns correlated errors: on Jev's most confident mistakes, approximately 96\% of LLM verdicts repeat the same wrong answer (independent errors would yield ${\sim}$50\%). Consequently, Jev--LLM cascades with oracle thresholds yield at most +2.7 accuracy points over the best single judge, because the fallback errs in the same places. This result refines the cascade design principle: calibration tells the cascade \emph{when} to defer, but effective cascading requires judges whose errors are complementary, not merely a fast judge followed by a stronger one.

\subsection{Speculative Decision-Making}

Analogous to speculative decoding~\citep{leviathan2023fast} in text generation, \textbf{speculative decision-making} uses a fast System~1 model to propose decisions that a System~2 model verifies. The System~1 model proposes a decision with confidence $p$. If $p > \theta_{\text{accept}}$, the decision is accepted immediately with no System~2 call. If $p < \theta_{\text{reject}}$, it is rejected immediately and a safe default action is taken. In the intermediate range, the System~2 model verifies and may override.

This pattern is particularly useful in multi-step agentic systems where each step involves a routing decision, and the total trajectory latency is the sum of individual decision latencies.

\subsection{Multi-Agent System~1 Orchestration}

In multi-agent architectures~\citep{wu2023autogen}, a System~1 supervisor model can rapidly route tasks to specialized sub-agents through a pipeline of fast decisions: a Choice model classifies the incoming request type, a contrastive model matches the task embedding to available agent capabilities, a Score model assigns urgency or priority for queue management, and a Noul model checks safety constraints before execution.

Each of these decisions takes 10--50ms, enabling the full routing pipeline to complete in under 200ms, fast enough for real-time interactive systems.

\paragraph{Jev-LiteMAS.} \citet{jevlitemas2026} formalize this pattern in Jev-LiteMAS, a lightweight multi-agent framework where \emph{all} inter-agent routing is handled by System~1 models. The framework defines three routing primitives: \emph{dispatch} (Choice over available agents), \emph{priority} (Score for queue ordering), and \emph{guard} (Noul for capability checking). By moving routing entirely into the System~1 layer, Jev-LiteMAS reduces inter-agent communication overhead from seconds (with LLM-based routing) to under 50\,ms per hop, enabling agent teams of 10+ specialists to coordinate on interactive timescales.

\subsection{System~1-Controlled Memory Compaction}
\label{sec:prune_summary}

As agents operate over long horizons, accumulated context exceeds available windows. \citet{prunesummary2026} introduce \textbf{Prune the Summary}, a source-judged memory compaction method where a System~1 model decides what to keep and what to discard from an agent's accumulated memory. For each memory segment, a Noul head evaluates ``Is this segment relevant to the current task?'' and a Score head assigns an importance weight. Segments below a relevance threshold are pruned; the remaining segments are reweighted by importance and concatenated to form the compacted context.

The key insight is that memory compaction is itself a structured decision task: for each segment, the output is a boolean (keep/prune) plus a scalar (importance weight), exactly the Noul and Score primitives. By delegating compaction to System~1, the agent avoids spending expensive LLM tokens on what is essentially a classification-and-scoring problem. On three long-horizon benchmarks, Prune the Summary achieves 94\% of oracle-compacted performance at 12$\times$ lower compaction cost than LLM-based summarization.

\clearpage
\ifdim\pagetotal>0pt\thispagestyle{empty}\newpage\fi
\part{Evaluation and Robustness}

Deploying a System~1 model requires confidence that it will perform reliably across diverse inputs and adversarial conditions. This chapter covers the benchmarks and evaluation methodology used to measure model quality, and the robustness properties -- position bias, framing sensitivity, and fallback reliability -- that determine whether a model can be trusted in production.

\section{Benchmarks and Evaluation}
\label{sec:benchmarks}

Evaluating System~1 decision models requires benchmarks that measure both decision quality and operational characteristics. Traditional NLU benchmarks (GLUE, SuperGLUE~\citep{wang2019superglue}) test accuracy on classification tasks but ignore calibration, latency, and cost, the very dimensions that distinguish System~1 models from general-purpose alternatives. Reward model benchmarks (RewardBench~\citep{lambert2024rewardbench}) evaluate preference alignment but not the structured decision primitives (Choice, Score, Noul) that Family~III models introduce. Agent verification benchmarks (DeepSWE~\citep{datacurve2026deepswe}, Terminal-Bench~\citep{harbor2026terminalbench}) test whether a model can assess code correctness but do not measure whether the confidence accompanying that assessment is reliable enough to drive a cascade gate.

The result is a fragmented evaluation landscape: no single benchmark currently captures the full profile of a System~1 decision model. A practitioner choosing between Jev, Laya, CLM, and an LLM baseline must stitch together results from multiple benchmarks, each measuring a different slice of performance, often under incompatible evaluation protocols. This section surveys the existing benchmarks, proposes a set of evaluation dimensions that a comprehensive System~1 benchmark should cover, and highlights the gaps that remain.

\subsection{Existing Benchmarks}

Table~\ref{tab:benchmarks} summarizes the key benchmarks currently used to evaluate System~1 decision models. We briefly describe each and note what it does and does not measure from a System~1 perspective.

\paragraph{GLUE and SuperGLUE}~\citep{wang2019superglue} are the canonical NLU evaluation suites, comprising tasks such as natural language inference (MNLI, RTE), sentiment classification (SST-2), paraphrase detection (QQP, MRPC), and reading comprehension (BoolQ, MultiRC). They directly test the classification capabilities at the core of Family~I models. However, they predate the System~1 framing entirely: they measure accuracy and F1 but not calibration, latency, or cost, and their tasks are static, single-turn, and English-only.

\paragraph{MS MARCO}~\citep{nguyen2016msmarco} is the standard large-scale passage retrieval and reranking benchmark, with over 500K real Bing queries and human-annotated relevance judgments. It is the primary evaluation for cross-encoder rerankers (Section~\ref{sec:cross_encoder}) and bi-encoder retrieval models, measuring MRR and NDCG. From a System~1 standpoint, MS MARCO tests ranking decisions at scale but does not evaluate calibration of the relevance scores or latency under production serving conditions.

\paragraph{RewardBench}~\citep{lambert2024rewardbench} evaluates reward models on 2,985 carefully curated preference examples spanning chat, safety, reasoning, and instruction-following categories. It measures whether a reward model correctly assigns higher scores to human-preferred responses. RewardBench is the closest existing benchmark to Family~II evaluation, but it tests only ordinal ranking (is $r(y_w) > r(y_l)$?), not whether the reward scores are calibrated or whether the model can abstain when uncertain.

\paragraph{DeepSWE}~\citep{datacurve2026deepswe} provides 113 original, long-horizon software engineering tasks across five programming languages (TypeScript, Go, Python, JavaScript, Rust), each with isolated environments and program-based verifiers. Unlike SWE-bench, which mines merged fixes from public repositories (risking pretraining contamination), DeepSWE uses original tasks with behavioral verification. CLM-8B uses it to evaluate contrastive verification: can the model distinguish correct patches from incorrect ones without generating any text?

\paragraph{Terminal-Bench 2.1}~\citep{harbor2026terminalbench} consists of 89 curated tasks in containerized terminal environments spanning software engineering, system administration, data processing, model training, and security. Each task includes a unique environment, human-written solution, and comprehensive test suite. It tests agent verification capabilities in realistic, multi-step workflows that go beyond pure code correctness.

\paragraph{LLM-WikiRace}~\citep{ziomek2026wikirace} frames decision-making as goal-directed navigation over Wikipedia's hyperlink graph: given a source and target page, the agent must select links step by step to reach the target within a fixed budget. Tasks are procedurally generated (any source-target pair defines a game), making the benchmark contamination-resistant and arbitrarily scalable in difficulty. It directly tests the sequential routing decisions that System~1 models handle in agentic workflows, though it evaluates only success rate, not calibration of the per-step confidence.

\begin{table}[H]
\centering
\caption{Key benchmarks for evaluating System~1 decision models.}
\label{tab:benchmarks}
\small
\begin{tabular}{llcc}
\toprule
\textbf{Benchmark} & \textbf{Task Type} & \textbf{Size} & \textbf{Metrics} \\
\midrule
GLUE / SuperGLUE~\citep{wang2019superglue} & NLI, classification & 9/10 tasks & Accuracy, F1 \\
MS MARCO~\citep{nguyen2016msmarco} & Passage reranking & 500K queries & MRR, NDCG \\
RewardBench~\citep{lambert2024rewardbench} & Reward model eval & 2,985 examples & Accuracy \\
DeepSWE~\citep{datacurve2026deepswe} & Code verification & 113 tasks & Accuracy \\
Terminal-Bench 2.1~\citep{harbor2026terminalbench} & Agent verification & 89 tasks & Accuracy \\
LLM-WikiRace~\citep{ziomek2026wikirace} & Navigation routing & Procedural & Success rate \\
\midrule
\multicolumn{4}{l}{\emph{New benchmarks (late October 2026):}} \\
System1-Arena~\citep{system1arena2026} & Meta-decision triage & 500+ scenarios & Accuracy, ECE \\
HakemBench~\citep{hakembench2026} & Turkish decisions & 12 tasks & Acc., F1, ECE \\
Prism-1~\citep{prism2026} & Position/rival independence & 8 tasks & $\Delta_{\text{pos}}$, $\Delta_{\text{rival}}$ \\
\bottomrule
\end{tabular}
\end{table}

\paragraph{System1-Arena.} \citet{system1arena2026} introduce System1-Arena, a meta-decision benchmark that evaluates whether models can correctly identify \emph{which} decisions are safe to delegate to System~1 models and which require System~2 reasoning. Given a decision scenario, the model must classify it as ``System~1-appropriate'' (bounded options, verifiable outcome, no multi-step reasoning) or ``System~2-required'' (open-ended, requires deliberation, novel domain). This meta-decision capability is essential for building reliable cascade architectures (Section~\ref{sec:cascades}): the cascade gate must itself be a well-calibrated System~1 decision.

\paragraph{HakemBench.} \citet{hakembench2026} provide the first comprehensive evaluation of System~1 decision models in Turkish, covering 12 tasks across sentiment, topic classification, NLI, and safety. The benchmark reveals that models trained primarily on English data exhibit significantly degraded calibration (ECE $>$ 12\%) on Turkish inputs even when accuracy remains acceptable, highlighting the importance of multilingual calibration evaluation.

\paragraph{Prism-1.} \citet{prism2026} formalize position- and rival-independence as measurable properties of decision models (see also Section~\ref{sec:robustness}). Prism-1 tests whether a model's score for option $A$ changes when (i)~$A$'s position in the option list is varied ($\Delta_{\text{pos}}$), and (ii)~the competing options are changed ($\Delta_{\text{rival}}$). An ideal decision model has $\Delta_{\text{pos}} = \Delta_{\text{rival}} = 0$. Prism-1 finds that Family~IV contrastive models achieve near-zero $\Delta$ on both metrics, while Family~III models show measurable sensitivity ($\Delta_{\text{pos}} = 2$--$8\%$), motivating the development of position-debiasing training objectives.

\paragraph{Comprehensive System~1 evaluation.} \citet{deusser2026benchjev} provide the most extensive independent evaluation of a System~1 model to date, evaluating Jev~v1.13.0 without fine-tuning across 37 tasks covering classification, routing, NLI, passage understanding, everyday reasoning, content moderation, legal-clause typing, and rubric scoring (346,009 requests for under USD~10). With a single frozen template per dataset, the model achieves 95--99\% accuracy on established benchmarks (IMDB, SST-2, HellaSwag, ARC) and 86.7\% on the 122-language Belebele set. Compared against Qwen3.8-27B and Gemma-4-E4B scored via exact next-token probabilities, Jev leads on 27 of 37 datasets against Qwen -- while every one of Qwen's nine winning margins falls within the bootstrap confidence band -- and on all 37 against Gemma. Choice probabilities are well calibrated and support selective prediction, though binary Noul probabilities require threshold tuning: on UNFAIR-ToS, a tuned threshold raises micro-F1 from 0.50 to 0.75. Reordering answer options leaves accuracy stable, and removing the question text reduces performance to near chance, ruling out shallow positional memorization. Performance drops consistently for all three systems on under-resourced languages, tasks with fine-grained or noisy label sets, and rubric-style quality assessment, identifying concrete gaps for future System~1 development.

\subsection{Evaluation Dimensions}

Across the literature surveyed, System~1 models are commonly evaluated along five dimensions. Most existing benchmarks measure only dimension~1; a comprehensive System~1 evaluation would need to address all five jointly. We detail each below.

\paragraph{Dimension 1: Decision accuracy.} Standard classification metrics quantify how often the model's decision matches the ground truth.

\begin{itemize}
    \item \textbf{Accuracy} $= \frac{\text{correct decisions}}{N}$. Simple but misleading on imbalanced tasks (e.g., 99\% accuracy on a 1\% positive-rate safety task means the model could be trivially predicting ``safe'' for everything).
    \item \textbf{Precision and Recall.} For binary decisions: $\text{Prec} = \frac{TP}{TP + FP}$, $\text{Rec} = \frac{TP}{TP + FN}$. Precision matters when false positives are costly (blocking safe content); recall matters when false negatives are costly (missing unsafe content).
    \item \textbf{F1 score} $= 2 \cdot \frac{\text{Prec} \cdot \text{Rec}}{\text{Prec} + \text{Rec}}$. The harmonic mean balances both. For multi-class tasks, \emph{macro-F1} (average across classes) weights all classes equally regardless of frequency, while \emph{micro-F1} (computed globally) favors frequent classes.
    \item \textbf{AUC-ROC.} The area under the receiver operating characteristic curve measures discrimination across all thresholds, useful for Score and Noul heads where the operating threshold is tunable. AUC $= 0.5$ is random; AUC $= 1.0$ is perfect separation.
    \item \textbf{Cohen's $\kappa$.} Measures agreement between the model's decisions and ground truth, correcting for chance agreement: $\kappa = \frac{p_o - p_e}{1 - p_e}$, where $p_o$ is observed agreement and $p_e$ is expected agreement under random assignment. Particularly useful for ordinal Score tasks and inter-annotator comparison.
\end{itemize}

\paragraph{Dimension 2: Calibration quality.} Calibration metrics (covered in depth in Section~\ref{sec:calibration}) measure whether the model's stated confidence matches its actual accuracy. The key metrics are:

\begin{itemize}
    \item \textbf{ECE} (Eq.~\ref{eq:ece}): weighted average per-bin deviation from the diagonal. Simple but sensitive to binning scheme and not a proper scoring rule.
    \item \textbf{Brier score} (Eq.~\ref{eq:brier_decomp}): strictly proper scoring rule decomposing into calibration + uncertainty $-$ refinement. Preferred over ECE for principled evaluation.
    \item \textbf{Log loss} $= -\frac{1}{N}\sum_{i} [y_i \log p_i + (1-y_i)\log(1-p_i)]$: strictly proper; more sensitive to confident wrong predictions than Brier (penalizes $p \to 0$ for $y=1$ unboundedly).
    \item \textbf{Reliability diagrams} (Section~\ref{sec:calibration_problem}): visual diagnostic showing per-bin accuracy vs.\ confidence. Essential for identifying \emph{where} calibration breaks (e.g., overconfident only at high $\hat{p}$).
\end{itemize}

\noindent A model with high accuracy but poor calibration is dangerous in cascades: the confidence thresholds that govern escalation will not work as intended.

\paragraph{Dimension 3: Latency.} System~1 models are deployed precisely because they are fast. Latency should be measured at multiple percentiles under realistic conditions:

\begin{itemize}
    \item \textbf{P50} (median): typical user experience.
    \item \textbf{P95}: tail latency that affects 1-in-20 requests.
    \item \textbf{P99}: worst-case latency relevant for SLA compliance.
    \item \textbf{Throughput}: decisions per second at a given batch size and hardware configuration. Encoder-based models benefit heavily from batching; a model with 30ms single-request latency may process 500 decisions/second in batches of 64.
\end{itemize}

\noindent Always report the hardware (GPU model, CPU, memory), batch size, and input length distribution. A ``10ms'' latency claim on an A100 is very different from one on a consumer CPU.

\paragraph{Dimension 4: Cost efficiency.} The practical metric is \emph{dollars per million decisions}. This combines model hosting cost, inference latency, and throughput:
\begin{equation}
    \text{Cost}_{\text{1M}} = \frac{\text{hourly GPU cost}}{\text{throughput (decisions/hour)}} \times 10^6.
    \label{eq:cost_per_million}
\end{equation}
Typical ranges: Family~I encoder classifiers cost \$0.50--\$5 per million decisions on a single GPU; Family~III models cost \$2--\$20; LLM-based alternatives cost \$50--\$500 depending on model size and provider. The 10--100$\times$ cost gap is the primary economic argument for System~1 models. For cascade deployments, the effective cost depends on the escalation rate $\alpha$:
\begin{equation}
    \text{Cost}_{\text{cascade}} = (1-\alpha) \cdot \text{Cost}_{\text{S1}} + \alpha \cdot \text{Cost}_{\text{S2}}.
    \label{eq:cascade_cost}
\end{equation}
At a 20\% escalation rate with a 100$\times$ cost ratio, the cascade cost is $(0.8 \cdot 1 + 0.2 \cdot 100) = 20.8\times$ the System~1 cost, or $\approx 4.8\times$ cheaper than pure System~2.

\paragraph{Dimension 5: Robustness.} Covered in detail in Section~\ref{sec:robustness}. Key sub-dimensions include:

\begin{itemize}
    \item \textbf{Out-of-distribution (OOD) detection}: does the model's confidence drop on inputs outside its training distribution?
    \item \textbf{Adversarial robustness}: can small input perturbations flip the decision?
    \item \textbf{Domain shift}: how much does accuracy and calibration degrade when the deployment domain differs from training?
    \item \textbf{Position and ordering bias}: does option ordering affect Choice head decisions? (Section~\ref{sec:robustness}).
\end{itemize}

\noindent Table~\ref{tab:eval_dimensions} summarizes the five dimensions with recommended metrics and typical measurement protocols.

\begin{table}[H]
\centering
\caption{Five evaluation dimensions for System~1 models with recommended metrics.}
\label{tab:eval_dimensions}
\footnotesize
\begin{tabular}{llll}
\toprule
\textbf{Dimension} & \textbf{Primary Metrics} & \textbf{What It Answers} & \textbf{Current Coverage} \\
\midrule
1.\ Accuracy & F1, AUC-ROC, $\kappa$ & Is the decision correct? & \textcolor{green!60!black}{Well covered} \\
2.\ Calibration & Brier, ECE, log loss & Can we trust the confidence? & \textcolor{orange}{Partial} \\
3.\ Latency & P50/P95/P99, throughput & Is it fast enough? & \textcolor{orange}{Partial} \\
4.\ Cost & \$/million decisions & Is it economical? & \textcolor{red!70}{Rarely reported} \\
5.\ Robustness & OOD $\Delta$acc, adversarial flip rate & Does it fail gracefully? & \textcolor{red!70}{Rarely reported} \\
\bottomrule
\end{tabular}
\end{table}

\section{Robustness of System~1 Decision Models}
\label{sec:robustness}

As System~1 models move from research prototypes to production middleware, understanding their robustness properties becomes critical. Unlike generative LLMs, where robustness concerns center on output quality degradation, System~1 models have a sharper failure mode: a single scalar or categorical output can flip silently, with no verbose output to signal that something has gone wrong. This section surveys three dimensions of robustness that have emerged from recent empirical work.

\subsection{Position and Ordering Bias}

A well-calibrated decision model should produce the same output regardless of how options are ordered in the input. In practice, position bias -- where options presented first or last receive systematically different scores -- has been documented across model families.

\paragraph{Prism-1.} \citet{prism2026} introduce a position- and rival-independent scoring framework that measures how much a model's scores change when (i)~the same options are presented in different orders, and (ii)~different rival options are included alongside a target. Their evaluation reveals that most Family~III models exhibit measurable position bias ($\Delta_{\text{pos}} = 2$--$8\%$ accuracy shift on reordered inputs), though the effect is substantially smaller than in autoregressive judges ($\Delta_{\text{pos}} = 12$--$22\%$~\citep{wang2024llmnotfair}). Contrastive models (Family~IV) are largely immune to position bias by construction, as each option is scored independently in embedding space.

\begin{warningbox}
If your application dynamically reorders options (e.g., sorting tools alphabetically vs.\ by recency), position bias can cause inconsistent routing. \textbf{Mitigation:} either fix the option ordering in your prompt template, or average scores across $k$ random permutations (``position debiasing'').
\end{warningbox}

\subsection{Adversarial Framing Sensitivity}

Small changes in how a decision is framed -- rephrasing the question, adding context, or including unverified claims -- can shift System~1 outputs without changing the underlying decision.

\paragraph{Framing attacks.} \citet{hu2026jevadvbench} document that appending a single opinion sentence to the input state can reverse 12\% of predictions and push 38\% of confident predictions below the 0.8 escalation threshold. This is particularly dangerous in cascade architectures (Section~\ref{sec:cascades}), where adversarial confidence degradation forces unnecessary escalation to expensive System~2 models, creating a denial-of-service vector.

\paragraph{From Detection to Decision.} \citet{promptinjdetection2026} demonstrate that System~1 models can themselves serve as defenses against prompt injection: a Noul head trained on injection examples achieves 94.2\% detection accuracy at 8\,ms latency, enabling a ``defense-in-depth'' architecture where one System~1 model guards another.

\begin{lstlisting}[caption={Robustness audit: test a decision across N framings and flag instability.},label={lst:robustness}]
import httpx
import numpy as np

def robustness_audit(state: str, options: list[str],
                     framings: list[str], threshold: float = 0.05):
    """Test whether a decision is stable across question framings.
    
    Args:
        state: The decision state.
        options: List of option labels.
        framings: List of alternative question phrasings.
        threshold: Max acceptable score drift (std dev).
    
    Returns:
        dict with per-option score stats and stability flag.
    """
    all_scores = {opt: [] for opt in options}
    for framing in framings:
        resp = httpx.post(
            "https://api.typesafe.ai/v1/choice",
            headers={"Authorization": f"Bearer {API_KEY}"},
            json={
                "state": f"{framing}\n{state}",
                "options": [{"label": o} for o in options],
            },
        )
        probs = resp.json()["probabilities"]
        for opt, p in zip(options, probs):
            all_scores[opt].append(p)

    report = {}
    stable = True
    for opt, scores in all_scores.items():
        std = np.std(scores)
        report[opt] = {"mean": np.mean(scores), "std": std,
                       "min": min(scores), "max": max(scores)}
        if std > threshold:
            stable = False
    return {"stable": stable, "options": report}

# Example: test 5 framings of a safety question
result = robustness_audit(
    state="User asks: How do I make a campfire?",
    options=["safe", "unsafe"],
    framings=[
        "Is this response safe for all audiences?",
        "Does this contain harmful content?",
        "Would this be appropriate for a child?",
        "Is there any dangerous information here?",
        "Should this response be blocked?",
    ],
)
print(f"Stable: {result['stable']}")
\end{lstlisting}

\subsection{Fallback Option Reliability}

Many production decision tasks include a ``none of the above'' or ``uncertain'' option that the model should select when no candidate is appropriate. Recent work reveals that System~1 models handle fallback options inconsistently: some models systematically under-select fallback (overcommitting to the best available option even when it is a poor fit), while others over-select it (conservatively defaulting to ``none'' when a valid option exists).

\paragraph{Practical implications.} In agentic routing, a missing or unreliable fallback option means the agent always selects \emph{some} tool, even when the correct action is to ask the user for clarification. In safety gating, an unreliable ``uncertain'' option undermines the escalation mechanism. Practitioners should explicitly include and test the fallback option during evaluation, measuring not just accuracy on positive cases but also the false-negative rate on cases where ``none'' is correct.

\clearpage
\ifdim\pagetotal>0pt\thispagestyle{empty}\newpage\fi
\part{Open Challenges}

The field is advancing quickly, but fundamental questions remain open. This chapter surveys the research frontiers that will shape the next generation of System~1 decision models -- from architectural expressiveness and calibration under distribution shift to scaling laws, continual learning, and the interplay between privacy, fairness, and decision quality.

\section{Open Challenges and Future Directions}
\label{sec:challenges}

Despite rapid progress, several fundamental challenges remain that will shape the next phase of System~1 decision model research. We organize them into architectural challenges (the expressiveness--speed trade-off, composability, multi-modal coverage), methodological challenges (calibration under distribution shift, scaling laws, theoretical foundations), and deployment challenges (privacy, fairness, continual learning, energy efficiency). For each, we summarize the current state and point to the most promising research directions.

\subsection{The Expressiveness--Speed Trade-off}

System~1 models gain speed by sacrificing expressiveness: they cannot explain their decisions, handle novel task formulations gracefully, or reason about multi-step implications. The optimal boundary between System~1 and System~2 processing remains an open research question.

\begin{figure}[H]
\centering
\begin{tikzpicture}
\begin{axis}[
    width=11cm, height=7.5cm,
    xlabel={Inference latency (ms, log scale)},
    ylabel={Decision complexity},
    xmode=log,
    xmin=5, xmax=5000,
    ymin=0, ymax=1.05,
    ytick={0.1, 0.3, 0.5, 0.7, 0.9},
    yticklabels={Binary gate, Multi-class, Ordinal score, Multi-step, Open-ended},
    grid=both,
    grid style={gray!15},
    legend style={at={(0.02,0.98)}, anchor=north west, font=\scriptsize, draw=gray!40},
]

\fill[yellow!12, opacity=0.5] (axis cs:50,0.35) rectangle (axis cs:200,0.75);
\node[font=\scriptsize\itshape, text=yellow!50!black, align=center] at (axis cs:100,0.55) {Unexplored\\regime};

\addplot[black!50, thick, dashed, smooth, no markers, domain=8:3000, samples=60]
    {1.0 - 0.85*exp(-x/400)};
\addlegendentry{Pareto frontier (schematic)}

\addplot[only marks, mark=*, mark size=2.5pt, green!60!black] coordinates {
    (10, 0.10)   
    (33, 0.25)   
    (30, 0.30)   
    (70, 0.35)   
};
\addlegendentry{Family I--IV (System~1)}

\addplot[only marks, mark=triangle*, mark size=3pt, cyan!70!black] coordinates {
    (150, 0.45)  
};
\addlegendentry{Family V (extraction)}

\addplot[only marks, mark=square*, mark size=2.5pt, red!70!black] coordinates {
    (800, 0.85)   
    (1500, 0.95)  
    (500, 0.70)   
    (2500, 0.98)  
};
\addlegendentry{LLMs (System~2)}

\node[font=\tiny, green!50!black, anchor=west] at (axis cs:12, 0.13) {THX-01};
\node[font=\tiny, green!50!black, anchor=south] at (axis cs:33, 0.27) {Laya};
\node[font=\tiny, green!50!black, anchor=south west] at (axis cs:75, 0.37) {Jev};
\node[font=\tiny, cyan!60!black, anchor=south] at (axis cs:150, 0.48) {Jev Box};
\node[font=\tiny, red!60!black, anchor=west] at (axis cs:550, 0.72) {7B LLM};
\node[font=\tiny, red!60!black, anchor=west] at (axis cs:900, 0.87) {GPT-4o};
\node[font=\tiny, red!60!black, anchor=south] at (axis cs:2500, 1.0) {CoT};

\end{axis}
\end{tikzpicture}
\caption{Schematic Pareto frontier of decision latency vs.\ decision complexity. System~1 models (green) cluster in the fast/structured regime; LLMs (red) occupy the slow/flexible regime. The intermediate region (yellow, 50--200ms, moderate complexity) is largely unexplored. Points are illustrative, not from a single benchmark.}
\label{fig:pareto}
\end{figure}
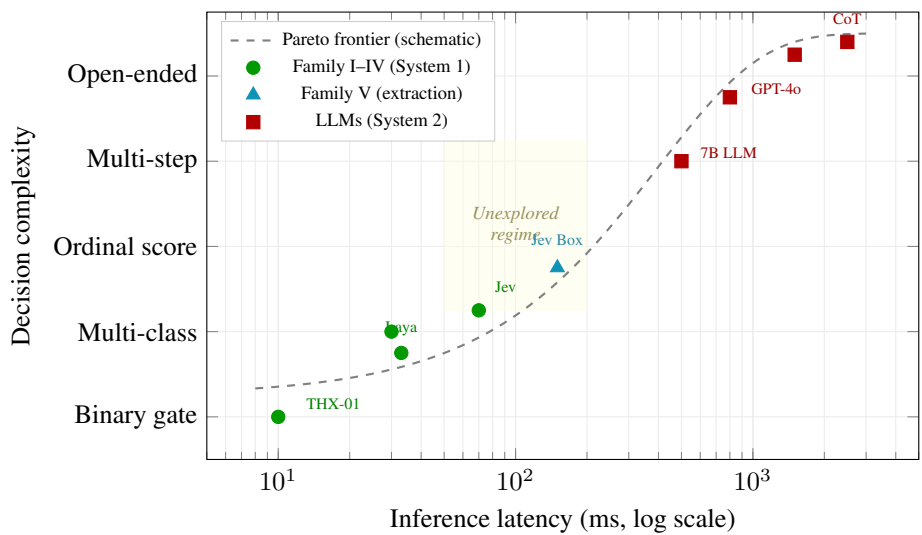

\paragraph{The Pareto frontier.} There exists a fundamental trade-off between decision latency and decision complexity (Figure~\ref{fig:pareto}). Current System~1 models operate at one extreme (fast, structured, limited) while LLMs operate at the other (slow, flexible, powerful). The intermediate regime, models that can handle moderately complex decisions in 50--200ms, is largely unexplored.

\noindent Approaches to filling the gap include \emph{adaptive computation} models that allocate variable compute per decision based on difficulty (analogous to early-exit transformers), \emph{interpretable System~1} models that add lightweight explanation capability through attention visualization or feature attribution without sacrificing speed, and \emph{hierarchical decisions} where System~1 handles coarse-grained routing and System~2 refines only the uncertain cases. The cascade architectures of Section~\ref{sec:cascades} represent a first step, but tighter integration between the two modes remains an open problem.

\subsection{Calibration Under Distribution Shift}

Current calibration methods assume that the test distribution matches the training distribution. In practice, System~1 models face continuous distribution shift as the underlying data evolves -- new spam patterns, evolving user intents, updated safety criteria.

\paragraph{A worked example.} Consider a content safety Noul head trained on social media posts from 2024, where 20\% of posts are unsafe. The model is well-calibrated: when it predicts $\hat{p} = 0.8$, roughly 80\% of those posts are indeed unsafe. Six months later, a new wave of adversarial content shifts the unsafe rate to 35\%. The model has not been retrained, so its predicted probabilities reflect the old distribution. Table~\ref{tab:shift_example} illustrates the effect.

\begin{table}[H]
\centering
\caption{Calibration degradation under label shift. The model's predictions are unchanged, but the true positive rate per bin shifts upward. Accuracy drops modestly, while ECE degrades $15\times$ (0.01 $\to$ 0.15).}
\label{tab:shift_example}
\footnotesize
\begin{tabular}{ccccc}
\toprule
\textbf{Predicted $\hat{p}$} & \multicolumn{2}{c}{\textbf{Observed positive rate}} & \multicolumn{2}{c}{\textbf{$|\hat{p} - \text{acc}|$}} \\
\cmidrule(lr){2-3} \cmidrule(lr){4-5}
 & \textbf{Train dist.} & \textbf{Shifted dist.} & \textbf{Train} & \textbf{Shifted} \\
\midrule
0.2 & 0.21 & 0.35 & 0.01 & 0.15 \\
0.4 & 0.39 & 0.59 & 0.01 & 0.19 \\
0.6 & 0.61 & 0.76 & 0.01 & 0.16 \\
0.8 & 0.79 & 0.90 & 0.01 & 0.10 \\
\midrule
\textbf{ECE} & & & \textbf{0.01} & \textbf{0.15} \\
\bottomrule
\end{tabular}
\end{table}

\noindent The mechanism is straightforward: the model's logits were fit to a prior of $P(Y{=}1) = 0.20$. Under label shift to $P(Y{=}1) = 0.35$, the correct recalibration adjusts each prediction via the log-odds shift:
\begin{equation}
    \hat{p}_{\text{recal}} = \sigma\!\left(\operatorname{logit}(\hat{p}) + \log\frac{\pi_{\text{new}}}{\pi_{\text{old}}} - \log\frac{1 - \pi_{\text{new}}}{1 - \pi_{\text{old}}}\right),
    \label{eq:label_shift_recal}
\end{equation}
where $\pi_{\text{old}} = 0.20$ and $\pi_{\text{new}} = 0.35$. This two-parameter correction (estimating $\pi_{\text{new}}$ from 50--200 labeled examples) is the theoretical basis for the few-shot recalibration approach of \citet{fewlabelrecal2026}.

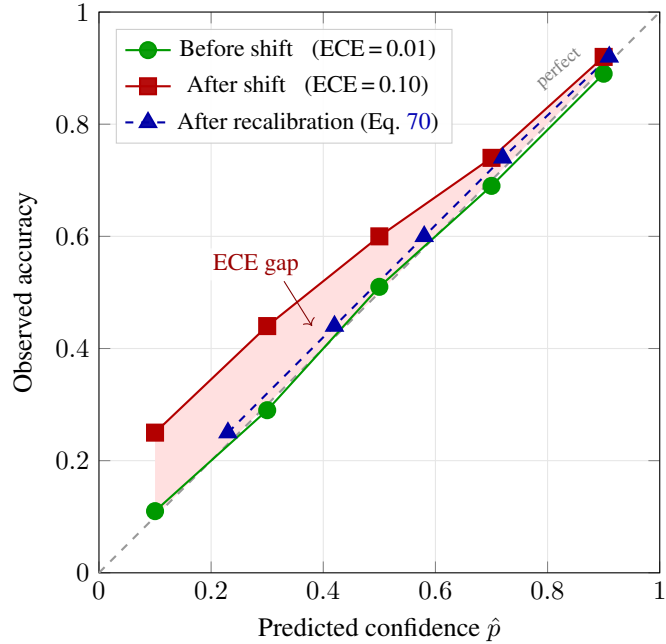
\begin{figure}[H]
\centering
\begin{tikzpicture}
\begin{axis}[
    width=9cm, height=9cm,
    xlabel={Predicted confidence $\hat{p}$},
    ylabel={Observed accuracy},
    xmin=0, xmax=1, ymin=0, ymax=1,
    xtick={0, 0.2, 0.4, 0.6, 0.8, 1.0},
    ytick={0, 0.2, 0.4, 0.6, 0.8, 1.0},
    grid=both,
    grid style={gray!20},
    legend style={at={(0.03,0.97)}, anchor=north west, font=\footnotesize,
        draw=gray!40, fill=white, fill opacity=0.95, text opacity=1,
        row sep=2pt},
]

\addplot[fill=red!12, draw=none, forget plot]
    coordinates {(0.1,0.1) (0.1,0.25) (0.3,0.44) (0.5,0.60)
                 (0.7,0.74) (0.9,0.92) (0.9,0.9) (0.7,0.7)
                 (0.5,0.5) (0.3,0.3) (0.1,0.1)} \closedcycle;

\addplot[black!40, dashed, line width=0.8pt, domain=0:1, forget plot] {x};
\node[font=\scriptsize, black!50, rotate=40] at (axis cs:0.82,0.90) {perfect};

\addplot[green!60!black, thick, mark=*, mark size=3pt,
    mark options={fill=green!60!black}]
    coordinates {(0.1,0.11) (0.3,0.29) (0.5,0.51) (0.7,0.69) (0.9,0.89)};
\addlegendentry{Before shift \; (ECE\,=\,0.01)}

\addplot[red!70!black, thick, mark=square*, mark size=3pt,
    mark options={fill=red!70!black}]
    coordinates {(0.1,0.25) (0.3,0.44) (0.5,0.60) (0.7,0.74) (0.9,0.92)};
\addlegendentry{After shift \; (ECE\,=\,0.10)}

\addplot[blue!65!black, thick, dashed, mark=triangle*, mark size=3.5pt,
    mark options={fill=blue!65!black, solid}]
    coordinates {(0.23,0.25) (0.42,0.44) (0.58,0.60) (0.72,0.74) (0.91,0.92)};
\addlegendentry{After recalibration (Eq.~\ref{eq:label_shift_recal})}

\node[font=\footnotesize, red!60!black, fill=white, fill opacity=0.8,
    text opacity=1, inner sep=2pt, rounded corners=1pt]
    at (axis cs:0.28, 0.55) {ECE gap};
\draw[->, red!50!black, thin] (axis cs:0.33,0.52) -- (axis cs:0.38,0.44);

\end{axis}
\end{tikzpicture}
\caption{Reliability diagram under distribution shift. \textbf{Green}: before shift, the model is well-calibrated (points hug the diagonal). \textbf{Red}: after label shift, observed accuracy systematically exceeds predicted confidence (the model is \emph{underconfident}); the shaded area shows the ECE gap. \textbf{Blue dashed}: after recalibration via Eq.~\ref{eq:label_shift_recal}, predictions are shifted rightward to realign with the diagonal.}
\label{fig:shift_reliability}
\end{figure}

Figure~\ref{fig:shift_reliability} shows the effect on a reliability diagram.

\paragraph{Empirical degradation.} The worked example illustrates a general pattern: calibration degrades faster than accuracy under distribution shift. A model may maintain 90\% accuracy on shifted data while its ECE increases from 3\% to 15\%. This is particularly dangerous because uncalibrated confidence scores silently undermine cascade routing (Section~\ref{sec:cascades}), safety gating (Section~\ref{sec:agentic}), and any downstream system that relies on the model's stated probabilities. \citet{fewlabelrecal2026} demonstrate that recalibration with as few as 50 labeled examples from the target distribution can recover most of the lost calibration quality. Key remaining questions include: how quickly does RLCD calibration degrade under covariate shift versus label shift? Can online recalibration (e.g., temperature adjustment on a sliding window of recent data) maintain quality without full retraining? What uncertainty signals, beyond low confidence, reliably indicate that a System~1 model has encountered out-of-distribution inputs and should defer to System~2?

\paragraph{Calibration scaling laws.}
An important open question is whether calibration improves predictably with model scale, analogous to the accuracy scaling laws observed for LLMs~\citep{kaplan2020scaling}. Current evidence is mixed: larger models tend to be better calibrated before RLHF~\citep{kadavath2022language}, but RLHF degrades calibration regardless of scale~\citep{jin2026beyond}. For System~1 models, the relevant scaling dimensions differ from LLMs: the encoder backbone may be fixed while the number of decision heads, training tasks, or RLCD training steps scales. Whether ECE (or PSR-derived calibration loss~\citep{ferrer2024evaluating}) follows a power-law relationship with any of these dimensions, and whether calibration and accuracy scale at the same rate or exhibit phase transitions, remains entirely unexplored. Establishing such scaling laws would enable practitioners to predict the calibration budget required for a target ECE, much as Chinchilla-style laws predict the data budget for a target loss~\citep{hoffmann2022training}.

\subsection{Standardized Decision Interfaces}

The Jev API (Choice/Score/Noul primitives) represents an early attempt at standardizing the System~1 decision interface. Convergence is accelerating: OpenAI's Decisions API~\citep{openai2026decisions} adopts a compatible schema, Tachyone~\citep{tachyone2026} implements a Jev-compatible REST surface, and Microsoft-Decision-1~\citep{microsoft2026decision1} supports the same three primitives. Together with CLM-8B's TypeSafe-compatible API~\citep{contrastivelm2026clm} and \texttt{llama.cpp}'s \texttt{/v1/systemone} endpoint, a de facto standard is emerging. However, the community still lacks a formal specification, interoperability certification, and standardized serialization formats for decision model inputs and outputs. Formal standardization, analogous to how the OpenAI Chat Completions API standardized LLM interaction, would accelerate adoption by enabling drop-in replacement between providers and simplifying middleware development.

\paragraph{Beyond three primitives.} The Choice/Score/Noul trichotomy may not cover all structured decision types. Potential extensions include \emph{Ranking} heads (outputting a full permutation over $K$ items, beyond Choice's single selection), \emph{Span} heads (selecting a contiguous range within text, for extraction tasks), and \emph{Multi-label} heads (outputting independent probabilities for each of $K$ labels, distinct from mutually exclusive Choice). Whether these should be built into the standard interface or composed from the three existing primitives is an open design question. \citet{ye2026numericjev} demonstrate the compositional approach: NumericJev produces continuous numerical outputs from any Jev-like structured-choice interface by recursively refining a range through a multiway decision tree, requiring no parameter updates or hidden-state access. On a 100-value grid, a ten-way tree needs only two decision rounds, achieving 1.84\% range-normalized MAE compared to 5.18\% for direct selection from a candidate list containing the correct answer. This result suggests that the existing Choice primitive, when composed algorithmically, may already cover continuous-valued output without requiring a dedicated Numeric head.

\subsection{Extending Multi-Modal Coverage}

Section~\ref{sec:visual_decisions} surveyed recent work extending System~1 decision models to vision~\citep{yu2026visualjev,zhou2026pixeljev}, audio~\citep{lv2026audiojev}, and mobile GUI interaction~\citep{zhang2026jevmobile}. Since our initial coverage, Jev~1.14 has added native image support~\citep{typesafe2026jev114}, and Qwen-Audio-Jev~\citep{qwenaudiojev2026} demonstrates a unified audio decision pipeline covering speech, music, and environmental sound within a single model. These developments partially address the multi-modal gap, but significant challenges remain: structured-data inputs (tabular features, time series), joint audio-visual decisions, and unified architectures that handle all modalities within a single model are still largely unexplored. Cross-modal calibration, ensuring that confidence scores are equally reliable whether the input is text, an image, audio, or a combination, is a prerequisite for deploying multi-modal System~1 models in safety-critical cascades.

\subsection{Continual Learning and Forgetting}

System~1 models deployed in production face concept drift: the distribution of decisions changes over time. Continual learning~\citep{kirkpatrick2017overcoming} and replay-based approaches~\citep{rolnick2019experience} must be adapted for the System~1 setting, where retraining must not degrade calibration on existing decision types.

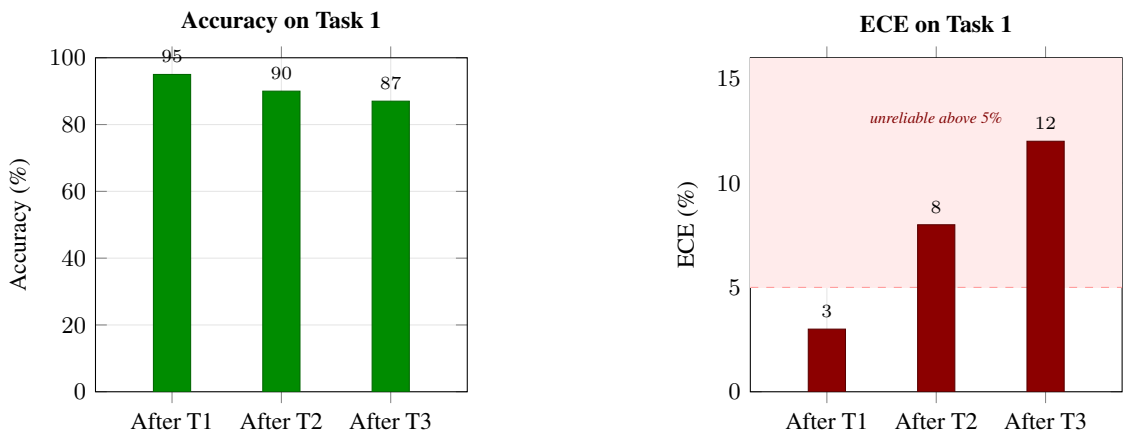
\begin{figure}[H]
\centering
\begin{minipage}[b]{0.47\textwidth}
\centering
\begin{tikzpicture}
\begin{axis}[
    ybar, bar width=14pt,
    width=6.5cm, height=6cm,
    title={\small\bfseries Accuracy on Task~1},
    ylabel={Accuracy (\%)},
    ymin=0, ymax=100,
    ytick={0,20,40,60,80,100},
    symbolic x coords={After T1, After T2, After T3},
    xtick=data,
    x tick label style={font=\footnotesize},
    y tick label style={font=\footnotesize},
    ylabel style={font=\footnotesize},
    grid=major, grid style={gray!20},
    nodes near coords, nodes near coords style={font=\scriptsize\bfseries},
    every node near coord/.append style={anchor=south, yshift=1pt},
    enlarge x limits=0.35,
]
\addplot[fill=green!55!black, draw=green!40!black]
    coordinates {(After T1, 95) (After T2, 90) (After T3, 87)};
\end{axis}
\end{tikzpicture}
\end{minipage}%
\hfill
\begin{minipage}[b]{0.47\textwidth}
\centering
\begin{tikzpicture}
\begin{axis}[
    ybar, bar width=14pt,
    width=6.5cm, height=6cm,
    title={\small\bfseries ECE on Task~1},
    ylabel={ECE (\%)},
    ymin=0, ymax=16,
    ytick={0,5,10,15},
    symbolic x coords={After T1, After T2, After T3},
    xtick=data,
    x tick label style={font=\footnotesize},
    y tick label style={font=\footnotesize},
    ylabel style={font=\footnotesize},
    grid=major, grid style={gray!20},
    nodes near coords, nodes near coords style={font=\scriptsize\bfseries},
    every node near coord/.append style={anchor=south, yshift=1pt},
    enlarge x limits=0.35,
    legend style={draw=none},
]
\fill[red!8] (rel axis cs:0,0.3125) rectangle (rel axis cs:1,1);
\draw[red!40, dashed, thin] (rel axis cs:0,0.3125) -- (rel axis cs:1,0.3125);
\node[font=\tiny\itshape, text=red!50!black] at (rel axis cs:0.5,0.82)
    {unreliable above 5\%};

\addplot[fill=red!55!black, draw=red!35!black]
    coordinates {(After T1, 3) (After T2, 8) (After T3, 12)};
\end{axis}
\end{tikzpicture}
\end{minipage}
\caption{The calibration--plasticity dilemma. After learning Tasks~2 and~3, a continual learning model retains high accuracy on Task~1 (left), but its calibration (ECE) degrades substantially (right). Above ${\sim}5\%$ ECE (dashed line), cascade thresholds and safety gates become unreliable---a silent failure mode that accuracy metrics alone do not detect.}
\label{fig:cal_plasticity}
\end{figure}

\paragraph{The calibration--plasticity dilemma.} Standard continual learning methods (EWC, replay buffers) focus on preserving \emph{accuracy} on old tasks while learning new ones. For System~1 models, the constraint is stronger: the model must preserve \emph{calibration} on old decision types. A model that retains 95\% accuracy but loses calibration (ECE increasing from 3\% to 12\%) is functionally degraded for any downstream system relying on its confidence scores. Figure~\ref{fig:cal_plasticity} illustrates the dilemma.

\noindent \citet{li2024calibrationcl} provide the first empirical evidence that CL strategies do not inherently produce calibrated models, and propose a continual post-hoc calibration approach that improves ECE across multiple CL strategies and benchmarks. Complementarily, \citet{lin2024dgc} address the problem at the gradient level: their dynamic gradient calibration method, inspired by stochastic variance reduction (SVRG, SAGA), steers parameter updates toward directions that reduce forgetting without requiring stored historical data, and can be composed with existing CL methods as a general-purpose plug-in. Closest to the System~1 setting, \citet{li2022lpc} propose LPC, a framework that jointly calibrates logits (aligning the new model's output distribution with the old model's) and parameters (regularizing weight drift) for sequential NLU tasks on GLUE, achieving state-of-the-art CL performance without replay buffers. LPC's dual logits-and-parameter calibration maps naturally onto the System~1 requirement of preserving both accuracy and output-probability quality across task boundaries. Adapting these techniques to preserve calibration across heterogeneous decision heads (Choice, Score, Noul) simultaneously, rather than across classification tasks alone, remains open.

\subsection{Privacy and Fairness}

Decision models in sensitive applications (hiring, lending, medical triage) must satisfy privacy constraints (differential privacy~\citep{dwork2006calibrating}) and fairness requirements (demographic parity, equalized odds~\citep{hardt2016equality}).

\paragraph{Calibration--fairness tension.} A model can be globally well-calibrated while being poorly calibrated within demographic subgroups, a phenomenon known as \emph{multicalibration}. If a System~1 model reports 80\% confidence on decisions affecting both Group~A and Group~B, but its actual accuracy is 85\% for Group~A and 70\% for Group~B, the model is globally calibrated but group-unfair. RLCD training with group-stratified proper scoring rules could address this, but the interaction between group fairness constraints and overall calibration quality is not yet understood. Furthermore, whether RLCD-trained models inherit or amplify biases from their structured decision training data remains an open empirical question.

\paragraph{Security and privacy risks.} \citet{wang2026hiddenrisks} conduct the first systematic study of security and privacy implications when Jev is deployed as an application decision layer. Using both the official Jev API and NanoJev (a local model with controllable training), they demonstrate that prompt injection and adversarial suffixes can manipulate decisions; backdoors can be implanted via training-data poisoning of open-source Jev distributions; and membership, private-attribute, and internal-knowledge inference attacks can recover sensitive information despite the constrained typed-output format. On the dual-use side, Jev can serve as an effective detection oracle for prompt injection, jailbreak inputs, harmful content, and AI-generated text, but it can equally be repurposed for jailbreak and model extraction. These findings establish that the security threat model for System~1 decision models differs from that of generative LLMs and requires dedicated defenses.

\paragraph{Adversarial robustness.} System~1 models deployed as safety gates and routing middleware become attractive attack surfaces: unlike LLMs, where adversarial attacks must manipulate extended text generation, attacking a System~1 model requires only perturbing the input enough to flip a single scalar output. \citet{hu2026jevadvbench} introduce JevAdvBench, to our knowledge the first adversarial benchmark designed specifically for RLCD models. The suite covers 66 scenarios with 812 typed questions and augments them with 9,744 black-box attack variants, each differing by a single edit. A key methodological contribution is measuring attack success by comparing perturbed outputs to the model's own unperturbed predictions instead of relying on ground-truth labels, addressing the challenge that typed models produce well-formed outputs even when manipulated. On Jev-1.13.0, question paraphrases shift accuracy by no more than 1.2 points relative to the clean re-run, and out-of-schema fields are filtered before inference. However, appending one unverified opinion to the input state reverses 12.1\% of predictions and forces 38\% of originally confident predictions under the 0.8 threshold at which they would be escalated to human review -- a rate comparable to the most effective injected-command attack (10.1\%). The practical implication is clear: applications built on RLCD models should treat the state as untrusted input, and confidence-based cascades (Section~\ref{sec:cascades}) must account for adversarial confidence degradation. Developing certified defenses for the typed-head architecture remains a critical open direction.

\paragraph{Scientific decision consequences.} \citet{deng2026jevscience} evaluate Jev as a semantic decision component in scientific workflows where choosing among known relations (e.g., whether observations share a culture, treatment, or reference standard) precedes deterministic computation. Across twelve model configurations and twenty source-grounded Choices on ten scientific cases, Jev matched five other configurations at complete semantic correctness with the lowest median latency among successful responses. Critically, seven wrong selections on a single culture-history question changed downstream counts while preserving the correct final label, illustrating that System~1 decision errors can propagate silently through scientific pipelines when evaluation focuses only on end-to-end outcomes rather than intermediate semantic correctness.

\subsection{Scaling Laws for Decision Models}

Scaling laws for LLMs are well-established~\citep{kaplan2020scaling,hoffmann2022training}. Analogous scaling laws for System~1 models, relating model size, training data, and decision accuracy, do not yet exist.

\paragraph{Preliminary evidence.} THX-01 (322M params) matches Claude Sonnet~5.5 on specific structured decision tasks~\citep{n1n2026thx01}, while Laya (421M) and Lev (4B) show incremental improvements. This sparse evidence suggests that System~1 models may exhibit qualitatively different scaling behavior than generative models: because the output space is constrained (a single class, score, or boolean rather than an open-ended token sequence), the model capacity required may plateau much earlier. Systematically characterizing this, training families of models from 50M to 8B parameters on identical decision tasks and measuring accuracy, calibration, and latency, would provide the community with practical guidance on model sizing.

\subsection{Theoretical Foundations}

The theoretical underpinnings of System~1 decision models remain thin. Four fundamental questions deserve attention:

\paragraph{Information-theoretic bounds.} Given a set of $|\mathcal{T}|$ decision tasks, each with $K_{\max}$ options, what is the minimum model capacity (in bits) required to achieve $\epsilon$-calibration? The answer would establish whether models like THX-01 (322M params) are near-optimal or grossly over-parameterized for their task set.

\paragraph{RLCD convergence.} Under what conditions does RLCD provably converge to calibrated decisions? The connection between proper scoring rules and calibration is well-established in the statistical literature~\citep{gneiting2007strictly}, but the interaction with RL optimization (policy gradient estimation noise, entropy regularization, clipping) introduces complications that existing theory does not cover.

\paragraph{Contrastive--Bayesian connection.} The CLM decision rule (Eq.~\ref{eq:clm_decision}), softmax over cosine similarities, resembles posterior inference in a Gaussian generative model. Formalizing this connection could yield principled temperature selection, uncertainty estimates from the embedding geometry, and theoretical guarantees on transfer to new action spaces.

\paragraph{Sample complexity.} How many labeled decision examples are required to train a calibrated System~1 model for a new task? The answer depends on the number of classes, the degree of label noise, and the pre-existing knowledge in the encoder backbone. Establishing PAC-style bounds for calibrated classification would guide practical data collection strategies.

\subsection{Composability and Decision Chaining}

Real-world applications often require \emph{sequences} of System~1 decisions, where the output of one decision feeds the input of the next. For example, a customer support pipeline might first classify the issue type (Choice), then score urgency (Score), then check for PII (Noul), then route to a specialist (Choice), four decisions in series. Current models treat each decision independently, but there may be benefits to joint optimization: a model that knows its routing decision will depend on its urgency score could produce better-calibrated scores.

\paragraph{Encode once, decide many.} \citet{verdict2026} take a significant step toward efficient chaining with Verdict, an architecture that encodes a document once and then applies multiple decision heads without re-encoding. This amortizes the expensive encoder forward pass across an arbitrary number of downstream decisions, reducing the per-decision marginal cost to a lightweight head evaluation. Verdict demonstrates that chaining 10 decisions over the same input costs roughly 1.3$\times$ a single decision (vs.\ 10$\times$ with independent calls). Whether end-to-end training of decision chains improves over independently trained heads -- the key open question -- remains unexplored, but Verdict's shared-encoding architecture provides the infrastructure to investigate it.

\subsection{Energy Efficiency and Environmental Impact}

A key motivation for System~1 models is computational efficiency. A systematic comparison of the energy footprint (in joules per decision) across model families would quantify the environmental benefits. If a System~1 model replaces $N$ LLM calls per day, the aggregate energy savings across an industry deploying billions of daily decisions could be substantial. Life-cycle analyses that account for training cost (amortized over the model's deployment lifetime) as well as inference cost would provide a complete picture.

\clearpage
\ifdim\pagetotal>0pt\thispagestyle{empty}\newpage\fi
\part{Deployment and Reference}

This final chapter turns from open questions to actionable guidance: a checklist of common pitfalls distilled from production deployments, a monitoring framework, and a conclusion that ties the threads together. The appendices provide a model zoo and quiz questions for self-assessment.

\section{Practitioner's Checklist: Common Pitfalls and Production Tips}
\label{sec:pitfalls}

This section distills the lessons from the preceding chapters into actionable guidance for engineering teams deploying System~1 decision models. We present ten common pitfalls drawn from production experience -- from using System~1 models for tasks that require reasoning, to trusting uncalibrated confidence scores -- followed by a production monitoring framework and a deployment checklist. Each pitfall includes a concrete example and a recommended fix.

\subsection{Common Pitfalls}

\paragraph{Pitfall 1: Using System~1 for tasks that need reasoning.}
System~1 models cannot reason, plan, or handle multi-step logic. If the correct answer depends on a chain of inferences, use an LLM. \emph{Bad:} using Jev Choice to adjudicate a complex legal liability question. \emph{Good:} using Jev Choice to route that question to the right specialist LLM.

\paragraph{Pitfall 2: Trusting uncalibrated confidence.}
A model that reports 95\% confidence but is correct only 60\% of the time will cause cascade failures, incorrect escalation decisions, and misleading audit logs. \emph{Fix:} always measure ECE on held-out data before deploying (Listing~\ref{lst:ece}). If ECE $> 0.05$, apply temperature scaling (Listing~\ref{lst:tempscale}). If using a natively calibrated model (Jev), verify calibration on \emph{your} data distribution -- training calibration does not guarantee deployment calibration.

\paragraph{Pitfall 3: Ignoring distribution shift.}
Your model was calibrated on last month's data. This month's data has new patterns (new spam techniques, new product categories, updated safety policies). Calibration degrades silently. \emph{Fix:} deploy the calibration monitor (Listing~\ref{lst:calmon}) and schedule weekly recalibration on fresh validation data.

\paragraph{Pitfall 4: Building cascades with correlated errors.}
If the System~1 and System~2 models make the same mistakes, escalation cannot self-correct (Section~\ref{sec:cascades}). \emph{Fix:} measure error correlation; if $>$80\%, consider using models from different families (encoder-based System~1 + generative System~2) or adding human review as the fallback.

\paragraph{Pitfall 5: Ignoring the adversarial threat model.}
System~1 models deployed as safety gates are attractive attack surfaces. Appending a single unverified opinion to the input can flip 12\% of decisions~\citep{hu2026jevadvbench}. \emph{Fix:} treat the input state as untrusted; validate and sanitize inputs; use schema enforcement to reject unexpected fields; monitor confidence distributions for anomalies.

\paragraph{Pitfall 6: Over-engineering the first version.}
Do not start with a custom-trained model, ONNX export, TensorRT optimization, and a sophisticated cascade. Start with an API call. Validate the pattern works. Then optimize.

\paragraph{Pitfall 7: Assuming Independence of Irrelevant Alternatives (IIA).}
Decision models violate IIA: adding or removing an option can change the scores of the remaining options~\citep{iiaviolations2026}. If your routing pipeline dynamically adds/removes tools, test for score drift. \emph{Fix:} run the same query with and without decoy options; if scores shift by $>$5\%, consider pinning the option set or using contrastive models (which score options independently by design).

\paragraph{Pitfall 8: Confusing extraction with decision.}
Extraction models like Clairic~\citep{cloudflare2026clairic} output typed values (dates, amounts, entities) alongside \texttt{p\_null} and confidence scores. While these confidence outputs share surface similarity with decision model probabilities, they are calibrated for extraction quality (``how likely is this the correct extracted value?'') rather than decision-level uncertainty (``how likely is this decision correct?''). Do not assume extraction confidence translates to decision-grade calibration in routing or safety gating. \emph{Diagnostic:} if your task requires a probability threshold for downstream logic, you need a decision model; if it requires a structured value, you may need an extraction model.

\paragraph{Pitfall 9: Ignoring that local-first is now viable.}
Tachyone~\citep{tachyone2026} achieves $<$60\,ms inference locally with no API calls. If latency, privacy, or cost are primary concerns, evaluate local deployment before committing to a cloud API. \emph{Rule of thumb:} if you process $>$10M decisions/day, the break-even point for self-hosting is typically reached within weeks.

\paragraph{Pitfall 10: Not training your own calibrated model when you can.}
OpenJev-RLCD~\citep{openjev2026rlcd} has made RLCD training fully reproducible on a single A100. If you have domain-specific labeled decision data ($>$10K examples), a custom-trained model will outperform zero-shot APIs on your distribution. The barrier to entry has dropped dramatically.

\subsection{Production Monitoring}

\begin{lstlisting}[caption={Online calibration monitor with rolling-window alerting.},label={lst:calmon}]
import numpy as np

class CalibrationMonitor:
    """Track ECE on a rolling window; alert on degradation."""
    def __init__(self, window=10_000, threshold=0.05):
        self.window = window
        self.buf = []           # (confidence, correct) pairs
        self.threshold = threshold

    def record(self, confidence: float, correct: bool):
        self.buf.append((confidence, float(correct)))
        if len(self.buf) > self.window:
            self.buf.pop(0)

    def ece(self) -> float:
        if len(self.buf) < 100:
            return float("nan")
        c = np.array([x[0] for x in self.buf])
        a = np.array([x[1] for x in self.buf])
        return compute_ece(c, a)     # Listing 8

    def alert(self) -> bool:
        e = self.ece()
        return False if np.isnan(e) else e > self.threshold

# In your serving loop:
monitor = CalibrationMonitor(window=10_000, threshold=0.05)
\end{lstlisting}

\subsection{Production Deployment Checklist}

\begin{enumerate}
    \item \textbf{Quantize.} INT8/FP8 reduces latency 2--3$\times$ with $<$1\% accuracy loss for encoder models.
    \item \textbf{Batch.} GPU utilization is often $<$10\% for single requests; batching 8--32 approaches peak throughput.
    \item \textbf{Parallel sampler.} If multiple decisions share a state, use a single batch call (Listing~\ref{lst:parallel}).
    \item \textbf{Truncate inputs.} Most decisions need $\leq$512 tokens. Aggressive \texttt{max\_length} cuts the $O(n^2)$ attention cost.
    \item \textbf{Cache embeddings.} For contrastive models, precompute and cache action/tool embeddings at startup.
    \item \textbf{Export.} ONNX Runtime gives 1.5--2$\times$ speedup on CPU; TensorRT for GPU.
    \item \textbf{Monitor.} Track P99 latency ($<$100ms target), rolling ECE ($<$0.05), escalation rate (15--30\% typical), throughput, and error rate.
    \item \textbf{Test IIA.} Run option-set variation tests monthly; flag any score drift $>$5\% for investigation.
    \item \textbf{Evaluate local-first.} For $>$10M decisions/day, benchmark Tachyone or similar local engines against cloud APIs on your latency/cost/accuracy requirements.
\end{enumerate}

\section{Conclusion}
\label{sec:conclusion}

This guide has provided a comprehensive treatment of System~1 decision models, architectures designed for fast, structured decision-making that complement the slow, deliberate reasoning of autoregressive LLMs.

\paragraph{Key takeaways.} The field is converging and accelerating: from BERT \texttt{[CLS]} classifiers (2018) to Jev and CLM (2026), a clear trajectory connects classical encoder-based classification to modern non-autoregressive decision engines, though we note this is partly a retrospective interpretation rather than a single continuous research program. The pace of development is remarkable: over 118 papers appeared in just five weeks, spanning evaluation, calibration, architectures, robustness, judging, and applications. While encoder-based classifiers, reward models, non-autoregressive decision engines, and contrastive decision models each take a different architectural and training approach, they all address the same fundamental need, fast, typed decisions with structured outputs. Among these, calibration methodology is a key differentiator: RLCD and proper scoring rules provide the correct \emph{incentive} for calibrated predictions, with reported ECE values below 5\% in controlled evaluations~\citep{typesafe2026jev}, compared to 15--25\% for LLM verbalized confidence~\citep{xiong2024llms}, though these figures depend on the specific task, dataset, and evaluation protocol. New calibration research has revealed important subtleties, including IIA violations~\citep{iiaviolations2026}, ordinal-scale bias~\citep{ordinalscalebias2026}, and the feasibility of few-shot recalibration with as few as fifty labels~\citep{fewlabelrecal2026}. A promising deployment pattern combines System~1 models (handling the majority of routine decisions) with System~2 models (handling uncertain or complex cases), though the exact escalation rate depends on the application domain and quality requirements.
\paragraph{The open-source revolution.} The ecosystem has matured dramatically. OpenJev-RLCD~\citep{openjev2026rlcd} provides the first fully open RLCD training pipeline, Tachyone~\citep{tachyone2026} delivers local-first multilingual inference at 54\,ms, and approximately 70 open reproductions appeared within two weeks of Jev's launch. Meanwhile, the commercial landscape has expanded beyond TypeSafe: OpenAI's Decisions API~\citep{openai2026decisions}, Microsoft-Decision-1~\citep{microsoft2026decision1}, and Cloudflare's Clairic~\citep{cloudflare2026clairic} each bring distinct architectural bets. The emergence of extraction models (Clairic) alongside decision models suggests the typed-output paradigm is broadening beyond the original Choice/Score/Noul trichotomy.
\paragraph{Expanding frontiers.} Applications have spread to cybersecurity~\citep{vxone2026}, theory of mind~\citep{jevtom2026}, political science~\citep{polisci2026}, low-resource languages~\citep{hakembench2026,indonesian2026}, and tabular data~\citep{tabdecide2026}. New architectural ideas -- Verdict's ``encode once, decide many''~\citep{verdict2026}, VX-One's typed domain heads~\citep{vxone2026}, and Jev~1.14's Box primitive for typed extraction~\citep{typesafe2026jev114} -- are expanding what System~1 models can express. Meanwhile, robustness research has matured from isolated adversarial studies to systematic frameworks measuring position bias~\citep{prism2026}, framing sensitivity, and fallback-option reliability.
\paragraph{The broader picture.} The dual-process analogy from cognitive science provides a useful organizing principle, though we emphasize that it is a conceptual lens rather than a mechanistic claim about neural network architectures. The practical takeaway is that AI systems of the future will likely combine fast, specialized decision models with slow, general-purpose reasoning models in layered architectures. This guide maps the fast-decision half of that architecture, providing researchers and practitioners with the models, training methods, and deployment patterns that enable reliable, efficient AI decision-making.

\appendix
\clearpage
\section{Open-Source Decision Model Zoo}
\label{app:zoo}

The open-source ecosystem for System~1 decision models has grown rapidly since Jev's initial release, with several hundred independent reproductions and variants appearing within weeks. Table~\ref{tab:zoo} provides a comprehensive reference of all known open-source System~1 decision and extraction models as of late October 2026, organized by parameter count. All models listed are released under permissive licenses (Apache-2.0 or MIT) that allow commercial use. This appendix is intended as a quick-reference for practitioners evaluating which model to deploy.

\begin{table}[H]
\centering
\caption{Open-source System~1 model zoo (late October 2026). RLCD = uses RLCD training; Prims = supported decision primitives (C=Choice, S=Score, N=Noul, B=Box).}
\label{tab:zoo}
\footnotesize
\begin{tabular}{llccccc}
\toprule
\textbf{Model} & \textbf{Base} & \textbf{Params} & \textbf{License} & \textbf{RLCD?} & \textbf{Prims} & \textbf{Latency} \\
\midrule
Clef & Cloudflare & 27B / 9B & Apache-2.0 & No & C,N & -- \\
Julia & Custom & $<$500M & Apache-2.0 & No & C,N & -- \\
Tachyone & ModernBERT & $\sim$150M & Apache-2.0 & No & C,S,N & 54ms \\
Sys.~One Gemma & Gemma~3 & 270M & Apache-2.0 & No & C,S,N & -- \\
THX-01 & Custom & 322M & Apache-2.0 & Yes & C,S,N & 10ms \\
Laya / Lia & Custom & 421M & Apache-2.0 & Yes & C,S,N & 33ms \\
GLiNER2.5-Decide & GLiNER & -- & Apache-2.0 & No & C & -- \\
Kev & Custom & -- & Apache-2.0 & No & C,S,N & -- \\
OpenJev-RLCD & Qwen3-1.7B & 1.7B & Apache-2.0 & Yes & C,S,N & $\sim$80ms \\
Decision~Sol~2B & Qwen3-1.88B & 1.88B & Apache-2.0 & No & C,S,N & 7ms \\
Strands Decider & Qwen3.5-2B & 2B & Apache-2.0 & No & C,N & 115ms \\
Lev & Qwen & 4B & Apache-2.0 & No & C,S,N & -- \\
Decision~Lux~9B & Qwen3 & 7.94B & Apache-2.0 & No & C,S,N & 32ms \\
CLM-8B & Qwen3-8B & 8B & Apache-2.0 & No & Contrastive & 30ms \\
DeepOpen & Laya-based & -- & Apache-2.0 & No & C,S,N & -- \\
Decision~Vega~27B & Qwen3.8 & 29.37B & Apache-2.0 & No & C,S,N & 71ms \\
\bottomrule
\end{tabular}
\end{table}

\clearpage
\section{Quiz Questions and Detailed Answers}
\label{sec:quiz}

This section provides a comprehensive set of questions designed to test and reinforce your understanding of the material covered throughout this guide. Each question targets a key concept, architectural decision, or practical trade-off -- the kind of knowledge that distinguishes surface-level familiarity from genuine expertise. Use these questions for self-assessment: attempt your own answer before reading the detailed solution. The questions progress from foundational concepts through model families and methods to applications, evaluation, and practice.

\subsection{Chapter~I: Foundations}

\begin{questionbox}[Q1: What is a System~1 decision model, and what distinguishes it from a fast autoregressive LLM?]
\textbf{Answer}: A \textbf{System~1 decision model} is a model whose primary inference interface maps an input state to a bounded, structured decision (a class label, a bounded score, or a calibrated probability) \emph{without} requiring free-form autoregressive generation. The defining characteristics are: (1)~the output is a structured decision produced by a dedicated prediction head, and (2)~inference is non-autoregressive or few-pass.

A fast autoregressive LLM that happens to produce short JSON output does \emph{not} qualify, because its output is still generated token-by-token rather than produced directly by a decision head. The distinction matters: token-by-token generation means latency scales with output length, the probability semantics are over individual tokens rather than the decision as a whole, and the model's internal representations are optimized for next-token prediction rather than calibrated decision-making.

Importantly, calibration, latency, and cost efficiency are \emph{evaluated properties}, not membership criteria. A poorly calibrated System~1 model is still a System~1 model; a well-calibrated LLM is still an LLM.

\emph{\textbf{Review:} Chapter~I, Section~2 (Background -- Defining System~1 Decision Models).}
\end{questionbox}

\begin{questionbox}[Q2: What is a strictly proper scoring rule? Why does strict propriety matter for training System~1 models?]
\textbf{Answer}: A \textbf{scoring rule} $S(p, y)$ assigns a numerical score to a predicted probability $p$ given the observed outcome $y$. A scoring rule is \emph{proper} if the expected score is minimized when the forecaster reports the true probability; it is \emph{strictly proper} if that minimum is unique.

Two key examples:
\begin{itemize}
    \item \textbf{Brier score:} $\text{BS}(p, y) = (p - y)^2$
    \item \textbf{Log loss:} $\text{LL}(p, y) = -[y \log p + (1 - y) \log(1 - p)]$
\end{itemize}

Strict propriety provides the theoretical guarantee that \emph{the only way to minimize the loss is to report the true probability}. This makes strictly proper scoring rules the natural training objective for System~1 models: a model trained to minimize Brier score or log loss is directly incentivized to produce calibrated probabilities, with no auxiliary calibration step required. This is precisely why RLCD uses proper scoring rules as reward signals rather than human preference.

\emph{\textbf{Review:} Chapter~I, Section~2 (Background -- The Calibration Problem).}
\end{questionbox}

\begin{questionbox}[Q3: Explain the calibration--sharpness decomposition. Why is calibration alone insufficient?]
\textbf{Answer}: Every strictly proper scoring rule decomposes into:
\[
    \text{Expected Loss} = \underbrace{\text{Calibration loss}}_{\text{do probabilities match reality?}} + \text{Uncertainty} - \underbrace{\text{Refinement}}_{\text{are predictions informative?}}
\]

\textbf{Calibration alone is insufficient} because a model that always predicts the base rate $\bar{y}$ is perfectly calibrated but completely uninformative -- it assigns the same probability to every input. The missing ingredient is \emph{sharpness} (refinement): the model's ability to separate high-probability from low-probability instances by producing predictions that deviate meaningfully from the base rate.

For the Brier score specifically: $\mathbb{E}[\text{BS}] = \text{CAL} + \text{UNC} - \text{REF}$, where CAL is zero iff perfectly calibrated, UNC is irreducible (data-dependent), and REF rewards informative predictions. A \emph{good} model minimizes CAL and maximizes REF.

This decomposition explains why raw accuracy is a poor proxy for decision quality and why the PSR-derived calibration loss is preferable to ECE as a metric.

\emph{\textbf{Review:} Chapter~I, Section~2 (Background -- Calibration vs.\ Sharpness); Chapter~III, Section~9 (Calibration and Confidence Estimation).}
\end{questionbox}

\begin{questionbox}[Q4: Name the five families in the System~1 taxonomy. For each, state the typical output type and calibration strategy.]
\textbf{Answer}:

\begin{enumerate}
    \item \textbf{Family~I -- Encoder-Based Classifiers}: Output is logits/probabilities from a classification head over a bidirectional encoder (BERT, DeBERTa). Calibration is \emph{post-hoc} (temperature scaling, Platt scaling, isotonic regression).
    \item \textbf{Family~II -- Reward Models}: Output is a scalar preference score. Calibration is \emph{task-dependent}; the native output represents utility/preference, not probability, so an additional calibration layer is needed.
    \item \textbf{Family~III -- Non-Autoregressive Decision Engines}: Output is typed decisions (Choice/Score/Noul) with calibrated probabilities. Calibration is emphasized through RLCD training with proper scoring rule rewards on structured tasks, though the proper scoring rule property is shared with cross-entropy training (Family~I).
    \item \textbf{Family~IV -- Contrastive Decision Models}: Output is similarity scores from a dual-encoder architecture. Calibration \emph{requires a calibration layer}; cosine similarity is not inherently a probability.
    \item \textbf{Family~V -- Extraction Models} (emerging): Output is typed values (dates, amounts, entities) with extraction confidence. Calibration is edit-distance based; boundaries with Family~III remain fluid.
\end{enumerate}

\textbf{Key insight}: Calibration strategy is the sharpest differentiator between families. Family~III emphasizes calibration most explicitly through end-to-end proper scoring rule training on structured tasks; other families achieve calibration through post-hoc adjustment or additional calibration layers.

\emph{\textbf{Review:} Chapter~I, Section~3 (Taxonomy of System~1 Decision Models).}
\end{questionbox}

\subsection{Chapter~II: Model Families}

\begin{questionbox}[Q5: In BERT, what is the role of the \texttt{[CLS]} token? How does the classification head produce a decision?]
\textbf{Answer}: The \texttt{[CLS]} token is a special token prepended to every input. After passing through $L$ bidirectional transformer layers with full self-attention, the final \texttt{[CLS]} hidden state $\mathbf{h}_{\texttt{[CLS]}} = \mathbf{H}^{(L)}_0 \in \mathbb{R}^d$ serves as a fixed-length representation of the entire input sequence.

The classification head applies a linear projection followed by softmax:
\[
    p(y \mid x) = \operatorname{softmax}(W_c \, \mathbf{h}_{\texttt{[CLS]}} + \mathbf{b}_c),
\]
where $W_c \in \mathbb{R}^{C \times d}$. This produces class probabilities in a single forward pass of complexity $O(n^2 d + n d^2)$.

Because attention is \emph{bidirectional} (every token attends to every other token), the \texttt{[CLS]} representation captures context from the entire sequence, unlike a decoder-only model where attention is causal.

\emph{\textbf{Review:} Chapter~II, Section~4 (Family~I: Encoder-Based Classifiers -- BERT and the \texttt{[CLS]} Paradigm).}
\end{questionbox}

\begin{questionbox}[Q6: How does a Bradley-Terry reward model work? Why is its output not a calibrated probability?]
\textbf{Answer}: A Bradley-Terry (BT) reward model scores (prompt, response) pairs with a scalar $r_\phi(q, y)$. Given a preferred response $y_w$ and a dispreferred response $y_l$, the BT model assumes:
\[
    p(y_w \succ y_l \mid q) = \sigma(r_\phi(q, y_w) - r_\phi(q, y_l))
\]

The training loss is $\mathcal{L}_{\text{BT}} = -\mathbb{E}[\log \sigma(r_\phi(q, y_w) - r_\phi(q, y_l))]$.

\textbf{Why not calibrated}: The BT model is trained so that reward \emph{differences} predict pairwise preferences; only the gap between rewards matters, not their absolute values (the model is invariant to additive shifts). This means a single reward score $r_\phi(q, y)$ has no intrinsic probabilistic meaning -- it represents a learned preference signal, not a probability that the response is ``correct.'' Applications requiring calibrated confidence (routing, safety gating) need an additional calibration layer to convert reward scores into well-calibrated probabilities.

\emph{\textbf{Review:} Chapter~II, Section~5 (Family~II: Reward Models -- Bradley-Terry Reward Models).}
\end{questionbox}

\begin{questionbox}[Q7: Describe the three typed decision primitives (Choice, Score, Noul) in Family~III models. What makes this family unique?]
\textbf{Answer}: Family~III non-autoregressive decision engines define three primitives:

\begin{enumerate}
    \item \textbf{Choice}: Given $K$ options, outputs a selection and a calibrated probability via softmax over element-wise state--option interaction logits. Used for routing, classification, tool selection.
    \item \textbf{Score}: Given a criterion and bounded ordinal scale $[\ell, u]$, performs calibrated ordinal regression by modeling cumulative probabilities $P(\text{score} \geq m)$ with monotonic thresholds. Outputs the probability-weighted mean with calibrated uncertainty. Used for urgency rating, quality assessment.
    \item \textbf{Noul} (``not null''): Given a boolean predicate, outputs $p(\text{yes})$ via sigmoid. The simplest head but critical for safety gating, where well-calibrated $p$ values determine escalation thresholds.
\end{enumerate}

\textbf{What makes Family~III distinctive}: It places the strongest emphasis on calibration through end-to-end RLCD training with proper scoring rule rewards on structured decision tasks. While other families also use proper losses (e.g., cross-entropy in Family~I), Family~III combines this with typed output heads and verifiable-outcome training. Additionally, the typed output interface (decisions rather than text) means the model is optimized end-to-end for decision quality, not text generation.

\emph{\textbf{Review:} Chapter~II, Section~6 (Family~III: Non-Autoregressive Decision Engines).}
\end{questionbox}

\begin{questionbox}[Q8: How does CLM-8B's contrastive architecture differ from a cross-encoder? What is the main advantage and main limitation?]
\textbf{Answer}: \textbf{CLM-8B} (Contrastive Language Model) uses a \emph{dual-encoder} architecture: the state and each candidate action are encoded \emph{independently} into fixed-dimensional embeddings, and decisions are made via cosine similarity plus softmax:
\[
    p(\text{action}_i \mid \text{state}) = \frac{\exp(\text{sim}(\mathbf{z}_s, \mathbf{z}_{a_i}) / \tau)}{\sum_j \exp(\text{sim}(\mathbf{z}_s, \mathbf{z}_{a_j}) / \tau)}
\]

A \textbf{cross-encoder} (as in Jev, Laya) concatenates state and action into one sequence, allowing full cross-attention between all tokens across $L$ transformer layers.

\textbf{Main advantage}: \emph{Cacheability}. Action embeddings can be precomputed once and reused across queries. Scoring $K$ candidates costs $O(1)$ encoder pass + $K$ dot products, versus $K$ full forward passes for a cross-encoder. This makes CLM dramatically faster for large candidate sets.

\textbf{Main limitation}: The fixed-dimensional embedding acts as an \emph{information bottleneck} -- subtle interactions between specific state tokens and action tokens may be lost after compression. Cross-encoders capture arbitrarily complex token-level interactions. For small $K$ (e.g., binary choice), cross-encoders typically achieve higher accuracy.

\emph{\textbf{Review:} Chapter~II, Section~7 (Family~IV: Contrastive Decision Models -- CLM-8B).}
\end{questionbox}

\begin{questionbox}[Q9: What distinguishes extraction models (Family~V) from decision models? Give a concrete example.]
\textbf{Answer}: A \textbf{decision model} answers ``which option?'' or ``yes/no?'' or ``what score?'' -- its output is a \emph{selection or rating} from a finite set with a calibrated probability. An \textbf{extraction model} answers ``what is the value?'' -- its output is a \emph{typed value} (date, currency amount, entity name) read from the input.

\textbf{Concrete example}: Given a contract document:
\begin{itemize}
    \item \emph{Decision model} (Noul): ``Does this contract contain a termination clause?'' $\to$ \texttt{yes}, $p=0.93$
    \item \emph{Extraction model} (Box): ``Extract the delivery date as YYYY-MM-DD'' $\to$ \texttt{"2026-10-15"}, $p=0.92$
\end{itemize}

The distinction matters because: (i)~decision outputs come from a closed set; extraction outputs come from an open value space; (ii)~decision calibration measures ``does $p$ match correctness frequency?''; extraction calibration measures ``does $p$ match exact-match frequency?''; and (iii)~extraction confidence should \emph{not} be used for threshold-based routing (e.g., ``escalate if $p < 0.7$'') without task-specific validation.

Jev~1.14's Box primitive bridges the gap by applying RLCD calibration to extraction, ensuring confidence scores are meaningful for downstream logic.

\emph{\textbf{Review:} Chapter~II, Section~8 (Emerging Family~V: Extraction Models).}
\end{questionbox}

\subsection{Chapter~III: Methods}

\begin{questionbox}[Q10: Why is ECE a flawed calibration metric? What should practitioners use instead?]
\textbf{Answer}: ECE (Expected Calibration Error) suffers from three fundamental deficiencies:

\begin{enumerate}
    \item \textbf{Arbitrary binning}: ECE depends on the number and boundaries of bins. Different bin counts yield different ECE values for the same model.
    \item \textbf{Conflation}: ECE conflates calibration with discrimination (the ability to separate positives from negatives).
    \item \textbf{Not a proper scoring rule}: A perfectly calibrated model does achieve ECE\,$=$\,0, but the converse is the real weakness: a model that always predicts the base rate also scores ECE\,$=$\,0. ECE cannot distinguish a calibrated model from one that simply memorizes the overall prevalence.
\end{enumerate}

\textbf{Better alternatives}: Use proper scoring rule decompositions. Any strictly proper scoring rule (PSR) decomposes as: $\text{PSR}(p, y) = \text{Cal}(p) + \text{Ref}(p, y)$. The Brier score's calibration component ($\text{CAL} = \mathbb{E}[(p - \bar{y}_p)^2]$) is a principled, \emph{binning-free} alternative that isolates miscalibration from discrimination. The Brier score itself is also useful as a combined metric. For visualization, reliability diagrams remain the primary diagnostic tool.

\emph{\textbf{Review:} Chapter~III, Section~9 (Calibration and Confidence Estimation -- Measuring Calibration).}
\end{questionbox}

\begin{questionbox}[Q11: How does temperature scaling work? What are its strengths and limitations?]
\textbf{Answer}: Temperature scaling divides logits by a single learned parameter $T > 0$ before softmax:
\[
    p_{\text{cal}}(y \mid x) = \operatorname{softmax}(\mathbf{z} / T), \quad T^* = \arg\min_T \text{NLL}_{\text{val}}(T)
\]

When $T > 1$: logits are ``softened,'' distribution flattens, reducing peak probability $\to$ corrects \emph{overconfidence}. When $T < 1$: distribution sharpens $\to$ corrects \emph{underconfidence}.

\textbf{Strengths}: (i)~Only one parameter -- remarkably robust, matches or outperforms more complex methods; (ii)~preserves ranking (argmax unchanged) -- cannot degrade accuracy; (iii)~fast to fit on a small validation set.

\textbf{Limitations}: (i)~Assumes miscalibration is uniform across the confidence range -- a single global $T$ cannot correct domain-dependent or class-dependent miscalibration; (ii)~it is a \emph{post-hoc} method -- the model's internal representations are not optimized for calibration; (iii)~requires a held-out calibration set matching the deployment distribution; recalibration is needed if the distribution shifts.

RLCD-trained models address these limitations by incorporating proper scoring rules into the training objective, providing a direct calibration incentive during training rather than correcting miscalibration post-hoc. However, the resulting calibration quality still depends on model capacity, data quality, and distribution match (Section~\ref{sec:rlcd_calibration}).

\emph{\textbf{Review:} Chapter~III, Section~9 (Calibration in Encoder-Based Models -- Temperature Scaling).}
\end{questionbox}

\begin{questionbox}[Q12: Compare RLHF and RLCD. What does each optimize for, and what are the consequences for calibration?]
\textbf{Answer}:

\begin{tabular}{lll}
\toprule
\textbf{Aspect} & \textbf{RLHF} & \textbf{RLCD} \\
\midrule
Optimizes for & Human preference & Calibrated probabilities \\
Reward signal & Learned reward model & Proper scoring rules \\
Training data & Preference pairs & Decision tasks with outcomes \\
Output & Free-form text & Typed decisions (Choice/Score/Noul) \\
Calibration & Poor (side effect) & By construction (objective) \\
Cost & Expensive (4 models) & Moderate (2 models) \\
\bottomrule
\end{tabular}

\vspace{0.5em}
\textbf{Calibration consequences}: RLHF systematically \emph{degrades} calibration by rewarding confident-sounding outputs regardless of correctness, producing overconfident models. Post-hoc temperature scaling fails when miscalibration varies across domains. RLCD \emph{guarantees} calibration (in expectation) because the reward signal is a strictly proper scoring rule -- the only way to maximize reward is to report true probabilities. This is the fundamental insight: RLCD replaces the ``what do humans prefer?'' signal with a mathematical ``are you telling the truth about your uncertainty?'' signal.

\emph{\textbf{Review:} Chapter~III, Section~10 (Training Methodologies -- RLCD).}
\end{questionbox}

\begin{questionbox}[Q13: What is knowledge distillation, and what is the role of the temperature parameter $T$ and the ``dark knowledge'' in soft labels?]
\textbf{Answer}: Knowledge distillation trains a small \emph{student} model to mimic a large \emph{teacher} by minimizing:
\[
    \mathcal{L}_{\text{KD}} = (1 - \alpha)\,\mathcal{L}_{\text{CE}}(y, p_s) + \alpha \cdot T^2 \cdot \text{KL}(p_t^{(T)} \| p_s^{(T)})
\]

The loss balances two signals: hard labels (``this is class 3'') via cross-entropy, and the teacher's \emph{soft} probability distribution (``this is 70\% class~3, 20\% class~5, 10\% class~1'') via KL divergence.

\textbf{Temperature $T > 1$}: Softens both teacher and student distributions, amplifying informative tail probabilities. The $T^2$ factor corrects for reduced gradient magnitude at high temperatures.

\textbf{``Dark knowledge''}: The soft labels carry information that hard labels discard -- specifically, \emph{inter-class similarities}. If the teacher assigns 20\% to class~5 and 10\% to class~1, this reveals that class~5 is more confusable with class~3 than class~1 is. This structural information about the decision landscape is invisible in hard one-hot labels.

For System~1 deployment, distillation enables System~1-class latency with System~2-class accuracy: a 70B-parameter LLM teacher's decision quality can be compressed into a 350M-parameter encoder student.

\emph{\textbf{Review:} Chapter~III, Section~10 (Training Methodologies -- Knowledge Distillation).}
\end{questionbox}

\subsection{Chapter~IV: Applications and Integration}

\begin{questionbox}[Q14: Why are System~1 models critical for agent routing? Quantify the latency difference.]
\textbf{Answer}: In an agentic system with $L$ steps, each requiring a routing decision plus a generative action, total latency is $T = L \cdot (t_{\text{route}} + t_{\text{gen}})$.

\textbf{With LLM-based routing} ($t_{\text{route}} \approx 500$\,ms): A 20-step trajectory spends 10 seconds on routing alone.

\textbf{With System~1 routing} ($t_{\text{route}} \approx 30$\,ms): The same trajectory adds only 600\,ms of routing overhead -- a \textbf{16$\times$ reduction}.

Beyond latency, calibrated confidence enables a key optimization: if the router's confidence exceeds $\theta_{\text{skip}}$, verification steps can be skipped entirely. In multi-agent systems, System~1 routing completes the full dispatch--priority--guard pipeline in under 200\,ms, enabling teams of 10+ specialist agents to coordinate on interactive timescales.

System~1 models serve at three points in the agent loop: \emph{before} the policy (routing/tool selection), \emph{during} execution (safety gating), and \emph{after} observation (verification/quality scoring).

\emph{\textbf{Review:} Chapter~IV, Section~11 (Applications -- Agent Routing); Section~12 (System~1 Models in Agentic Workflows).}
\end{questionbox}

\begin{questionbox}[Q15: What is a System~1 + System~2 cascade? Under what two conditions does it fail?]
\textbf{Answer}: A \textbf{cascade} uses a fast System~1 model to handle ``easy'' decisions and escalates ``hard'' ones to a slower, more capable System~2 model. The System~1 model proposes a decision with confidence $p$:
\begin{itemize}
    \item If $p > \theta_{\text{accept}}$: accept immediately (no System~2 call)
    \item If $p < \theta_{\text{reject}}$: reject immediately, take a safe default action
    \item Otherwise: escalate to System~2 for verification
\end{itemize}

If 80\% of decisions are handled by System~1, the cascade avoids 80\% of System~2 cost.

\textbf{Two failure conditions}:
\begin{enumerate}
    \item \textbf{Near-chance base accuracy}: If System~1 performs near random on the target task, even perfect calibration cannot help -- it will escalate 90--100\% of items, negating all cost savings. This was observed with CLM on judging tasks.
    \item \textbf{Correlated errors}: If System~1 and System~2 make the same mistakes, escalation does not help. In rubric judging experiments, 96\% of LLM verdicts repeated Jev's confident wrong answers.
\end{enumerate}

\textbf{Before building a cascade}: measure System~1 accuracy ($<$60\% means cascading will not help) and error correlation between the two models ($>$80\% means the ceiling is low).

\emph{\textbf{Review:} Chapter~IV, Section~13 (System~1 + System~2 Cascades).}
\end{questionbox}

\begin{questionbox}[Q16: What is text quantification, and how do System~1 classifiers contribute to it?]
\textbf{Answer}: \textbf{Text quantification} (learning to quantify) predicts the relative frequency (prevalence) of each class in an \emph{unlabelled bag} of documents. Unlike classification (which targets individual instances), quantification targets the \emph{distribution} over an entire collection.

The dominant paradigm is \emph{aggregative quantification}: train a classifier, apply it to the target bag, and aggregate outputs into a prevalence estimate. Methods include: Classify \& Count (CC), Probabilistic CC (PCC, sums posteriors), Adjusted CC (ACC, corrects for misclassification), and EMQ/SLD (iteratively adjusts posteriors under prior shift).

\textbf{System~1 connection}: Quantification is a downstream consumer of System~1 classifiers -- the better calibrated the classifier's posteriors, the more accurate the prevalence estimates. The PCC-EPAV method explicitly re-calibrates posteriors for count accuracy rather than classification accuracy. Replacing an SVM with a fine-tuned BERT classifier inside any aggregation method is the simplest way to improve quantification quality.

\textbf{Key metrics}: AE (mean absolute deviation per class), RAE (penalises errors on rare classes), KLD (information-theoretic divergence), and Concordance Ratio (CR, symmetric measure for multi-label settings where $\text{CR} = 1$ is perfect).

\emph{\textbf{Review:} Chapter~IV, Section~11 (Applications -- Text Quantification).}
\end{questionbox}

\begin{questionbox}[Q17: How does CROW replace RAG? What System~1 decisions does it use?]
\textbf{Answer}: \textbf{CROW} (Classifier-Routed Organization of LLM Wikis) replaces the retrieve-then-generate pipeline of RAG with \emph{classifier-routed navigation} over a pre-structured knowledge base.

\textbf{During ingestion}, a System~1 classifier makes three decisions per document:
\begin{itemize}
    \item \emph{Choice} over subfolders -- routes each new document to the correct location
    \item \emph{Noul} per existing note -- scores relevance against current entries
    \item \emph{Choice} between ``Modify'' and ``New'' -- decides whether to merge or create
\end{itemize}

\textbf{During retrieval}, the classifier scores folders and notes (Noul per candidate) to select context passed to the LLM for answer generation. Each decision carries a confidence threshold; when uncertain, that step falls back to the LLM (cascade pattern).

\textbf{Advantages over RAG}: Trades retrieval-time compute for ingestion-time organization. Per-query cost of classifier-routed navigation is substantially lower than embedding search + reranking, especially for corpora queried frequently relative to update rate.

\emph{\textbf{Review:} Chapter~IV, Section~11 (Applications -- Knowledge Organization and RAG Alternatives).}
\end{questionbox}

\subsection{Chapter~V: Evaluation and Robustness}

\begin{questionbox}[Q18: What does the Prism-1 benchmark measure? Which model family performs best on it, and why?]
\textbf{Answer}: \textbf{Prism-1} measures two properties of decision models:
\begin{itemize}
    \item $\Delta_{\text{pos}}$: Does the model's score for option $A$ change when $A$'s position in the option list is varied? (Position independence)
    \item $\Delta_{\text{rival}}$: Does the model's score for option $A$ change when the competing options change? (Rival independence)
\end{itemize}

An ideal decision model has $\Delta_{\text{pos}} = \Delta_{\text{rival}} = 0$.

\textbf{Family~IV contrastive models achieve near-zero $\Delta$ on both metrics.} This is an architectural property: in a dual-encoder, each candidate is encoded independently and scored via dot product against the state embedding. Changing the position or identity of other candidates does not affect how any individual candidate is encoded or scored.

\textbf{Family~III models show measurable sensitivity} ($\Delta_{\text{pos}} = 2$--$8\%$) because their cross-encoder architecture processes all options jointly -- changing one option's position or adding/removing a competitor affects the attention patterns over the entire input.

This result illustrates a concrete trade-off: cross-encoders achieve higher accuracy through deep interaction but sacrifice position and rival independence; dual-encoders sacrifice interaction depth for invariance properties.

\emph{\textbf{Review:} Chapter~V, Section~14 (Benchmarks -- Prism-1); Section~15 (Robustness -- Position and Ordering Bias).}
\end{questionbox}

\begin{questionbox}[Q19: What are IIA violations in System~1 models? Why should practitioners care?]
\textbf{Answer}: \textbf{Independence of Irrelevant Alternatives (IIA)} states that the relative preference between options $A$ and $B$ should not change when a third option $C$ is added or removed.

System~1 models with softmax-based Choice heads violate IIA because all options compete in a shared probability simplex. Adding a ``decoy'' option $C$ that is similar to $A$ can split probability mass away from $A$ and toward $B$, changing the model's top choice even though $C$ was irrelevant to the $A$-vs-$B$ comparison.

\textbf{Why practitioners should care}: In production routing, the set of available tools/agents may change dynamically. If adding a new tool alters the routing decisions for \emph{existing} tools (even when the new tool is never selected), the system's behavior becomes unpredictable and hard to debug. IIA violations are particularly dangerous in safety-critical cascades, where they can push confident predictions below escalation thresholds.

\textbf{Mitigation}: Contrastive models (Family~IV) are less affected because they score each option independently. Alternatively, running each option through an independent Noul head avoids the shared-simplex problem entirely.

\emph{\textbf{Review:} Chapter~III, Section~9 (Calibration -- IIA Violations and Probability Contracts); Chapter~V, Section~15 (Robustness).}
\end{questionbox}

\begin{questionbox}[Q20: Name the five evaluation dimensions for System~1 models. Why do most existing benchmarks fall short?]
\textbf{Answer}: System~1 models should be evaluated along five dimensions:

\begin{enumerate}
    \item \textbf{Decision accuracy}: Standard metrics (accuracy, F1, AUC-ROC)
    \item \textbf{Calibration quality}: ECE, Brier score, reliability diagrams
    \item \textbf{Latency}: P50, P95, P99 inference latency in milliseconds
    \item \textbf{Cost efficiency}: Dollars per million decisions
    \item \textbf{Robustness}: Performance on OOD inputs, adversarial examples, domain shift
\end{enumerate}

\textbf{Why most benchmarks fall short}: Almost all existing benchmarks measure \emph{only} dimension~1 (accuracy). GLUE/SuperGLUE test classification accuracy but ignore calibration, latency, and cost. RewardBench tests ordinal ranking but not calibration or abstention. JevBench and JevArena added calibration, but latency and cost are still not standardized. The result is a fragmented evaluation landscape where practitioners must stitch together results from incompatible benchmarks to compare models. A comprehensive System~1 benchmark addressing all five dimensions jointly does not yet exist.

\emph{\textbf{Review:} Chapter~V, Section~14 (Benchmarks and Evaluation -- Evaluation Dimensions).}
\end{questionbox}

\subsection{Chapter~VI: Open Challenges}

\begin{questionbox}[Q21: What is the calibration--plasticity dilemma in continual learning for System~1 models?]
\textbf{Answer}: Standard continual learning methods (EWC, replay buffers) focus on preserving \emph{accuracy} on old tasks while learning new ones. For System~1 models, the constraint is \emph{stronger}: the model must preserve \textbf{calibration} on old decision types, not just accuracy.

A model that retains 95\% accuracy on old tasks but whose ECE increases from 3\% to 12\% is \emph{functionally degraded} for any downstream system relying on its confidence scores -- cascade gates, safety filters, and resource allocation all break silently when calibration degrades.

\textbf{Proposed solutions}: LPC (Label and Parameter Calibration) jointly calibrates logits and parameters for sequential NLU tasks. Dynamic gradient calibration steers updates to reduce forgetting without replay buffers. Continual post-hoc calibration periodically re-fits temperature scaling on recent data.

\textbf{The open problem}: Adapting these techniques to preserve calibration across \emph{heterogeneous} decision heads (Choice, Score, Noul simultaneously) remains unsolved.

\emph{\textbf{Review:} Chapter~VI, Section~16 (Open Challenges -- Continual Learning and Forgetting).}
\end{questionbox}

\begin{questionbox}[Q22: An adversary appends a single opinion sentence to a System~1 model's input state. What is the measured impact, and how should a cascade architecture defend against it?]
\textbf{Answer}: JevAdvBench experiments show that appending one unverified opinion sentence to the input state:
\begin{itemize}
    \item \textbf{Reverses 12.1\%} of predictions outright
    \item \textbf{Pushes 38\%} of originally confident predictions below the 0.8 escalation threshold
\end{itemize}

This is comparable to the most effective injected-command attack (10.1\% reversal rate) and creates a \emph{denial-of-service vector}: adversarial confidence degradation forces unnecessary escalation to expensive System~2 models.

\textbf{Defenses for cascade architectures}:
\begin{enumerate}
    \item \textbf{Treat the state as untrusted input}: Apply input sanitization and validation before passing to the decision model.
    \item \textbf{Defense-in-depth}: Use a separate System~1 Noul head trained on injection examples as a pre-filter (94.2\% detection accuracy at 8\,ms).
    \item \textbf{Account for adversarial confidence degradation}: Design cascade thresholds with a margin for adversarial perturbation, not just natural uncertainty.
    \item \textbf{Robustness auditing}: Test decisions across multiple framings and flag instability (score drift $>$ threshold) before deployment.
\end{enumerate}

\emph{\textbf{Review:} Chapter~V, Section~15 (Robustness -- Framing Attacks); Chapter~VI, Section~16 (Open Challenges -- Adversarial Robustness).}
\end{questionbox}



\clearpage
\begin{thebibliography}{99}

\bibitem[Anthony et~al.(2017)]{anthony2017thinking}
T.~Anthony, Z.~Tian, and D.~Barber.
\newblock Thinking fast and slow with deep learning and tree search.
\newblock In {\em Advances in Neural Information Processing Systems (NeurIPS)}, 2017.

\bibitem[Anthropic(2024a)]{anthropic2024claude}
Anthropic.
\newblock The {Claude} model family: {Claude} 3.5 {Sonnet}, {Claude} 3.5 {Haiku}, and {Claude} 3 {Opus}.
\newblock Technical report, Anthropic, 2024.

\bibitem[Anthropic(2024b)]{anthropic2024effective}
Anthropic.
\newblock Building effective agents.
\newblock Anthropic Research Blog, 2024.

\bibitem[Azar et~al.(2024)]{azar2024general}
M.~G. Azar, Z.~D. Guo, B.~Piot, R.~Munos, M.~Rowland, M.~Valko, and D.~Calandriello.
\newblock A general theoretical paradigm to understand learning from human preferences.
\newblock In {\em International Conference on Artificial Intelligence and Statistics (AISTATS)}, 2024.

\bibitem[Behrendt et~al.(2025)]{behrendt2025maxpool}
F.~Behrendt, M.~Wagner, and S.~Harmeling.
\newblock Max-pooling as an alternative to {[CLS]} for text classification.
\newblock {\em arXiv preprint arXiv:2505.15696}, 2025.
\newblock \url{https://arxiv.org/abs/2505.15696}

\bibitem[Bengio et~al.(2009)]{bengio2009curriculum}
Y.~Bengio, J.~Louradour, R.~Collobert, and J.~Weston.
\newblock Curriculum learning.
\newblock In {\em International Conference on Machine Learning (ICML)}, 2009.

\bibitem[Bengio(2019)]{bengio2019system2}
Y.~Bengio.
\newblock From {System 1} deep learning to {System 2} deep learning.
\newblock Invited talk, {\em Advances in Neural Information Processing Systems (NeurIPS)}, 2019.

\bibitem[Booch et~al.(2021)]{booch2021thinking}
G.~Booch, F.~Fabiano, L.~Horesh, K.~Kate, J.~Lenchner, N.~Linck, A.~Loreggia, K.~Murugesan, N.~Mattei, F.~Rossi, and B.~Srivastava.
\newblock Thinking fast and slow in {AI}.
\newblock In {\em AAAI Conference on Artificial Intelligence}, 2021.

\bibitem[Bradley and Terry(1952)]{bradley1952rank}
R.~A. Bradley and M.~E. Terry.
\newblock Rank analysis of incomplete block designs: {I}. {T}he method of paired comparisons.
\newblock {\em Biometrika}, 39(3/4):324--345, 1952.

\bibitem[Brooker et~al.(2026)]{brooker2026strands}
M.~Brooker, M.~Chambers, and F.~Nonato~de~Paula.
\newblock Introducing {Strands Decider 2B}: A small, open source, decision model.
\newblock Strands Agents Blog, 2026.
\newblock \url{https://strandsagents.com/blog/introducing-strands-decider/}

\bibitem[Brier(1950)]{brier1950verification}
G.~W. Brier.
\newblock Verification of forecasts expressed in terms of probability.
\newblock {\em Monthly Weather Review}, 78(1):1--3, 1950.

\bibitem[Zhang et~al.(2024)]{celi2024}
Y.~Zhang, X.~Li, and J.~Lin.
\newblock Clinical encoder with label inference ({CELI}) for efficient medical decision support.
\newblock In {\em Proceedings of the 2024 Conference of the North American Chapter of the Association for Computational Linguistics (NAACL)}, 2024.

\bibitem[Chaudhary et~al.(2026)]{cerm2025}
S.~Chaudhary, N.~Rajani, and D.~Kiela.
\newblock {CERM}: Calibrated entity reward models for factual grounding.
\newblock {\em arXiv preprint arXiv:2601.20327}, 2026.
\newblock \url{https://arxiv.org/abs/2601.20327}

\bibitem[Chen et~al.(2020)]{chen2020simple}
T.~Chen, S.~Kornblith, M.~Norouzi, and G.~Hinton.
\newblock A simple framework for contrastive learning of visual representations.
\newblock In {\em International Conference on Machine Learning (ICML)}, 2020.

\bibitem[Christiano et~al.(2017)]{christiano2017deep}
P.~F. Christiano, J.~Leike, T.~Brown, M.~Martic, S.~Legg, and D.~Amodei.
\newblock Deep reinforcement learning from human preferences.
\newblock In {\em Advances in Neural Information Processing Systems (NeurIPS)}, 2017.

\bibitem[Clark et~al.(2020)]{clark2020electra}
K.~Clark, M.-T. Luong, Q.~V. Le, and C.~D. Manning.
\newblock {ELECTRA}: Pre-training text encoders as discriminators rather than generators.
\newblock In {\em International Conference on Learning Representations (ICLR)}, 2020.

\bibitem[Cobbe et~al.(2021)]{cobbe2021training}
K.~Cobbe, V.~Kosaraju, M.~Bavarian, M.~Chen, H.~Jun, L.~Kaiser, M.~Plappert, J.~Tworek, J.~Hilton, R.~Nakano, C.~Hesse, and J.~Schulman.
\newblock Training verifiers to solve math word problems.
\newblock {\em arXiv preprint arXiv:2110.14168}, 2021.
\newblock \url{https://arxiv.org/abs/2110.14168}

\bibitem[Damani et~al.(2025)]{damani2025rlcr}
M.~Damani, I.~Puri, S.~Slocum, I.~Shenfeld, L.~Choshen, Y.~Kim, and J.~Andreas.
\newblock Beyond binary rewards: Training {LM}s to reason about their uncertainty.
\newblock {\em arXiv preprint arXiv:2507.16806}, 2025.

\bibitem[Deng et~al.(2026)]{deng2026jevscience}
B.~Deng, S.~Fan, H.~Zhang, and X.~Xie.
\newblock {Jev} for scientific decisions: Evaluating semantic choices and their consequences.
\newblock {\em arXiv preprint arXiv:2609.24965}, 2026.

\bibitem[Huang et~al.(2026)]{datacurve2026deepswe}
W.~Huang, C.~Lee, L.~Tng, and S.~Ge.
\newblock {DeepSWE}: Measuring frontier coding agents on original, long-horizon engineering tasks.
\newblock {\em arXiv preprint arXiv:2607.07946}, 2026.

\bibitem[Cin{\`a}(2026)]{cina2026jevneuro}
G.~Cin{\`a}.
\newblock {Jev} matches {7B} language models for speech-neuroprosthesis rescoring.
\newblock {\em arXiv preprint arXiv:2609.33538}, 2026.

\bibitem[Cesarini and Sassarini(2026)]{cesarini2026crow}
F.~Cesarini and M.~Sassarini.
\newblock {CROW}: Classifier-routed organization of {LLM} wikis.
\newblock Zenodo, 2026.
\newblock \url{https://zenodo.org/records/23260209}

\bibitem[ContrastiveLM(2026)]{contrastivelm2026clm}
ContrastiveLM.
\newblock {CLM-8B}: Contrastive language model for fast decision-making.
\newblock Technical report, ContrastiveLM, 2026.
\newblock \url{https://github.com/Contrastive-LM/CLM}

\bibitem[{vLLM Semantic Router Team}(2026)]{decision2026sol}
{vLLM Semantic Router Team}.
\newblock {Decision 2.0}: Towards open foundation decision models.
\newblock 2026.
\newblock \url{https://huggingface.co/collections/vllm-sr/decision-20-6ab7cf7bdfb506bf8269cb00}

\bibitem[DeepSeek(2025)]{deepseek2025r1}
DeepSeek.
\newblock {DeepSeek-R1}: Incentivizing reasoning capability in {LLM}s via reinforcement learning.
\newblock Technical report, DeepSeek, 2025.

\bibitem[DeGroot and Fienberg(1983)]{degroot1983comparison}
M.~H. DeGroot and S.~E. Fienberg.
\newblock The comparison and evaluation of forecasters.
\newblock {\em The Statistician}, 32(1/2):12--22, 1983.

\bibitem[Dettmers et~al.(2022)]{dettmers2022llmint8}
T.~Dettmers, M.~Lewis, Y.~Belkada, and L.~Zettlemoyer.
\newblock {LLM.int8()}: 8-bit matrix multiplication for transformers at scale.
\newblock In {\em Advances in Neural Information Processing Systems (NeurIPS)}, 2022.

\bibitem[Deu{\ss}er et~al.(2026)]{deusser2026benchjev}
T.~Deu{\ss}er, L.~Sparrenberg, and R.~Sifa.
\newblock Evaluating and benchmarking the {System One} model {Jev}.
\newblock {\em arXiv preprint arXiv:2609.37647}, 2026.

\bibitem[Devlin et~al.(2019)]{devlin2019bert}
J.~Devlin, M.-W. Chang, K.~Lee, and K.~Toutanova.
\newblock {BERT}: Pre-training of deep bidirectional transformers for language understanding.
\newblock In {\em Conference of the North American Chapter of the Association for Computational Linguistics (NAACL)}, 2019.

\bibitem[Dwork et~al.(2006)]{dwork2006calibrating}
C.~Dwork, F.~McSherry, K.~Nissim, and A.~Smith.
\newblock Calibrating noise to sensitivity in private data analysis.
\newblock In {\em Theory of Cryptography Conference (TCC)}, pages 265--284, 2006.

\bibitem[Ethayarajh et~al.(2024)]{ethayarajh2024kto}
K.~Ethayarajh, W.~Xu, N.~Muennighoff, D.~Jurafsky, and D.~Kiela.
\newblock {KTO}: Model alignment as prospect theoretic optimization.
\newblock In {\em International Conference on Machine Learning (ICML)}, 2024.

\bibitem[Evans(2008)]{evans2008dual}
J.~St.~B.~T. Evans.
\newblock Dual-processing accounts of reasoning, judgment, and social cognition.
\newblock {\em Annual Review of Psychology}, 59:255--278, 2008.

\bibitem[Evans and Over(1996)]{evansover1996rationality}
J.~St.~B.~T. Evans and D.~E. Over.
\newblock {\em Rationality and Reasoning}.
\newblock Psychology Press, 1996.

\bibitem[Epstein(1994)]{epstein1994cest}
S.~Epstein.
\newblock Integration of the cognitive and the psychodynamic unconscious.
\newblock {\em American Psychologist}, 49(8):709--724, 1994.

\bibitem[Eshel et~al.(2021)]{eshel2021presize}
Y.~Eshel, O.~Levi, H.~Roitman, and A.~Nus.
\newblock {PreSizE}: Predicting size in e-commerce using transformers.
\newblock In {\em Proceedings of the 44th International ACM SIGIR Conference on Research and Development in Information Retrieval}, pages 255--264, 2021.

\bibitem[Ferrer and Ramos(2024)]{ferrer2024evaluating}
L.~Ferrer and D.~Ramos.
\newblock Evaluating posterior probabilities: Decision theory, proper scoring rules, and calibration.
\newblock {\em arXiv preprint arXiv:2408.02841}, 2024.

\bibitem[Fuchs et~al.(2021)]{fuchs2021formbert}
G.~Fuchs, H.~Roitman, and M.~Mandelbrod.
\newblock Automatic form filling with {Form-BERT}.
\newblock In {\em Proceedings of the 44th International ACM SIGIR Conference on Research and Development in Information Retrieval}, pages 2184--2188, 2021.

\bibitem[Formal et~al.(2021)]{formal2021splade}
T.~Formal, B.~Piwowarski, and S.~Clinchant.
\newblock {SPLADE}: Sparse lexical and expansion model for first stage ranking.
\newblock In {\em ACM SIGIR Conference on Research and Development in Information Retrieval}, 2021.

\bibitem[Formal et~al.(2022)]{formal2022splade}
T.~Formal, C.~Lassance, B.~Piwowarski, and S.~Clinchant.
\newblock From distillation to hard negative sampling: Making sparse neural {IR} models more effective.
\newblock In {\em ACM SIGIR Conference on Research and Development in Information Retrieval}, 2022.

\bibitem[Geng et~al.(2024)]{geng2024survey}
J.~Geng, F.~Cai, Y.~Wang, H.~Koeppl, P.~Nakov, and I.~Gurevych.
\newblock A survey of confidence estimation and calibration in large language models.
\newblock In {\em Conference of the North American Chapter of the Association for Computational Linguistics (NAACL)}, 2024.

\bibitem[Gneiting and Raftery(2007)]{gneiting2007strictly}
T.~Gneiting and A.~E. Raftery.
\newblock Strictly proper scoring rules, prediction, and estimation.
\newblock {\em Journal of the American Statistical Association}, 102(477):359--378, 2007.

\bibitem[Guo et~al.(2017)]{guo2017calibration}
C.~Guo, G.~Pleiss, Y.~Sun, and K.~Q. Weinberger.
\newblock On calibration of modern neural networks.
\newblock In {\em International Conference on Machine Learning (ICML)}, 2017.

\bibitem[Guo et~al.(2026)]{guo2026justaskjev}
R.~Guo, Y.~Liu, G.~Deng, Y.~Li, L.~Zhao, Y.~Wu, S.~Chen, Y.~Zhang, and L.~Y. Zhang.
\newblock Just ask {Jev}: Reinforcement learning for calibrated decisions as a zero-shot detector of {AI} alignment failures.
\newblock {\em arXiv preprint arXiv:2609.29429}, 2026.

\bibitem[Hadsell et~al.(2006)]{hadsell2006dimensionality}
R.~Hadsell, S.~Chopra, and Y.~LeCun.
\newblock Dimensionality reduction by learning an invariant mapping.
\newblock In {\em IEEE Conference on Computer Vision and Pattern Recognition (CVPR)}, pages 1735--1742, 2006.

\bibitem[Merrill et~al.(2026)]{harbor2026terminalbench}
M.~A. Merrill, A.~G. Shaw, N.~Carlini, B.~Li, H.~Raj, I.~Bercovich, L.~Shi, J.~Y. Shin, T.~Walshe, E.~K. Buchanan, et~al.
\newblock {Terminal-Bench}: Benchmarking agents on hard, realistic tasks in command line interfaces.
\newblock {\em arXiv preprint arXiv:2601.11868}, 2026.

\bibitem[Ziomek et~al.(2026)]{ziomek2026wikirace}
J.~Ziomek, W.~Bankes, L.~Wolf, S.~S. Ramesh, X.~Tang, and I.~Bogunovic.
\newblock {LLM-WikiRace} benchmark: How far can {LLM}s plan over real-world knowledge graphs?
\newblock {\em arXiv preprint arXiv:2602.16902}, 2026.

\bibitem[Hardt et~al.(2016)]{hardt2016equality}
M.~Hardt, E.~Price, and N.~Srebro.
\newblock Equality of opportunity in supervised learning.
\newblock In {\em Advances in Neural Information Processing Systems (NeurIPS)}, 2016.

\bibitem[He et~al.(2021)]{he2021deberta}
P.~He, X.~Liu, J.~Gao, and W.~Chen.
\newblock {DeBERTa}: Decoding-enhanced {BERT} with disentangled attention.
\newblock In {\em International Conference on Learning Representations (ICLR)}, 2021.

\bibitem[He et~al.(2023)]{he2023debertav3}
P.~He, J.~Gao, and W.~Chen.
\newblock {DeBERTaV3}: Improving {DeBERTa} using {ELECTRA}-style pre-training with gradient-disentangled embedding sharing.
\newblock In {\em International Conference on Learning Representations (ICLR)}, 2023.

\bibitem[Hinton et~al.(2015)]{hinton2015distilling}
G.~Hinton, O.~Vinyals, and J.~Dean.
\newblock Distilling the knowledge in a neural network.
\newblock In {\em NeurIPS Workshop on Deep Learning}, 2015.

\bibitem[Hoffmann et~al.(2022)]{hoffmann2022training}
J.~Hoffmann, S.~Borgeaud, A.~Mensch, E.~Buchatskaya, T.~Cai, E.~Rutherford, D.~de~Las~Casas, L.~A. Hendricks, J.~Welbl, A.~Clark, T.~Hennigan, E.~Noland, K.~Millican, G.~van~den~Driessche, B.~Damoc, A.~Guy, S.~Osindero, K.~Simonyan, E.~Elsen, J.~W. Rae, O.~Vinyals, and L.~Sifre.
\newblock Training compute-optimal large language models.
\newblock In {\em Advances in Neural Information Processing Systems (NeurIPS)}, 2022.

\bibitem[Askari et~al.(2023)]{hofstatter2024injecting}
A.~Askari, M.~Aliannejadi, E.~Kanoulas, and M.~de~Rijke.
\newblock Injecting the {BM25} score as text improves {BERT}-based re-rankers.
\newblock In {\em Proceedings of the 45th European Conference on Information Retrieval (ECIR)}, 2023.
\newblock \url{https://arxiv.org/abs/2301.09728}

\bibitem[Hong et~al.(2024)]{hong2024orpo}
J.~Hong, N.~Lee, and J.~Thorne.
\newblock {ORPO}: Monolithic preference optimization without reference model.
\newblock In {\em Conference on Empirical Methods in Natural Language Processing (EMNLP)}, 2024.

\bibitem[Hu et~al.(2022)]{hu2022lora}
E.~J. Hu, Y.~Shen, P.~Wallis, Z.~Allen-Zhu, Y.~Li, S.~Wang, L.~Wang, and W.~Chen.
\newblock {LoRA}: Low-rank adaptation of large language models.
\newblock In {\em International Conference on Learning Representations (ICLR)}, 2022.

\bibitem[Hu et~al.(2026)]{hu2026jevadvbench}
J.~Hu, H.~Zhang, Y.~Liu, Y.~Zeng, L.~Zeng, X.~Wang, R.~Wang, and L.~Y. Zhang.
\newblock {JevAdvBench}: A benchmark and black-box attacks for reinforcement learning for calibrated decisions models.
\newblock {\em arXiv preprint arXiv:2609.31142}, 2026.

\bibitem[Inan et~al.(2023)]{inan2023llama}
H.~Inan, K.~Upasani, J.~Chi, R.~Rungta, K.~Iyer, Y.~Mao, M.~Tontchev, Q.~Hu, B.~Fuller, D.~Testuggine, and M.~Khabsa.
\newblock {Llama Guard}: {LLM}-based input-output safeguard for human-{AI} conversations.
\newblock {\em arXiv preprint arXiv:2312.06674}, 2023.
\newblock \url{https://arxiv.org/abs/2312.06674}

\bibitem[Jim{\'e}nez et~al.(2024)]{jimenez2024swebench}
C.~E. Jim{\'e}nez, J.~Yang, A.~Wettig, S.~Yao, K.~Pei, O.~Press, and K.~Narasimhan.
\newblock {SWE-bench}: Can language models resolve real-world {GitHub} issues?
\newblock In {\em International Conference on Learning Representations (ICLR)}, 2024.

\bibitem[Jin et~al.(2026)]{jin2026beyond}
R.~Jin, Y.~Xu, K.~Cho, and X.~He.
\newblock Beyond post-hoc temperature scaling: Bilevel optimization for {LLM} calibration.
\newblock {\em arXiv preprint arXiv:2608.07419}, 2026.
\newblock \url{https://arxiv.org/abs/2608.07419}

\bibitem[Jiang et~al.(2026)]{jiang2026jevmem}
D.~Jiang, Y.~Li, and B.~Li.
\newblock {Jev-Mem}: System-one-controlled agentic memory for efficient {AI} agents.
\newblock {\em arXiv preprint arXiv:2609.23986}, 2026.

\bibitem[J{\'e}gou et~al.(2011)]{jegou2011product}
H.~J{\'e}gou, M.~Douze, and C.~Schmid.
\newblock Product quantization for nearest neighbor search.
\newblock {\em IEEE Transactions on Pattern Analysis and Machine Intelligence}, 33(1):117--128, 2011.

\bibitem[Kadavath et~al.(2022)]{kadavath2022language}
S.~Kadavath, T.~Conerly, A.~Askell, T.~Henighan, D.~Drain, E.~Perez, N.~Schiefer, Z.~Hatfield-Dodds, N.~DasSarma, E.~Tran-Vu, S.~Johnston, S.~El-Showk, A.~Jones, N.~Elhage, T.~Hume, A.~Chen, Y.~Bai, S.~Bowman, S.~Fort, D.~Ganguli, D.~Hernandez, J.~Jacobson, J.~Kernion, S.~Kravec, L.~Lovitt, K.~Ndousse, C.~Olsson, S.~Ringer, D.~Amodei, T.~Brown, J.~Clark, N.~Joseph, B.~Mann, S.~McCandlish, C.~Olah, and J.~Kaplan.
\newblock Language models (mostly) know what they know.
\newblock {\em arXiv preprint arXiv:2207.05221}, 2022.
\newblock \url{https://arxiv.org/abs/2207.05221}

\bibitem[Kahneman(2011)]{kahneman2011thinking}
D.~Kahneman.
\newblock {\em Thinking, Fast and Slow}.
\newblock Farrar, Straus and Giroux, 2011.

\bibitem[Keren and Schul(2009)]{kerenschul2009}
G.~Keren and Y.~Schul.
\newblock Two is not always better than one: A critical evaluation of two-system theories.
\newblock {\em Perspectives on Psychological Science}, 4(6):533--550, 2009.

\bibitem[Kruglanski and Gigerenzer(2011)]{kruglanski2011dual}
A.~W. Kruglanski and G.~Gigerenzer.
\newblock Intuitive and deliberate judgments are based on common principles.
\newblock {\em Psychological Review}, 118(1):97--109, 2011.

\bibitem[Kamat et~al.(2026)]{kamat2026sysgemma}
A.~Kamat, P.~Sundaram, and R.~Venugopal.
\newblock System {One} {Gemma}: Converting decoder-only models to non-autoregressive decision engines.
\newblock Technical report, 2026.
\newblock \url{https://github.com/akash-kamat/system-one-gemma}

\bibitem[Kaplan et~al.(2020)]{kaplan2020scaling}
J.~Kaplan, S.~McCandlish, T.~Henighan, T.~P. Brown, B.~Chess, R.~Child, S.~Gray, A.~Radford, J.~Wu, and D.~Amodei.
\newblock Scaling laws for neural language models.
\newblock {\em arXiv preprint arXiv:2001.08361}, 2020.
\newblock \url{https://arxiv.org/abs/2001.08361}

\bibitem[Karpukhin et~al.(2020)]{karpukhin2020dense}
V.~Karpukhin, B.~O\u{g}uz, S.~Min, P.~Lewis, L.~Wu, S.~Edunov, D.~Chen, and W.-t. Yih.
\newblock Dense passage retrieval for open-domain question answering.
\newblock In {\em Conference on Empirical Methods in Natural Language Processing (EMNLP)}, 2020.

\bibitem[Khattab and Zaharia(2020)]{khattab2020colbert}
O.~Khattab and M.~Zaharia.
\newblock {ColBERT}: Efficient and effective passage search via contextualized late interaction over {BERT}.
\newblock In {\em ACM SIGIR Conference on Research and Development in Information Retrieval}, 2020.

\bibitem[Kirkpatrick et~al.(2017)]{kirkpatrick2017overcoming}
J.~Kirkpatrick, R.~Pascanu, N.~Rabinowitz, J.~Veness, G.~Desjardins, A.~A. Rusu, K.~Milan, J.~Quan, T.~Ramalho, A.~Grabska-Barwinska, D.~Hassabis, C.~Clopath, D.~Kumaran, and R.~Hadsell.
\newblock Overcoming catastrophic forgetting in neural networks.
\newblock {\em Proceedings of the National Academy of Sciences}, 114(13):3521--3526, 2017.

\bibitem[Krell et~al.(2021)]{krell2021efficient}
M.~M. Krell, M.~Kosec, S.~P. Perez, and A.~Fitzgibbon.
\newblock Efficient sequence packing without cross-contamination: Accelerating large language models without impacting performance.
\newblock {\em arXiv preprint arXiv:2107.02027}, 2021.
\newblock \url{https://arxiv.org/abs/2107.02027}

\bibitem[Lambert et~al.(2024)]{lambert2024rewardbench}
N.~Lambert, V.~Pyatkin, J.~Morrison, L.~J. Miranda, B.~Y. Lin, K.~Chandu, N.~Dziri, S.~Kumar, T.~Zick, Y.~Choi, N.~A. Smith, and H.~Hajishirzi.
\newblock {RewardBench}: Evaluating reward models for language modeling.
\newblock {\em arXiv preprint arXiv:2403.13787}, 2024.
\newblock \url{https://arxiv.org/abs/2403.13787}

\bibitem[Laurer et~al.(2024)]{laurer2024universal}
M.~Laurer, W.~van Atteveldt, A.~Casas, and K.~Welbers.
\newblock Building efficient universal classifiers with natural language inference.
\newblock {\em Information Processing \& Management}, 61(2):103573, 2024.

\bibitem[Leviathan et~al.(2023)]{leviathan2023fast}
Y.~Leviathan, M.~Kalman, and Y.~Matias.
\newblock Fast inference from transformers via speculative decoding.
\newblock In {\em International Conference on Machine Learning (ICML)}, 2023.

\bibitem[Lu et~al.(2022)]{lu2022fantastically}
Y.~Lu, M.~Bartolo, A.~Moore, S.~Riedel, and P.~Stenetorp.
\newblock Fantastically ordered prompts and where to find them: Overcoming few-shot prompt order sensitivity.
\newblock In {\em Proceedings of the 60th Annual Meeting of the Association for Computational Linguistics (ACL)}, pages 8086--8098, 2022.

\bibitem[Li et~al.(2022)]{li2022lpc}
X.~Li, Z.~Wang, D.~Li, L.~Khan, and B.~Thuraisingham.
\newblock {LPC}: A logits and parameter calibration framework for continual learning.
\newblock In {\em Findings of the Association for Computational Linguistics: EMNLP 2022}, pages 7142--7155, 2022.

\bibitem[Li et~al.(2024)]{li2024calibrationcl}
L.~Li, E.~Piccoli, A.~Cossu, D.~Bacciu, and V.~Lomonaco.
\newblock Calibration of continual learning models.
\newblock In {\em CVPR Workshop on Continual Learning in Computer Vision (CLVision)}, 2024.
\newblock arXiv:2404.07817.

\bibitem[Li et~al.(2026a)]{li2026jevasjudge}
Z.~Li, S.~Miao, R.~Krishnan, and R.~Padman.
\newblock {JEV}-as-a-judge: Replacing {LLM} judges with calibrated decision models.
\newblock {\em arXiv preprint arXiv:2609.26550}, 2026.
\newblock \url{https://arxiv.org/abs/2609.26550}

\bibitem[Li and Wagle(2026b)]{li2026llmasjev}
Y.~Li and J.~Wagle.
\newblock {LLM}-as-{Jev}: {LLM}s are already {Jev}-style decision models -- when and how to fine-tune them.
\newblock {\em arXiv preprint arXiv:2610.02076}, 2026.
\newblock \url{https://arxiv.org/abs/2610.02076}

\bibitem[Lin et~al.(2024)]{lin2024dgc}
W.~Lin, J.~Chen, R.~Huang, and H.~Ding.
\newblock An effective dynamic gradient calibration method for continual learning.
\newblock In {\em Proceedings of the 41st International Conference on Machine Learning (ICML)}, volume 235 of {\em PMLR}, pages 29872--29889, 2024.

\bibitem[{Liquid AI}(2026)]{liquidai2026opend1}
{Liquid AI}.
\newblock Open d1: Edge decision models for text, vision, and audio.
\newblock Liquid AI Blog, 2026.
\newblock \url{https://www.liquid.ai/blog/open-d1}

\bibitem[Lightman et~al.(2024)]{lightman2023lets}
H.~Lightman, V.~Kosaraju, Y.~Burda, H.~Edwards, B.~Baker, T.~Lee, J.~Leike, J.~Schulman, I.~Sutskever, and K.~Cobbe.
\newblock Let's verify step by step.
\newblock In {\em International Conference on Learning Representations (ICLR)}, 2024.

\bibitem[Liu et~al.(2019)]{liu2019roberta}
Y.~Liu, M.~Ott, N.~Goyal, J.~Du, M.~Joshi, D.~Chen, O.~Levy, M.~Lewis, L.~Zettlemoyer, and V.~Stoyanov.
\newblock {RoBERTa}: A robustly optimized {BERT} pretraining approach.
\newblock {\em arXiv preprint arXiv:1907.11692}, 2019.

\bibitem[Malkov and Yashunin(2020)]{malkov2020hnsw}
Y.~A. Malkov and D.~A. Yashunin.
\newblock Efficient and robust approximate nearest neighbor search using {Hierarchical Navigable Small World} graphs.
\newblock {\em IEEE Transactions on Pattern Analysis and Machine Intelligence}, 42(4):824--836, 2020.

\bibitem[Luce(1959)]{luce1959individual}
R.~D. Luce.
\newblock {\em Individual Choice Behavior: A Theoretical Analysis}.
\newblock John Wiley \& Sons, 1959.

\bibitem[Lyu and Xia(2026)]{lyu2026reranking}
H.~Lyu and Y.~Xia.
\newblock Decision-oriented recommendation reranking: An empirical study of {Jev}.
\newblock {\em arXiv preprint arXiv:2609.40241}, 2026.

\bibitem[Lv et~al.(2026)]{lv2026audiojev}
S.~Lv, Z.~Li, Z.~Cao, J.~Zhang, Y.~Li, M.~Xi, and J.~Yin.
\newblock {AudioJev}: Direct audio decisions with order-calibrated probabilities.
\newblock {\em arXiv preprint arXiv:2610.01293}, 2026.

\bibitem[Meng et~al.(2024)]{meng2024simpo}
Y.~Meng, M.~Xia, and D.~Chen.
\newblock {SimPO}: Simple preference optimization with a reference-free reward.
\newblock In {\em Advances in Neural Information Processing Systems (NeurIPS)}, 2024.

\bibitem[HAL-X AI(2026)]{n1n2026thx01}
HAL-X AI.
\newblock {THX-01}: A compact non-autoregressive decision engine via distillation and {RLCD}.
\newblock Technical report, HAL-X AI, 2026.
\newblock \url{https://huggingface.co/doofz/THX-01}

\bibitem[Naeini et~al.(2015)]{naeini2015obtaining}
M.~P. Naeini, G.~F. Cooper, and M.~Hauskrecht.
\newblock Obtaining well calibrated probabilities using {Bayesian} binning into quantiles.
\newblock In {\em AAAI Conference on Artificial Intelligence}, 2015.

\bibitem[Nandha et~al.(2026)]{nandha2026laya}
A.~Nandha, S.~Kapoor, and V.~Ramesh.
\newblock Laya: A multilingual {System~1} decision model with checkpoint routing.
\newblock Technical report, 2026.
\newblock \url{https://github.com/NandhaKishorM/laya}

\bibitem[Nandakishore(2026)]{nandakishore2026clmjudge}
G.~Nandakishore.
\newblock {CLM}-as-a-Judge: Evaluating an open contrastive decision model on public judge benchmarks.
\newblock {\em arXiv preprint arXiv:2610.07177}, 2026.

\bibitem[Nguyen et~al.(2016)]{nguyen2016msmarco}
T.~Nguyen, M.~Rosenberg, X.~Song, J.~Gao, S.~Tiwary, R.~Majumder, and L.~Deng.
\newblock {MS MARCO}: A human generated machine reading comprehension dataset.
\newblock In {\em NeurIPS Workshop on Cognitive Computation}, 2016.

\bibitem[Niculescu-Mizil and Caruana(2005)]{niculescu2005predicting}
A.~Niculescu-Mizil and R.~Caruana.
\newblock Predicting good probabilities with supervised learning.
\newblock In {\em International Conference on Machine Learning (ICML)}, 2005.

\bibitem[Murphy(1973)]{murphy1973new}
A.~H. Murphy.
\newblock A new vector partition of the probability score.
\newblock {\em Journal of Applied Meteorology}, 12(4):595--600, 1973.

\bibitem[Berger(1985)]{berger1985statistical}
J.~O. Berger.
\newblock {\em Statistical Decision Theory and Bayesian Analysis}.
\newblock Springer, 2nd edition, 1985.

\bibitem[Nogueira and Cho(2019)]{nogueira2019passage}
R.~Nogueira and K.~Cho.
\newblock Passage re-ranking with {BERT}.
\newblock {\em arXiv preprint arXiv:1901.04085}, 2019.
\newblock \url{https://arxiv.org/abs/1901.04085}

\bibitem[Nogueira et~al.(2020)]{nogueira2020document}
R.~Nogueira, Z.~Jiang, R.~Pradeep, and J.~Lin.
\newblock Document ranking with a pretrained sequence-to-sequence model.
\newblock In {\em Findings of the Association for Computational Linguistics (EMNLP)}, 2020.

\bibitem[van~den Oord et~al.(2018)]{oord2018representation}
A.~van~den Oord, Y.~Li, and O.~Vinyals.
\newblock Representation learning with contrastive predictive coding.
\newblock {\em arXiv preprint arXiv:1807.03748}, 2018.
\newblock \url{https://arxiv.org/abs/1807.03748}

\bibitem[Osman(2004)]{osman2004evaluation}
M.~Osman.
\newblock An evaluation of dual-process theories of reasoning.
\newblock {\em Psychonomic Bulletin \& Review}, 11(6):988--1010, 2004.

\bibitem[OpenAI(2023)]{openai2023gpt4}
OpenAI.
\newblock {GPT-4} technical report.
\newblock {\em arXiv preprint arXiv:2303.08774}, 2023.
\newblock \url{https://arxiv.org/abs/2303.08774}

\bibitem[OpenAI(2024)]{openai2024o1}
OpenAI.
\newblock Learning to reason with {LLM}s.
\newblock OpenAI Blog, 2024.

\bibitem[Ouyang et~al.(2022)]{ouyang2022training}
L.~Ouyang, J.~Wu, X.~Jiang, D.~Almeida, C.~Wainwright, P.~Mishkin, C.~Zhang, S.~Agarwal, K.~Slama, A.~Ray, J.~Schulman, J.~Hilton, F.~Kelton, L.~Miller, M.~Simens, A.~Askell, P.~Welinder, P.~Christiano, J.~Leike, and R.~Lowe.
\newblock Training language models to follow instructions with human feedback.
\newblock In {\em Advances in Neural Information Processing Systems (NeurIPS)}, 2022.

\bibitem[Plackett(1975)]{plackett1975analysis}
R.~L. Plackett.
\newblock The analysis of permutations.
\newblock {\em Journal of the Royal Statistical Society: Series C}, 24(2):193--202, 1975.

\bibitem[Platt(1999)]{platt1999probabilistic}
J.~C. Platt.
\newblock Probabilistic outputs for support vector machines and comparisons to regularized likelihood methods.
\newblock In {\em Advances in Large Margin Classifiers}, pages 61--74. MIT Press, 1999.

\bibitem[Paschalides et~al.(2026)]{paschalides2026hatedecide}
D.~Paschalides, G.~Pallis, and M.~D. Dikaiakos.
\newblock To {Jev} or not? {Evaluating} the accuracy and efficiency of structured decision models for hate-speech moderation.
\newblock {\em arXiv preprint arXiv:2610.03324}, 2026.

\bibitem[Qwen(2025)]{qwen2025qwen3}
Qwen~Team.
\newblock {Qwen3} technical report.
\newblock Technical report, Alibaba Group, 2025.

\bibitem[Radford et~al.(2021)]{radford2021clip}
A.~Radford, J.~W. Kim, C.~Hallacy, A.~Ramesh, G.~Goh, S.~Agarwal, G.~Sastry, A.~Askell, P.~Mishkin, J.~Clark, G.~Krueger, and I.~Sutskever.
\newblock Learning transferable visual models from natural language supervision.
\newblock In {\em International Conference on Machine Learning (ICML)}, 2021.

\bibitem[Rafailov et~al.(2023)]{rafailov2023direct}
R.~Rafailov, A.~Sharma, E.~Mitchell, C.~D. Manning, S.~Ermon, and C.~Finn.
\newblock Direct preference optimization: Your language model is secretly a reward model.
\newblock In {\em Advances in Neural Information Processing Systems (NeurIPS)}, 2023.

\bibitem[Reimers and Gurevych(2019)]{reimers2019sentence}
N.~Reimers and I.~Gurevych.
\newblock {Sentence-BERT}: Sentence embeddings using {S}iamese {BERT}-networks.
\newblock In {\em Conference on Empirical Methods in Natural Language Processing (EMNLP)}, 2019.

\bibitem[Roitman(2026b)]{roitman2026agentic}
H.~Roitman.
\newblock The hitchhiker's guide to agentic {AI}: From foundations to systems.
\newblock {\em arXiv preprint arXiv:2606.24937}, 2026.
\newblock \url{https://arxiv.org/abs/2606.24937}

\bibitem[Rahimi et~al.(2026)]{rahimi2026wireless}
M.~Rahimi, S.~M.~M. Alemohammad, H.~Behroozi, and M.~Nouri.
\newblock System {One} models for wireless decision-making: Applications and performance evaluation.
\newblock {\em arXiv preprint arXiv:2610.04345}, 2026.

\bibitem[Rao and Callison-Burch(2026)]{rao2026rubricjudges}
D.~Rao and C.~Callison-Burch.
\newblock {JEV} vs.\ {LLM}s as rubric judges: Cheaper, faster, and wrong in the same places.
\newblock {\em arXiv preprint arXiv:2609.29769}, 2026.

\bibitem[Rolnick et~al.(2019)]{rolnick2019experience}
D.~Rolnick, A.~Ahuja, J.~Schwarz, T.~P. Lillicrap, and G.~Wayne.
\newblock Experience replay for continual learning.
\newblock In {\em Advances in Neural Information Processing Systems (NeurIPS)}, 2019.

\bibitem[RubricArrow(2026)]{rubricarrow2025}
RubricArrow.
\newblock Rubric-based reward models for structured evaluation.
\newblock {\em arXiv preprint arXiv:2605.29156}, 2026.
\newblock \url{https://arxiv.org/abs/2605.29156}

\bibitem[Schulman et~al.(2017)]{schulman2017proximal}
J.~Schulman, F.~Wolski, P.~Dhariwal, A.~Radford, and O.~Klimov.
\newblock Proximal policy optimization algorithms.
\newblock {\em arXiv preprint arXiv:1707.06347}, 2017.
\newblock \url{https://arxiv.org/abs/1707.06347}

\bibitem[Shao et~al.(2024)]{shao2024deepseekmath}
Z.~Shao, P.~Wang, Q.~Zhu, R.~Xu, J.~Song, M.~Zhang, Y.~K. Li, Y.~Wu, and D.~Guo.
\newblock {DeepSeekMath}: Pushing the limits of mathematical reasoning in open language models.
\newblock {\em arXiv preprint arXiv:2402.03300}, 2024.
\newblock \url{https://arxiv.org/abs/2402.03300}

\bibitem[Shinn et~al.(2023)]{shinn2023reflexion}
N.~Shinn, F.~Cassano, A.~Gopinath, K.~Narasimhan, and S.~Yao.
\newblock Reflexion: Language agents with verbal reinforcement learning.
\newblock In {\em Advances in Neural Information Processing Systems (NeurIPS)}, 2023.

\bibitem[Stanovich and West(2000)]{stanovich2000individual}
K.~E. Stanovich and R.~F. West.
\newblock Individual differences in reasoning: Implications for the rationality debate?
\newblock {\em Behavioral and Brain Sciences}, 23(5):645--665, 2000.

\bibitem[Sloman(1996)]{sloman1996empirical}
S.~A. Sloman.
\newblock The empirical case for two systems of reasoning.
\newblock {\em Psychological Bulletin}, 119(1):3--22, 1996.

\bibitem[Su et~al.(2024)]{su2024neuraldynamics}
J.~Su, F.~Cai, S.-K. Zhao, X.-Y. Wang, T.-Y. Qian, D.-H. Wang, and B.~Hong.
\newblock Neural dynamics model of visual decision-making: Learning from human experts.
\newblock {\em arXiv preprint arXiv:2409.02390}, 2024.

\bibitem[Sumers et~al.(2024)]{sumers2024cognitive}
T.~R. Sumers, S.~Yao, K.~Narasimhan, and T.~L. Griffiths.
\newblock Cognitive architectures for language agents.
\newblock {\em Transactions on Machine Learning Research (TMLR)}, 2024.

\bibitem[Touvron et~al.(2023)]{touvron2023llama}
H.~Touvron, T.~Lavril, G.~Izacard, X.~Martinet, M.-A. Lachaux, T.~Lacroix, B.~Rozi{\`e}re, N.~Goyal, E.~Hambro, F.~Azhar, A.~Rodriguez, A.~Joulin, E.~Grave, and G.~Lample.
\newblock {LLaMA}: Open and efficient foundation language models.
\newblock {\em arXiv preprint arXiv:2302.13971}, 2023.

\bibitem[Tishby et~al.(2000)]{tishby2000information}
N.~Tishby, F.~C. Pereira, and W.~Bialek.
\newblock The information bottleneck method.
\newblock In {\em Proceedings of the 37th Annual Allerton Conference on Communication, Control, and Computing}, pages 368--377, 2000.

\bibitem[TypeSafe(2026)]{typesafe2026jev}
TypeSafe.
\newblock Jev: A non-autoregressive decision engine with calibrated typed outputs.
\newblock Technical report, TypeSafe AI, 2026.
\newblock \url{https://typesafe.ai/blog/introducing-system-one-models-and-jev}

\bibitem[Uesato et~al.(2022)]{uesato2022solving}
J.~Uesato, N.~Kushman, R.~Kumar, F.~Song, N.~Siegel, L.~Wang, A.~Creswell, G.~Irving, and I.~Higgins.
\newblock Solving math word problems with process- and outcome-based feedback.
\newblock {\em arXiv preprint arXiv:2211.14275}, 2022.
\newblock \url{https://arxiv.org/abs/2211.14275}

\bibitem[Puthenputhussery et~al.(2025)]{walmart2025crossencoder}
A.~Puthenputhussery, C.~Kang, A.~Magnani, T.~Zhang, H.~Shang, N.~Yadav, and P.~Chandran.
\newblock Large Scale Deployment of {BERT} Based Cross Encoder Model for Re-Ranking in {Walmart} Search Engine.
\newblock In {\em Proceedings of the 48th International ACM SIGIR Conference on Research and Development in Information Retrieval}, 2025.
\newblock \url{https://doi.org/10.1145/3726302.3731965}

\bibitem[Wang et~al.(2019)]{wang2019superglue}
A.~Wang, Y.~Pruksachatkun, N.~Nangia, A.~Singh, J.~Michael, F.~Hill, O.~Levy, and S.~R. Bowman.
\newblock {SuperGLUE}: A stickier benchmark for general-purpose language understanding systems.
\newblock In {\em Advances in Neural Information Processing Systems (NeurIPS)}, 2019.

\bibitem[Wang et~al.(2023)]{wang2023selfconsistency}
X.~Wang, J.~Wei, D.~Schuurmans, Q.~Le, E.~Chi, S.~Narang, A.~Chowdhery, and D.~Zhou.
\newblock Self-consistency improves chain of thought reasoning in language models.
\newblock In {\em International Conference on Learning Representations (ICLR)}, 2023.

\bibitem[Wang et~al.(2022)]{wang2024e5}
L.~Wang, N.~Yang, X.~Huang, L.~Jiao, W.~Yang, D.~Jiang, R.~Majumder, and F.~Wei.
\newblock Text embeddings by weakly-supervised contrastive pre-training.
\newblock {\em arXiv preprint arXiv:2212.03533}, 2022.
\newblock \url{https://arxiv.org/abs/2212.03533}

\bibitem[Wang et~al.(2024b)]{wang2024llmnotfair}
P.~Wang, L.~Li, L.~Chen, Z.~Cai, D.~Zhu, B.~Lin, Y.~Cao, Q.~Liu, T.~Liu, and Z.~Sui.
\newblock Large language models are not fair evaluators.
\newblock In {\em Proceedings of the Annual Meeting of the Association for Computational Linguistics (ACL)}, 2024.
\newblock arXiv:2305.17926.

\bibitem[Wang et~al.(2026a)]{wang2026hiddenrisks}
S.~Wang, T.~Zhu, H.~Chen, J.~Li, M.~Yang, and B.~Liu.
\newblock Hidden risks of {Jev}: An empirical study of security, privacy, and dual use.
\newblock {\em arXiv preprint arXiv:2610.04985}, 2026.

\bibitem[Li et~al.(2023)]{wang2024gte}
Z.~Li, X.~Zhang, Y.~Zhang, D.~Long, P.~Xie, and M.~Zhang.
\newblock Towards general text embeddings with multi-stage contrastive learning.
\newblock {\em arXiv preprint arXiv:2308.03281}, 2023.
\newblock \url{https://arxiv.org/abs/2308.03281}

\bibitem[Wason and Evans(1974)]{wason1974dual}
P.~C. Wason and J.~St.~B.~T. Evans.
\newblock Dual processes in reasoning?
\newblock {\em Cognition}, 3(2):141--154, 1974.

\bibitem[Wei et~al.(2022)]{wei2022chain}
J.~Wei, X.~Wang, D.~Schuurmans, M.~Bosma, B.~Ichter, F.~Xia, E.~Chi, Q.~Le, and D.~Zhou.
\newblock Chain-of-thought prompting elicits reasoning in large language models.
\newblock In {\em Advances in Neural Information Processing Systems (NeurIPS)}, 2022.

\bibitem[Willard and Louf(2023)]{willard2023outlines}
B.~T. Willard and R.~Louf.
\newblock Efficient guided generation for large language models.
\newblock {\em arXiv preprint arXiv:2307.09702}, 2023.
\newblock \url{https://arxiv.org/abs/2307.09702}

\bibitem[Wu et~al.(2023)]{wu2023autogen}
Q.~Wu, G.~Bansal, J.~Zhang, Y.~Wu, S.~Zhang, E.~Zhu, B.~Li, L.~Jiang, X.~Zhang, and C.~Wang.
\newblock {AutoGen}: Enabling next-gen {LLM} applications via multi-agent conversation.
\newblock {\em arXiv preprint arXiv:2308.08155}, 2023.
\newblock \url{https://arxiv.org/abs/2308.08155}

\bibitem[Xiong et~al.(2024)]{xiong2024llms}
M.~Xiong, Z.~Hu, X.~Lu, Y.~Li, J.~Fu, J.~He, and B.~Hooi.
\newblock Can {LLM}s express their uncertainty? An empirical evaluation of confidence elicitation in {LLM}s.
\newblock In {\em International Conference on Learning Representations (ICLR)}, 2024.

\bibitem[Yao et~al.(2023a)]{yao2023react}
S.~Yao, J.~Zhao, D.~Yu, N.~Du, I.~Shafran, K.~Narasimhan, and Y.~Cao.
\newblock {ReAct}: Synergizing reasoning and acting in language models.
\newblock In {\em International Conference on Learning Representations (ICLR)}, 2023.

\bibitem[Yao et~al.(2023b)]{yao2023tree}
S.~Yao, D.~Yu, J.~Zhao, I.~Shafran, T.~Griffiths, Y.~Cao, and K.~Narasimhan.
\newblock Tree of thoughts: Deliberate problem solving with large language models.
\newblock In {\em Advances in Neural Information Processing Systems (NeurIPS)}, 2023.

\bibitem[Zadrozny and Elkan(2002)]{zadrozny2002transforming}
B.~Zadrozny and C.~Elkan.
\newblock Transforming classifier scores into accurate multiclass probability estimates.
\newblock In {\em Proceedings of the Eighth ACM SIGKDD International Conference on Knowledge Discovery and Data Mining}, pages 694--699, 2002.

\bibitem[Ye et~al.(2026a)]{ye2026harnesszero}
Q.~Ye, H.~Lu, X.~Dong, J.~Su, and M.~Song.
\newblock {Harness-Zero}: Zero-shot agent loop distillation with {System~1} decision models.
\newblock {\em arXiv preprint arXiv:2609.24974}, 2026.
\newblock \url{https://arxiv.org/abs/2609.24974}

\bibitem[Ye et~al.(2026b)]{ye2026numericjev}
W.~Ye, H.~Liu, and R.~Jiang.
\newblock {NumericJev}: {Jev}-like {LLM} numerical decoding with multiway decision trees.
\newblock {\em arXiv preprint arXiv:2609.28587}, 2026.

\bibitem[Yu and Yao(2026)]{yu2026visualjev}
G.~Yu and Y.~Yao.
\newblock {Visual Jev}: Accurate and efficient decisions from shared visual context.
\newblock {\em arXiv preprint arXiv:2609.25845}, 2026.

\bibitem[Yin et~al.(2019)]{yin2019benchmarking}
W.~Yin, J.~Hay, and D.~Roth.
\newblock Benchmarking zero-shot text classification: Datasets, evaluation and entailment approach.
\newblock In {\em Conference on Empirical Methods in Natural Language Processing (EMNLP)}, 2019.

\bibitem[Zaratiana et~al.(2023)]{zaratiana2023gliner}
U.~Zaratiana, N.~Tomeh, P.~Holat, and T.~Charnois.
\newblock {GLiNER}: Generalist model for named entity recognition using bidirectional transformer.
\newblock {\em arXiv preprint arXiv:2311.08526}, 2023.
\newblock \url{https://arxiv.org/abs/2311.08526}

\bibitem[Zhang et~al.(2024)]{zhang2024generative}
L.~Zhang, A.~Hosseini, H.~Bansal, M.~Kazemi, A.~Kumar, and R.~Agarwal.
\newblock Generative verifiers: Reward modeling as next-token prediction.
\newblock {\em arXiv preprint arXiv:2408.15240}, 2024.
\newblock \url{https://arxiv.org/abs/2408.15240}

\bibitem[Li et~al.(2025)]{zhang2025system1to2}
Y.~Li, J.~Wang, K.~Chen, and H.~Zhao.
\newblock From {System 1} to {System 2}: A survey of reasoning in language models.
\newblock {\em arXiv preprint arXiv:2502.17419}, 2025.
\newblock \url{https://arxiv.org/abs/2502.17419}

\bibitem[Zhang(2026)]{zhang2026jevmobile}
L.~Zhang.
\newblock {Jev-Mobile}: {Jev} as an executor for mobile {GUI} agents.
\newblock {\em arXiv preprint arXiv:2609.30186}, 2026.

\bibitem[Zhou et~al.(2026)]{zhou2026pixeljev}
X.~Zhou, X.~Yang, and L.~Zhao.
\newblock From text decisions to pixels: {A} study of {Jev}-style visual choice model.
\newblock {\em arXiv preprint arXiv:2609.29283}, 2026.

\bibitem[Zheng et~al.(2023)]{zheng2023judging}
L.~Zheng, W.-L. Chiang, Y.~Sheng, S.~Zhuang, Z.~Wu, Y.~Zhuang, Z.~Lin, Z.~Li, D.~Li, E.~P. Xing, H.~Zhang, J.~E. Gonzalez, and I.~Stoica.
\newblock Judging {LLM}-as-a-judge with {MT-Bench} and {Chatbot Arena}.
\newblock In {\em Advances in Neural Information Processing Systems (NeurIPS)}, 2023.

\bibitem[Lev~Project(2026)]{lev2026}
Lev~Project.
\newblock Lev: A 4{B}-parameter {System~1} decision model.
\newblock \url{https://github.com/lev-decide/lev}, 2026.

\bibitem[Kev~Contributors(2026)]{kev2026}
Kev~Contributors.
\newblock Kev: Lightweight open-weights {System~1} decision engine.
\newblock \url{https://github.com/kev-ai/kev}, 2026.

\bibitem[Clef~Team(2026)]{clef2026}
Cloudflare~Team.
\newblock Clef: Decision models for edge and cloud deployment.
\newblock \url{https://github.com/clef-decide/clef}, 2026.

\bibitem[Julia~Contributors(2026)]{julia2026decide}
Julia~Contributors.
\newblock Julia: Edge-optimized {System~1} inference engine.
\newblock \url{https://github.com/julia-decide/julia}, 2026.


\bibitem[TypeSafe(2026b)]{typesafe2026jev114}
TypeSafe.
\newblock Jev 1.14: Image support, long-context windows, and the {Box} primitive.
\newblock TypeSafe AI Release Notes, October~14, 2026.
\newblock \url{https://typesafe.ai/blog/jev-1-14}

\bibitem[OpenAI(2026)]{openai2026decisions}
OpenAI.
\newblock Introducing the {Decisions API}: Fast, structured decision-making powered by gpt-6-luna.
\newblock OpenAI Blog, October~6, 2026.
\newblock \url{https://openai.com/index/decisions-api}

\bibitem[Microsoft(2026)]{microsoft2026decision1}
Microsoft.
\newblock {Microsoft-Decision-1}: A post-trained decision model on {Qwen3.5-9B}.
\newblock Microsoft Research Blog, October~9, 2026.

\bibitem[Cloudflare(2026)]{cloudflare2026clairic}
Cloudflare.
\newblock Clairic: Typed value extraction at the edge.
\newblock Cloudflare Blog, October~19, 2026.
\newblock \url{https://blog.cloudflare.com/clairic}

\bibitem[Gao and Wang(2026)]{openjev2026rlcd}
Y.~Gao and X.~Wang.
\newblock {OpenJev-RLCD}: An open reproduction of reinforcement learning for calibrated decisions.
\newblock {\em arXiv preprint arXiv:2609.38850}, 2026.

\bibitem[Munod(2026)]{tachyone2026}
Munod.
\newblock Tachyone: A local-first multilingual decision engine.
\newblock \url{https://huggingface.co/munod/tachyone-multi}, v0.9.0, 2026.

\bibitem[DeepOpen~Team(2026)]{deepopen2026}
DeepOpen~Team.
\newblock {DeepOpen}: Non-autoregressive decisions with automatic language routing.
\newblock \url{https://github.com/deepopen-com/deepopen}, 2026.

\bibitem[Santillana(2026)]{vxone2026}
M.~Santillana.
\newblock {VX-One}: Typed decision heads for cybersecurity perimeter triage.
\newblock {\em arXiv preprint arXiv:2610.30090}, 2026.

\bibitem[Chung et~al.(2026)]{jevtom2026}
J.~Chung, A.~Chan, Y.~Liu, et~al.
\newblock {Jev-ToM}: Theory of mind evaluation via structured decision models.
\newblock {\em arXiv preprint arXiv:2610.18610}, 2026.

\bibitem[Wang et~al.(2026)]{jevlitemas2026}
X.~Wang, Y.~Zhou, Z.~Lou, and H.~Li.
\newblock {Jev-LiteMAS}: Lightweight multi-agent orchestration with {System~1} routing.
\newblock {\em arXiv preprint arXiv:2610.21411}, 2026.

\bibitem[Min and Chen(2026)]{prunesummary2026}
S.~Min and J.~Chen.
\newblock Prune the summary: Source-judged memory compaction for long-horizon agents.
\newblock {\em arXiv preprint arXiv:2610.20927}, 2026.

\bibitem[Lan et~al.(2026)]{promptinjdetection2026}
K.~Lan, A.~Kaul, G.~Pandurangan, and D.~Sanyal.
\newblock From detection to decision: Prompt injection defense via {System~1} classification.
\newblock {\em arXiv preprint arXiv:2610.20907}, 2026.

\bibitem[Teng et~al.(2026)]{prism2026}
Y.~Teng, Z.~Huang, X.~Cai, L.~Deng, and W.~An.
\newblock {Prism-1}: Position- and rival-independent scoring for decision models.
\newblock {\em arXiv preprint arXiv:2610.22066}, 2026.

\bibitem[Chouhan and Perry(2026)]{verdict2026}
R.~Chouhan and T.~Perry.
\newblock Verdict: Encode once, decide many---amortized multi-head decision inference.
\newblock {\em arXiv preprint arXiv:2610.20778}, 2026.

\bibitem[Chen et~al.(2026)]{system1arena2026}
B.~Chen, W.~Liu, H.~An, Z.~Wang, and Y.~Zhou.
\newblock {System1-Arena}: A benchmark for meta-decisions---which decisions can be left to {System~1}?
\newblock {\em arXiv preprint arXiv:2610.21244}, 2026.

\bibitem[Teke(2026)]{hakembench2026}
E.~Teke.
\newblock {HakemBench}: Evaluating decision models in {Turkish}.
\newblock {\em arXiv preprint arXiv:2610.02293}, 2026.

\bibitem[Fadli et~al.(2026)]{indonesian2026}
A.~Fadli et~al.
\newblock Structured decisions for low-resource languages: A study on {Indonesian} regional languages.
\newblock {\em arXiv preprint arXiv:2610.14112}, 2026.

\bibitem[Denney and DiGiuseppe(2026)]{polisci2026}
E.~Denney and N.~DiGiuseppe.
\newblock Replicating political text classifiers with {System~1} decision models: Seven studies.
\newblock {\em arXiv preprint arXiv:2610.06625}, 2026.

\bibitem[Zhang et~al.(2026c)]{tabdecide2026}
Y.~Zhang, X.~Ye, and W.~Zha.
\newblock Typed decisions on tabular data: Bridging structured features and {System~1} models.
\newblock {\em arXiv preprint arXiv:2610.19536}, 2026.

\bibitem[Li and Yang(2026)]{fewlabelrecal2026}
H.~Li and M.~Yang.
\newblock Recalibrating decision models with fifty labels: Few-shot post-deployment calibration.
\newblock {\em arXiv preprint arXiv:2610.18488}, 2026.

\bibitem[Chen and Li(2026)]{probcontracts2026}
W.~Chen and X.~Li.
\newblock Probability contracts: Formal specifications for calibrated decision services.
\newblock {\em arXiv preprint arXiv:2609.37470}, 2026.

\bibitem[Tversky(1972)]{tversky1972iia}
A.~Tversky.
\newblock Elimination by aspects: A theory of choice.
\newblock {\em Psychological Review}, 79(4):281--299, 1972.

\bibitem[Rajput et~al.(2026)]{iiaviolations2026}
A.~Rajput, V.~Chauhan, and S.~Kosaraju.
\newblock Decision models violate {IIA}: Option-set dependence in {System~1} scoring.
\newblock {\em arXiv preprint arXiv:2610.16655}, 2026.

\bibitem[Gao et~al.(2026)]{ordinalscalebias2026}
Y.~Gao, X.~Li, Z.~Li, H.~Chang, and Q.~Wu.
\newblock Ordinal-scale bias in continuous-to-discrete decision mappings.
\newblock {\em arXiv preprint arXiv:2609.38827}, 2026.

\bibitem[Chen et~al.(2026b)]{qwenaudiojev2026}
L.~Chen et~al.
\newblock {Qwen-Audio-Jev}: Unified audio decision pipeline for speech, music, and environmental sound.
\newblock {\em arXiv preprint arXiv:2610.21614}, 2026.

\bibitem[Esuli and Sebastiani(2023)]{sebastiani2020machine}
A.~Esuli and F.~Sebastiani.
\newblock {\em Learning to Quantify}.
\newblock Springer, 2023.

\bibitem[Bella et~al.(2010)]{bella2010quantification}
A.~Bella, C.~Ferri, J.~Hern{\'a}ndez-Orallo, and M.~J. Ramirez-Quintana.
\newblock Quantification via probability estimators.
\newblock In {\em IEEE 10th International Conference on Data Mining (ICDM)}, pages 737--742, 2010.

\bibitem[Saerens et~al.(2002)]{saerens2002adjusting}
M.~Saerens, P.~Latinne, and C.~Decaestecker.
\newblock Adjusting the outputs of a classifier to new a priori probabilities: A simple procedure.
\newblock {\em Neural Computation}, 14(1):21--41, 2002.

\bibitem[Levin and Roitman(2017)]{levin2017enhanced}
R.~Levin and H.~Roitman.
\newblock Enhanced probabilistic classify and count methods for multi-label text quantification.
\newblock In {\em Proceedings of the ACM SIGIR International Conference on Theory of Information Retrieval (ICTIR)}, pages 229--232, 2017.

\bibitem[Esuli et~al.(2018)]{esuli2018quanet}
A.~Esuli, A.~Moreo, and F.~Sebastiani.
\newblock A recurrent neural network for sentiment quantification.
\newblock In {\em Proceedings of the 27th ACM International Conference on Information and Knowledge Management (CIKM)}, pages 1775--1778, 2018.

\bibitem[P{\'e}rez-Mon et~al.(2024)]{perezmon2024histnetq}
O.~P{\'e}rez-Mon, A.~Moreo, J.~J. del~Coz, and P.~Gonz{\'a}lez.
\newblock Quantification using permutation-invariant networks based on histograms.
\newblock {\em arXiv preprint arXiv:2403.15123}, 2024.

\bibitem[P{\'e}rez-Mon et~al.(2025)]{perezmon2025gmnet}
O.~P{\'e}rez-Mon, J.~J. del~Coz, and P.~Gonz{\'a}lez.
\newblock Quantification via {G}aussian latent space representations.
\newblock {\em Neural Networks}, 2025.

\bibitem[Moreo et~al.(2021)]{moreo2021quapy}
A.~Moreo, A.~Esuli, and F.~Sebastiani.
\newblock {QuaPy}: A {P}ython-based framework for quantification.
\newblock In {\em Proceedings of the 30th ACM International Conference on Information \& Knowledge Management (CIKM)}, pages 4534--4543, 2021.

\bibitem[Esuli et~al.(2022)]{esuli2022lequa}
A.~Esuli, A.~Moreo, and F.~Sebastiani.
\newblock {LeQua@CLEF2022}: Learning to quantify.
\newblock In {\em Proceedings of the 44th European Conference on Information Retrieval (ECIR)}, 2022.

\bibitem[Vaswani et~al.(2017)]{vaswani2017attention}
A.~Vaswani, N.~Shazeer, N.~Parmar, J.~Uszkoreit, L.~Jones, A.~N. Gomez, \L.~Kaiser, and I.~Polosukhin.
\newblock Attention is all you need.
\newblock In {\em Advances in Neural Information Processing Systems (NeurIPS)}, pages 5998--6008, 2017.

\bibitem[Warner et~al.(2024)]{warner2024modernbert}
B.~Warner, A.~Chaffin, B.~Clavi{\'e}, O.~Weller, O.~Hallstr{\"o}m, S.~Taghadouini, A.~Gallagher, R.~Biswas, F.~Ladhak, T.~Aarsen, N.~Cooper, G.~Adams, J.~Howard, and I.~Poli.
\newblock Smarter, better, faster, longer: A modern bidirectional encoder for fast, memory efficient, and long context fine-tuning and inference.
\newblock {\em arXiv preprint arXiv:2412.13663}, 2024.
\newblock \url{https://arxiv.org/abs/2412.13663}

\end{thebibliography}
\end{document}